\documentclass[10pt,twocolumn,letterpaper]{article}

\usepackage[pagenumbers]{cvpr} 
\usepackage{graphicx}
\usepackage{amsmath}

\usepackage{bm}
\usepackage{multirow}
\usepackage{caption}
\usepackage{array}
\usepackage{float}
\usepackage{placeins}
\usepackage{cuted}

\usepackage[utf8]{inputenc} 
\usepackage[T1]{fontenc}    

\usepackage{booktabs}       
\usepackage{amsfonts}       
\usepackage{nicefrac}       
\usepackage{microtype}      
\usepackage{xcolor} 
\usepackage{soul}
\usepackage{mathtools}
\usepackage{subcaption}
\usepackage{adjustbox}
\usepackage{algorithm}
\usepackage{algpseudocode}
\usepackage{lipsum}
\usepackage{pifont}

\newcommand{\yesmark}{\ding{51}}
\newcommand{\nomark}{\ding{55}}

\newcommand{\lossren}{\mathcal{L}_\mathcal{R}}
\newcommand{\lossgeo}{\mathcal{L}_\mathcal{G}}
\newcommand{\mesh}{\bm{\Theta}}

\newcommand{\vdisp}{\bm{D}_{\text{vert}}}
\newcommand{\vnorm}{\bm{\vec{n}}}

\newcommand{\imgt}{\bm{I}_{\text{gt}}}
\newcommand{\depthgt}{\bm{D}_{\text{gt}}}
\newcommand{\pressuregt}{\bm{P}_{\text{gt}}}
\newcommand{\maskgt}{\bm{M}_{\text{gt}}}

\newcommand{\tran}{\bm{t}}
\newcommand{\camrot}{\bm{R}_{\text{cam}}^i}
\newcommand{\camtran}{\bm{t}_{\text{cam}}^i}
\newcommand{\pose}{\bm{\theta}}

\newcommand{\texture}{\bm{\mathcal{T}}}
\newcommand{\textpres}{\mathcal{T}_P}

\newcommand{\bigv}[1]{\overset{\raisebox{-0.5ex}{\scalebox{0.9}[0.3]{\textbf{v}}}}{#1}}
\newcommand{\vrenfs}{\bigv{\mathcal{R}}}
\newcommand{\renf}{\mathcal{R}_F}
\newcommand{\rens}{\mathcal{R}_M}
\newcommand{\rend}{\mathcal{R}_D}
\newcommand{\vrenp}{\vrenfs_P(\Theta^*,\textpres)}
\newcommand{\vrend}{\vrenfs_D(\Theta^*)[z]}

\def\poseoptimization{\textsc{Pose Optimization}}
\def\shaperefinement{\textsc{Shape Refinement}}

\newcolumntype{C}[1]{>{\centering\arraybackslash}p{#1}}

\definecolor{cvprblue}{rgb}{0.21,0.49,0.74}
\usepackage[pagebackref,breaklinks,colorlinks,allcolors=cvprblue]{hyperref}

\def\paperID{***} 
\def\confName{CVPR}
\def\confYear{2025}

\newcommand\blfootnote[1]{%
  \begingroup
  \renewcommand\thefootnote{}\footnote{#1}%
  \addtocounter{footnote}{-1}%
  \endgroup
}
\def\dataset{EgoPressure}

\title{EgoPressure: A Dataset for Hand Pressure and Pose Estimation \\in Egocentric Vision}
\author{ Yiming Zhao$^{1*}$ ~~~~~~
  Taein Kwon$^{1*}$ ~~~~~~
  Paul Streli$^{1*}$ ~~~~~~
  Marc Pollefeys$^{1,2}$ ~~~~~~
  Christian Holz$^{1}$ 
  \smallskip \\ \\
$^1$ETH Z\"urich~~~~~~$^2$Microsoft Spatial AI Lab, Z\"urich
}

\begin{document}

\twocolumn[{%
\renewcommand\twocolumn[1][]{#1}%
\maketitle

\begin{center}
\includegraphics[width=1.0\textwidth]{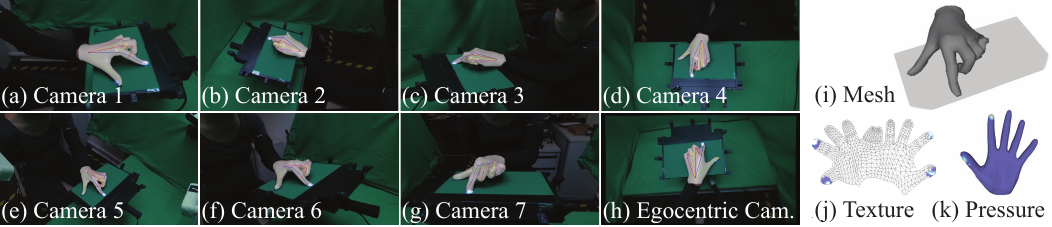}
    \captionof{figure}{\textbf{The EgoPressure dataset.} We introduce a novel egocentric pressure dataset with hand poses. 
   We label hand poses using our proposed optimization method across all static camera views (Cameras 1–7).
   The annotated hand mesh aligns well with the egocentric camera's view, indicating the high fidelity of our annotations. 
   We project the pressure intensity and annotated hand mesh~(Fig. \textit{i}) to all camera views (Fig. \textit{a} to \textit{h}), and further provide the pressure applied over the hand as a UV texture map~(Fig. \textit{j} and \textit{k}).
   }
   \label{fig:teasor}
 
\end{center}
}]
\maketitle
\blfootnote{*Equal contribution.}

\begin{abstract}

Touch contact and pressure are essential for understanding how humans interact with and manipulate objects, insights which can significantly benefit applications in mixed reality and robotics.
However, estimating these interactions from an egocentric camera perspective is challenging, largely due to the lack of comprehensive datasets that provide both accurate hand poses on contacting surfaces and detailed annotations of pressure information.
In this paper, we introduce \dataset, a novel egocentric dataset that captures detailed touch contact and pressure interactions.
\dataset{} provides high-resolution pressure intensity annotations for each contact point and includes accurate hand pose meshes obtained through our proposed multi-view, sequence-based optimization method processing data from an 8-camera capture rig.
Our dataset comprises 5 hours of recorded interactions from 21 participants captured simultaneously by one head-mounted and seven stationary Kinect cameras, which acquire RGB images and depth maps at 30\,Hz.
To support future research and benchmarking, we present several baseline models for estimating applied pressure on external surfaces from RGB images, with and without hand pose information.
We further explore the joint estimation of the hand mesh and applied pressure.
Our experiments demonstrate that pressure and hand pose are complementary for understanding hand-object interactions. Project page: \url{ https://yiming-zhao.github.io/EgoPressure/}.

\end{abstract}

\section{Introduction}
\label{sec:intro}

Having a sense of touch contact and pressure during hand-object interaction is crucial for a large variety of tasks in Augmented Reality (AR)~\cite{wilson2005playanywhere, opportunisticcontrols2010henderson}, Virtual Reality (VR)~\cite{grady2024pressurevision++, mrtouch2018xiao}, and robotic manipulation~\cite{mandikal2021learning, christen2022d, collins2023visual}.
In particular, estimating these physical properties from an egocentric perspective is a central enabler to support real-world tasks~\cite{streli2024touchinsight, meier2021tapid}. 
In AR/VR environments, touch contact and pressure information allow for more precise control and feedback~\cite{ismar2022-comfortableUIs}. 
For example, when users play a virtual piano on a table, the sound can change based on the pressure applied to the keys, providing a more refined feedback experience that current AR/VR systems lack~\cite{meier2021tapid}.
The sense of pressure is also crucial for enabling robots to accurately replicate human manipulation, as determining the precise pressure required to grasp objects remains a significant challenge~\cite{christen2022d, lambeta2024digitizing, collins2023visual}.

Previous approaches have used gloves~\cite{luo2024adaptive, luo2021learning} and robots with tactile sensors~\cite{lambeta2024digitizing, zlokapa2022integrated} to capture pressure measurements during object manipulation.
However, this instrumentation interferes with natural touch by obstructing tactile feedback.
In contrast, vision-based estimation methods require no instrumentation of the hands, and cameras are already integrated into devices like smart glasses and mixed-reality headsets, which are widely used to study human behavior from an egocentric perspective~\cite{grauman2023ego, grauman2022ego4d}.
Despite this potential, advancements in state-of-the-art (SOTA) models have been limited by the lack of datasets that provide contact and pressure data.
A notable exception is the PressureVision dataset~\cite{grady2022pressurevision} that comprises RGB footage from four static cameras of hands interacting with a pressure-sensitive surface and corresponding projected pressure images.

In this paper, we present a natural yet significant extension to this prior work~\cite{grady2022pressurevision, grady2024pressurevision++} by introducing a novel dataset, \dataset, which captures hand-surface interactions from an egocentric perspective, complete with accurate hand pose annotations and pressure maps projected onto the hand mesh.
Our capture platform combines a Sensel Morph touchpad with a head-mounted camera and seven synchronized Azure Kinect cameras, all recording RGB-D data at 30\,Hz (Figure~\ref{fig:teasor}).
The dataset includes 5~hours of footage from 21~participants, each performing 64~interaction sequences with an average length of 420~frames---making it the first bare-handed egocentric pressure dataset with pose and mesh annotations.

We further provide baseline models to demonstrate the potential of our dataset and establish a benchmark for future research.
First, we set PressureVisionNet~\cite{grady2022pressurevision} as a baseline on our egocentric dataset and compare it to adapted models that incorporate hand pose as additional input.
The model using hand poses estimated from the RGB images via the HaMeR~\cite{hamer} estimator outperforms PressureVisionNet by more than 5\% in volumetric IoU error, with improvements of over 7\% when using ground-truth hand poses.
Additionally, we introduce the first model to jointly estimate hand pose, hand mesh, and pressure both over the mesh and on the surface from an egocentric RGB camera, thereby localizing contact and pressure in 3D space.

We summarize our key contributions as follows:
\begin{enumerate}
\item \dataset\ is the first high-quality egocentric touch contact and pressure dataset with 3D hand poses. This enables the development of models that can generalize to movable cameras such as head-mounted and body-worn cameras.
We will make our dataset and annotations publicly available upon acceptance.
\item We present an optimization method to annotate hand poses from our multi-view capture setup using MANO~\cite{MANO:SIGGRAPHASIA:2017}.
 Our mesh-based hand pose annotations account for vertex displacement, supporting accurate hand manipulation analysis by projecting pressure inversely from the Sensel Morph touchpad onto the hand meshes.
\item We establish two novel benchmarks: (1)~estimating contact pressure from egocentric RGB images with and without hand pose information, and (2)~jointly reconstructing 3D hand poses and pressure, including the localization of pressure on a user's hand mesh.
\end{enumerate}

\noindent \dataset\, thus offers new opportunities for future models to address the unique challenges of egocentric perspectives and to precisely localize pressure on a user’s hand.

\section{Related Work}

Our work is related to hand-object pose estimation, contact estimation and pressure sensing.

\begin{table*}[]
\caption{Comparison between \dataset{} and selected hand-contact datasets. The overwhelming majority of prior datasets infer contacts based on hand and object pose. ContactLabelDB and PressureVisionDB also include ground-truth touch pressure but are limited to static cameras and do not provide accurate hand poses and meshes. Please see appendix for the full table.}
\resizebox{\textwidth}{!}{
\begin{tabular}{l|ccccccccccccc}

\textbf{Dataset} & \textbf{frames} & \textbf{participants} & \textbf{hand pose} & \textbf{hand mesh} & \textbf{markerless}  & \textbf{real} & \textbf{egocentric} & \textbf{multiview} & \textbf{RGB} & \textbf{depth} & \textbf{contact} & \multicolumn{2}{c}{\textbf{pressure}} \\
 &  &  &  &  &   &  &  &  &  &  &  &  {surface} & {hand}   \\ \hline
\textbf{EgoPressure (ours)}                                   & 4.3M & 21 & \yesmark{} & \yesmark{} & \yesmark{}  & \yesmark{} & \yesmark{} & \yesmark{} & \yesmark{} & \yesmark{} & Pressure sensor & \yesmark{}           &  \yesmark{}    \\ 
ContactLabelDB~\cite{grady2024pressurevision++} & 2.9M & 51 &  \nomark{} & \nomark{} & \yesmark{} & \yesmark{}  & \nomark{} & \yesmark{} & \yesmark{} & \nomark{} & Pressure sensor & \yesmark{}                       		&  \nomark{}   \\
PressureVisionDB~\cite{grady2022pressurevision} & 3.0M & 36 & \nomark{} & \nomark{} & \yesmark{} & \yesmark{}  & \nomark{} & \yesmark{} & \yesmark{} & \nomark{} & Pressure sensor & \yesmark{}                          	&  \nomark{}   \\
ContactPose~\cite{brahmbhatt2020contactpose} & 3.0M & 50 & \yesmark{} & \yesmark{} & \yesmark{}  & \yesmark{}  & \nomark{} & \yesmark{} & \yesmark{} & \yesmark{} & Thermal imprint & \nomark{}								&  \nomark{}   \\
GRAB~\cite{taheri2020grab} & 1.6M & 10 & \yesmark{} & \yesmark{} & \nomark{} & \yesmark{}  & \nomark{} & \nomark{} & \nomark{} & \nomark{} & Inferred from Pose & \nomark{}                              					&  \nomark{}   \\
ARCTIC~\cite{fan2023arctic} & 2.1M & 10 & \yesmark{} & \yesmark{} & \nomark{} & \yesmark{} & \yesmark{} & \yesmark{} & \yesmark{} & \yesmark{} & Inferred from Pose & \nomark{}                               				&  \nomark{}   \\
H2O~\cite{kwon2021h2o}  & 571k & 4 & \yesmark{} & \yesmark{} & \yesmark{} & \yesmark{} & \yesmark{} & \yesmark{} & \yesmark{} & \yesmark{} & Inferred from Pose & \nomark{}                             					&  \nomark{}   \\
OakInk~\cite{yang2022oakink} & 230k & 12 & \yesmark{} & \yesmark{} & \yesmark{}  & \yesmark{} & \nomark{} & \yesmark{} & \yesmark{}  & \yesmark{} & Inferred from Pose & \nomark{}                            				&  \nomark{}   \\
OakInk-2~\cite{zhan2024oakink2} & 4.01M & 9 & \yesmark{} & \yesmark{} & \yesmark{}  & \yesmark{} & \yesmark{} & \yesmark{} & \yesmark{} & \nomark{} & Inferred from Pose & \nomark{}                              			&  \nomark{}   \\
DexYCB~\cite{chao2021dexycb} & 582k & 10 & \yesmark{} & \yesmark{} & \yesmark{}  & \yesmark{} & \nomark{} & \yesmark{} & \yesmark{} & \yesmark{} & Inferred from Pose & \nomark{}                              				&  \nomark{}   \\
HO-3D~\cite{hampali2020honnotate} & 103k & 10 & \yesmark{} & \yesmark{} & \yesmark{}  & \yesmark{} & \nomark{} & \yesmark{} & \yesmark{} & \yesmark{} & Inferred from Pose & \nomark{}                             			&  \nomark{}   \\
TACO~\cite{liu2024taco} & 5.2M & 14 & \yesmark{} & \yesmark{} & \yesmark{}  & \yesmark{} & \yesmark{} & \yesmark{} & \yesmark{} & \yesmark{} & Inferred from Pose & \nomark{}                              					&  \nomark{}   \\

\end{tabular}
}
\label{tab:priorwork}
\end{table*}

\paragraph{Vision-based hand-object pose estimation} Hand tracking has been a long-standing challenge in computer vision, with applications in robotics~\cite{christen2023synh2r, li2020mobile}, human-computer interaction~\cite{han2022umetrack, han2020megatrack, rehg1994digiteyes}, and medicine~\cite{boato2009hand, lugo2017virtual, hein2021towards}.
Over the past decade, significant progress has been made, largely due to advancements in deep learning techniques~\cite{mueller2017real, hamer} and the collection of relevant datasets~\cite{zimmermann2019freihand, yuan2017bighand2, moon2020interhand2}.
While egocentric hand tracking for gesture recognition and direct input has advanced to the point of integration into modern commercial devices such as AR and VR headsets~\cite{han2022umetrack, han2020megatrack}, understanding hand interactions with external objects remains an active area of research~\cite{garcia2018first, fan2023arctic, kwon2021h2o, grauman2023ego}.
Datasets gathered to aid machine understanding of such hand-object interactions rely on additional instrumentation of the users' hands~\cite{garcia2018first}, motion capture systems with hand-attached markers~\cite{taheri2020grab, fan2023arctic}, or multi-view camera rigs~\cite{kwon2021h2o, yang2022oakink, zhan2024oakink2, chao2021dexycb, hampali2020honnotate} to capture accurate ground-truth poses of users' hands under the higher degree of occlusion caused by the object.

\paragraph{Hand-object contact estimation} In addition to object-relative hand pose, prior work has aimed to estimate contact points between the users' hands and external objects~\cite{taheri2020grab, fan2023arctic}.
Research has shown that when used as input proxies, real-world physical objects improve input control and provide haptic feedback~\cite{ismar2022-comfortableUIs}.
For interactive research purposes, external tracking systems~\cite{ismar2022-comfortableUIs, richardson2020decoding} and wearable sensors such as acoustic sensors~\cite{Acoustic} and inertial measurement units were used to estimate contact~\cite{Takayashi_telemetring, Yizheng_IMU, finger_mounted, meier2021tapid, Gupta_acustico, chi2022-taptype}.
Additionally, vision-based techniques have been developed that use fiducial markers~\cite{lee2003arkb}, active illumination for shadow creation~\cite{liang2023shadowtouch, wilson2005playanywhere}, vibration detection~\cite{streli2023structured}, or depth sensing~\cite{depth_estimate, Dante_vision, Imaginary_phone, FarOut, Ian_depth, Xu_latency, mrtouch2018xiao, xiao2016direct}.
More recent work estimates touch using passive cameras without additional instrumentation on the user's hand or surface, enabling deployment on commercial mixed reality headsets~\cite{streli2024touchinsight, richardson2024stegotype}.
More detailed contact maps are inferred based on the intersection of tracked hand and object meshes~\cite{taheri2020grab, fan2023arctic, kwon2021h2o, zhan2024oakink2, grady2021contactopt}, requiring sub-millimeter accuracy—a challenging task for complex gestures due to soft tissue dynamics.
To address this, Brahmbhatt et al.~\cite{brahmbhatt2020contactpose} used thermal imaging to obtain accurate contact maps.
Additionally, prior efforts have utilized simulations to obtain more granular labels about contacting tissue~\cite{zhu2023contactart, corona2020ganhand, hasson2019learning}.

\paragraph{Hand pressure estimation}

Moving beyond the mere detection of contact, prior work has estimated the pressure forces applied during hand interactions, which is crucial for robotic grasping tasks~\cite{mandikal2021learning, christen2022d} and provides an additional control dimension for input~\cite{quinn2021deep}.
To estimate pressure from monocular images, visual cues such as fingernail alterations~\cite{chen2020estimating, mascaro2004measurement} or surface deformations~\cite{mollyn2024egotouch, hwang2017inferring} during press events have been used.
Changes in object trajectory and interaction forces~\cite{ehsani2020use, li2019estimating, pham2017hand} also offer insights but are ineffective with static objects like tables and walls.
Accurate pressure labels for training usually require instrumenting the user's hands with gloves~\cite{buscher2015flexible, sundaram2019learning, tactileGlove} or the surface with force sensors~\cite{pham2017hand, grady2022pressurevision, tekscan}, ideally flexible or conforming to various shapes~\cite{kim2011capacitive, bhirangi2021reskin, luo2021learning}. However, this alters the visual appearance and tactile features of the hands and surface, affecting interaction and limiting generalization to bare hands and uninstrumented surfaces.
Grady et al.~\cite{grady2022pressurevision, grady2024pressurevision++} collected two datasets with ground-truth pressure maps using a Sensel Morph~\cite{senselmorph} pressure sensor to train a neural network for estimating contact regions on surfaces from single RGB images.
However, their method relies on an external static camera and good visibility of the corresponding fingertips.

With \dataset{}, we aim to bridge this gap by offering a dataset that includes egocentric views, utilizing head-mounted cameras to better understand human interactions from this perspective. Our dataset also captures accurate hand poses and meshes from multiple camera views without the use of markers. To the best of our knowledge, we provide the first dataset containing egocentric and multi-view RGB\nobreakdash-D images of a bare hand interacting with a surface, along with synchronized pressure data, hand poses, and meshes (see Table~\ref{tab:priorwork}).

\section{Marker-less Annotation Method}

\label{sec:method}

\begin{figure*}[h]

\centering

\includegraphics[width=1.0\linewidth]{"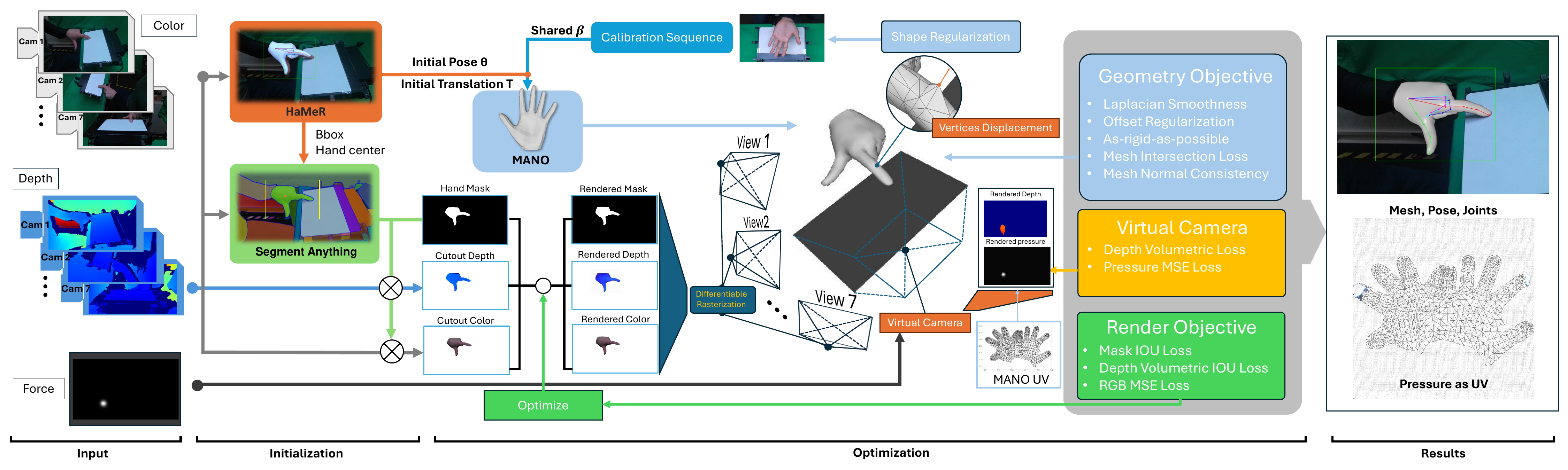"}\\
\caption{\textbf{Method overview.} The input for our annotation method consists of RGB-D images captured by 7 static Azure Kinect cameras and the pressure frame from a Sensel Morph touchpad. We leverage Segment-Anything~\cite{sam} and HaMeR~\cite{hamer} to obtain initial hand poses and masks. We refine the initial hand pose and shape estimates through differentiable rasterization~\cite{dibr} optimization across all static camera views. Using an additional virtual orthogonal camera placed below the touchpad, we reproject the captured pressure frame onto the hand mesh by optimizing the pressure as a texture feature of the corresponding UV map, while ensuring contact between the touchpad and all contact vertices.}
\label{fig:pipeline}
\end{figure*}

To capture accurate hand poses and meshes during hand-surface interactions without markers, we developed a multi-camera hand pose annotation method using the MANO hand model~\cite{MANO:SIGGRAPHASIA:2017}, differentiable rendering and multi-objective optimization. Figure~\ref{fig:pipeline} shows an overview of our method, which relies on $C$ static cameras and a pressure-sensitive touchpad.
Please see the supplementary material for a detailed evaluation of our annotation method.

\subsection{Automatic hand pose initialization}

We use HaMeR~\cite{hamer} to estimate an initial MANO hand pose \(\pose_{\text{init}}\) and translation \(\tran_{\text{init}}\) for each static camera.
Since HaMeR's prediction is based on a single RGB image, there is a scale-translation ambiguity, which we resolve by triangulating the root joints from the 7 static camera views.
The orientation and hand pose are then initialized based on the output of a single camera view.
HaMeR also provides a bounding box, which we use along with the 2D projected hand root as input to Segment-Anything (SAM)~\cite{sam}, from which we obtain an annotated segmentation mask \(\maskgt\) for the hand in each static camera image.

\subsection{Annotation refinement}

Based on the initial hand pose, we obtain refined hand pose annotations via the following optimization using the input from the $C$ \emph{static} cameras. We use the MANO~\cite{MANO:SIGGRAPHASIA:2017,manopth} model for mesh representation with 25 PCA components and employ the DIB-R~\cite{dibr} differentiable renderer. The annotations include the hand pose \(\pose\), hand translation \(\tran\), vertex displacement \(\vdisp\) in world coordinates, and the pressure over the hand mesh in the form of a texture map \(\textpres\). All static cameras are pre-calibrated, allowing us to project the hand mesh into the frame of each static camera $i$ using the extrinsic parameters $[\camrot | \camtran]$. 
\paragraph{$\beta$-calibration}
For the MANO shape parameters \(\beta\), we use separate calibration sequences for each hand of each participant, during which the participant slowly turns their hand to be visible from all cameras, with fingers spread.
For these sequences, we also optimize the MANO shape parameters \(\beta\) with \(l_2\) regularization in the previous optimization.
The shape parameters are then reused for all other sequences for the given participant, with \(\beta\) remaining fixed during subsequent optimizations.

Following HARP~\cite{harp}, our annotation algorithm consists of two stages: (1) \poseoptimization{} and (2) \shaperefinement{}, with a rendering objective \(\lossren\) and a geometry objective \(\lossgeo\).

Beginning with the first stage, \poseoptimization{}, the focus is on annotating the hand pose \(\pose\) and translation \(\tran\). Consequently, the hand mesh \(\mesh\) can be derived directly from the MANO model~\cite{manopth}, expressed as \(\mesh = \mathrm{MANO}(\pose, \beta) + \tran\). 
We note that certain parts of the hand, such as fingers, may not be visible from all camera angles---for instance, fingers obscured by the palm in a curled gesture. To address this, we incorporate the mesh intersection loss \(\mathcal{L}_{\text{insec}}\)~\cite{Karras:2012:MPC:2383795.2383801,Tzionas:IJCV:2016}. The objective function is then defined as:
\begin{equation}
\mathcal{L}_{\text{pose}}(\mesh) = \lossren(\mesh) + \mathcal{L}_{\text{insec}}(\mesh)
\label{eq:loss_pose_opt}
\end{equation}
The rendering objective \(\lossren\) and the mesh intersection loss \(\mathcal{L}_{\text{insec}}\) will be detailed in the supplementary material.
In the \shaperefinement{} stage, the pose \(\pose\) and translation \(\tran\) of the hand remain fixed. The optimization process introduces vertex displacement \(\vdisp\). Each vertex is adjusted by an offset along its normal vector \(\vnorm\), which is computed from the last epoch of the \poseoptimization{} stage, to minimize the rendering loss \(\lossren(\mesh^*)\). Consequently, the refined hand mesh \(\mesh^*\) can be expressed as \(\mesh^* = \mesh + \vnorm \cdot \vdisp\). To ensure a reasonable mesh, the geometry objective \(\lossgeo\) is also included in the optimization. Additionally, we introduce a virtual render \(\vrenfs\) to optimize pressure as a UV map \(\textpres\) and minimize the distance between the hand mesh \(\mesh^*\) and the contact area on the surface of the touchpad.
The objective function \(\mathcal{L}_{\text{shape}}\) for this stage is as follows:
\begin{equation}
\mathcal{L}_{\text{shape}}(\mesh^*) = \lossren(\mesh^*) + \lossgeo(\mesh^*) + \mathcal{L}_{\vrenfs}(\mesh^*)
\label{eq:loss_shape_opt}
\end{equation}

The virtual render \(\vrenfs\), and its objective \(\mathcal{L}_{\vrenfs}\) will be explained in the next section and the other terms in the geometry objective \(\lossgeo\) will be detailed in the supplementary material.

\subsubsection{Virtual Render for Contact and Pressure}
As shown in Figure~\ref{fig:pipeline}, we also incorporate the captured pressure data in the optimization as a hand mesh texture feature for our proposed virtual rendering method.
For this, we position a virtual orthogonal camera \(\vrenfs\) under the touchpad, oriented upwards in the world coordinate system. The render size matches the resolution of the touchpad, and the camera's plane overlaps with the touchpad's sensing surface. The goal is for the rendered pressure \(\vrenp\) on the hand mesh, with texture mapping of an optimized pressure UV map \(\textpres\), to align with the input pressure \(\pressuregt\).

Additionally, we infer the contact area from \(\pressuregt\) using a simple pressure threshold. Using this contact area as a mask, we ensure that the masked rendered z-axis depth \(\vrend\) aligns with the distance \(Z_{v2p}\) from the camera to the touchpad, thereby ensuring physical contact.

The objective function \(\mathcal{L}_{\vrenfs}(\mesh^*)\) for the virtual render is:
\begin{align}
   \mathcal{L}_{\vrenfs}(\mesh^*) & =  \quad \mathrm{MSE}(\vrenp, \pressuregt)  \notag \\
         & \quad + \left| \mathbb{I}(\pressuregt > 0) \odot (\vrend - Z_{v2p}) \right|_1.  \notag\\
         &  \hfill
    \label{eq:virtual_render}
\end{align}

\section{\dataset{} Dataset}

\dataset{} comprises 4.3M RGB-D frames (2560\,$\times$\,1440 for static camera, 1920\,$\times$\,1080 for egocentric camera) capturing interactions of both left and right hands with a touch and pressure-sensitive planar surface.
The dataset features 21 participants performing 31 distinct gestures, such as touch, drag, pinch, and press, with each hand.
It includes a total of 5.0 hours of hand gesture footage comprised of synchronized RGB-D frames from seven calibrated static cameras and one head-mounted camera, along with ground-truth pressure maps from the pressure-sensitive surface captured at a frame rate of 30\,fps.
We used four different surface textures for the data capture rig, which also includes a green wall to facilitate synthetic background augmentation.
Additionally, we provide high-fidelity hand pose and mesh data for the hands during interactions based on our proposed annotation method (see Section~\ref{sec:method}), as well as the tracked pose of the head-mounted camera.
With \dataset{}, we aim to offer a substantial dataset for egocentric hand pose and pressure estimation during interactions with rigid surfaces, thereby advancing machine understanding of human interaction with their surroundings through the fundamental modality of touch.

\begin{figure}[h]

    \centering
    \begin{minipage}[b]{0.49\textwidth}
        \centering
        \includegraphics[width=\textwidth]{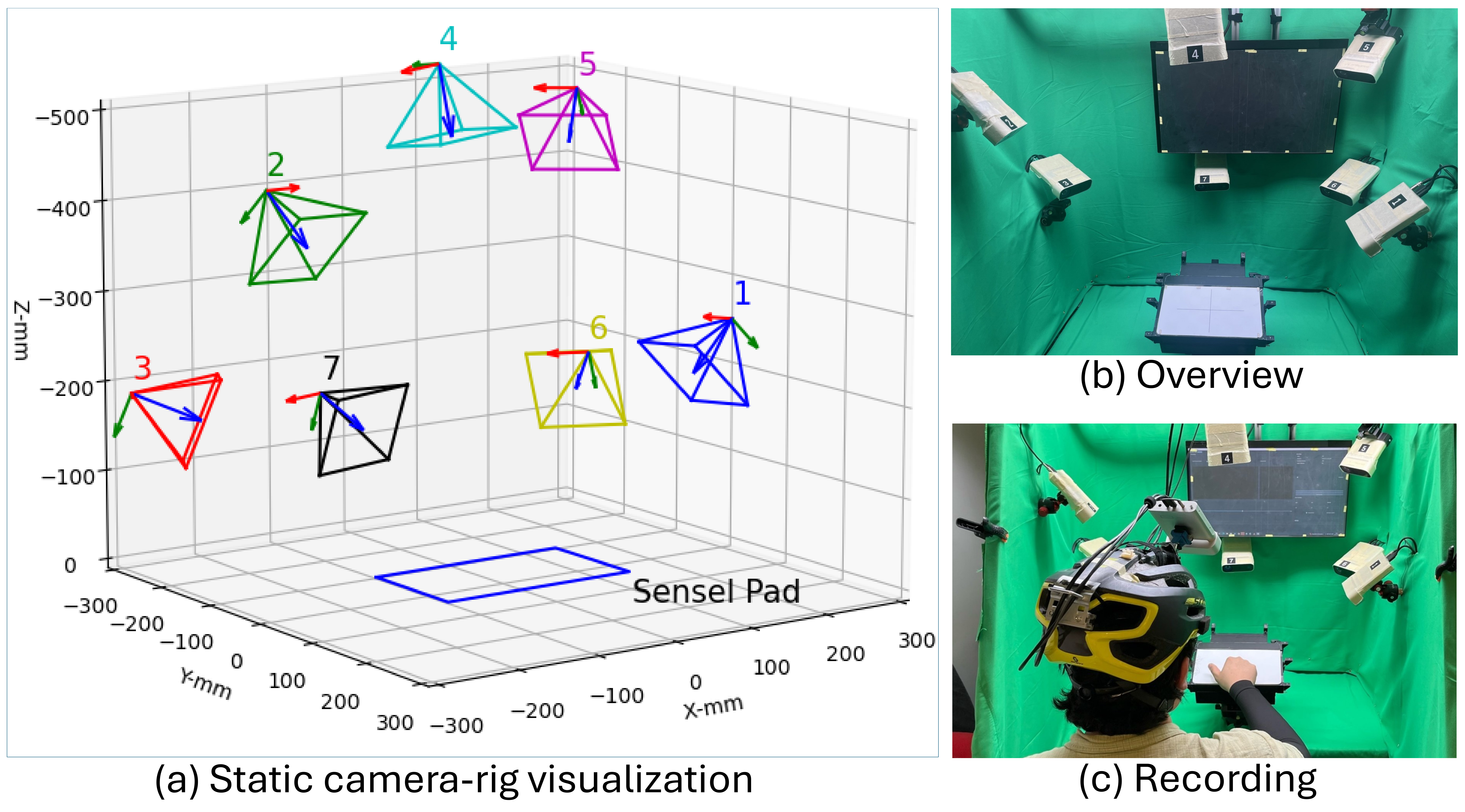}
        \caption{7 static\,+\,1 egocentric camera rig}
        \label{fig:capturerig}
    \end{minipage}
    \hfill
    \begin{minipage}[b]{0.49\textwidth}
        \centering
        \includegraphics[width=\textwidth]{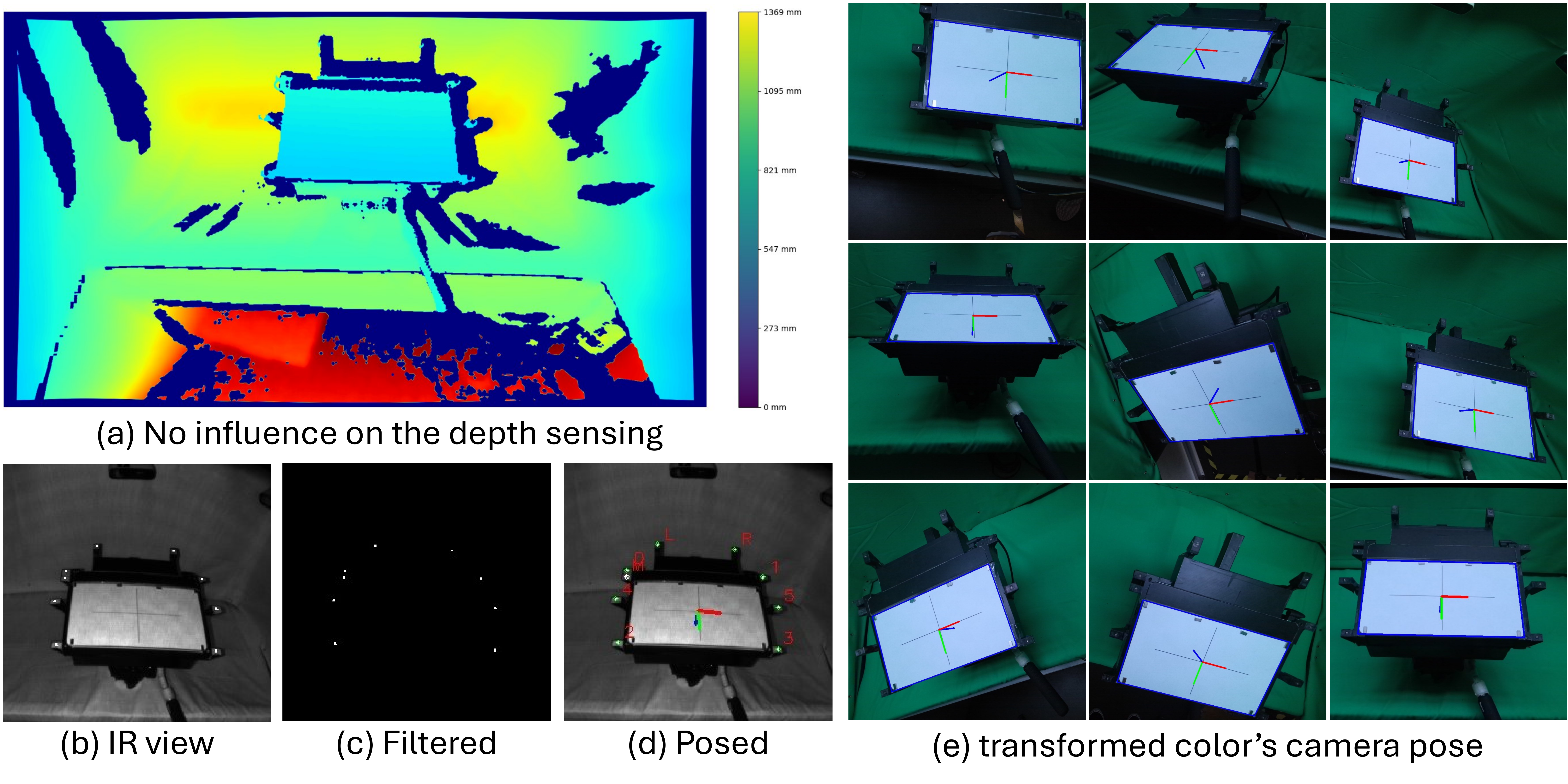}
        \caption{Camera pose tracking with IR makers}
        \label{fig:posetrack}
    \end{minipage}
\vspace{-3em}
\end{figure}

\begin{figure*}
\centering
   \includegraphics[width=0.9\textwidth]{"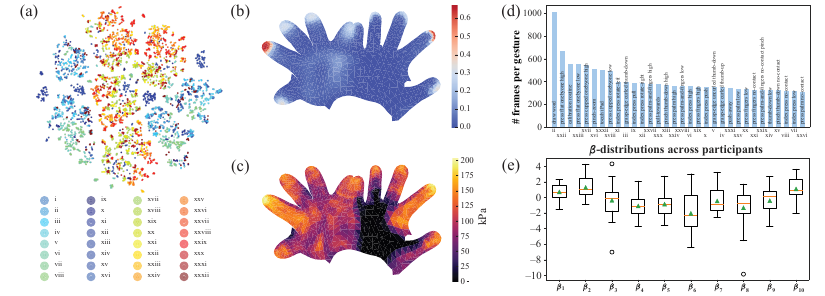"}
~   ~ \caption{\textbf{(a)}~t-SNE~\cite{van2008visualizing} visualization of hand pose frames $\theta$ over our dataset, with color coding for different gestures. All gestures are listed in Table \ref{tab:gestures} of the supplementary material.  \textbf{(b)}~Ratio of touch frames with contact for each vertex. \textbf{(c)}~Maximum pressure over hand vertices across dataset. \textbf{(d)}~Mean length of performed gestures. \textbf{(e)}~Distribution of $\beta$ values across participants.}
   \label{fig:statistics}
\end{figure*}
\subsection{Data capture setup}

\begin{figure}[h]
\centering
                \includegraphics[width=0.5\textwidth]{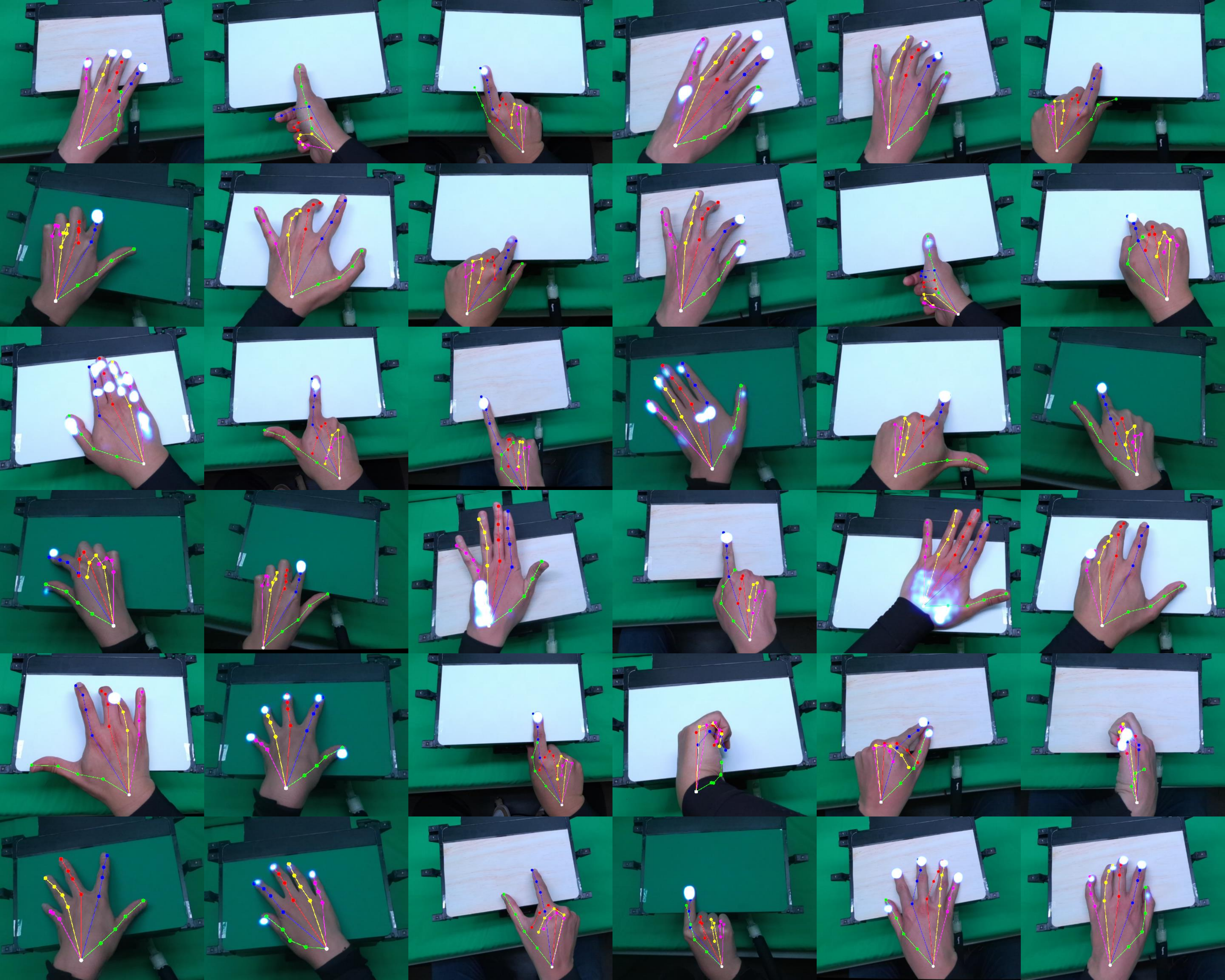}  
                           \caption{\footnotesize  Thumbnail of different poses in egocentric views}
                \label{fig:thumbnail}
\end{figure}

\begin{figure}[h]
    \centering
    \vspace{-1em}

        \begin{minipage}{0.38\textwidth}
            \includegraphics[width=\textwidth]{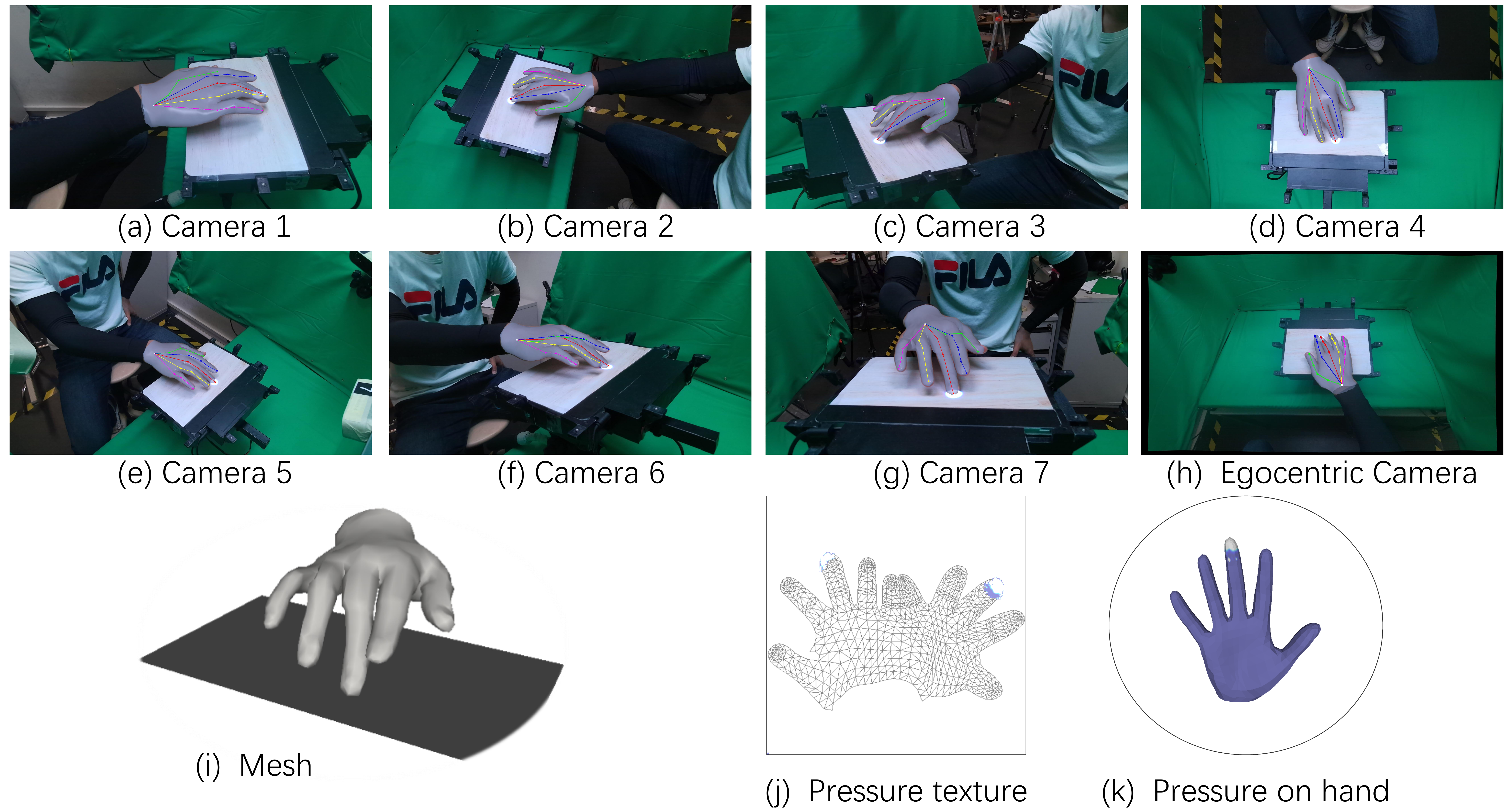} 
            \subcaption{\footnotesize Annotation examples for a right hand}
        \end{minipage}
       \hfill
        \begin{minipage}{0.38\textwidth}
            \includegraphics[width=\textwidth]{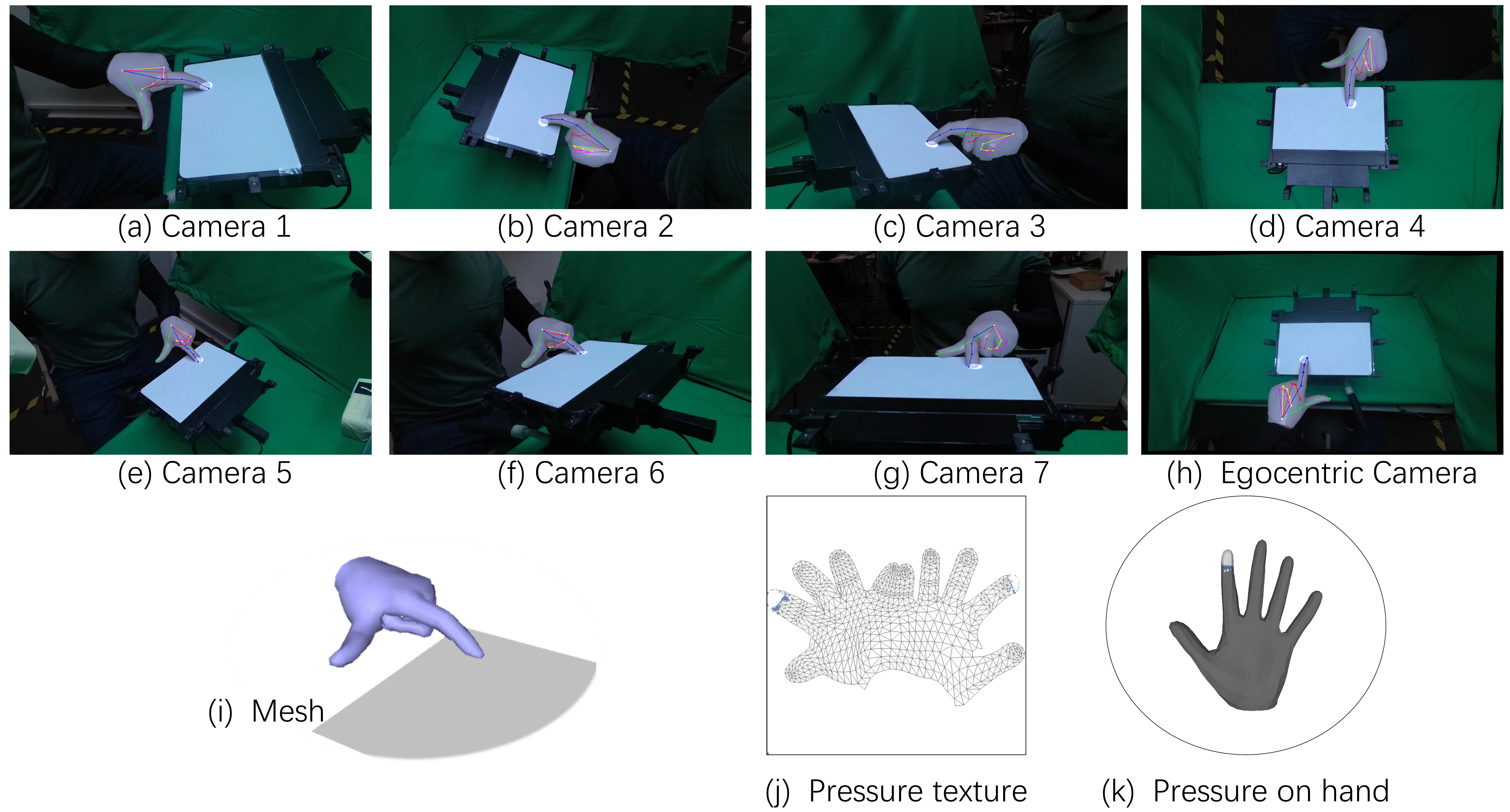} 
            \subcaption{\footnotesize Annotation examples for a left hand}

        \end{minipage}
    \caption{Sample data from EgoPressure}
    \label{fig:sample_data}
    \vspace{-6mm}
        
\end{figure}

To capture accurate ground-truth labels for hand pose and pressure from egocentric views, we constructed a data capture rig that integrates a pressure-sensitive touchpad (Sensel Morph~\cite{senselmorph}) for touch and pressure sensing, along with seven static and one head-mounted RGB-D camera (Azure Kinect~\cite{microsoft_azure}) to capture RGB and depth images (see Figure~\ref{fig:capturerig}).
The touchpad (Sensel Morph), measuring 240\,$\times$\,169.5\,mm, is mounted on a tripod head.  
We use four different texture overlays (white, green, dark wood, light wood) printed on paper and placed over the Sensel Morph pad across participants.
The seven static Azure Kinect cameras are attached to the aluminum frame, and the head-mounted camera is fixed on a helmet.
The frame also holds a computer display and is surrounded by a green screen.

All cameras and the touchpad are connected to two workstations (Intel Core i7-9700K, Nvidia GeForce RTX 3070), their timestamps are synchronized via a Raspberry Pi CM4 using PTP, which also triggers all Azure Kinect cameras simultaneously at a frame rate of 30\,fps.

\paragraph{Head-mounted camera tracking} To obtain accurate poses of the head-mounted camera, we attach nine active infrared markers around the Sensel Morph pad in an asymmetric layout (see Figure~\ref{fig:posetrack}).  These markers, controlled by the Raspberry Pi CM4, are identifiable in the Azure Kinect's infrared image using simple thresholding (saturating the range of values of the infrared camera).
The markers are turned on simultaneously, allowing for the computation of the camera pose via Perspective-N-Points and enabling an accurate evaluation of the temporal synchronization between cameras and the touchpad.

\subsection{Participants}

We recruited 21 participants from our institution (6 female, 15 male, ages 23--32 years, mean age\,=\,26 years), ensuring a broad representation to cover anatomic differences in hand characteristics. Participants' heights ranged from 160--194\,cm (mean\,=\,174, SD\,=\,9), weights from 51--95\,kg (mean\,=\,69, SD\,=\,14), and middle finger lengths from 7.3--9.2\,cm (mean\,=\,7.9, SD\,=\,0.5) (see Figure~\ref{fig:statistics} for distribution of MANO $\beta$-values). 
Please find details on instructions given to participants, consent form, and IRB approval in the supplementary material.

\subsection{Data acquisition procedure}

Participants sat on an adjustable stool in front of the apparatus, wearing a helmet with a mounted camera pointing towards the Sensel Morph and a black hand stocking on each arm up to the wrist.
Before starting the data capture, the experimenter explained the task and the purpose of the study.
They then signed a consent form and provided demographic information.
The participants first performed a calibration gesture by slowly turning each hand, with fingers spread, within the camera rig.
After calibration, participants conducted 31 different gestures, including touch, press, and drag gestures of varying strength, with each hand on the Sensel Morph touchpad (see supplementary material for a description of gestures).
Each gesture was repeated 5 times if it involved a single touch action (e.g., press index finger) and 3 times if it involved a sequence of sequential touches (e.g., draw letters).
Before each gesture, participants watched a video demonstrating how to perform the corresponding gesture with written instructions on a computer monitor in front of them.
The experimenter guided the participants throughout the study, which took around 1 hour per participant.
Participants could take a break after each gesture and received a chocolate bar as gratitude for their participation.
In total, we recorded 6216 different gestures, i.e., 21\,participants $\times$\,2\,hands\,$\times$\,(1\,calibration + 27\,$\times$\,5 + 4\,$\times$\,3) gestures.

\subsection{Data statistics}

The average length of each motion sequence is 14 seconds, with an almost equal balance between frames capturing the left and right hands. Figure~\ref{fig:statistics} shows the mean sequence lengths across gestures.
Approximately 45.1\% of all frames capture the hand in contact with the pressure-sensitive pad.
Figure~\ref{fig:statistics}b visualizes the ratio of contact frames with a given vertex touching the surface, and Figure~\ref{fig:statistics}c shows the maximum pressure measured for each vertex.
Following Grady et al.~\cite{grady2022pressurevision}, we set a threshold of 0.5\,kPa as the minimum effective pressure to discard diffuse readings from the touchpad.

\section{Benchmark Evaluation}
\label{sec:Evaluation}
Previous work estimates applied pressure maps using only RGB images~\cite{grady2022pressurevision, grady2024pressurevision++}.
With \dataset, we explore the advantages of incorporating accurate hand poses as additional input, which naturally provide richer context about the interaction.
We introduce new benchmarks for estimating hand pressure using both RGB images and 3D hand poses.
Additionally, we propose a novel network architecture that jointly estimates, from a single RGB image, the pressure applied to both an external surface and across the hand, providing a deeper understanding of the regions of the hand involved throughout the interaction.

\subsection{Image-projected Pressure Baselines}
\label{sec:imageprojpressure}
We evaluate the RGB-based baseline, \emph{PressureVisionNet}~\cite{grady2022pressurevision}, on our dataset. Since the PressureVision dataset~\cite{grady2022pressurevision} includes only static camera views, we split our baseline experiments into egocentric and exocentric views. Specifically, we use camera views \emph{2, 3, 4, and 5} for our dataset, as these have comparable orientation and distance to the touchpad as the cameras in PressureVision.

We test our hypothesis that incorporating hand pose as additional input enhances pressure estimation.
To this end, we design a straightforward extension of PressureVisionNet~\cite{grady2022pressurevision}. We augment the encoder-decoder segmentation architecture, originally designed for RGB inputs, by adding an additional channel for 2.5D hand key points. This involves projecting the 21 3D hand joints onto the image plane and adding their depth (z-coordinate) from the egocentric camera's coordinate system, scaled to millimeters.

For evaluation, we use both the ground truth hand joints from our annotations and the predicted hand joints from HaMeR~\cite{hamer}. The HaMeR-estimated hand poses serve as a fair baseline, reflecting the performance of current SOTA RGB-based hand pose estimators, while the ground truth joints provide an upper bound, demonstrating the potential improvement achievable with more accurate hand poses.

\begin{figure}[h]
\centering
   \includegraphics[width=0.47\textwidth]{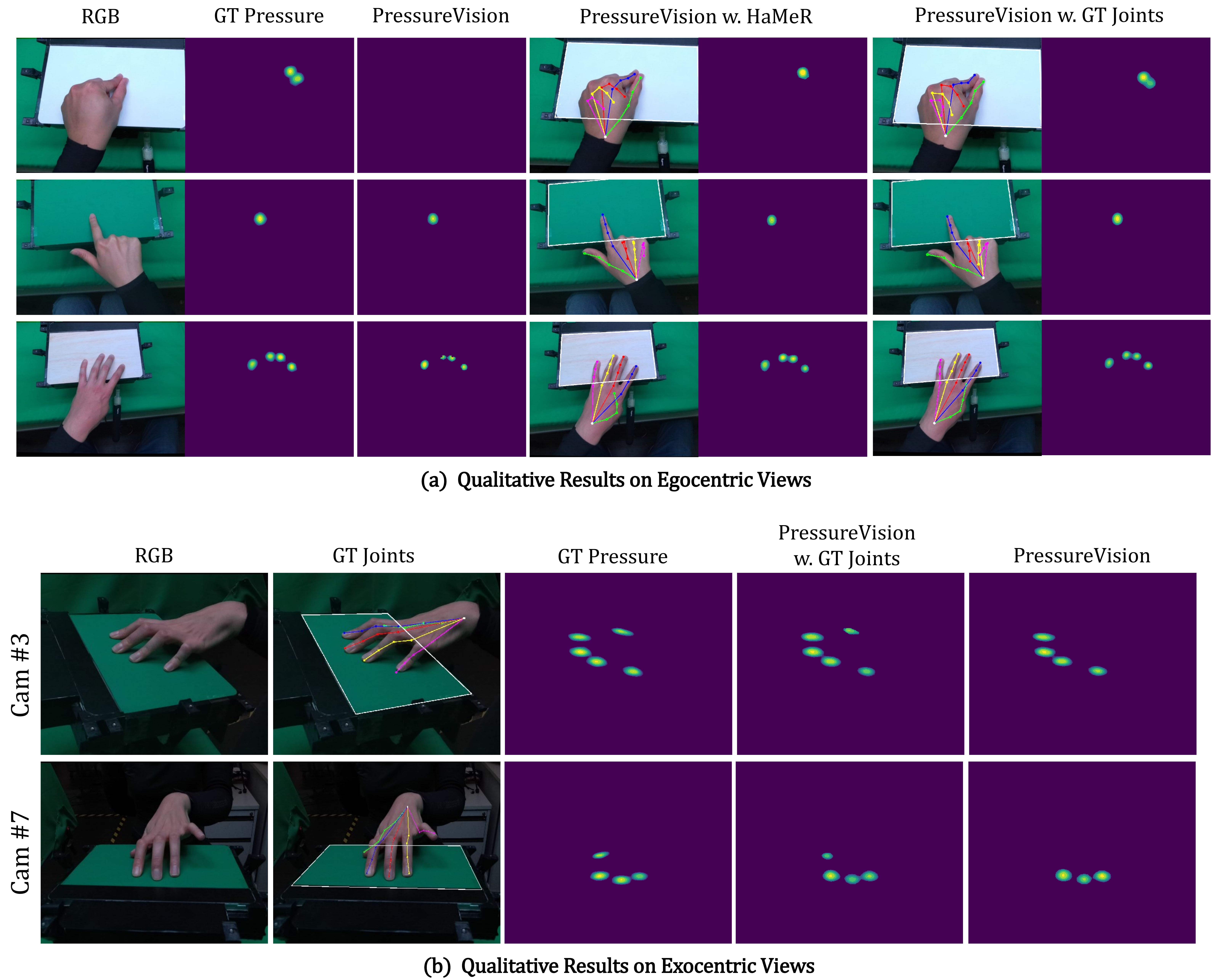}
   \caption{\textbf{Qualitative results}. 
  We present the egocentric experiment results in Subfigure (a). In Subfigure (b), both baseline models are trained using camera views 2, 3, 4, and 5. We display the results for one seen view and one unseen view. Additionally, we overlay the 2D keypoints predicted by HaMeR~\cite{hamer} and our annotated ground truth on the input image. For better visualization, the contour of the touch sensing area is also highlighted as a reference. 
   }
   \label{fig:qualitative_results}
   \vspace{-0.5em}
\end{figure}

The results are summarized in Table~\ref{table:baselines_results}.
We observe that the addition of the 2.5D hand joint layer improves performance for both egocentric and exocentric camera views.
Notably, the hand poses also enhance the model's generalization to unseen camera views. 
We trained the model on camera views \emph{2, 3, 4, and 5}, and evaluated it on views \emph{1, 6, and 7} (as shown in the third row of Table~\ref{table:baselines_results}). Qualitative results, presented in Figure~\ref{fig:qualitative_results}, further demonstrate the benefits of incorporating hand pose information.

\begin{table}[h]

\begin{center}
\caption{Pressure inference on the full dataset (21 participants) using different input modalities. Our high-fidelity hand pose annotations improve contact IoU [\%], volumetric IoU [\%], MAE [Pa], and temporal accuracy [\%] compared to using no hand poses or HaMeR~\cite{hamer} hand poses as additional input for novel exocentric and egocentric views. \label{table:baselines_results}
}
\vspace{-0.5em}
\begin{adjustbox}{width=1.0\columnwidth,center}
\begin{tabular}{l c c c c c c c}
\hline
Model           & Train          & Eval    & Modality          & Cont. IoU\,$\uparrow$\,& Vol. IoU\,$\uparrow$ &MAE\,$\downarrow$&Temp. Acc.\,$\uparrow$ \\
\hline\hline
PressureVisionNet~\cite{grady2022pressurevision}      & Ego  & Ego & RGB               & 55.73       & 38.64    & 53.60&91.68\\  
\cite{grady2022pressurevision} w. \cite{hamer} pose   & Ego  & Ego & RGB \& pred pose  & 56.25       & 40.52    & 55.23&91.67\\ 
\cite{grady2022pressurevision} w. GT pose              & Ego & Ego & RGB \& GT pose  &	58.80       &41.39     & 53.79& 92.17 \\

\hline
PressureVisionNet~\cite{grady2022pressurevision} & Exo (2,3,4,5)& Exo (2,3,4,5) & RGB               &62.11 & 44.73   &43.15&93.61\\  
\cite{grady2022pressurevision} w. \cite{hamer} pose    & Exo (2,3,4,5)  & Exo (2,3,4,5) & RGB \& pred pose  & 62,95     & 45,01  & 42,53& 93,83\\
\cite{grady2022pressurevision} w. GT pose               & Exo (2,3,4,5)& Exo  (2,3,4,5) & RGB \& GT pose              & 64.39       & 47.58   &41.72&94.18\\ 
\hline
PressureVisionNet~\cite{grady2022pressurevision} & Exo (2,3,4,5)& Exo (1,6,7) & RGB               &36.82 & 25.05   &62.22&83.40\\  
\cite{grady2022pressurevision} w. \cite{hamer} pose    & Exo (2,3,4,5)  & Exo (1,6,7) & RGB \& pred pose  & 38,46     & 28,10  & 51,50&86,34\\
\cite{grady2022pressurevision} w. GT pose               & Exo (2,3,4,5)& Exo  (1,6,7) & RGB \& GT pose              & 43.04       & 31.39   &49.45&89.78\\ 
\hline
\bottomrule
\end{tabular}
\end{adjustbox}
\end{center}

\vspace{-1em}

\end{table}

\subsection{First Hand-projected Pressure Baseline}
\label{sec:pressureformer}
\begin{figure*}[h]

\centering
\includegraphics[width=0.95\linewidth]{"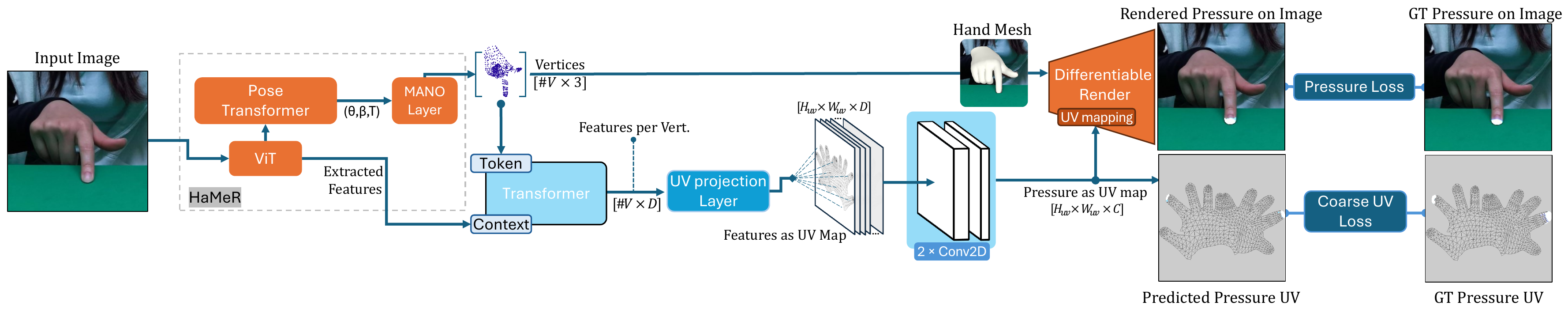"}
\vspace{-1em}
\caption{\textbf{PressureFormer} uses HaMeR's hand vertices and image feature tokens to estimate the pressure distribution over the UV map. We employ a differentiable renderer~\cite{dibr} to project the pressure back onto the image plane by texture-mapping it onto the predicted hand mesh.
}
\vspace{-1em}
\label{fig:pressure_former_pipeline}
\end{figure*}

Both the original PressureVision framework~\cite{grady2022pressurevision} and its subsequent iteration, PressureVision++~\cite{grady2024pressurevision++}, predict 2D hand pressure on the image plane.
However, this introduces ambiguity about the exact manifestation of this pressure between hands and objects within the 3D space.

To address this, we introduce a new baseline model, \emph{PressureFormer}, which estimates pressure as a UV map of the 3D hand mesh, enabling projection both as 3D pressure onto the hand surface and as 2D pressure onto the image space.

As illustrated in Figure~\ref{fig:pressure_former_pipeline}, our model builds upon HaMeR~\cite{hamer}. It processes the hand vertices \(V_{hand}\) in the camera frame and the image feature tokens from HaMeR's Vision Transformer (ViT)~\cite{dosovitskiy2020image}.
A transformer-based decoder receives \(V_{hand}\) as multiple input tokens while cross-attending to the image feature tokens from the ViT.
Each output token represents a \(D\)-dimensional feature for a corresponding mesh vertex, which we then map onto a UV feature map using the UV coordinates of the MANO model~\cite{MANO:SIGGRAPHASIA:2017}.
Given the sparsity of the UV feature map post-projection, we apply two convolutional layers for neural interpolation and reduce the dimensions to the number of force classes \(C\) to predict the quantized UV-pressure map \(U_{pred}\).

Initially, we compute the coarse UV-pressure loss \(\mathcal{L}_{c}\) between \(U_{pred}\) and the ground-truth UV-pressure map \(U_{gt}\), which is converted from the scalar UV pressure \(\textpres\) in our dataset.
Subsequently, we render the pressure \(P_{pred}\) back onto the original image plane based on the \(M_{hand}\) mesh of vertices \(V_{hand}\) and the predicted \(U_{pred}\) UV-pressure map. 
Using a differentiable renderer~\cite{dibr}, we invert the z-normal and z-axis of the face vertices to identify the mesh faces that are farthest from and invisible to the camera view, marking them as potential contact locations.
This allows us to compute the pressure loss \(\mathcal{L}_{p}\) with respect to the ground-truth pressure \(P_{gt}\).
We employ Cross-Entropy loss for both \(\mathcal{L}_{p}\) and \(\mathcal{L}_{c}\), resulting in the following loss function for PressureFormer:
\begin{equation}
     \mathcal{L_{PF}} = w_1 \mathcal{L}_{c} + w_2 \mathcal{L}_{p}
     \label{eq:loss_pressureformer}
\end{equation}

We trained the PressureFormer model and baseline models using hand-centered image crops across all camera views.
During training, we applied data augmentation techniques, including shifting, rescaling, and rotating.
The results are summarized in Table~\ref{table:pressureformer_results}, with visualizations provided in Figure~\ref{fig:pressureformer_validation}. Additional analyses are included in the supplementary material.

\begin{table}[h]

\begin{center}
\caption{Our method achieves the highest performance in terms of contact IoU and performs comparably to other approaches on additional evaluation metrics. Notably, it offers significant advantages over the image-projected pressure baselines by directly predicting pressure on the UV map, enabling the reconstruction of 3D pressure through projection onto the estimated hand surface.}
\label{table:pressureformer_results}
\begin{adjustbox}{width=0.5\textwidth,center}
\begin{tabular}{l c c c c }
\hline
Model            & Contact IoU\,$\uparrow$\,& Vol. IoU\,$\uparrow$ &MAE~[kPa]\,$\downarrow$&Temp. Acc.~[\%]\,$\uparrow$  \\
\hline\hline
PressureVisionNet~\cite{grady2022pressurevision} & 40.71  & 32.11  & 44& 90\\
\cite{grady2022pressurevision}  (w.\,HaMeR~\cite{hamer} pose)    & 42.52 & 35.40 &49& 92\\
PressureFormer & 43.04 & 31.57 &  71&  89\\
\hline
\bottomrule
\end{tabular}
\end{adjustbox}
\end{center}

\vspace{-1em}

\end{table}
\begin{figure}[t]
  \centering
  \fontsize{3.5}{5}\selectfont
  \setlength{\tabcolsep}{0.2pt}
  \newcommand{\psz}{0.125}
  \begin{tabular}{cccccccc}

      &                       &     & &     & \multicolumn{3}{c}{PressureFormer(Ours)}                        \\ 
      \cline{6-8}
  Input& Mesh & GT Pressure& PressureVision~\cite{grady2022pressurevision} & \cite{grady2022pressurevision} w. \cite{hamer}~pose  & Pred. Pressure  & UV Pressure & 3D Pressure
\\
  
  \includegraphics[width=\psz\linewidth]{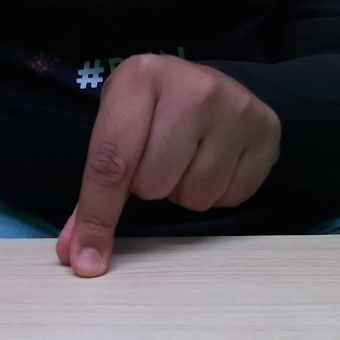} &
    \includegraphics[width=\psz\linewidth]{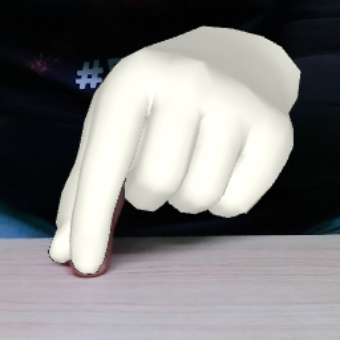} &
  \includegraphics[width=\psz\linewidth]{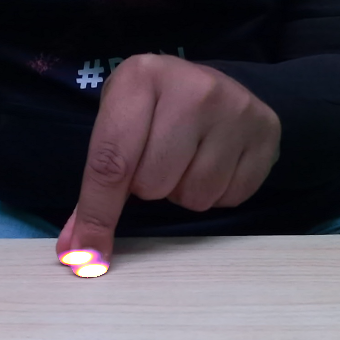} &
  \includegraphics[width=\psz\linewidth]{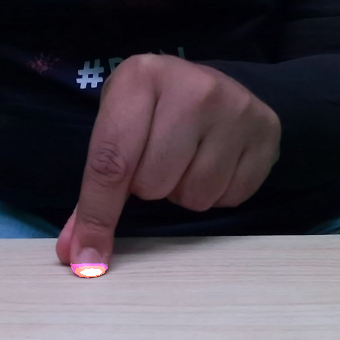} &
    \includegraphics[width=\psz\linewidth]{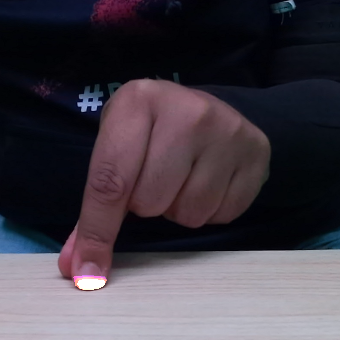} &

  \includegraphics[width=\psz\linewidth]{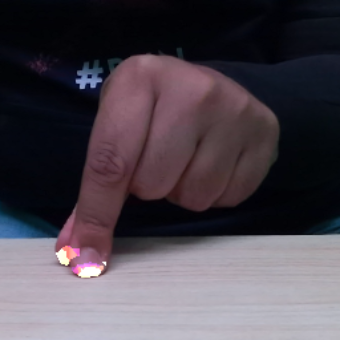} &
  \includegraphics[width=\psz\linewidth]{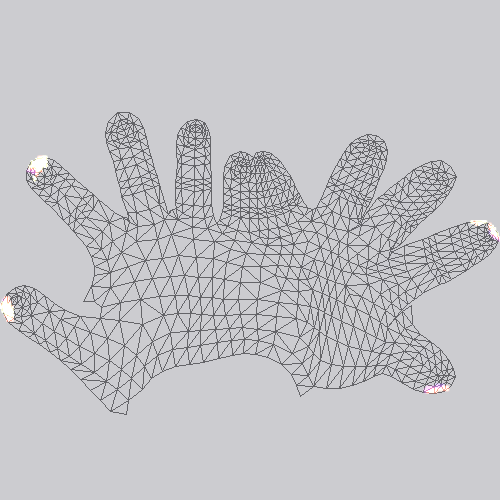} &
  \includegraphics[width=\psz\linewidth]{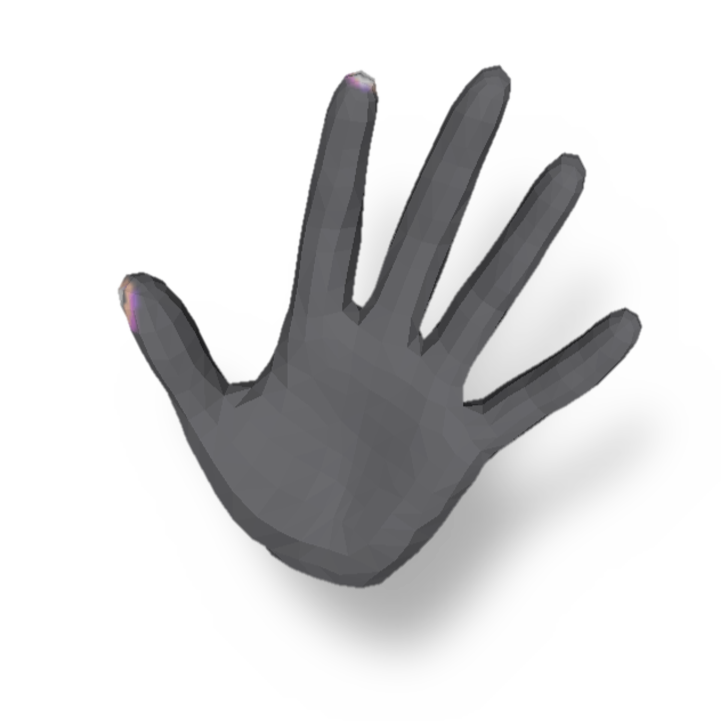} 
  \\
   \includegraphics[width=\psz\linewidth]{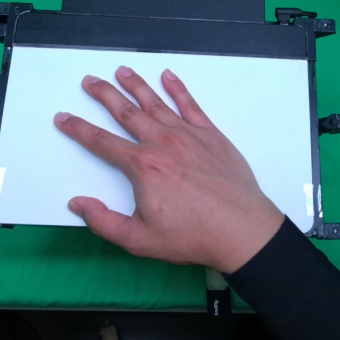} &
     \includegraphics[width=\psz\linewidth]{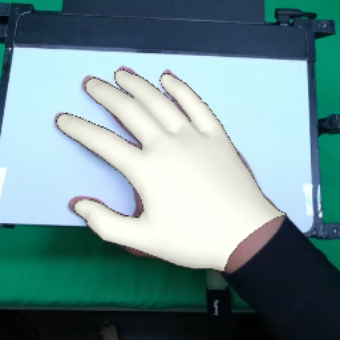} &
  \includegraphics[width=\psz\linewidth]{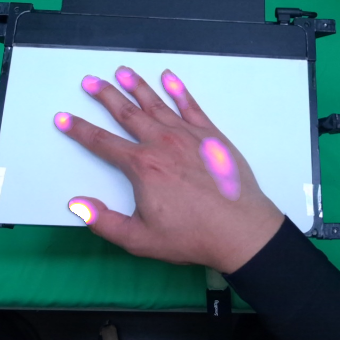} &
  \includegraphics[width=\psz\linewidth]{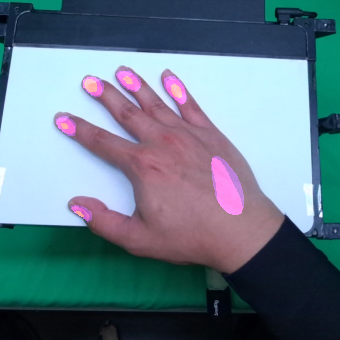} &
    \includegraphics[width=\psz\linewidth]{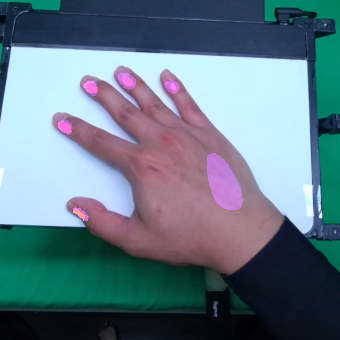} &

  \includegraphics[width=\psz\linewidth]{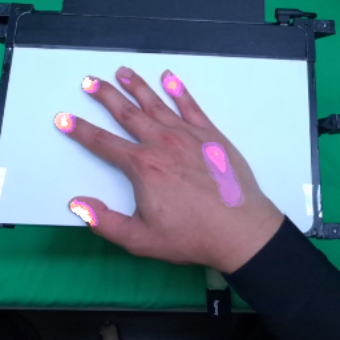} &
  \includegraphics[width=\psz\linewidth]{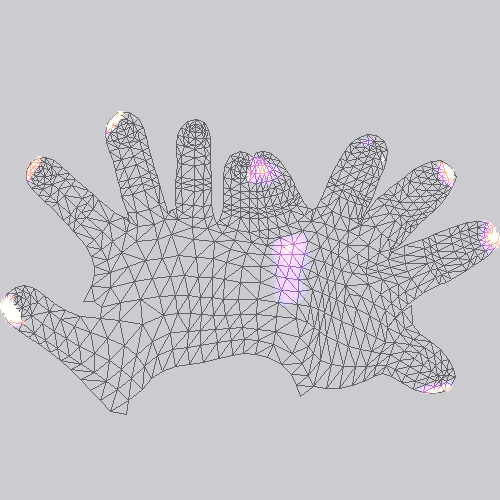} &
  \includegraphics[width=\psz\linewidth]{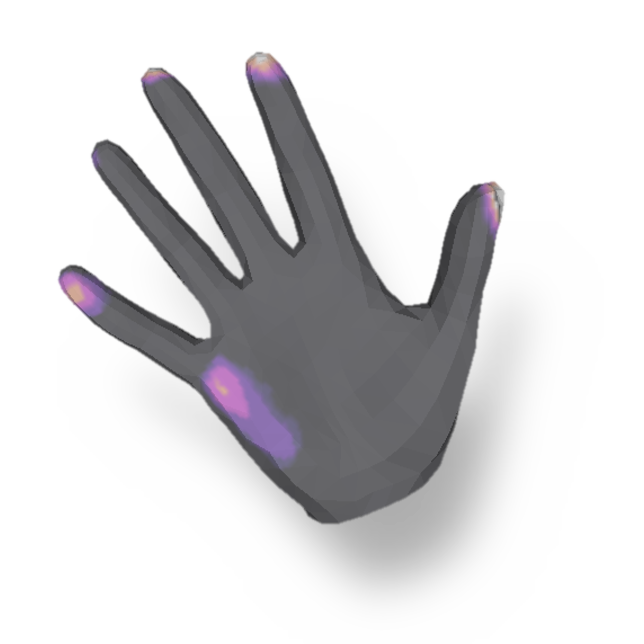} 
  \\   
  \includegraphics[width=\psz\linewidth]{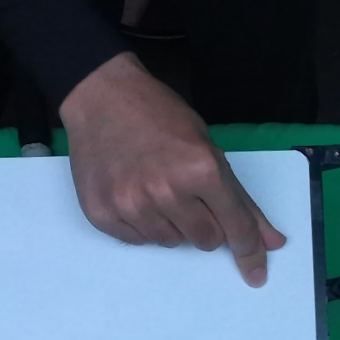} &
    \includegraphics[width=\psz\linewidth]{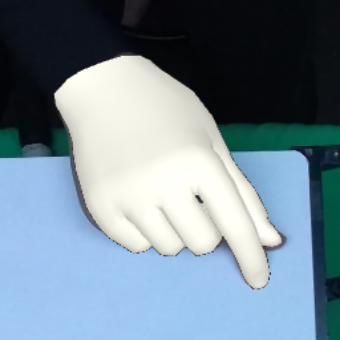} &
  \includegraphics[width=\psz\linewidth]{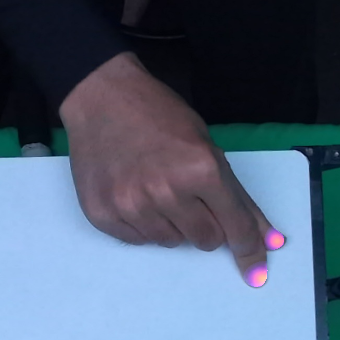} &
  \includegraphics[width=\psz\linewidth]{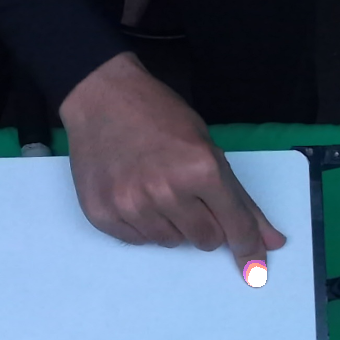} &
    \includegraphics[width=\psz\linewidth]{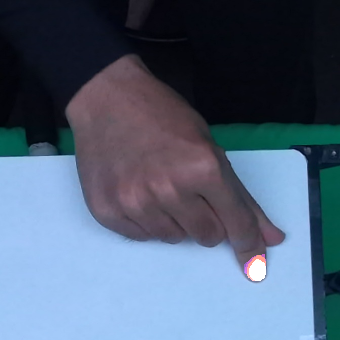} &

  \includegraphics[width=\psz\linewidth]{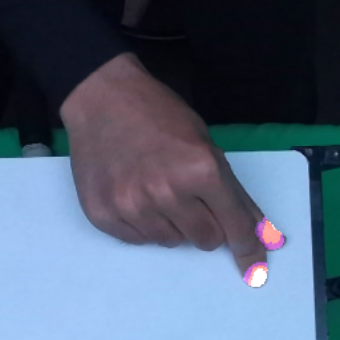} &
  \includegraphics[width=\psz\linewidth]{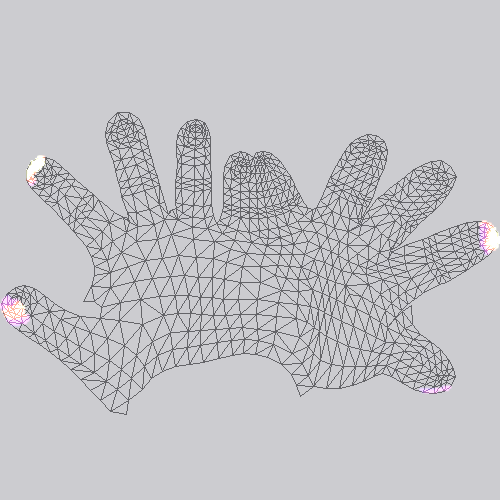} &
  \includegraphics[width=\psz\linewidth]{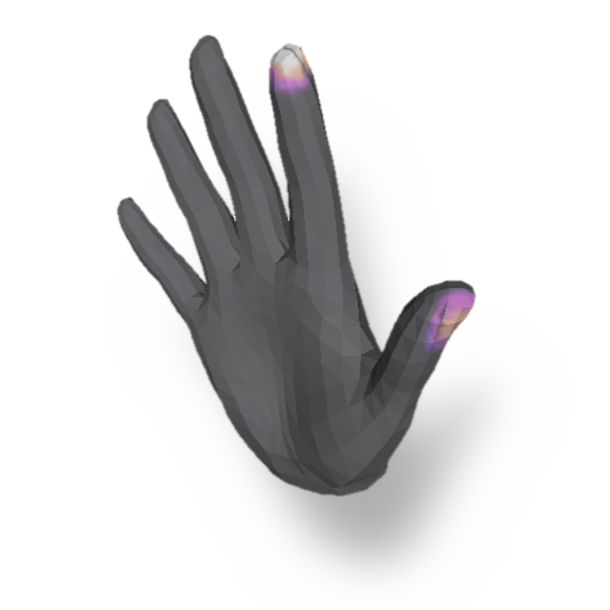} 
  \\
  \includegraphics[width=\psz\linewidth]{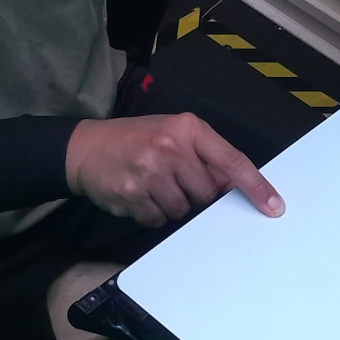} &
    \includegraphics[width=\psz\linewidth]{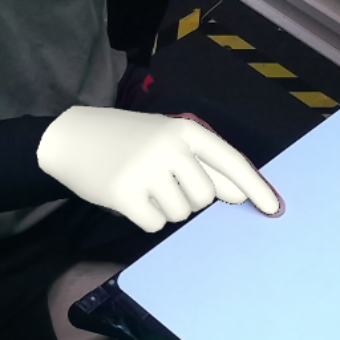} &
  \includegraphics[width=\psz\linewidth]{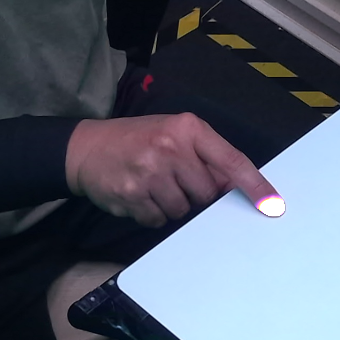} &
  \includegraphics[width=\psz\linewidth]{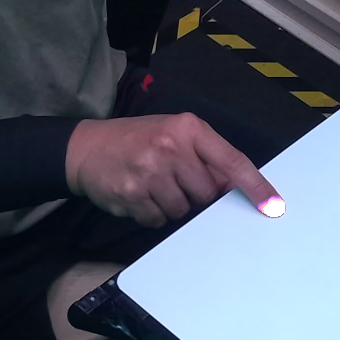} &
    \includegraphics[width=\psz\linewidth]{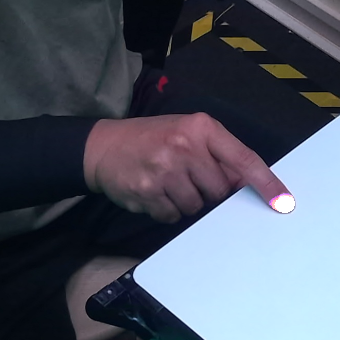} &

  \includegraphics[width=\psz\linewidth]{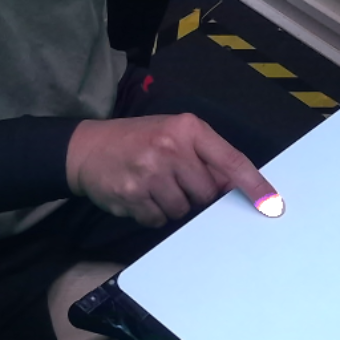} &
  \includegraphics[width=\psz\linewidth]{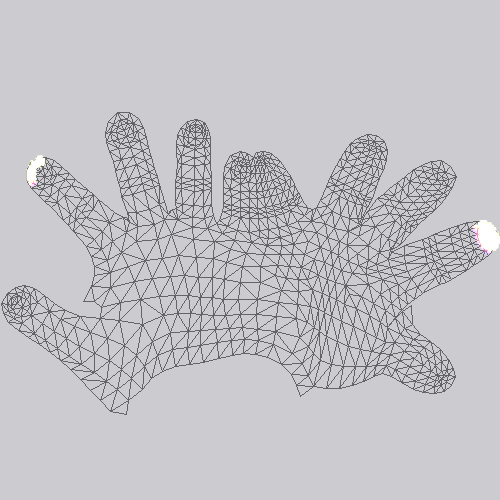} &
  \includegraphics[width=\psz\linewidth]{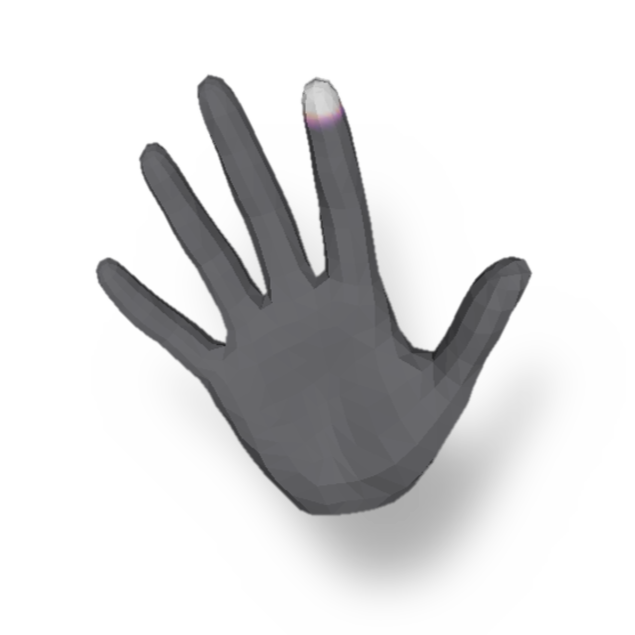} 
  \\
  \includegraphics[width=\psz\linewidth]{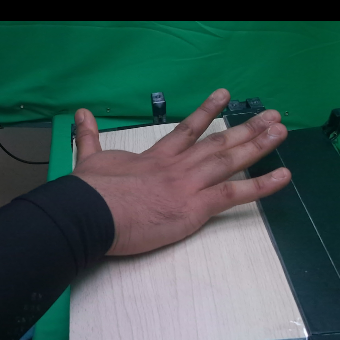} &
    \includegraphics[width=\psz\linewidth]{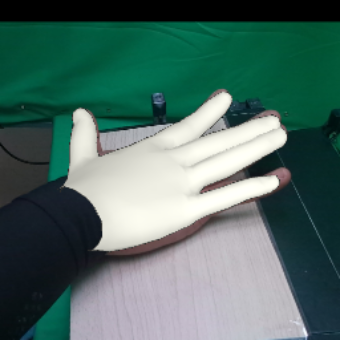} &
  \includegraphics[width=\psz\linewidth]{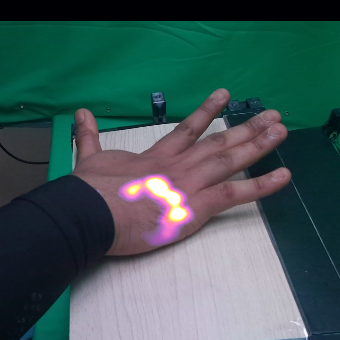} &
  \includegraphics[width=\psz\linewidth]{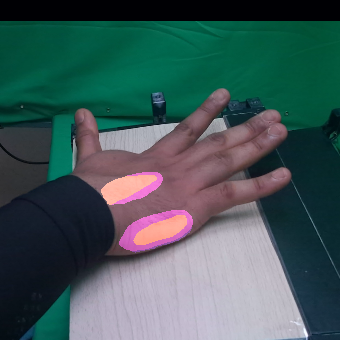} &
    \includegraphics[width=\psz\linewidth]{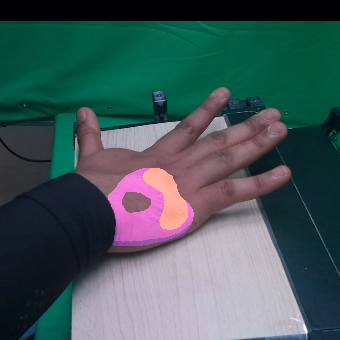} &

  \includegraphics[width=\psz\linewidth]{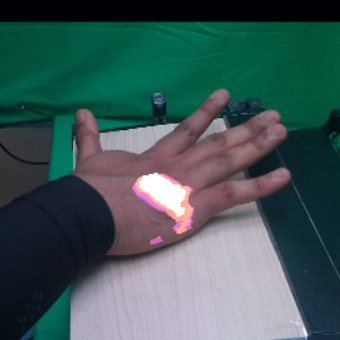} &
  \includegraphics[width=\psz\linewidth]{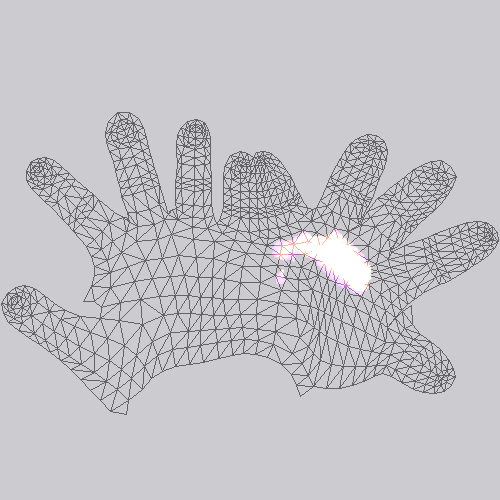} &
  \includegraphics[width=\psz\linewidth]{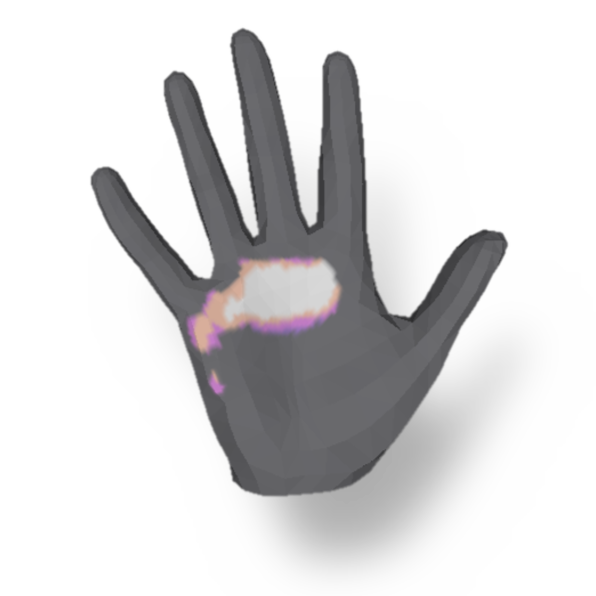} 
  \\
  \includegraphics[width=\psz\linewidth]{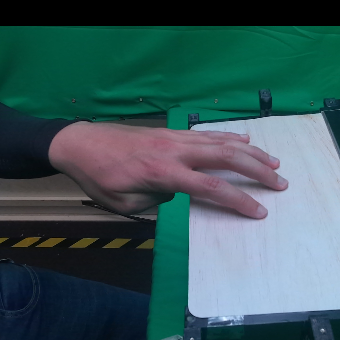} &
    \includegraphics[width=\psz\linewidth]{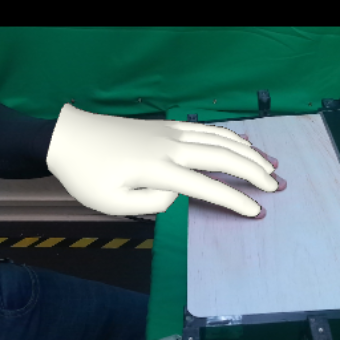} &
  \includegraphics[width=\psz\linewidth]{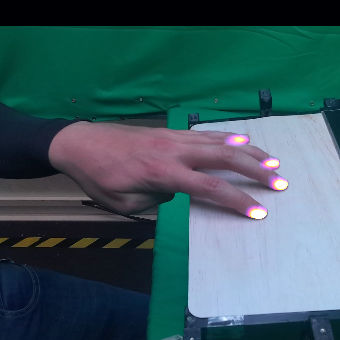} &
  \includegraphics[width=\psz\linewidth]{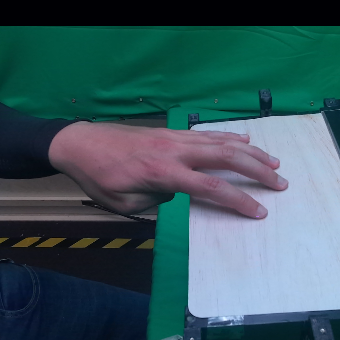} &
    \includegraphics[width=\psz\linewidth]{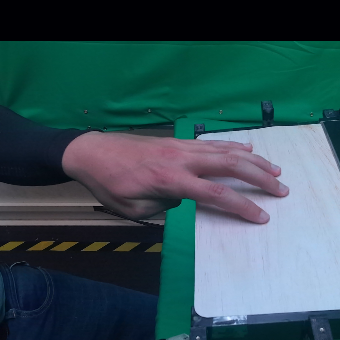} &

  \includegraphics[width=\psz\linewidth]{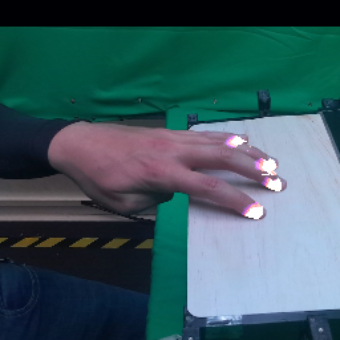} &
  \includegraphics[width=\psz\linewidth]{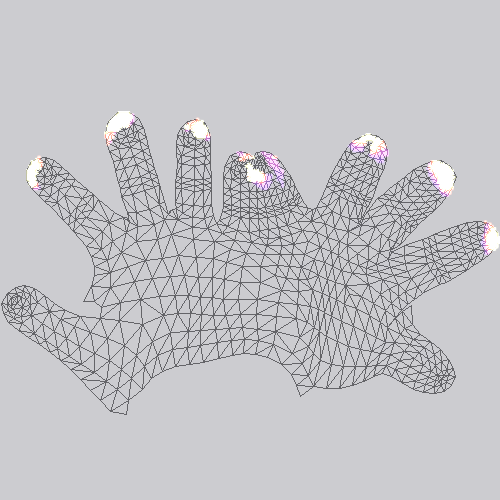} &
  \includegraphics[width=\psz\linewidth]{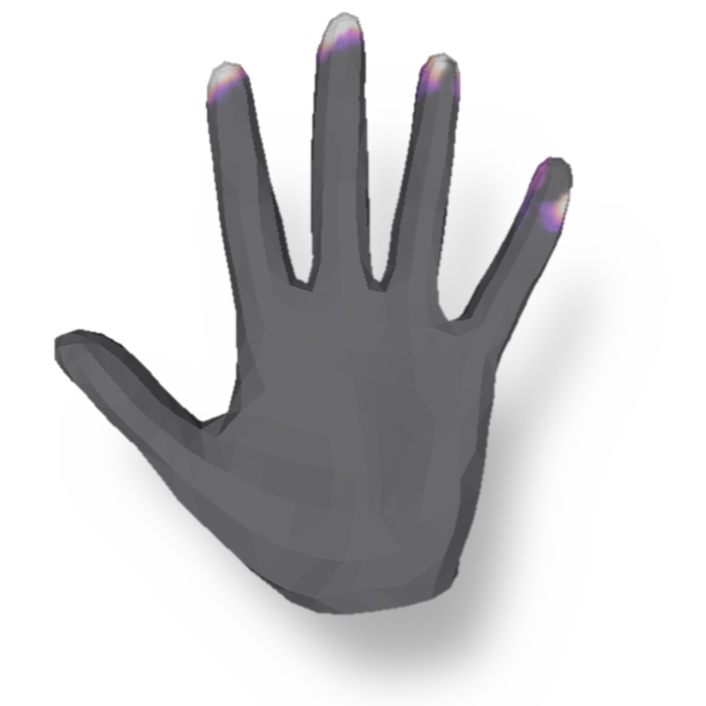} 
  \\
  \includegraphics[width=\psz\linewidth]{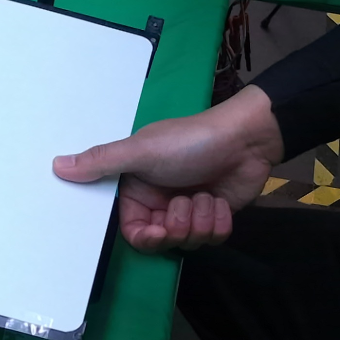} &
    \includegraphics[width=\psz\linewidth]{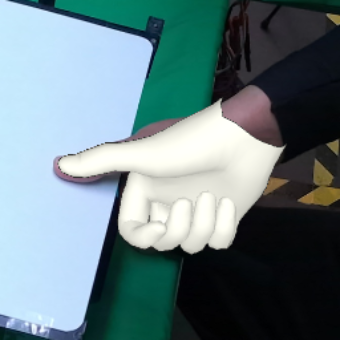} &
  \includegraphics[width=\psz\linewidth]{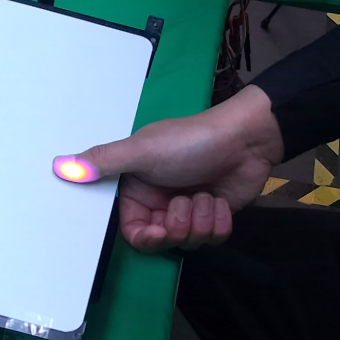} &
  \includegraphics[width=\psz\linewidth]{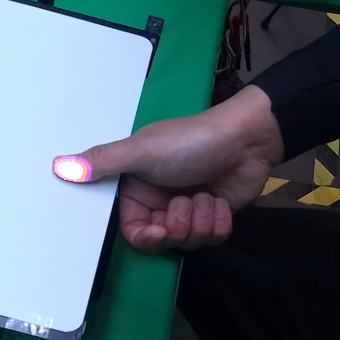} &
    \includegraphics[width=\psz\linewidth]{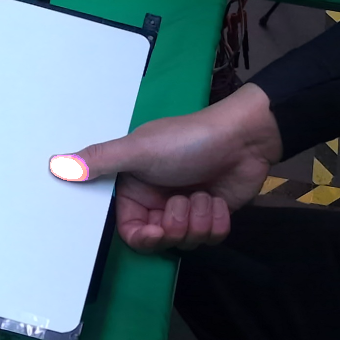} &

  \includegraphics[width=\psz\linewidth]{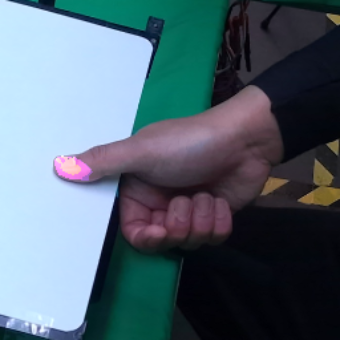} &
  \includegraphics[width=\psz\linewidth]{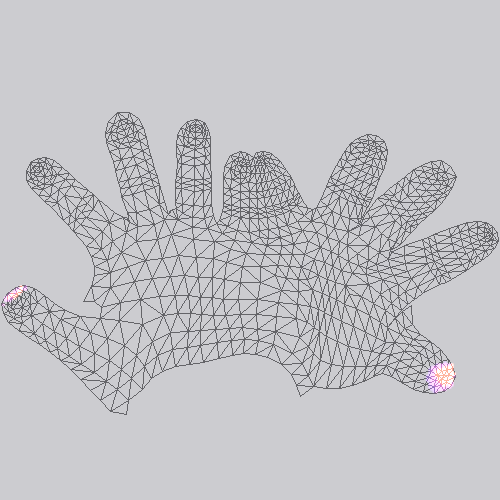} &
  \includegraphics[width=\psz\linewidth]{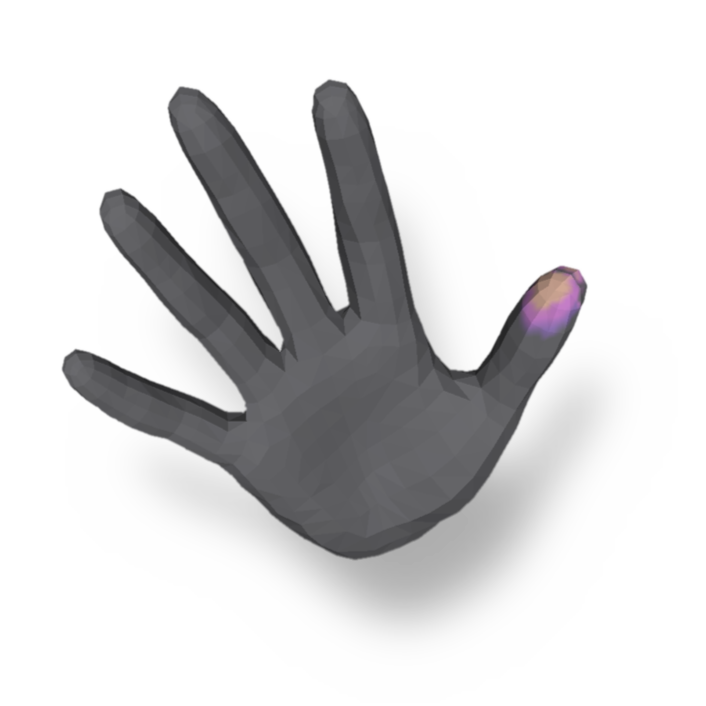} 
  \end{tabular}  
 \vspace{-2mm}
  \caption{\textbf{Qualitative Results PressureFormer on our dataset.} We compare our PressureFormer  with both PressureVision~\cite{grady2022pressurevision} and our extended baseline model with HaMeR-estimated~\cite{hamer} 2.5D joint positions. Additionally, we provide visualizations of the hand mesh estimated by HaMeR, alongside the 3D pressure distribution on the hand surface derived from our predicted UV-pressure in the last two columns. Note that we transform the left-hand UV maps into the right-hand format. }
    \label{fig:pressureformer_validation}
    \vspace{-1em}
\end{figure}

\section{Conclusion}
In this paper, we introduce EgoPressure, a novel egocentric hand pressure dataset paired with a multi-view hand pose estimation method. EgoPressure includes precise hand poses with meshes, multi-view RGB and depth images, egocentric view images, and high-quality pressure data. We establish a new benchmark and demonstrate the effectiveness of using hand pose data in pressure estimation. For future work, we plan to enhance our dataset by including objects to enable pressure estimation on more complex geometries. In conclusion, we believe that EgoPressure represents a significant step towards a deeper machine understanding of hand-object interactions from egocentric views.

 \clearpage
\setcounter{page}{1}
\maketitlesupplementary

\section{Details about Benchmark Evaluation}
In this section, we provide further details about the benchmark evaluation experiments from Section~5.

\subsection{Details for image-projected pressure baselines}

\subsubsection{Baseline model with Additional 2.5D Keypoints}
\begin{figure}[htbp]
\centering
\includegraphics[width=1.0\linewidth]{"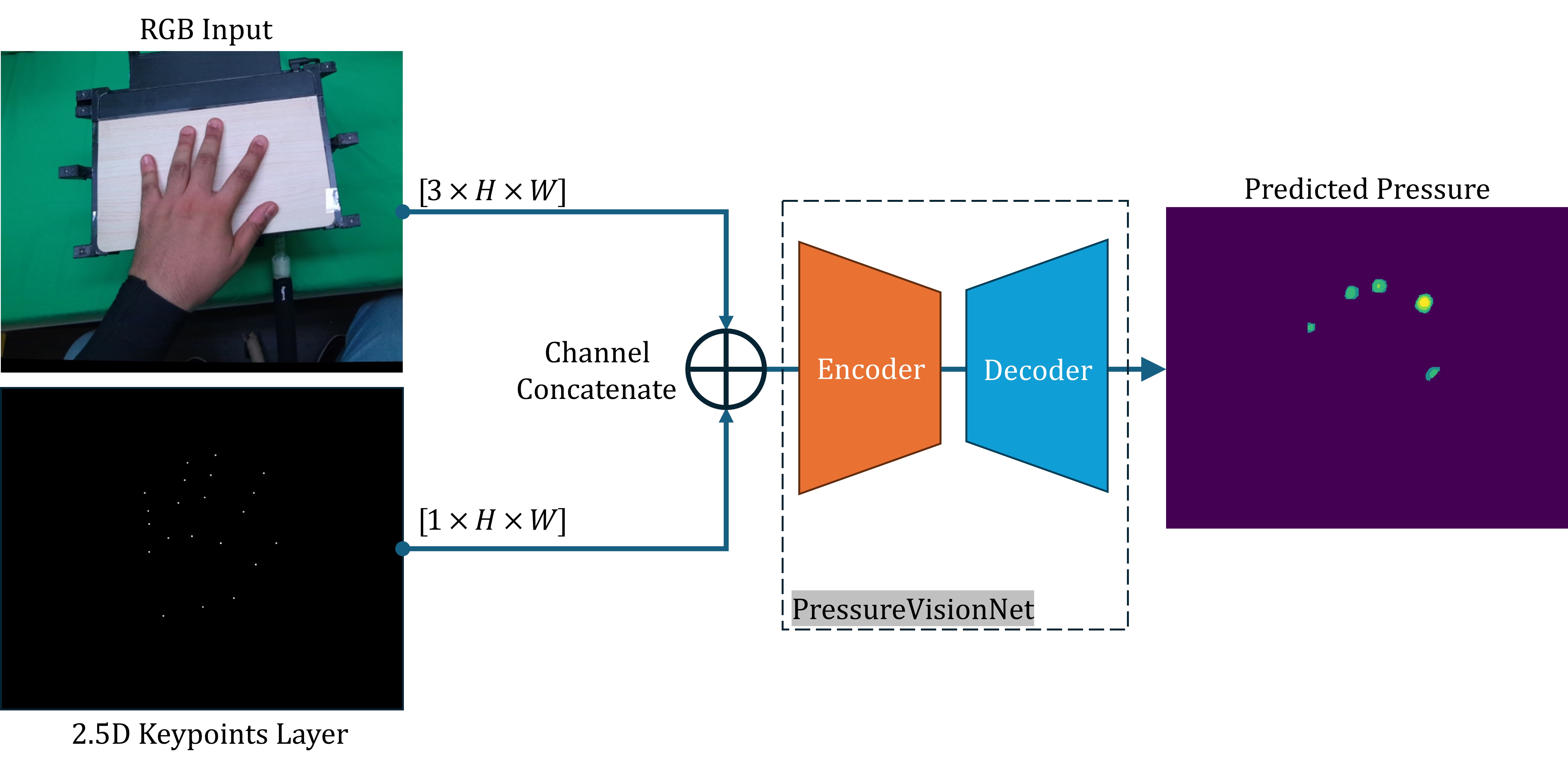"}
\vspace{-1em}
\caption{\textbf{Overview of the image-projected pressure baseline with additional hand pose input.} The baseline receives an RGB image and a 2.5D keypoint depth map as inputs to an encoder-decoder segmentation network for pressure estimation.}
\vspace{-1em}
\label{fig:baseline_keypoints_pipeline}
\end{figure}
The previous method~\cite{grady2022pressurevision} for predicting hand pressure relies solely on RGB images as inputs. In contrast, our new benchmark is designed to incorporate an additional modality, hand pose. To ensure a fair comparison between the baselines and our approach, we extend the existing method with additional hand pose inputs. In addition to the three RGB channels of PressureVision, we add a 2.5D depth map as an additional input channel to the segmentation network.

\paragraph{Encoder-decoder segmentation network architecture.} Similar to PressureVision, we employ an ImageNet-pretrained Squeeze-and-Excitation Network (SEResNeXt50)~\cite{hu2018squeeze, hu2019squeezeandexcitation} as the encoder, which takes both RGB and 3D hand pose inputs, and a feature pyramid network~\cite{lin2017feature, Iakubovskii:2019} as the decoder, which generates a pressure map.

\paragraph{Training.}
In all experiments, data from 15 participants is used for training and validation, while data from 6 participants is held out as the test set.
For training, we use the Adam optimizer with a batch size of 8.
The training process begins with a learning rate of 0.001 for 100k iterations, followed by 500k iterations with a learning rate of 0.0001.

\subsubsection{Evaluation Metrics}
For evaluation, we adopt the four metrics proposed in PressureVision~\cite{grady2022pressurevision}: Contact Intersection over Union (IoU), Volumetric IoU, Mean Absolute Error (MAE), and Temporal Accuracy.

Contact IoU measures the accuracy of contact surface predictions by calculating the IoU between the estimated and ground truth binarized pressure maps.
Volumetric IoU extends this by incorporating the accuracy of the predicted pressure magnitudes, calculated as the ratio of the sum of the minimum pressure values between the estimated and ground truth pressure maps at each pixel to the sum of the maximum values.
MAE quantifies the pressure prediction error in kilopascals (kPa) per pixel. Temporal Accuracy assesses the consistency of contact over time by verifying frame-by-frame contact consistency between the estimated and ground truth values.

\subsection{Additional Qualitative Results}

More qualitative results for the baselines are provided in Figure~\ref{fig:more_qualitative}. More qualitative examples for the annotations are shown in Figures~\ref{fig:ff1},~\ref{fig:ff2},~\ref{fig:ff3},~\ref{fig:ff4},~\ref{fig:ff5} and~\ref{fig:ff6}.

We also present qualitative results from the third-person view camera experiments (refer to Table~\ref{table:baselines_results} in the main paper). Figure~\ref{fig:cam2345val2345} and \ref{fig:cam2345val167} include visual comparisons between our model, which uses RGB and 2.5D hand keypoints, and PressureVisionNet~\cite{grady2022pressurevision} which uses only RGB input. Figure~\ref{fig:cam2345val2345} shows the models' qualitative performance on images from cameras \textit{2, 3, 4, and 5}, with both models trained on a separate training set from these views. In Figure~\ref{fig:cam2345val2345}, we evaluate the same models on novel views from cameras \textit{1, 6, and 7}, which were not included in the training set.

In the second column of Figure~\ref{fig:cam2345val2345} and Figure~\ref{fig:cam2345val167}, the reprojected touch sensing area is shown as a white outline to verify the camera pose. We also provide MAE and Contact IoU values for each sample. Notably, including additional hand pose information enhances the model's ability to estimate pressure and contact, especially for occluded hand parts (see examples \textit{04} in Figure~\ref{fig:cam2345val2345} and \textit{09, 11, 13} in Figure~\ref{fig:cam2345val167}).

\subsection{Additional Evaluation of PressureFormer}

PressureFormer improves upon the baselines from Section~\ref{sec:imageprojpressure} by estimating pressure directly on the UV map of the reconstructed hand mesh.
This approach extends the representation of pressure via the estimated 3D hand pose into 3D space.
While the hand mesh-based pressure representation can still be projected onto the image plane for benchmarking with prior methods~\cite{grady2022pressurevision, grady2024pressurevision++}, it offers additional insights about the specific hand regions applying pressure.
This capability is beneficial for scenarios involving complex hand-object interactions, such as when fingers are partially occluded or interacting with non-planar surfaces, where an image-projected pressure map may have limitations and introduce additional ambiguities.
These tactile hand dynamics are also helpful for enabling precise grasping and object manipulation in humanoid robotics.

\subsubsection{Accuracy of estimated UV Pressure Map}

In Section \ref{sec:pressureformer}, we compare PressureFormer with PressureVisionNet~\cite{grady2022pressurevision} and its 2.5D hand keypoint-augmented baseline, both of which directly estimate camera image-projected pressure maps. We make these comparisons based on the evaluation metrics established in PressureVision~(see Table~\ref{table:pressureformer_results_uv}).

\begin{figure}[t]
\centering
\includegraphics[width=1.0\linewidth]{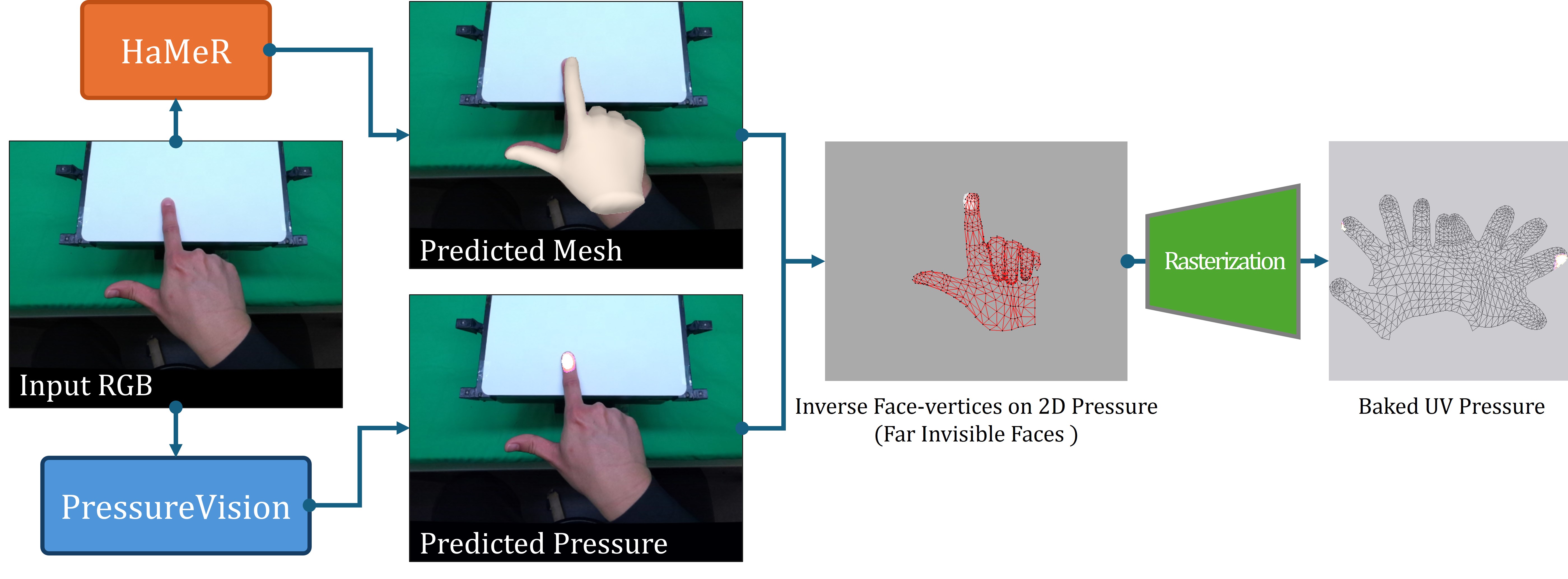}
\vspace{-2mm}
\caption{ \textbf{Pipeline for projecting the image-based pressure map (from PressureVision) onto the UV map:} Starting with the predicted hand mesh and 2D pressure map, the normals and z-axis are inverted to identify occluded (invisible) faces of the mesh. The pressure is then mapped onto the UV space using rasterization.
}

\label{fig:render_pressure_on_uv}
\end{figure}

We extend this evaluation by considering the hand mesh-projected pressure that PressureFormer directly estimates as a UV pressure map~(see Figure~\ref{fig:pressure_former_pipeline}).
For comparison, we project the image-based pressure maps from PressureVisionNet and its hand-pose-augmented baseline onto the corresponding hand mesh estimated from the same image using the HaMeR~\cite{hamer}.
This involves identifying the hand mesh faces furthest from the camera (i.e., occluded vertices) and rasterizing the 2D pressure map onto the UV map~(see Figure~\ref{fig:render_pressure_on_uv}).
Additionally, we evaluate a variant of PressureFormer trained without explicit UV loss supervision.

We thus introduce a novel benchmarking task that evaluates the accuracy of pressure on the hand surface and the performance of jointly estimating pressure and hand mesh.

\paragraph{Evaluation Metrics.}
To assess the accuracy of pressure estimation across the hand surface, we compute two metrics on the UV pressure map: Contact IoU and Volumetric IoU.

\paragraph{Training.}
The models are trained and evaluated using images from all camera views.  
We use 15 participants for training and validation, with 6 participants held out as the test set.  
During preprocessing, the images are cropped with a margin around the hand and resized to match the network's input dimensions.  
For evaluation, we ensure the hand remains centrally positioned in the frame throughout the cropping process.  
Data augmentation, including shifts, rescaling, and rotations, is applied across all methods.  
Training employs the Adam optimizer with a batch size of 8, using a learning rate of 0.001 for 100k iterations and 0.0001 for the subsequent 500k iterations. The loss function for PressureFormer~(see Eq.~\ref{eq:loss_pressureformer}) uses weighting parameters \( w_1 = 0.2 \) and \( w_2 = 0.05 \).

\paragraph{Results.} The results from Section~\ref{sec:pressureformer} and the UV map-based evaluation are summarized in Table~\ref{table:pressureformer_results_uv}.
PressureFormer outperforms all image-projected pressure baselines in terms of Contact IoU and Volumetric IoU on the UV pressure map.
It also attains the highest Contact IoU on the image-projected pressure map.
The hand-pose-augmented baseline, which directly predicts pressure onto the camera image, achieves the best Volumetric IoU on the image-based pressure map.
These results highlight the value of incorporating hand pose information for pressure estimation and underscore the potential of further research into the joint estimation of hand pose and pressure for more coherent interaction modeling.

Additionally, the results underline the value of the coarse UV-pressure loss in enhancing the accuracy of the pressure predictions on the UV map~(see Figure~13).

Figure~\ref{fig:uv_rendering} provides a qualitative comparison of the UV pressure maps estimated by the three baseline methods.

\begin{table*}[h!]
\caption{\textbf{Performance comparison} of our PressureFormer model against image-projected pressure baselines, evaluated using temporal accuracy [\%], image-based pressure metrics (Image Contact IoU, Image Vol. IoU, Image MAE [kPa]), and UV map-based pressure metrics (UV Pressure IoU, UV Pressure Vol. IoU). PressureFormer demonstrates superior performance in UV pressure IoU and UV Pressure Vol. IoU, while also achieving higher scores in image-based Contact IoU. By directly predicting pressure on the UV map, PressureFormer offers advantages, enabling accurate 3D pressure reconstruction by projecting the results onto the estimated hand surface.}
\label{table:pressureformer_results_uv}
\centering
\begin{adjustbox}{width=\textwidth,center}
\begin{tabular}{lccccccc}
\toprule
\textbf{Model} & \textbf{Im. Contact IoU\,$\uparrow$} & \textbf{Im. Vol. IoU\,$\uparrow$} & \textbf{Im. MAE\,$\downarrow$} & \textbf{UV Press. Contact IoU\,$\uparrow$} & \textbf{UV Press. Vol. IoU\,$\uparrow$} & \textbf{Temp. Acc.\,$\uparrow$} \\
\midrule
PressureVisionNet~\cite{grady2022pressurevision} & 40.71 & 32.11 & \textbf{44} & 21.53 & 16.41 & 90 \\
\cite{grady2022pressurevision} (w.\,HaMeR~\cite{hamer} pose) & 42.52 & \textbf{35.40} & 49 & 24.10 & 17.36 & \textbf{92} \\
PressureFormer (Ours) & \textbf{43.04} & 31.57 & 71 & \textbf{33.12} & \textbf{24.54} & 89 \\
PressureFormer w/o $\mathcal{L}_c$ & 41.27 & 29.57 & 74 & 26.24 & 18.61 & 88 \\
\bottomrule
\end{tabular}
\end{adjustbox}
\end{table*}

\begin{figure}[t]
  \centering
  \tiny
  \label{fig:uv_coarse_qualitative_figure}
  \setlength{\tabcolsep}{0.1pt}
  \newcommand{\csz}{0.09}
  \begin{tabular}{ccccc|ccc|ccc}

   \includegraphics[width=\csz\linewidth]{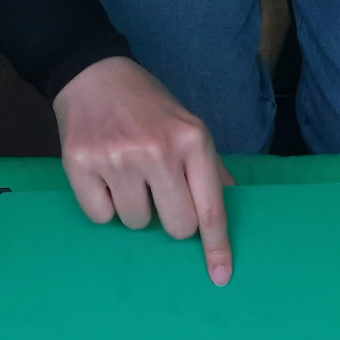} &
  \includegraphics[width=\csz\linewidth]{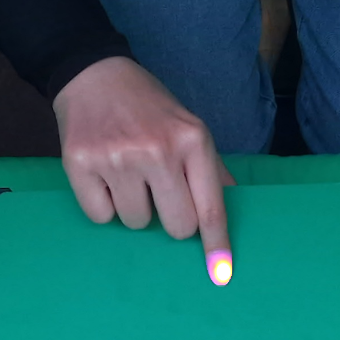} &
  \includegraphics[width=\csz\linewidth]{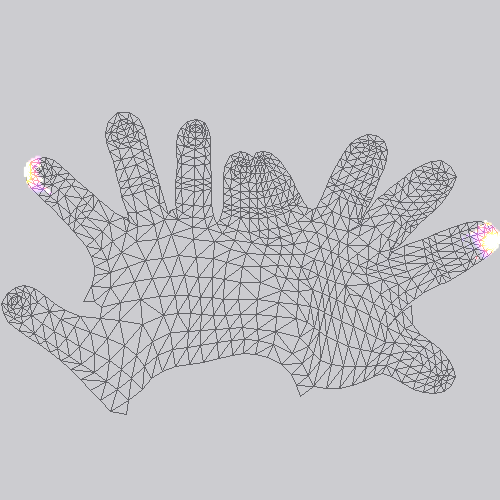} &
  \includegraphics[width=\csz\linewidth]{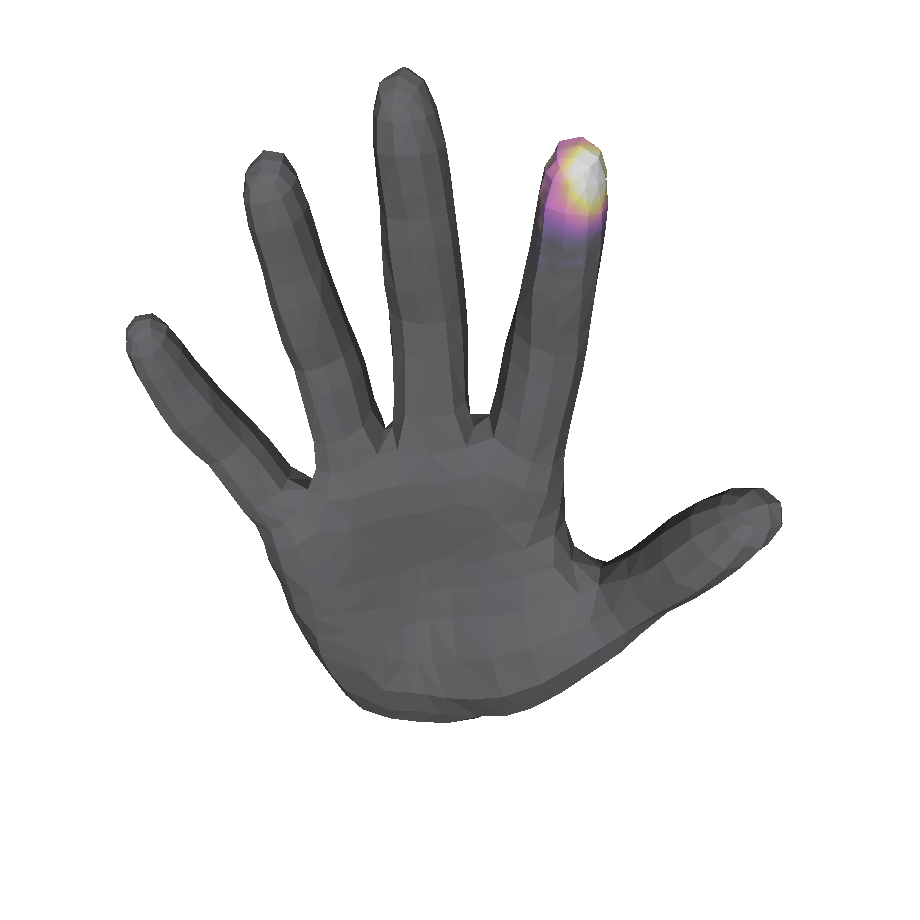} &
  \includegraphics[width=\csz\linewidth]{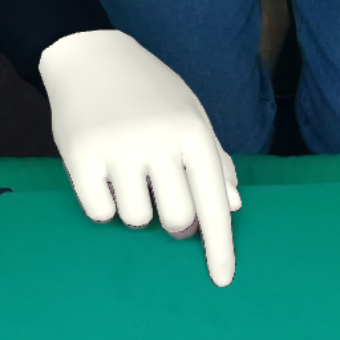} &
  
  \includegraphics[width=\csz\linewidth]{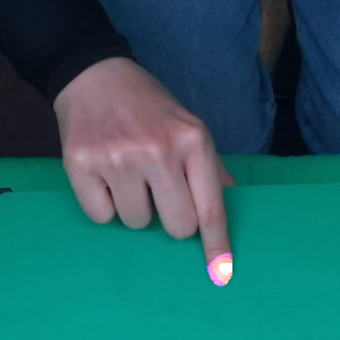} &
  \includegraphics[width=\csz\linewidth]{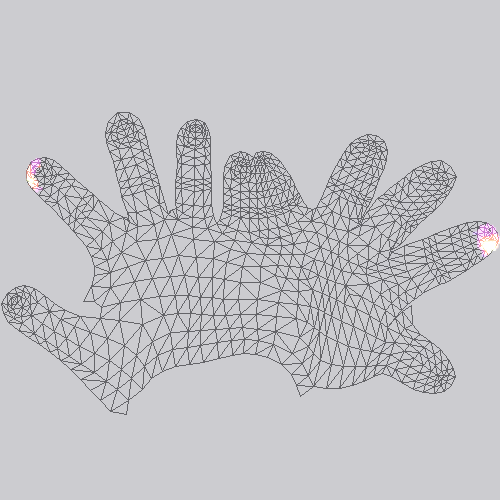} &
  \includegraphics[width=\csz\linewidth]{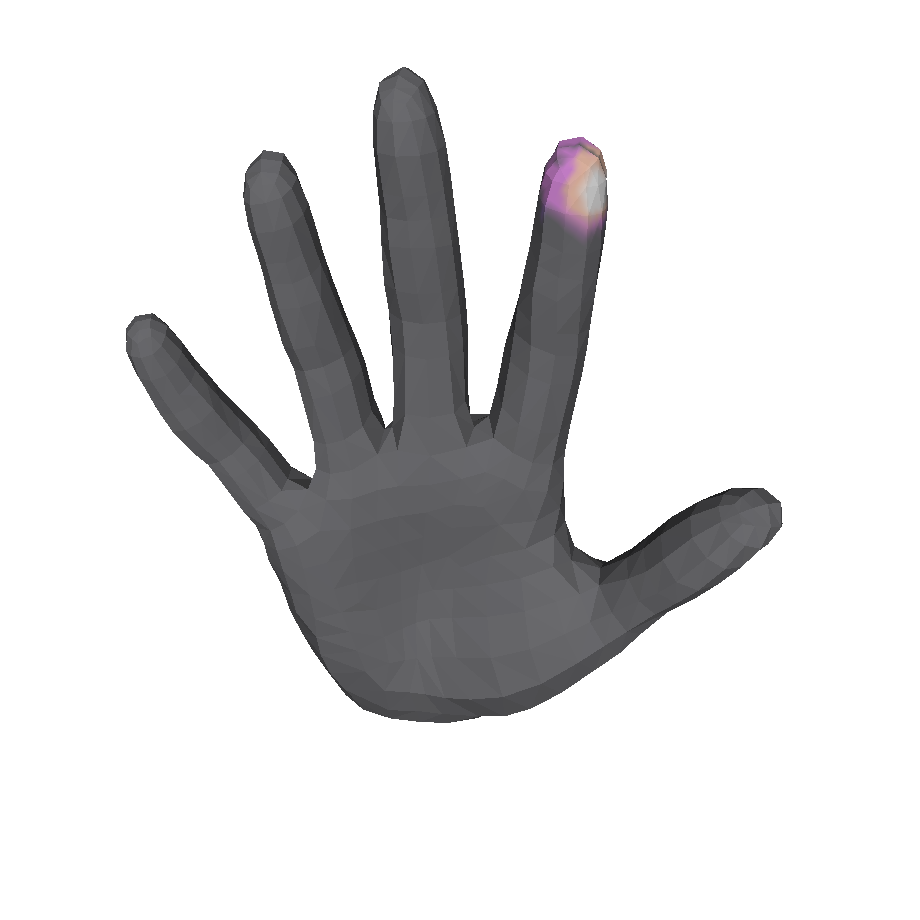} &

  \includegraphics[width=\csz\linewidth]{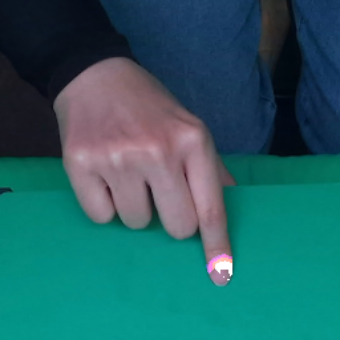} &
  \includegraphics[width=\csz\linewidth]{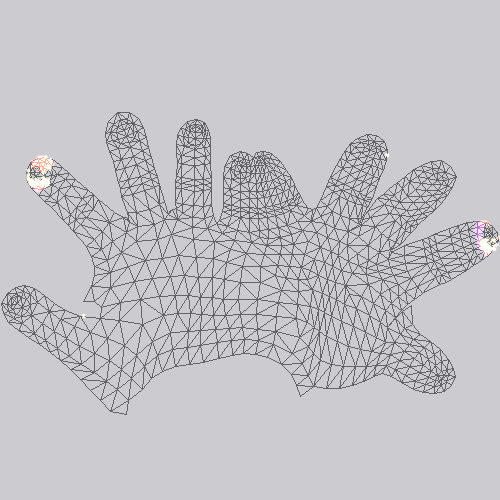} &
  \includegraphics[width=\csz\linewidth]{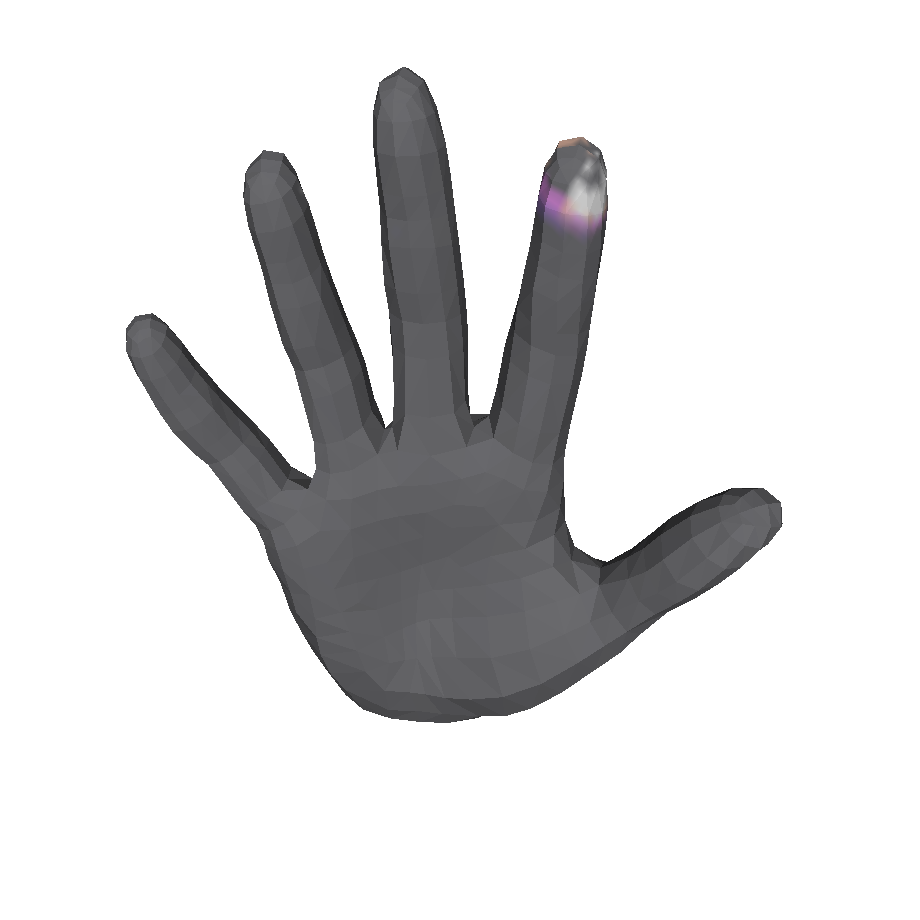} 

  \\

   \includegraphics[width=\csz\linewidth]{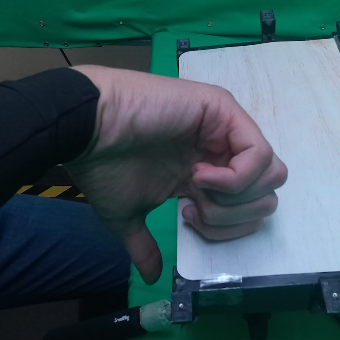} &
  \includegraphics[width=\csz\linewidth]{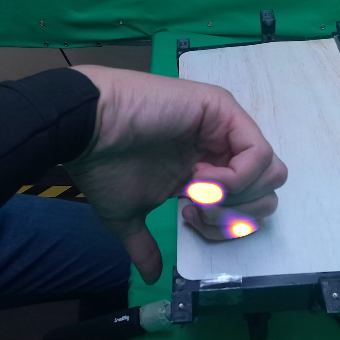} &
  \includegraphics[width=\csz\linewidth]{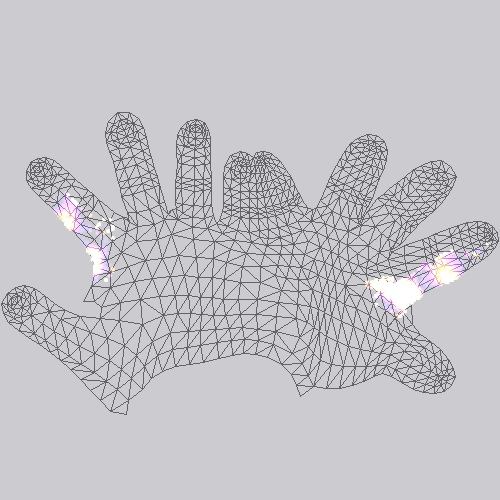} &
  \includegraphics[width=\csz\linewidth]{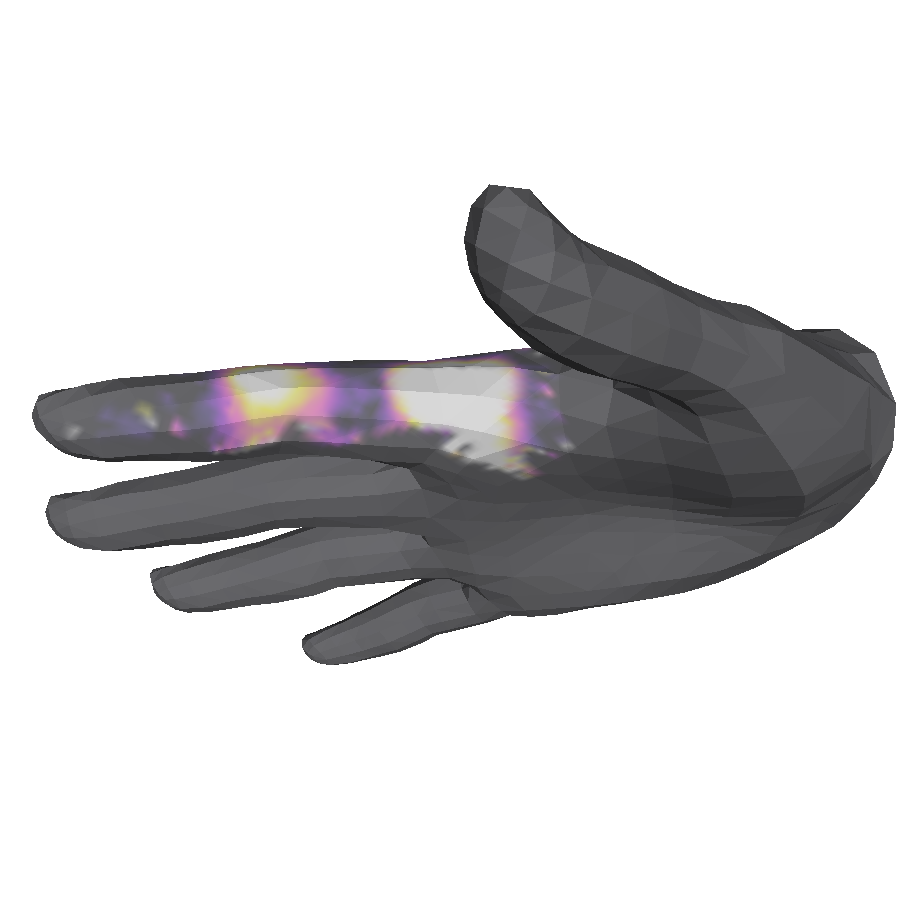} &
  \includegraphics[width=\csz\linewidth]{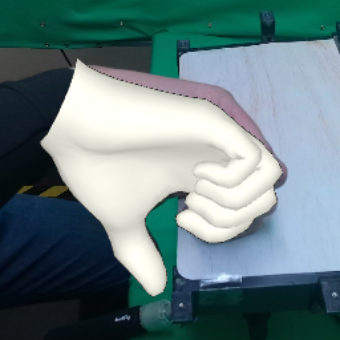} &
  
  \includegraphics[width=\csz\linewidth]{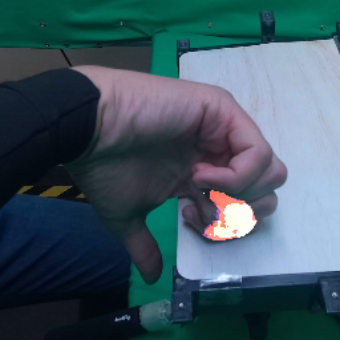} &
  \includegraphics[width=\csz\linewidth]{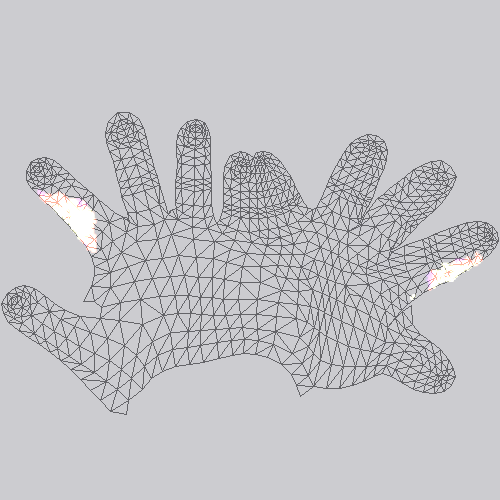} &
  \includegraphics[width=\csz\linewidth]{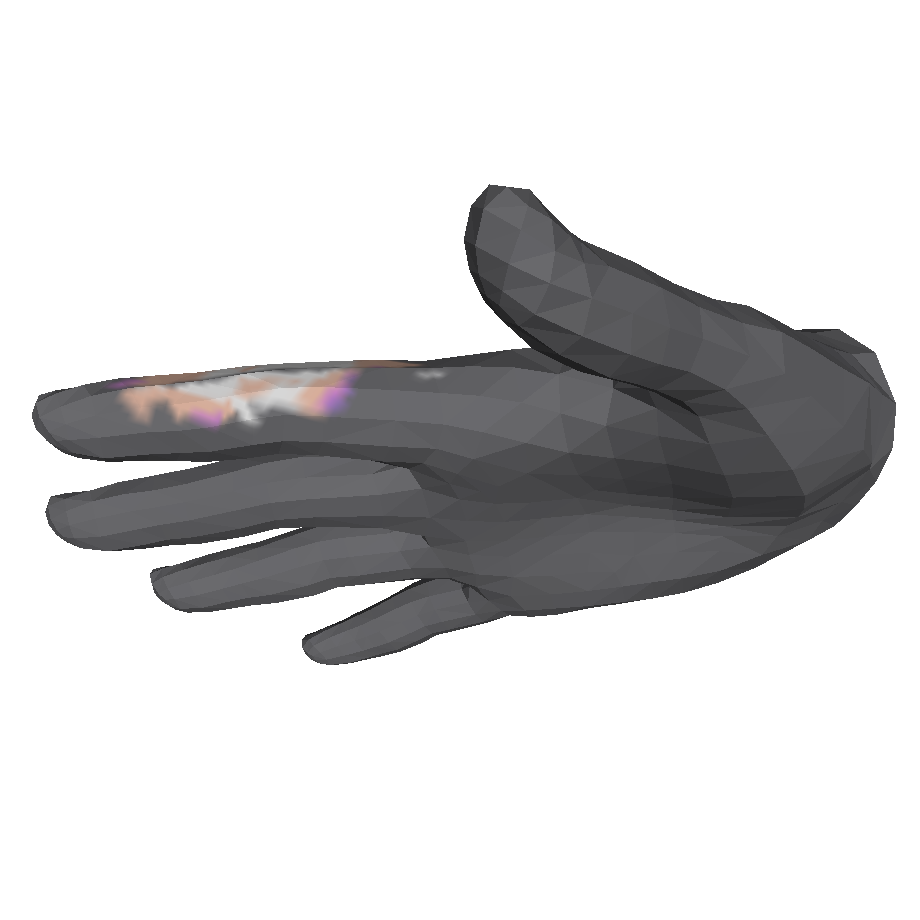} &

  \includegraphics[width=\csz\linewidth]{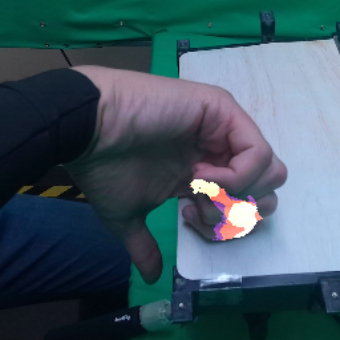} &
  \includegraphics[width=\csz\linewidth]{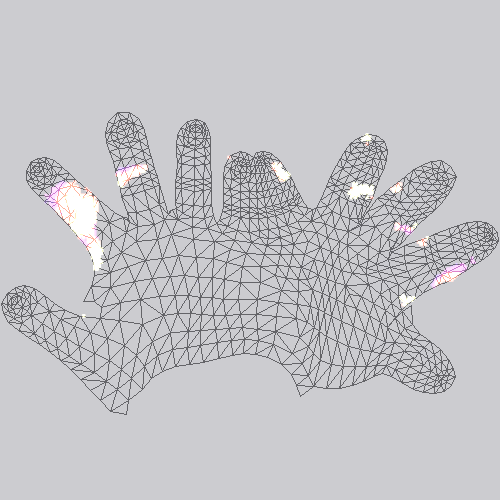} &
  \includegraphics[width=\csz\linewidth]{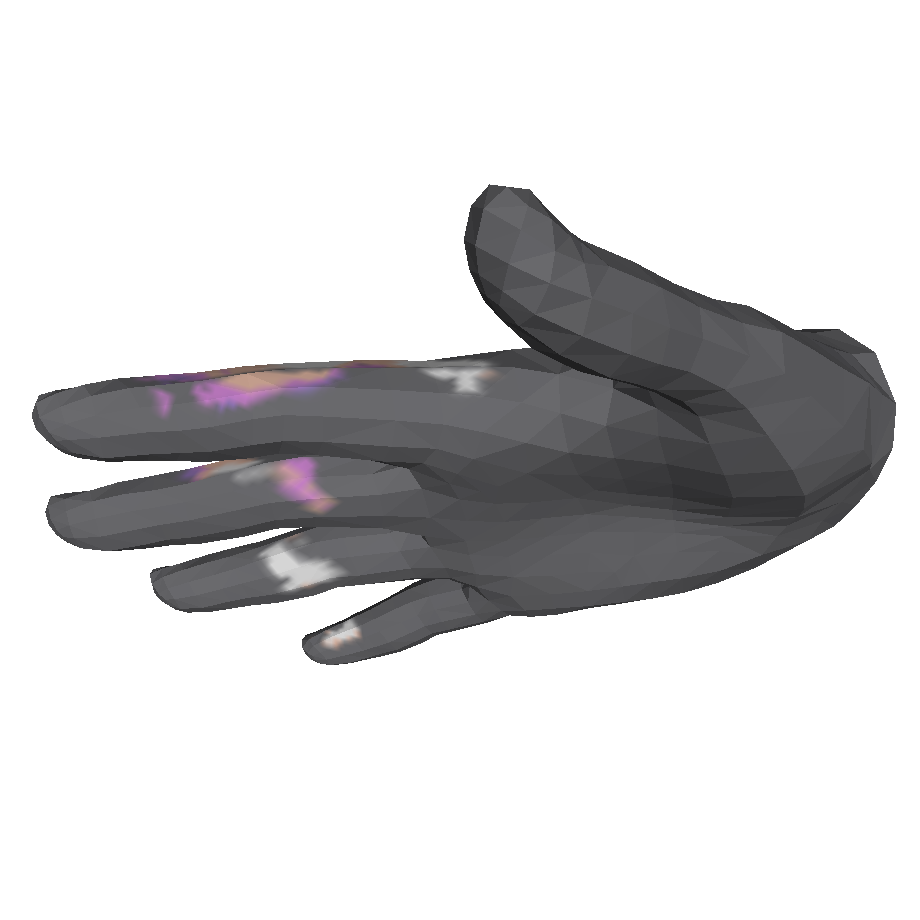} 

  \\
  
   \includegraphics[width=\csz\linewidth]{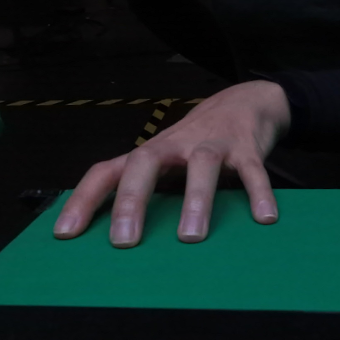} &
  \includegraphics[width=\csz\linewidth]{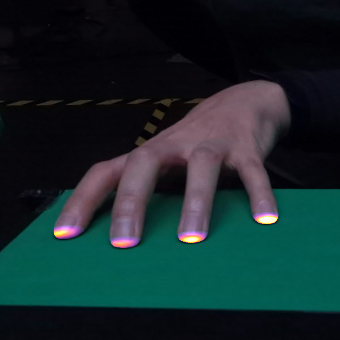} &
  \includegraphics[width=\csz\linewidth]{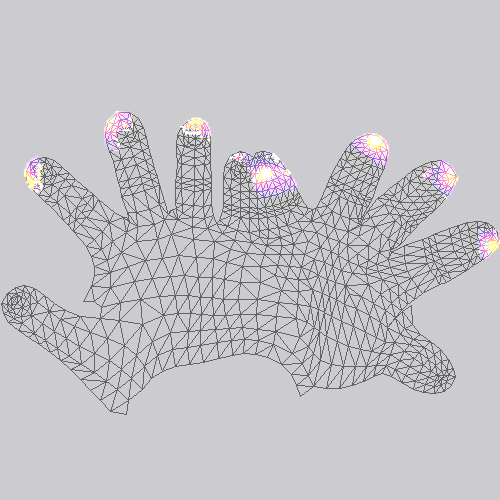} &
  \includegraphics[width=\csz\linewidth]{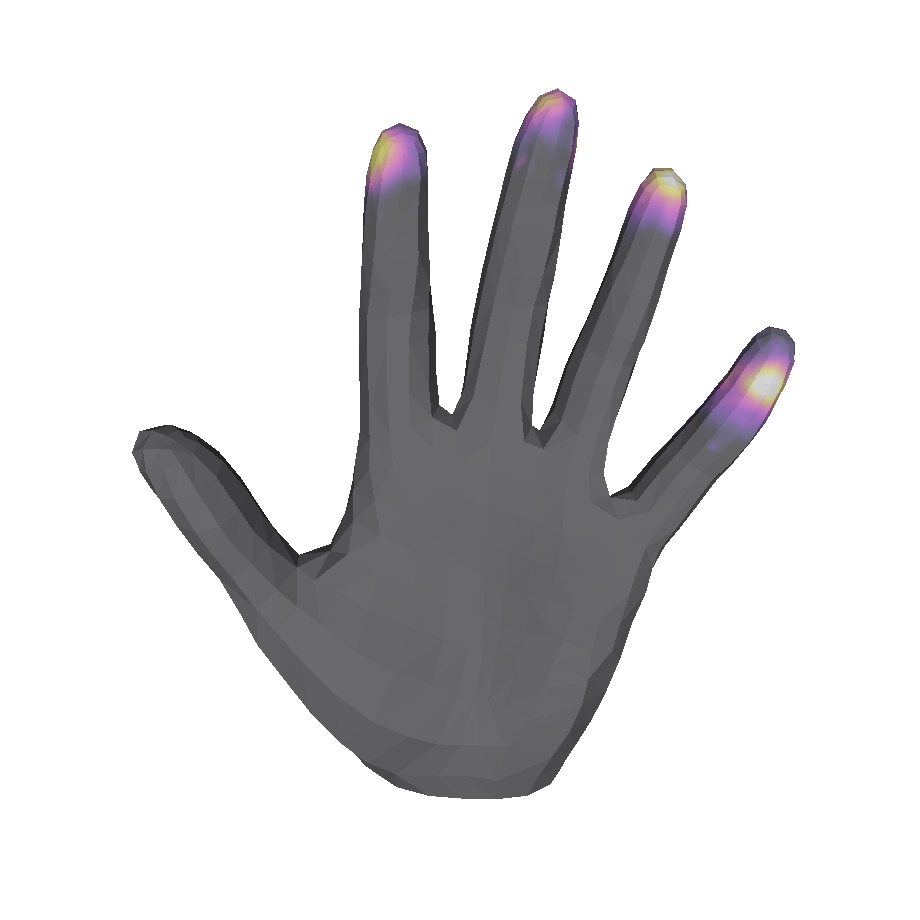} &
  \includegraphics[width=\csz\linewidth]{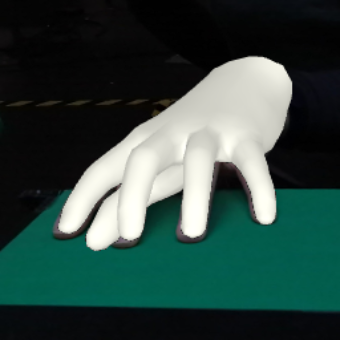} &
  
  \includegraphics[width=\csz\linewidth]{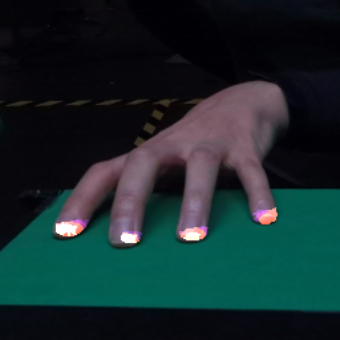} &
  \includegraphics[width=\csz\linewidth]{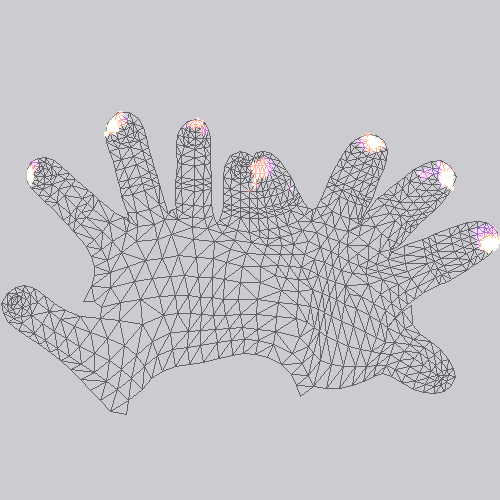} &
  \includegraphics[width=\csz\linewidth]{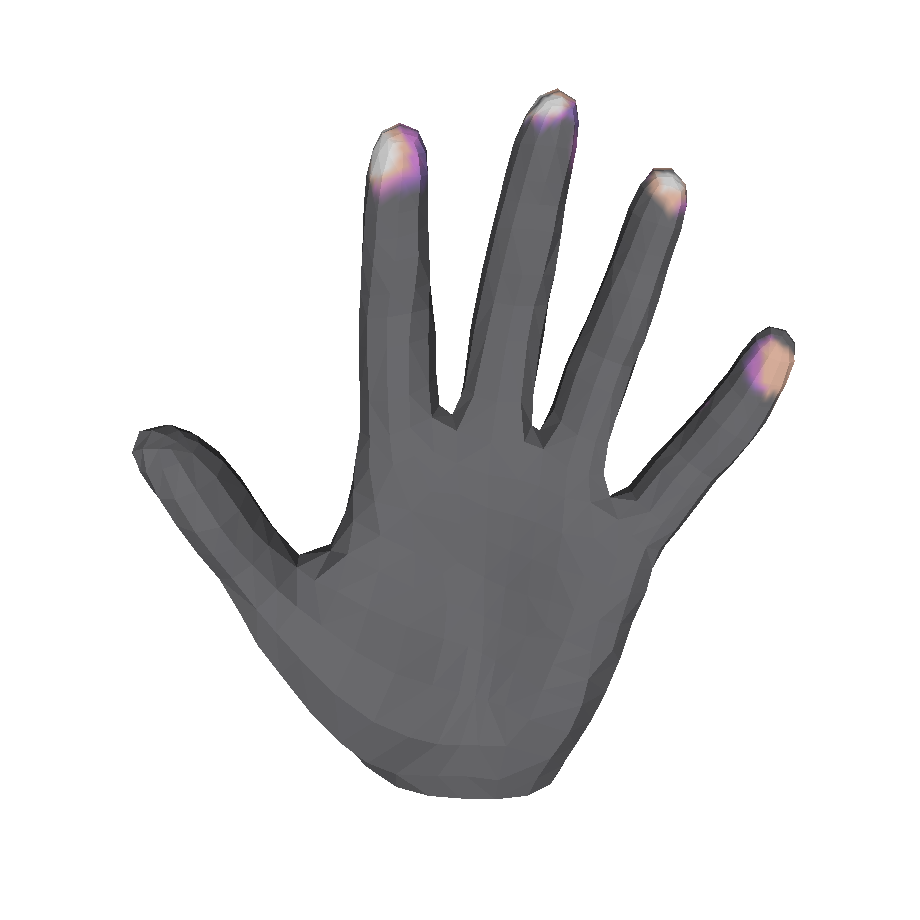} &

  \includegraphics[width=\csz\linewidth]{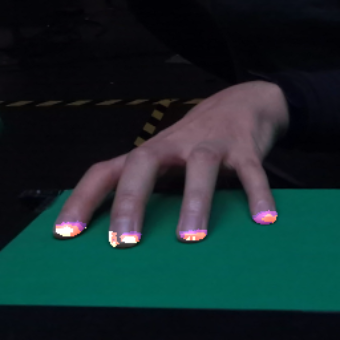} &
  \includegraphics[width=\csz\linewidth]{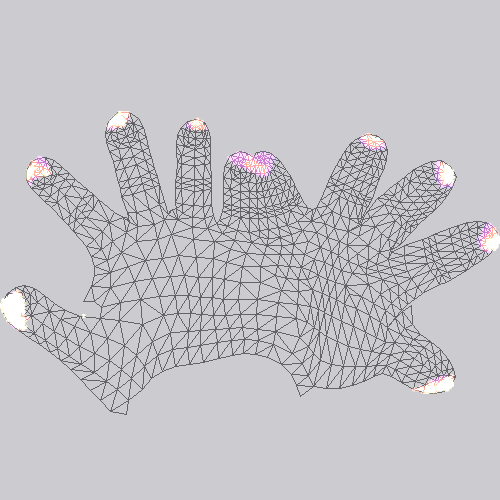} &
  \includegraphics[width=\csz\linewidth]{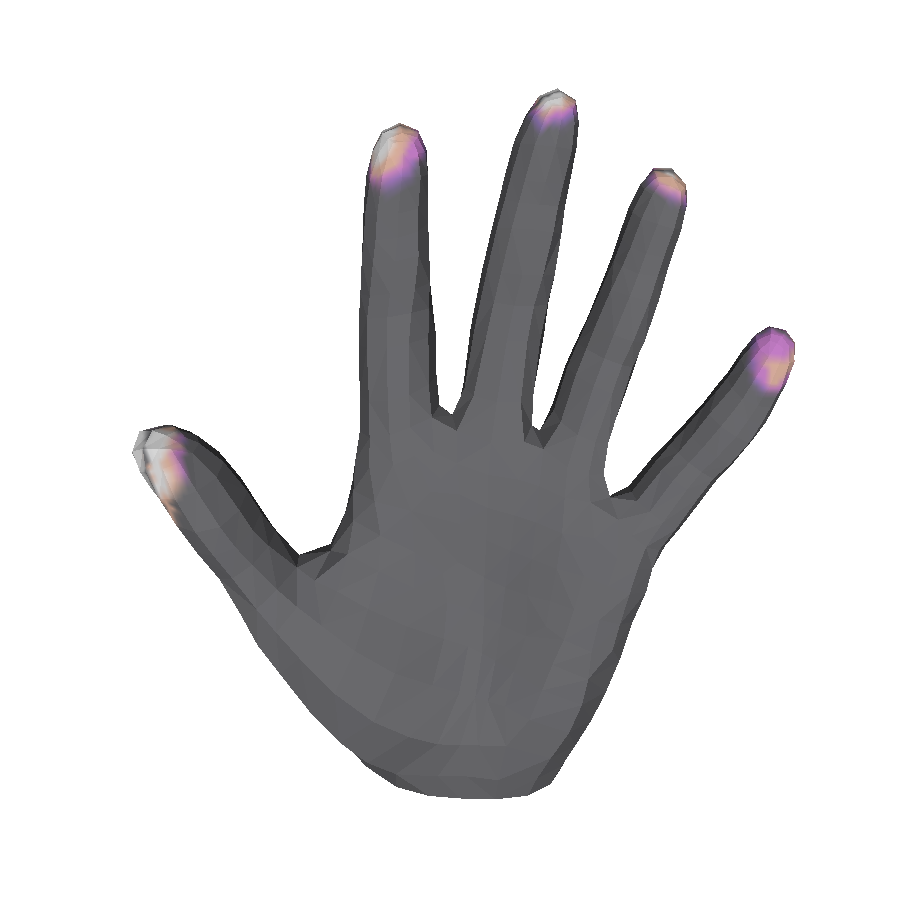} 

  \\

   \includegraphics[width=\csz\linewidth]{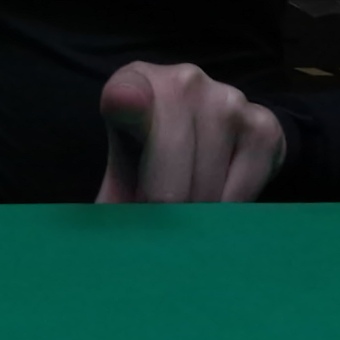} &
  \includegraphics[width=\csz\linewidth]{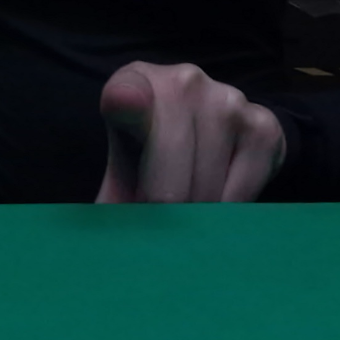} &
  \includegraphics[width=\csz\linewidth]{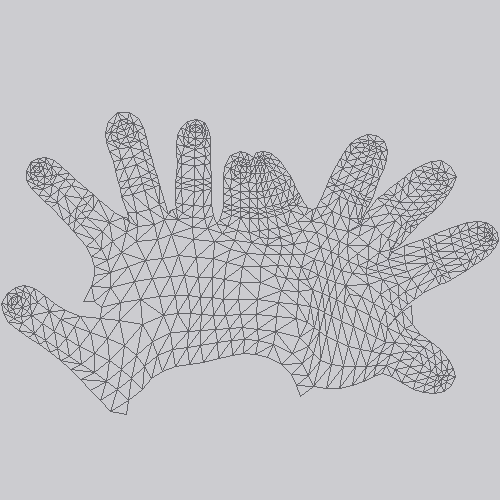} &
  \includegraphics[width=\csz\linewidth]{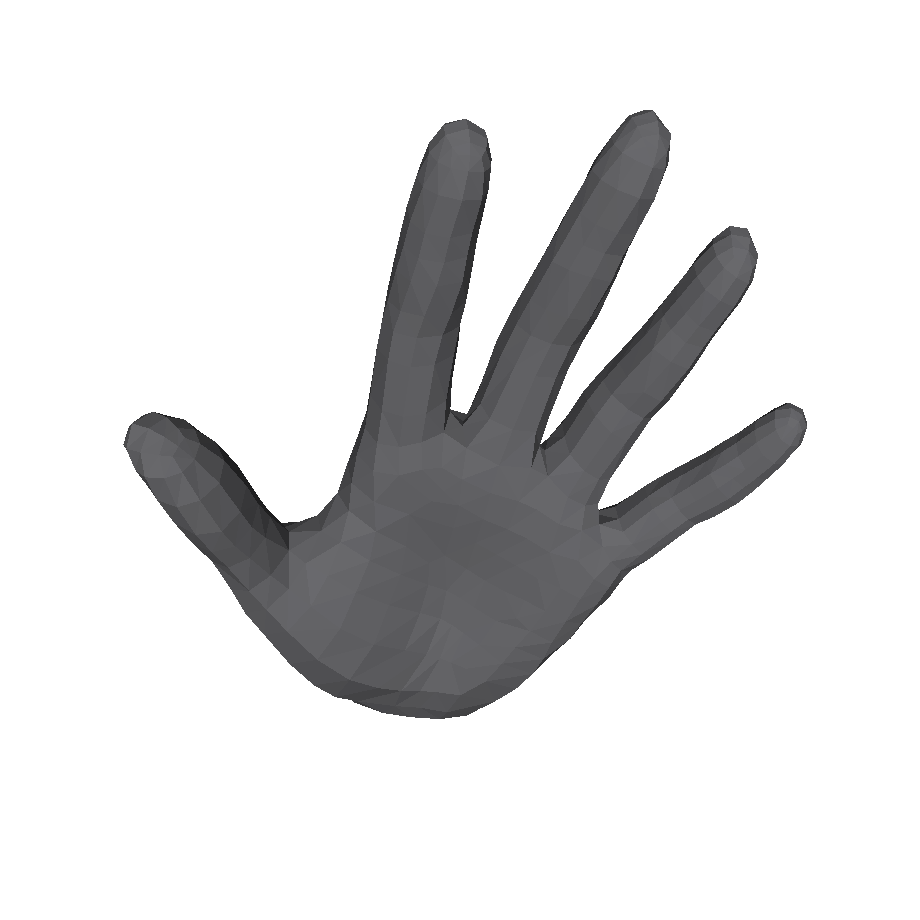} &
  \includegraphics[width=\csz\linewidth]{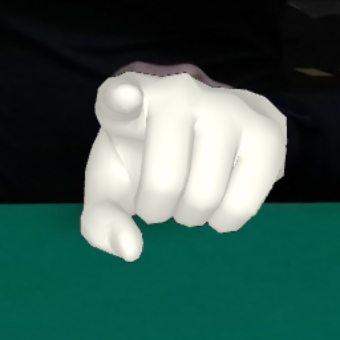} &
  
  \includegraphics[width=\csz\linewidth]{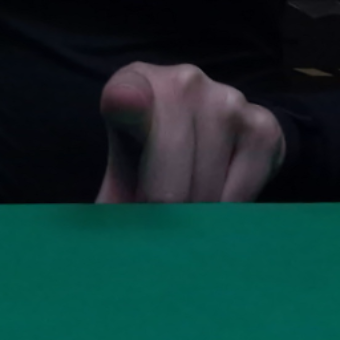} &
  \includegraphics[width=\csz\linewidth]{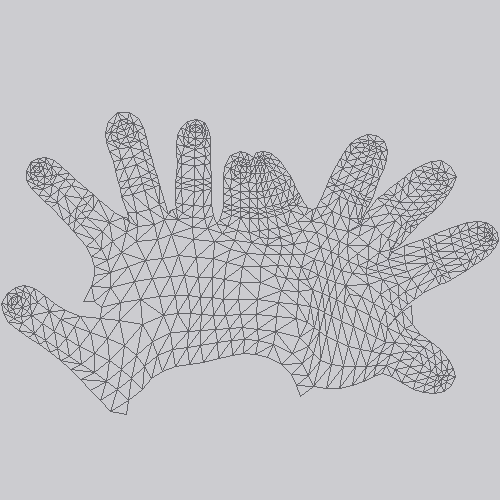} &
  \includegraphics[width=\csz\linewidth]{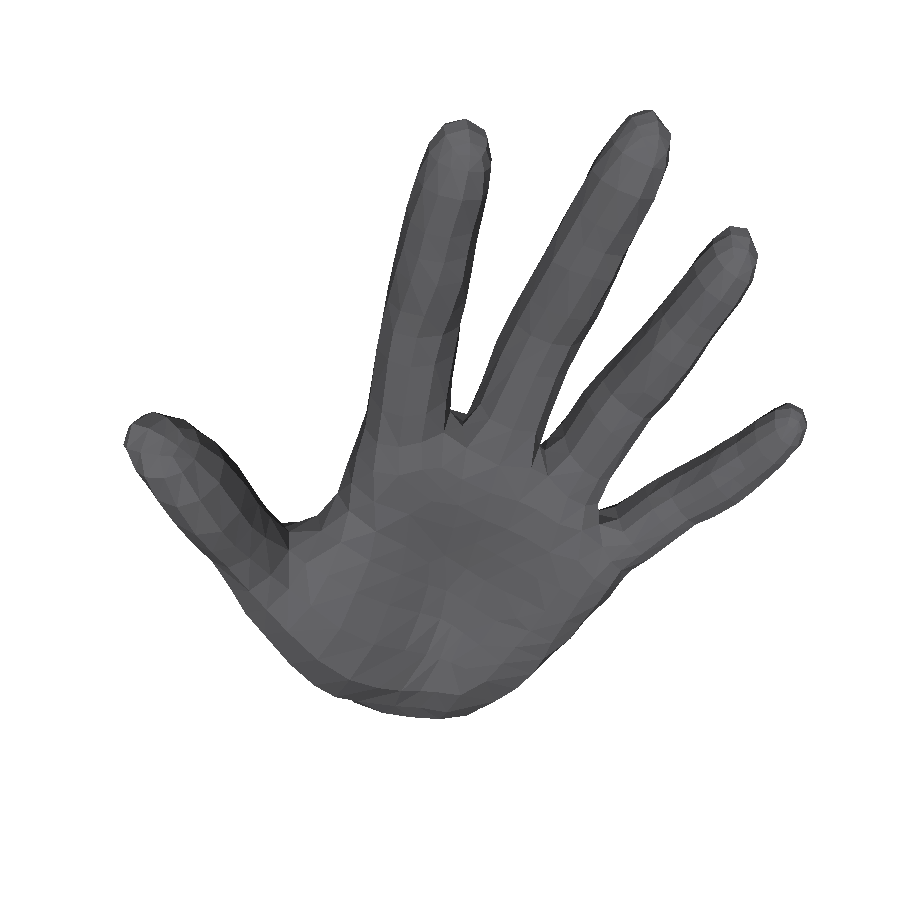} &

  \includegraphics[width=\csz\linewidth]{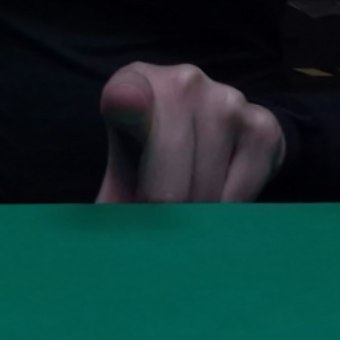} &
  \includegraphics[width=\csz\linewidth]{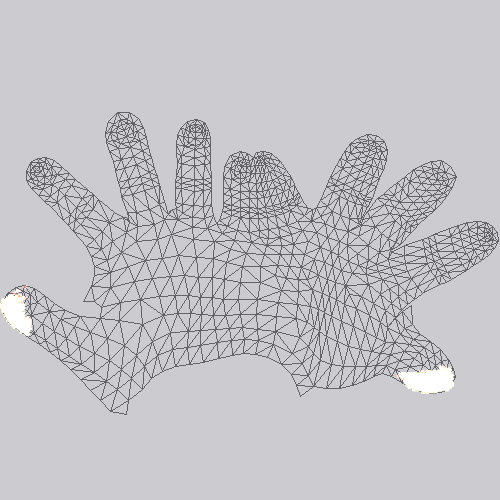} &
  \includegraphics[width=\csz\linewidth]{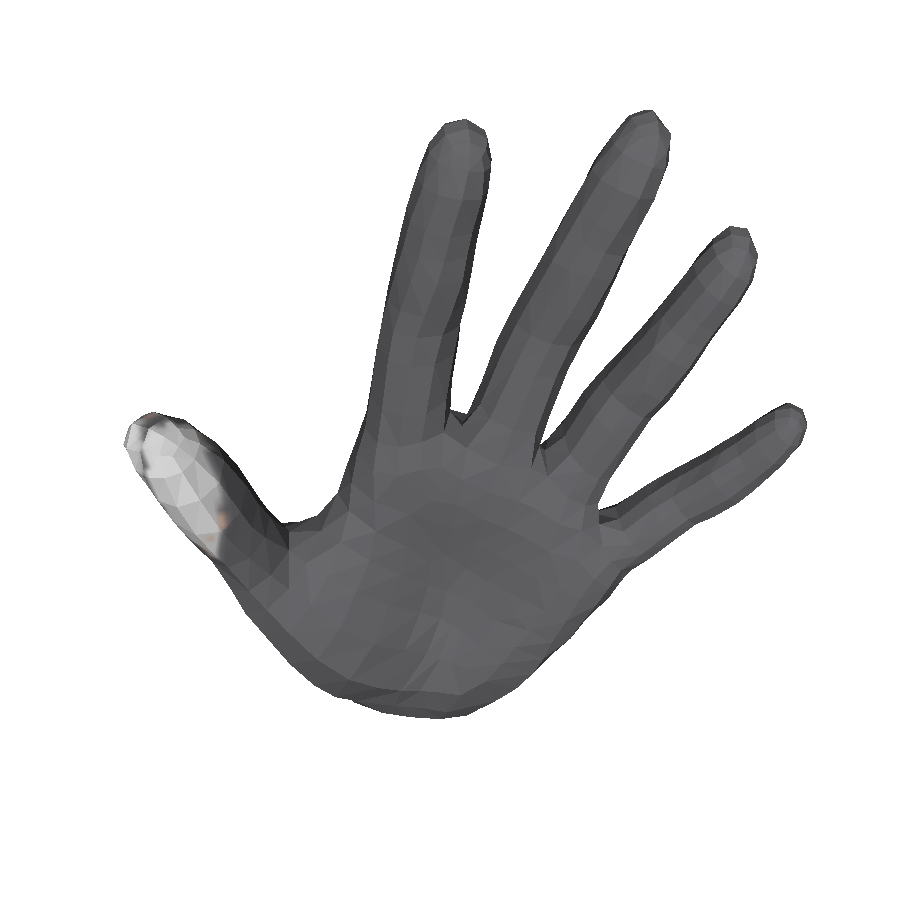} 

  \\

   \includegraphics[width=\csz\linewidth]{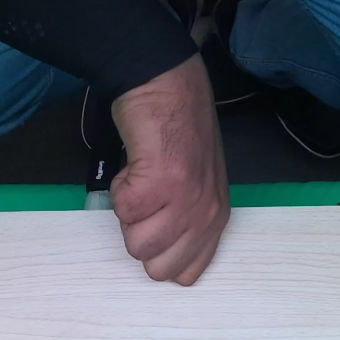} &
  \includegraphics[width=\csz\linewidth]{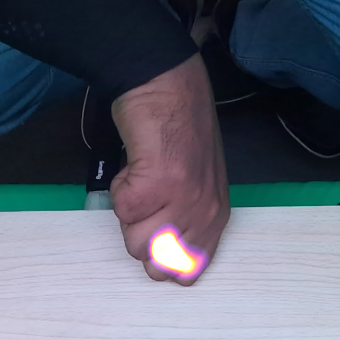} &
  \includegraphics[width=\csz\linewidth]{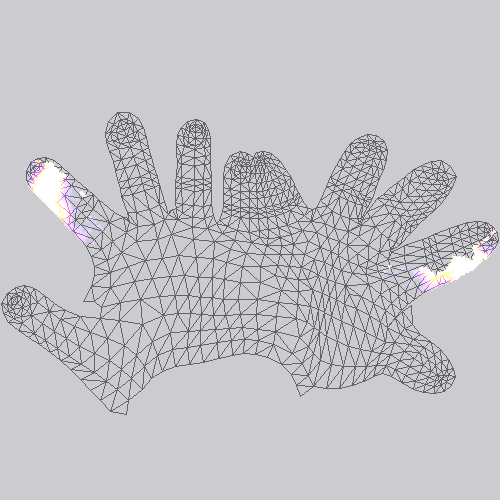} &
  \includegraphics[width=\csz\linewidth]{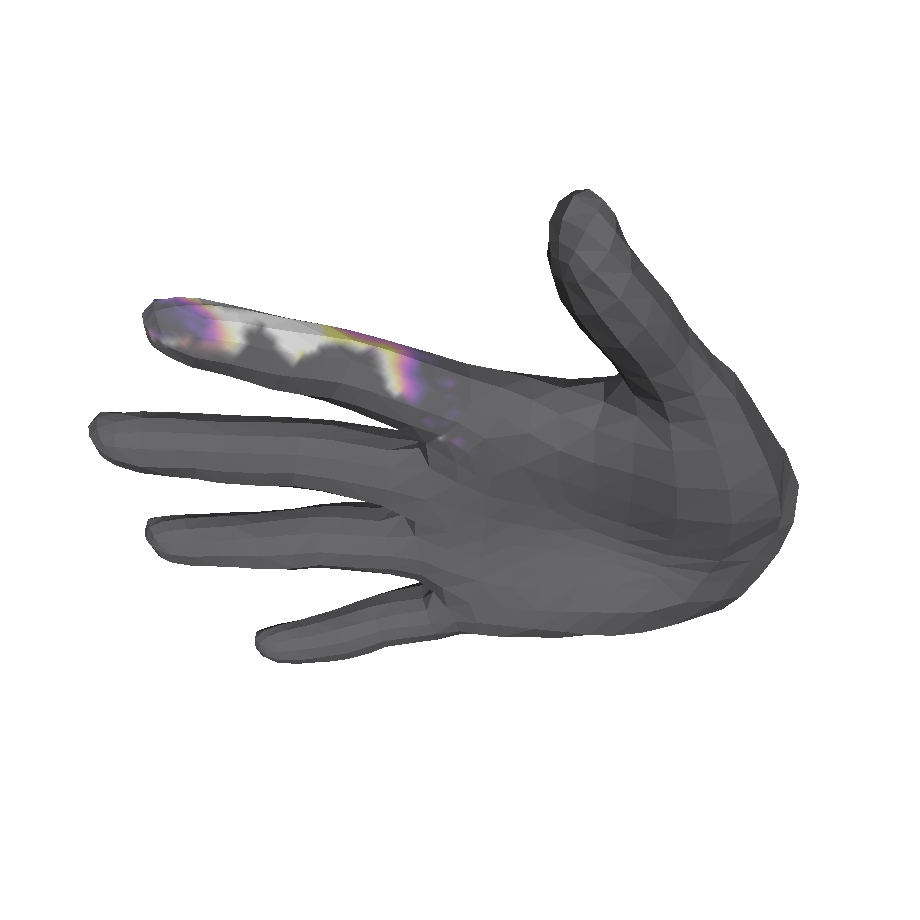} &
  \includegraphics[width=\csz\linewidth]{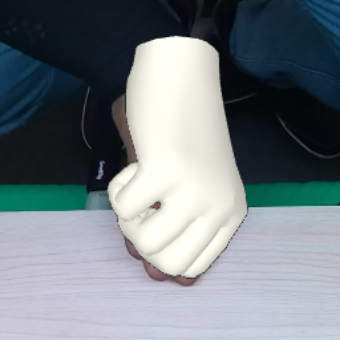} &
  
  \includegraphics[width=\csz\linewidth]{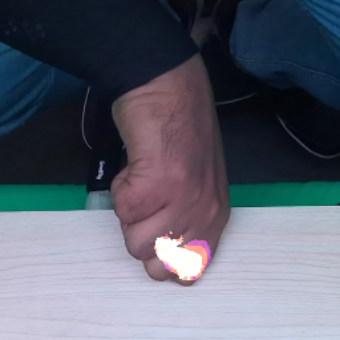} &
  \includegraphics[width=\csz\linewidth]{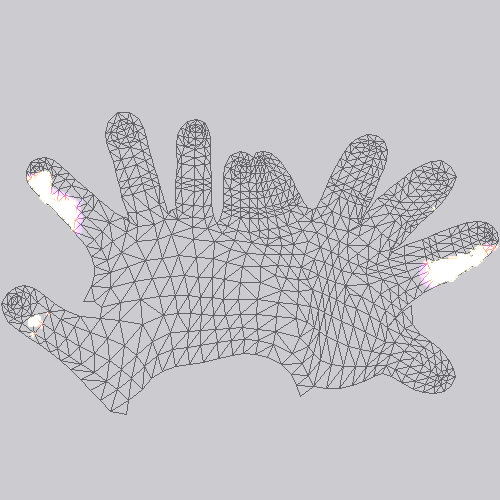} &
  \includegraphics[width=\csz\linewidth]{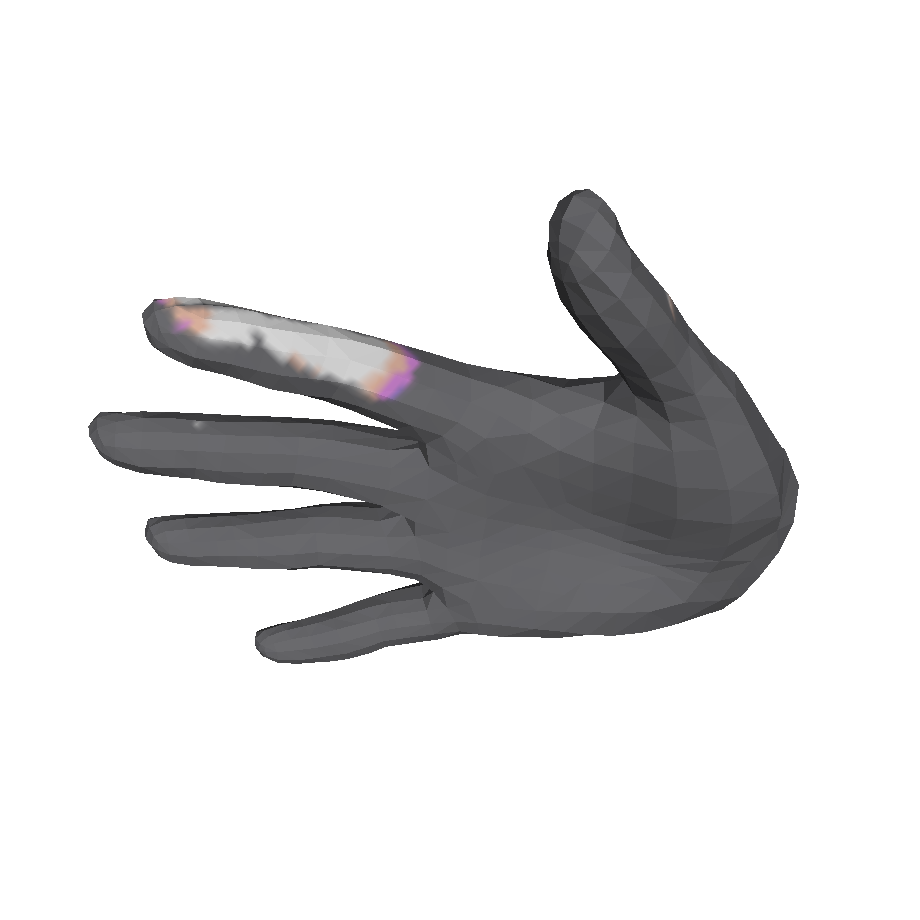} &

  \includegraphics[width=\csz\linewidth]{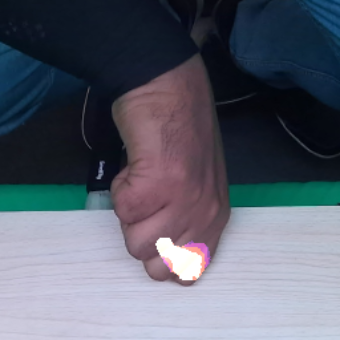} &
  \includegraphics[width=\csz\linewidth]{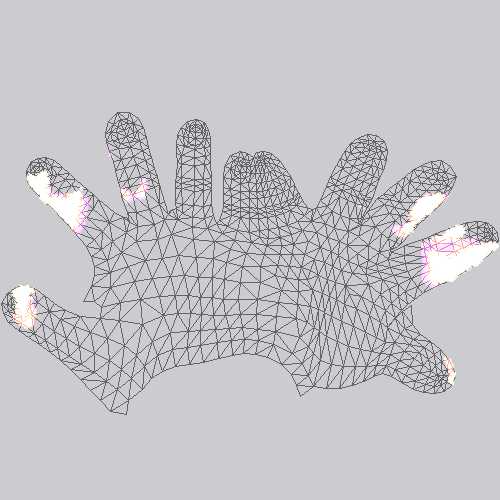} &
  \includegraphics[width=\csz\linewidth]{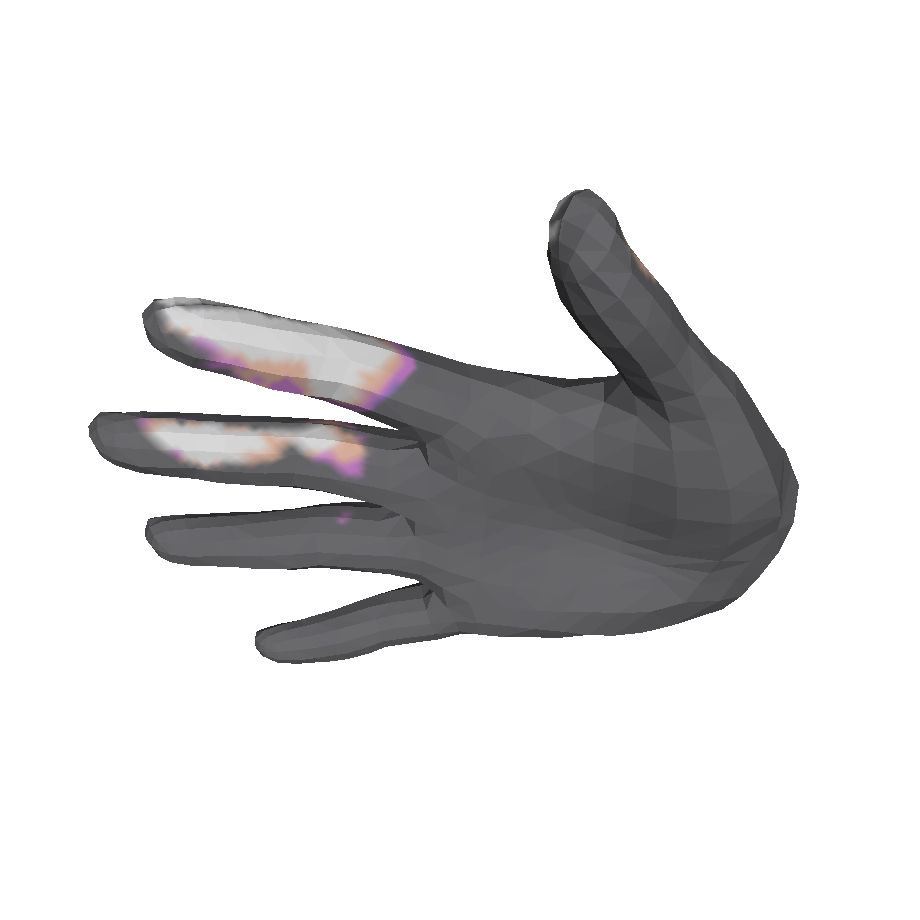} 

  \\

   \includegraphics[width=\csz\linewidth]{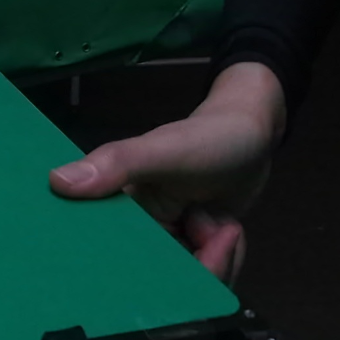} &
  \includegraphics[width=\csz\linewidth]{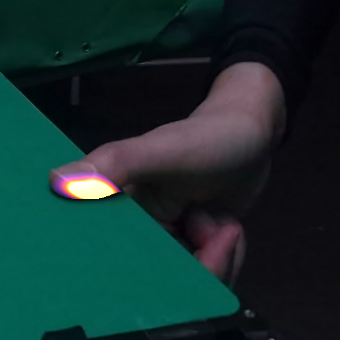} &
  \includegraphics[width=\csz\linewidth]{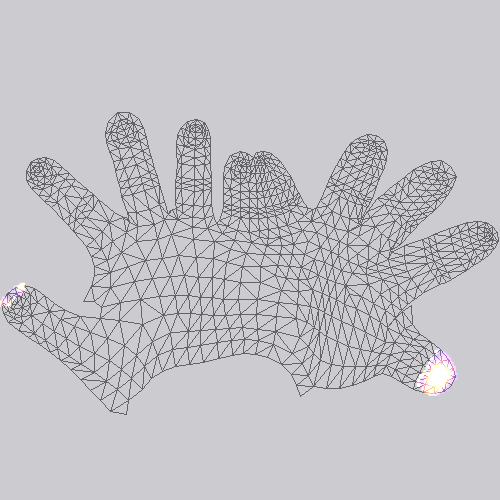} &
  \includegraphics[width=\csz\linewidth]{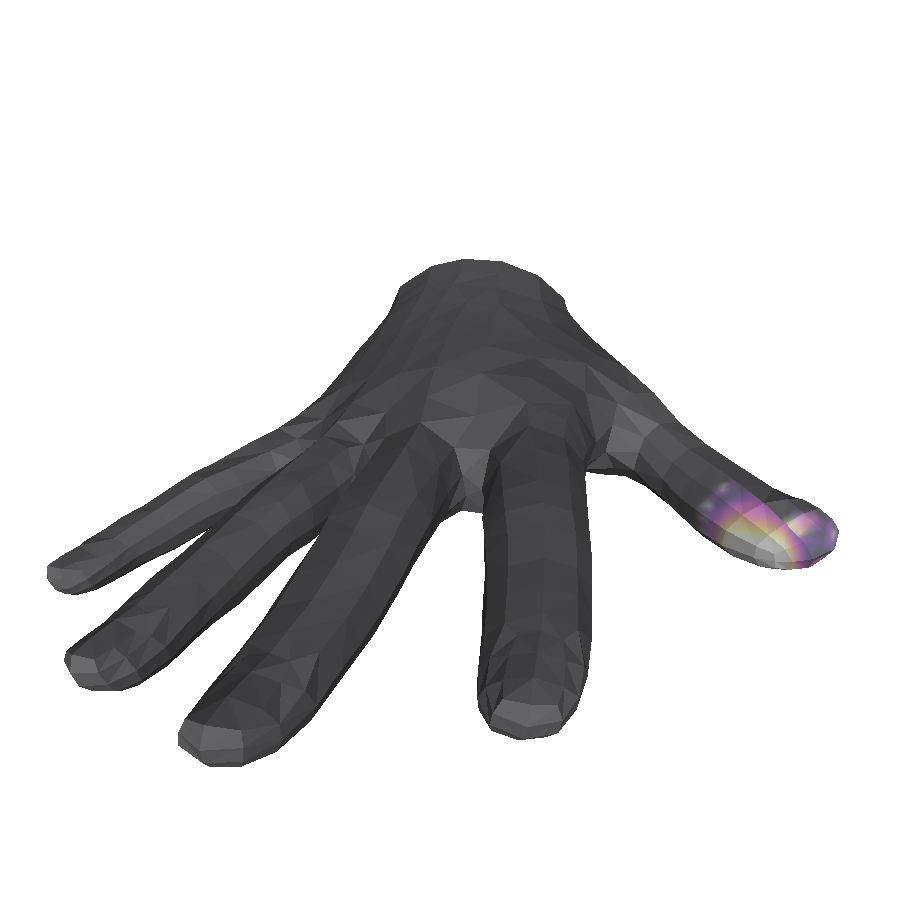} &
  \includegraphics[width=\csz\linewidth]{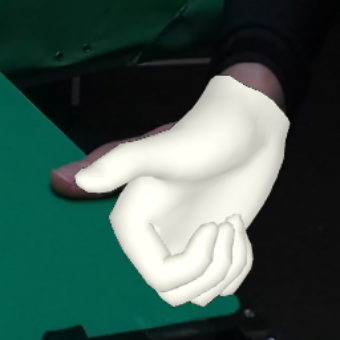} &
  
  \includegraphics[width=\csz\linewidth]{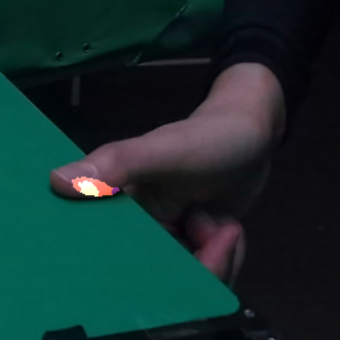} &
  \includegraphics[width=\csz\linewidth]{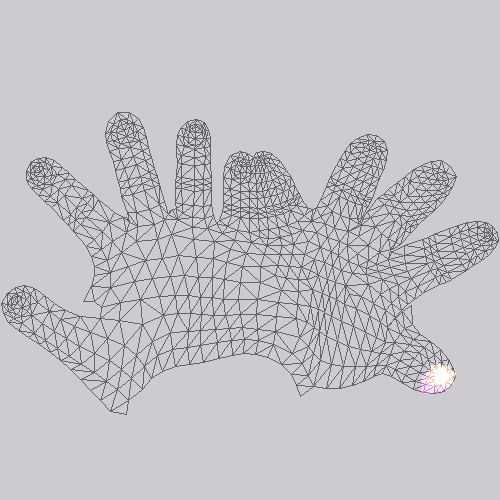} &
  \includegraphics[width=\csz\linewidth]{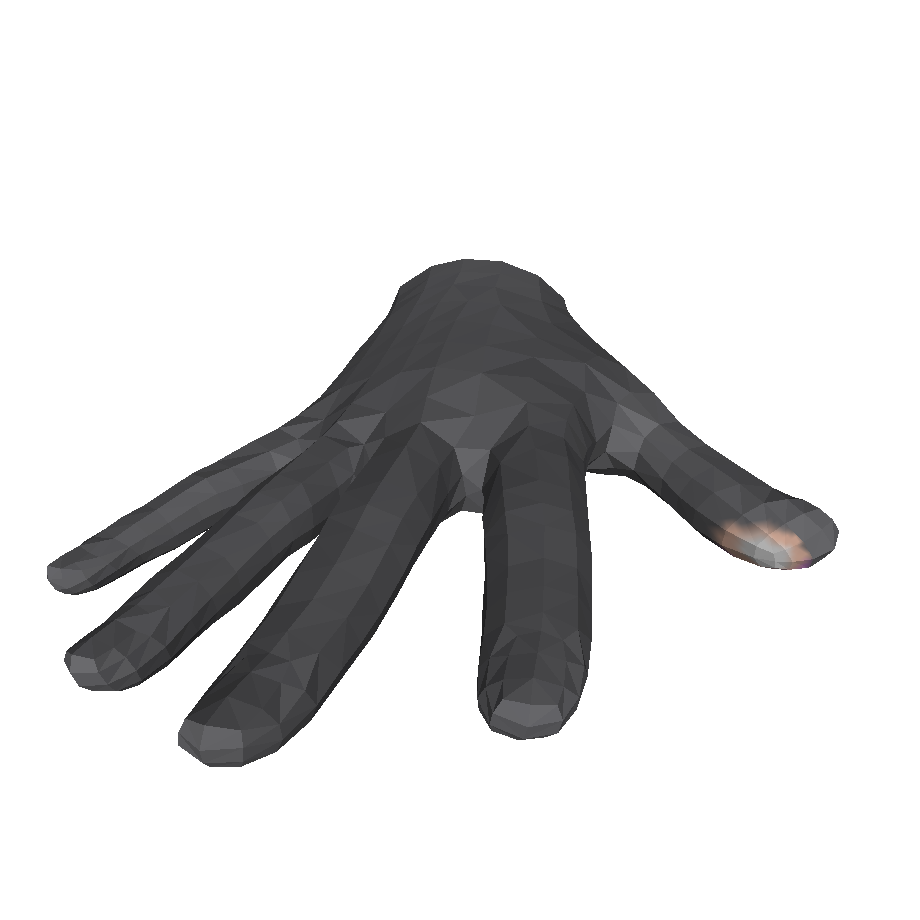} &

  \includegraphics[width=\csz\linewidth]{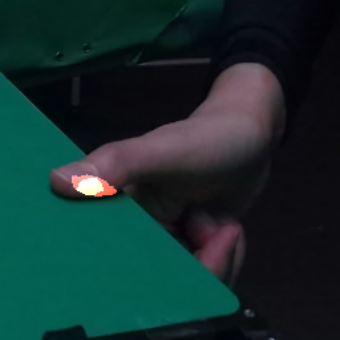} &
  \includegraphics[width=\csz\linewidth]{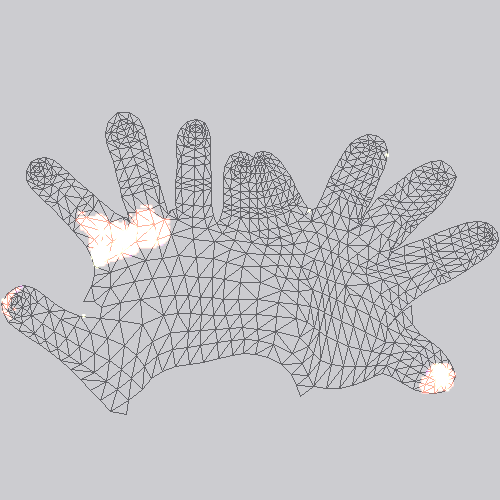} &
  \includegraphics[width=\csz\linewidth]{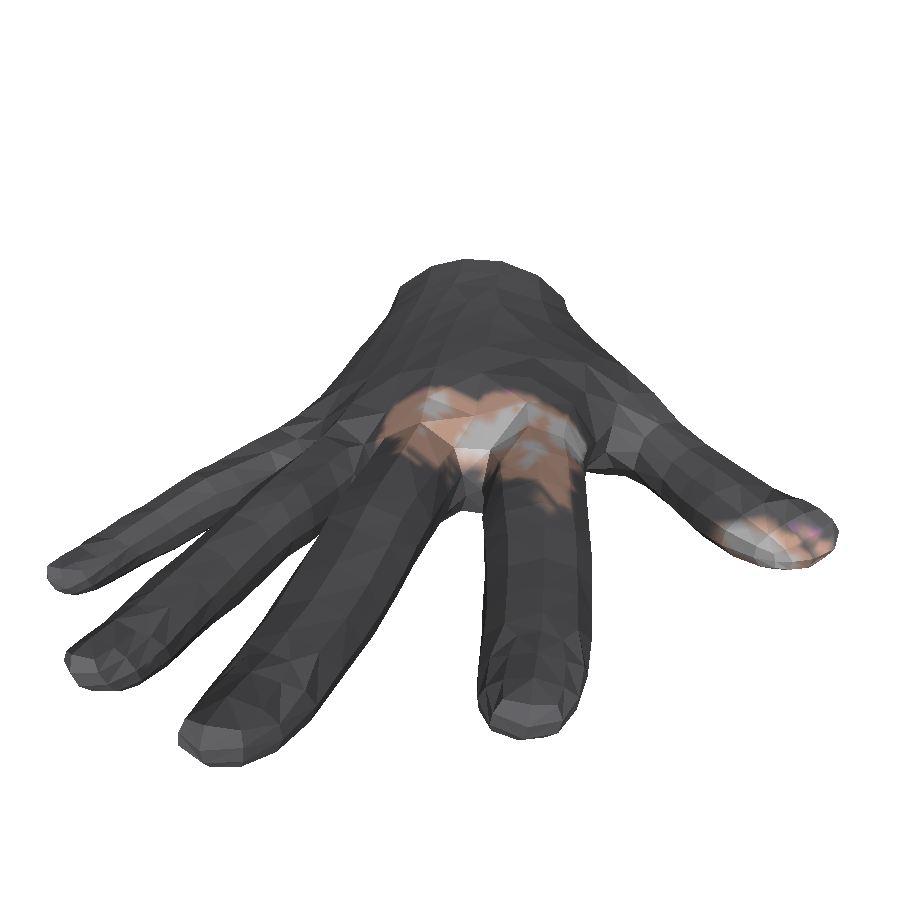} 

  \\

  Input &GT Pressure & GT UV & GT on Hand &  Mesh~\cite{hamer} & Pred. Pres. & Pred. UV &  On Hand
   &   Pred. Pres. & Pred. UV &  On Hand
  \\
     \multicolumn{5}{c}{}& \multicolumn{3}{c}{\tiny PressureFormer} &\multicolumn{3}{c}{\tiny PressureFormer w/o $\mathcal{L}_c$}

  \end{tabular}  
 \vspace{-2mm}
  \caption{\textbf{Qualitative examples demonstrating the impact of coarse UV loss supervision $\mathcal{L}_c$.} The coarse UV loss supervision $\mathcal{L}_c$ prevents the prediction of pressure in areas of the UV map that are not rendered on the image plane (see Section~\ref{sec:pressureformer}). These regions typically correspond to faces oriented toward the camera, where pressure and contact are not physically possible.}

\vspace{-5pt}
\end{figure} 
\begin{figure*}[t!]
  \centering
  \tiny
  \setlength{\tabcolsep}{0.2pt}
  \newcommand{\sz}{0.068}
  \newcommand{\sza}{1.8} 
  \begin{tabular}{cccc|c|ccc|ccc|ccc}

   \includegraphics[width=\sz\linewidth]{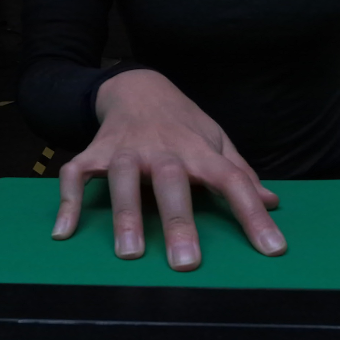} &
  \includegraphics[width=\sz\linewidth]{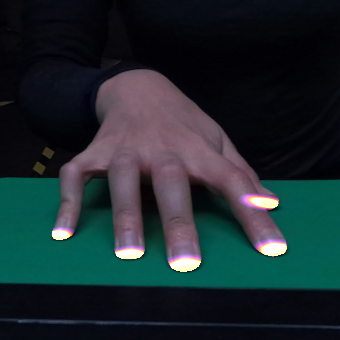} &
  \includegraphics[width=\sz\linewidth]{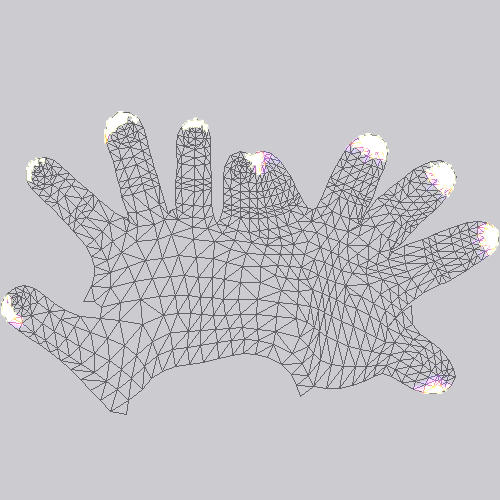} &
  \includegraphics[width=\sz\linewidth]{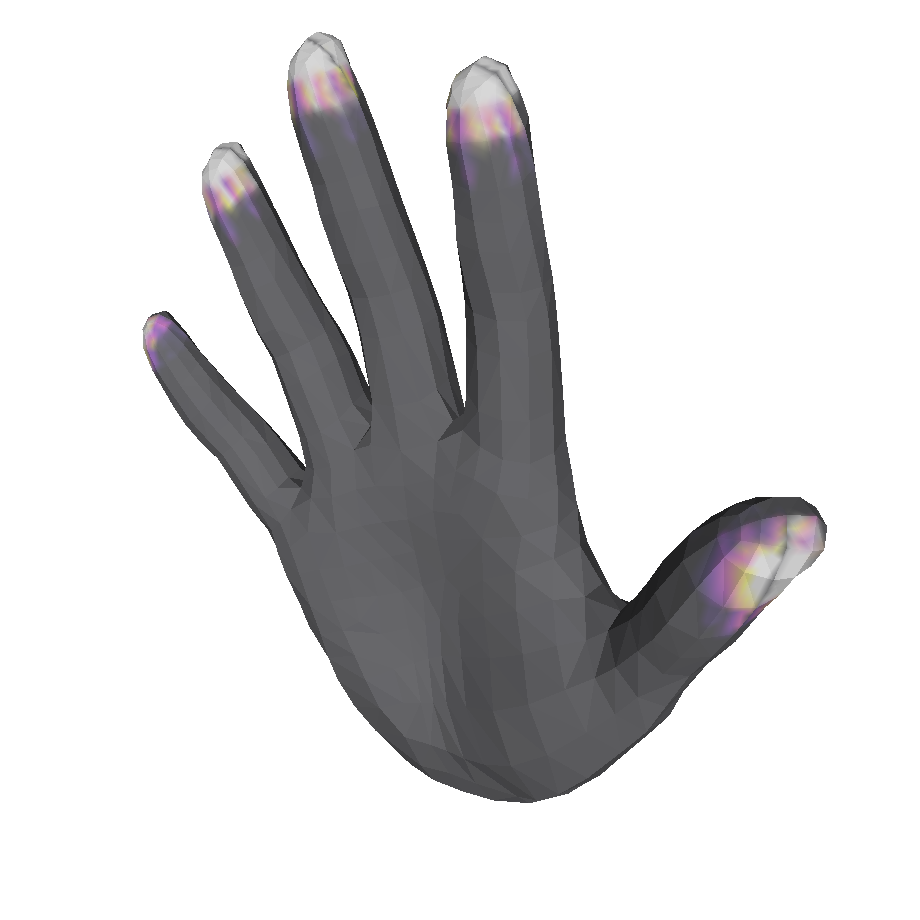} &
  \includegraphics[width=\sz\linewidth]{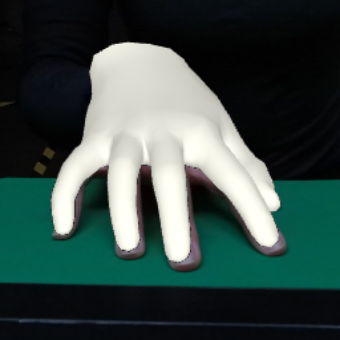} &
  
  \includegraphics[width=\sz\linewidth]{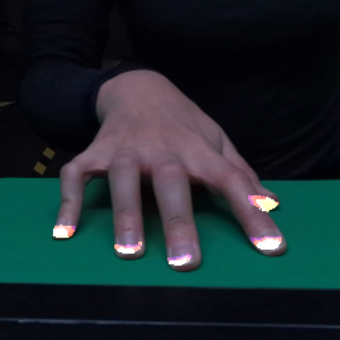} &
  \includegraphics[width=\sz\linewidth]{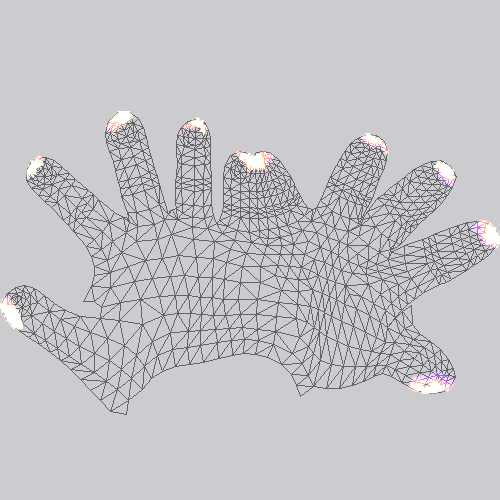} &
  \includegraphics[width=\sz\linewidth]{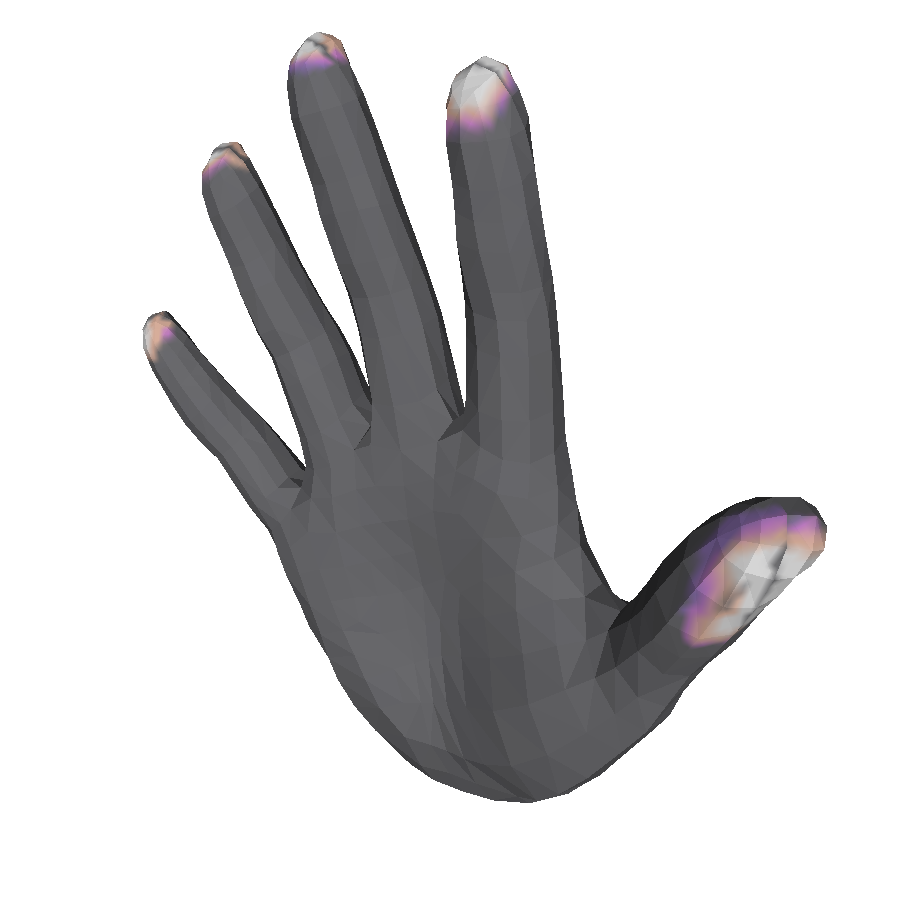} &

  \includegraphics[width=\sz\linewidth]{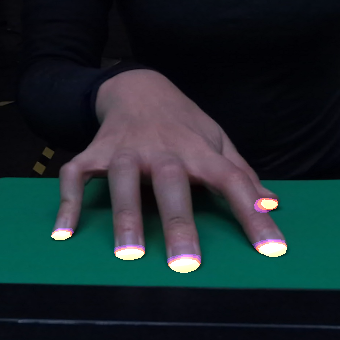} &
  \includegraphics[width=\sz\linewidth]{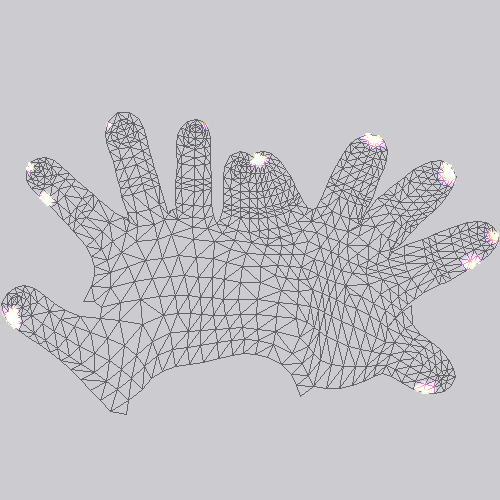} &
  \includegraphics[width=\sz\linewidth]{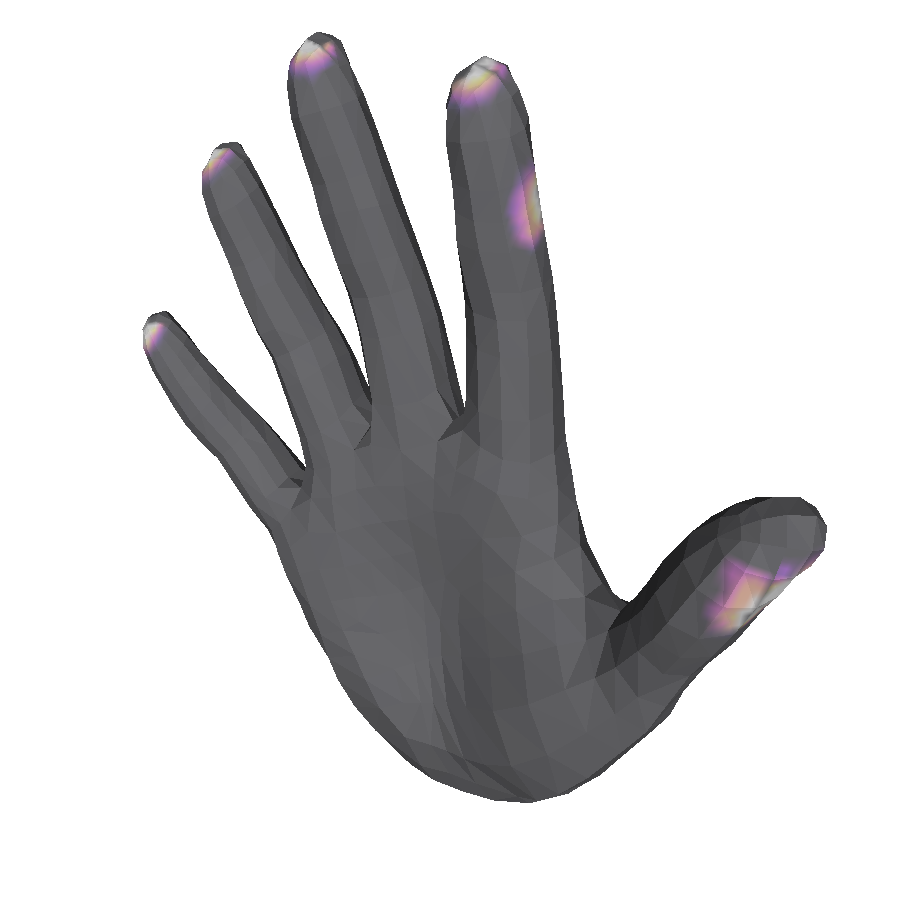} &
  
  \includegraphics[width=\sz\linewidth]{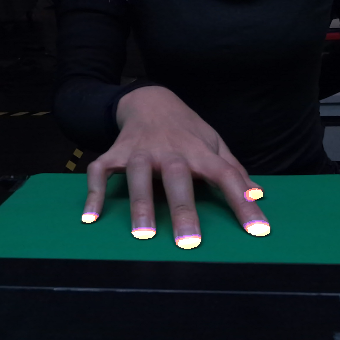} &
  \includegraphics[width=\sz\linewidth]{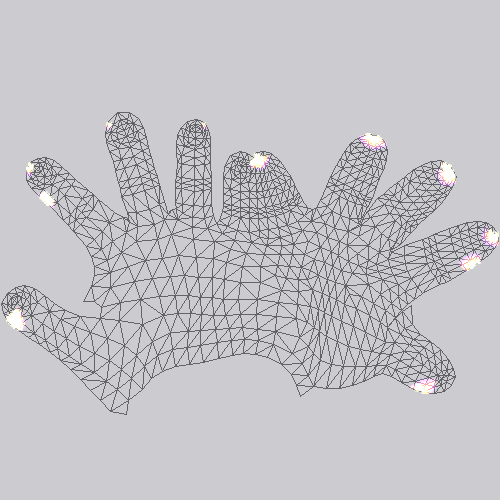} &
  \includegraphics[width=\sz\linewidth]{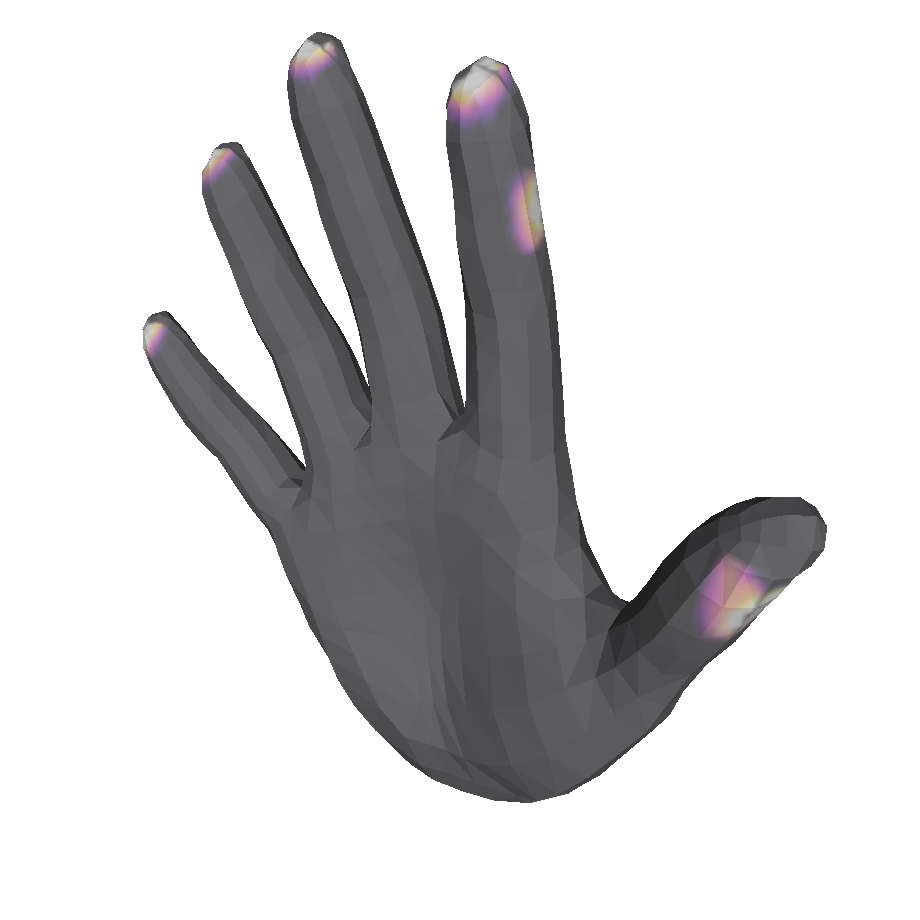} 
  \\
  
   \includegraphics[width=\sz\linewidth]{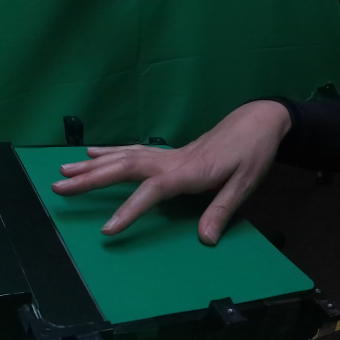} &
  \includegraphics[width=\sz\linewidth]{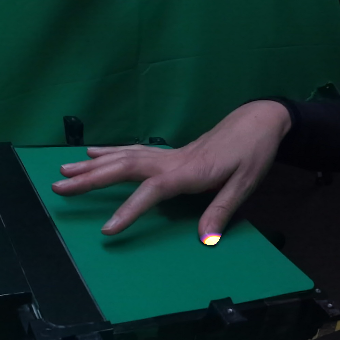} &
  \includegraphics[width=\sz\linewidth]{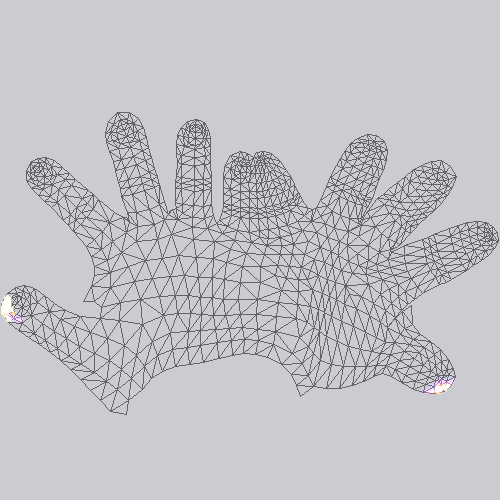} &
  \includegraphics[width=\sz\linewidth]{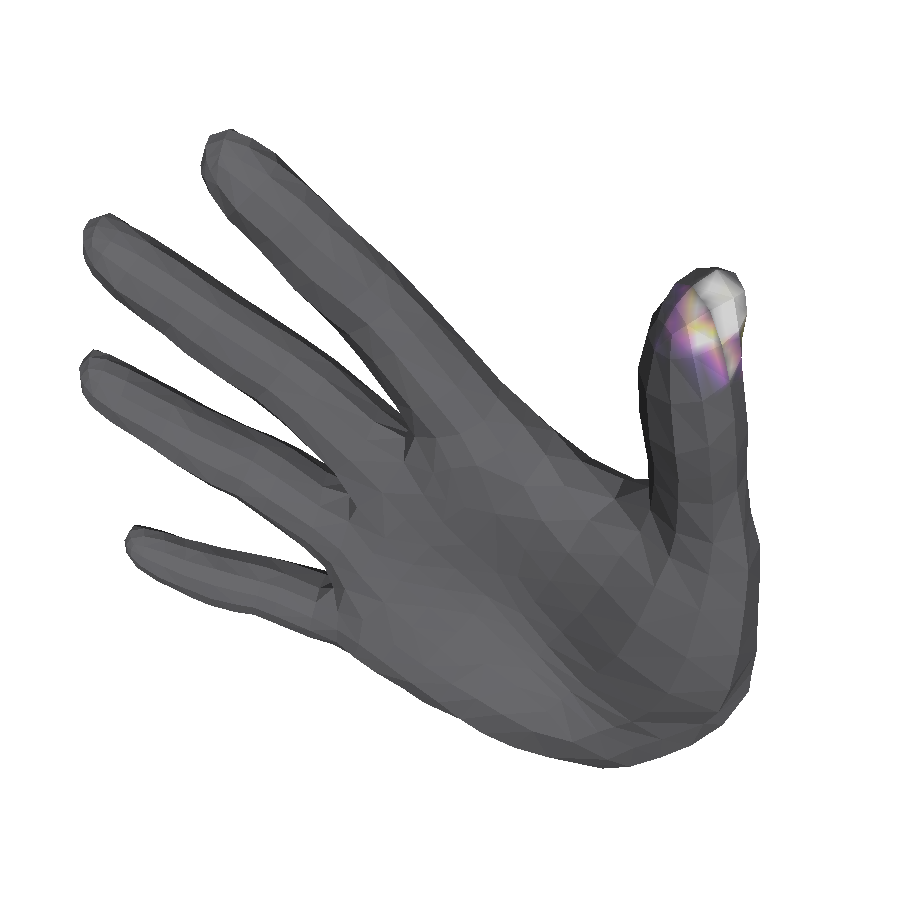} &
  \includegraphics[width=\sz\linewidth]{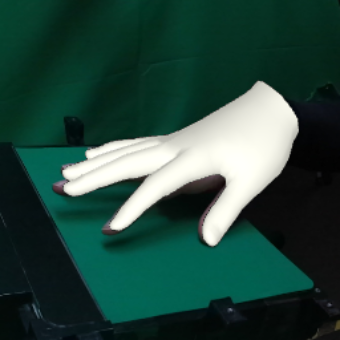} &
  
  \includegraphics[width=\sz\linewidth]{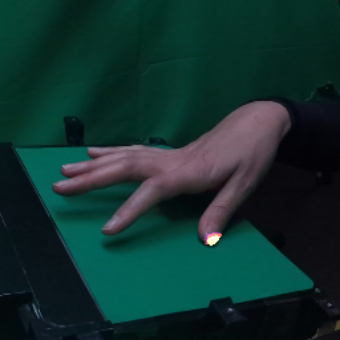} &
  \includegraphics[width=\sz\linewidth]{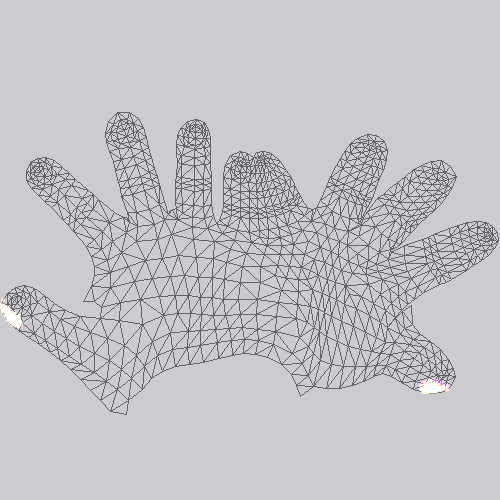} &
  \includegraphics[width=\sz\linewidth]{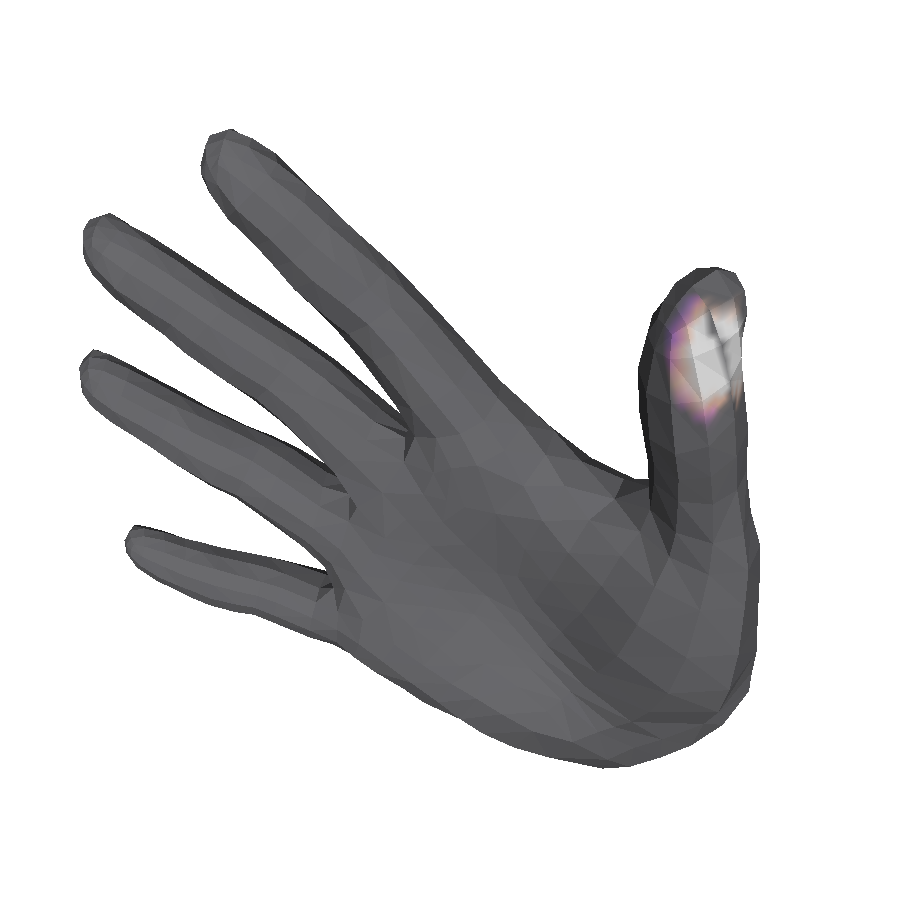} &

  \includegraphics[width=\sz\linewidth]{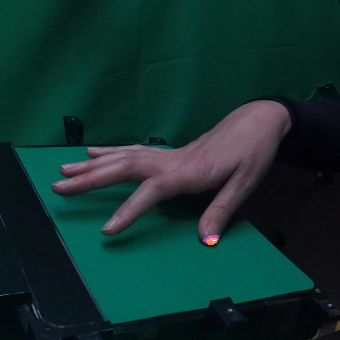} &
  \includegraphics[width=\sz\linewidth]{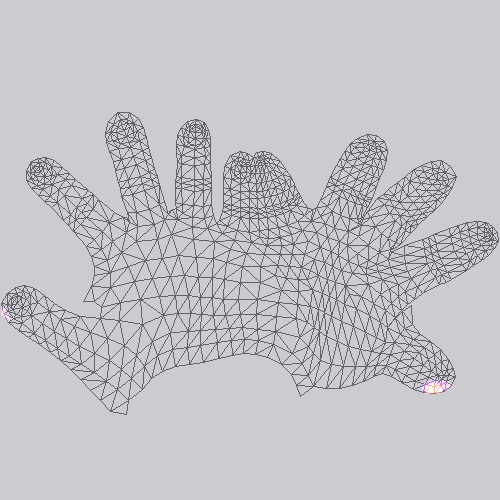} &
  \includegraphics[width=\sz\linewidth]{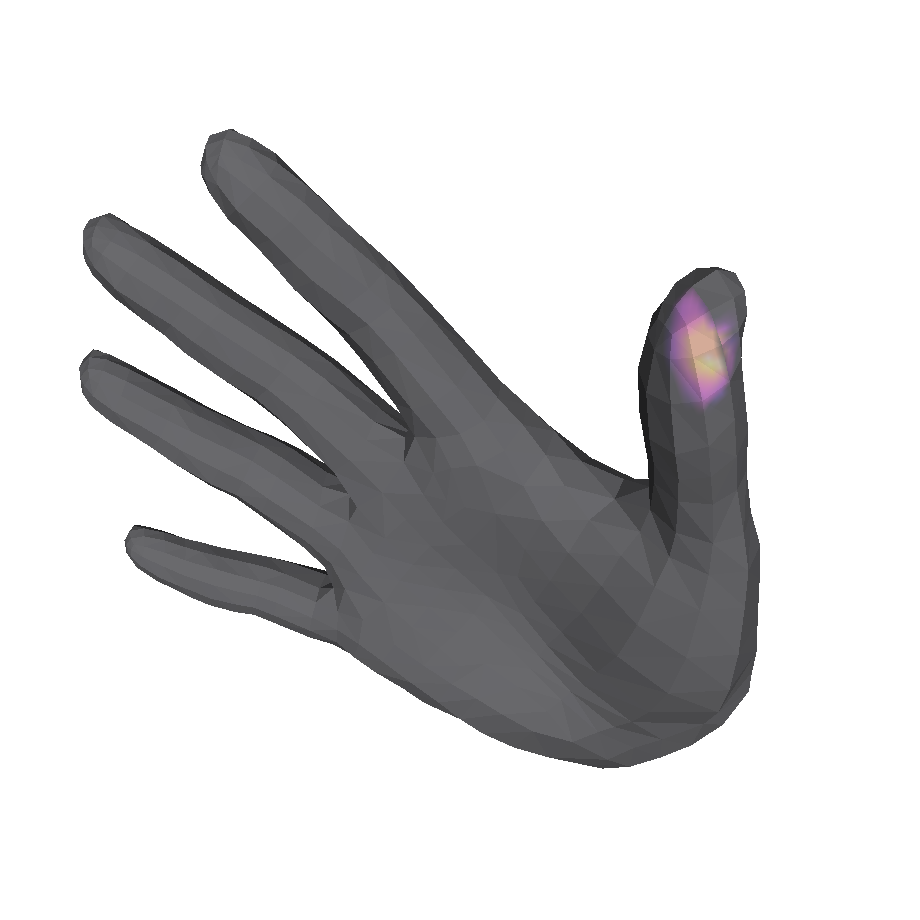} &
  
  \includegraphics[width=\sz\linewidth]{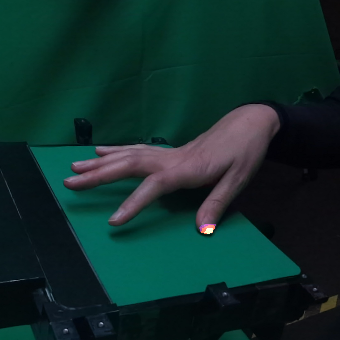} &
  \includegraphics[width=\sz\linewidth]{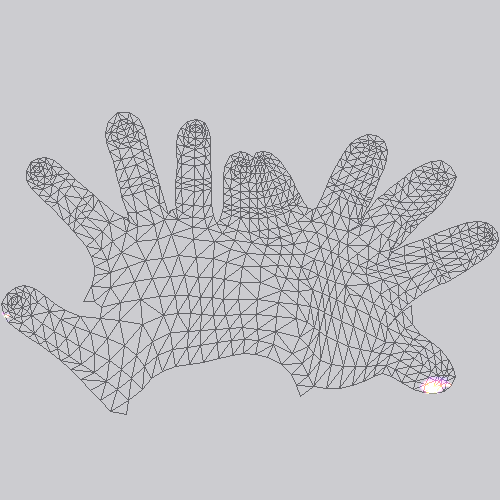} &
  \includegraphics[width=\sz\linewidth]{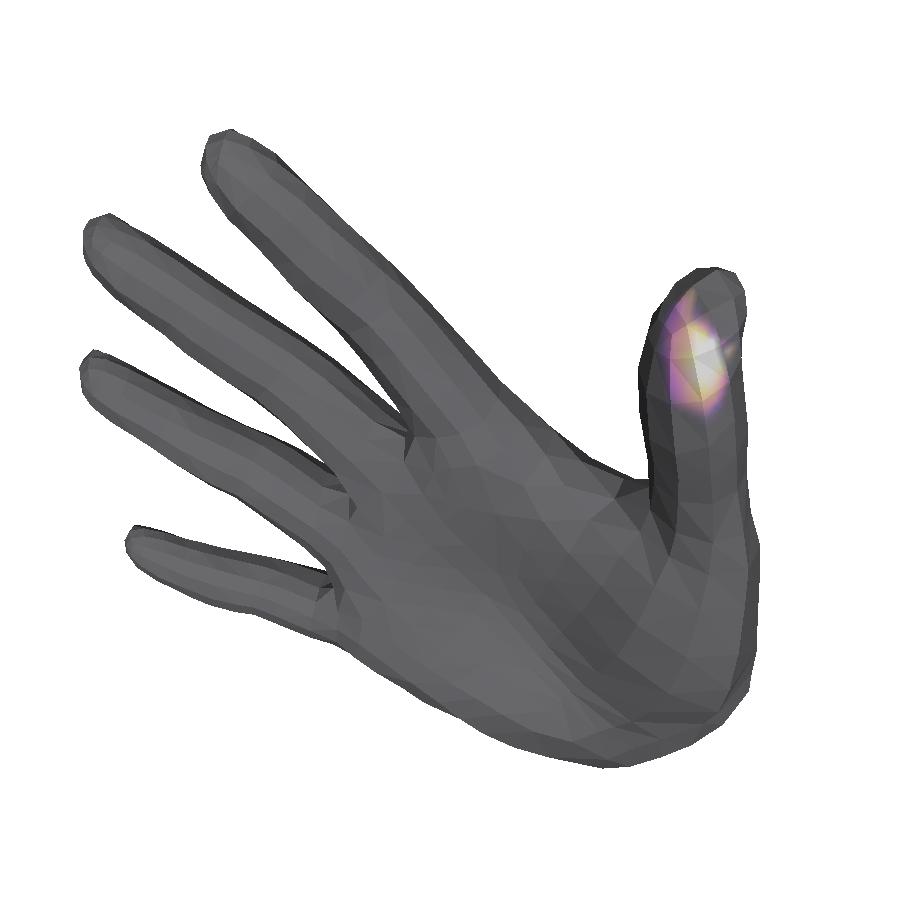} 
\\
  \includegraphics[width=\sz\linewidth]{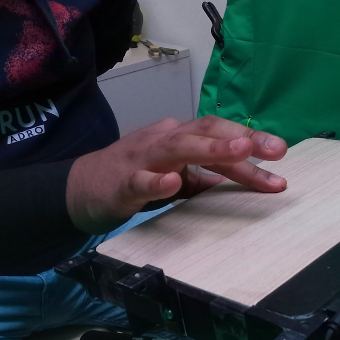} &
  \includegraphics[width=\sz\linewidth]{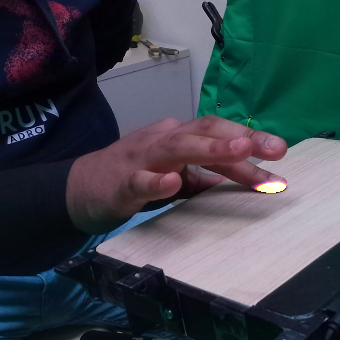} &
  \includegraphics[width=\sz\linewidth]{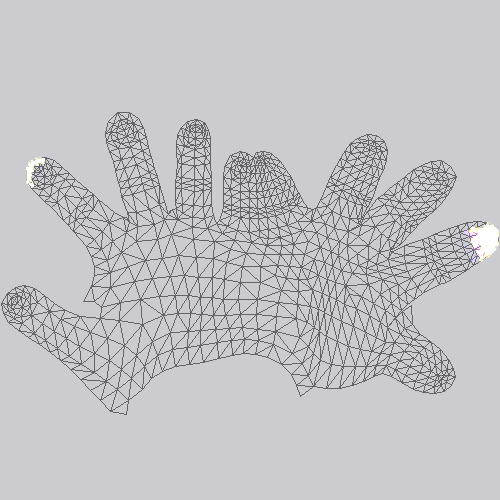} &
  \includegraphics[width=\sz\linewidth]{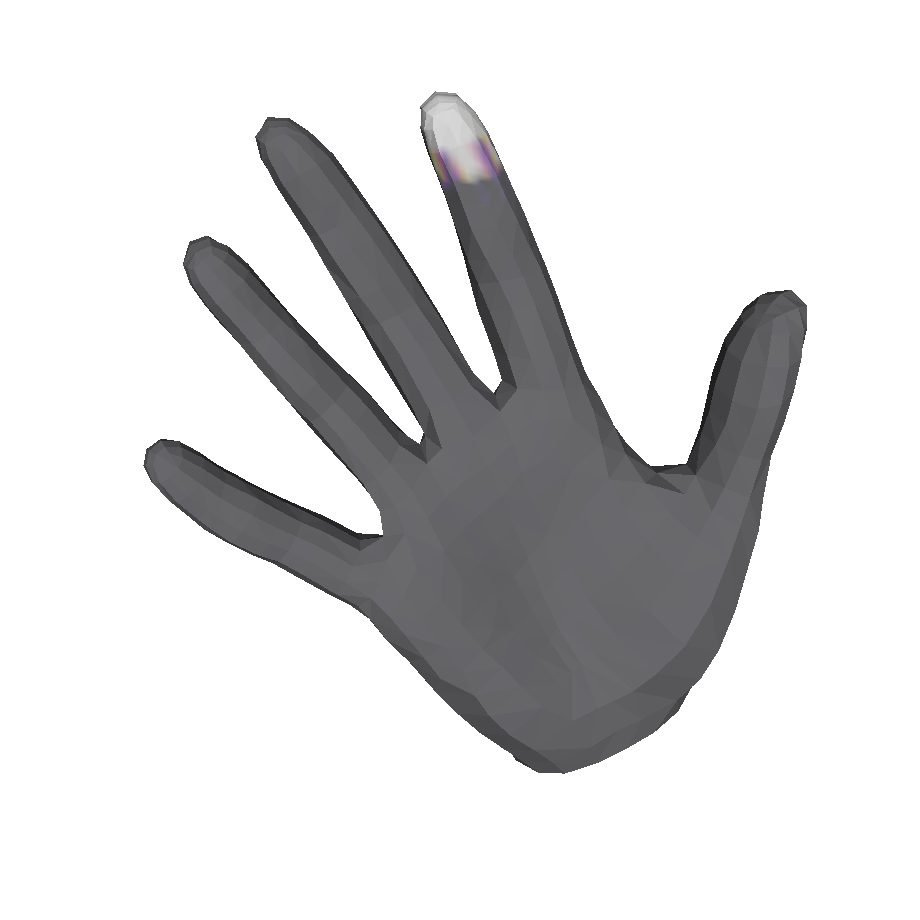} &
  \includegraphics[width=\sz\linewidth]{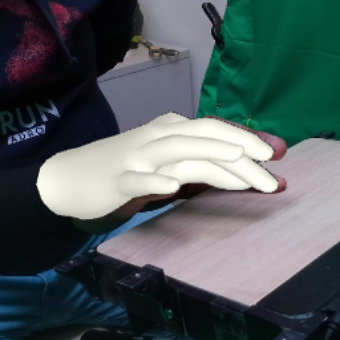} &
  
  \includegraphics[width=\sz\linewidth]{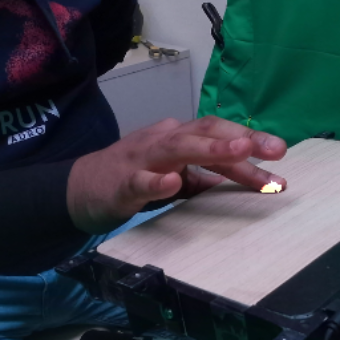} &
  \includegraphics[width=\sz\linewidth]{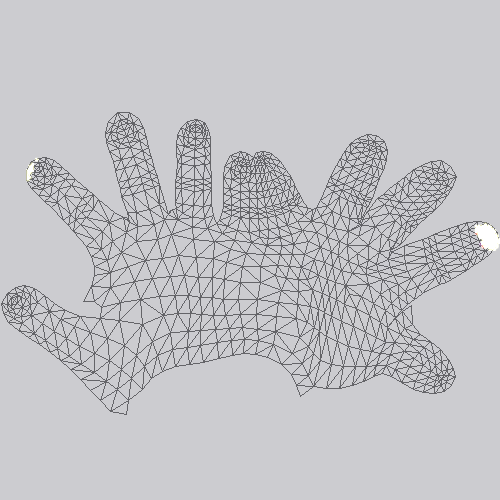} &
  \includegraphics[width=\sz\linewidth]{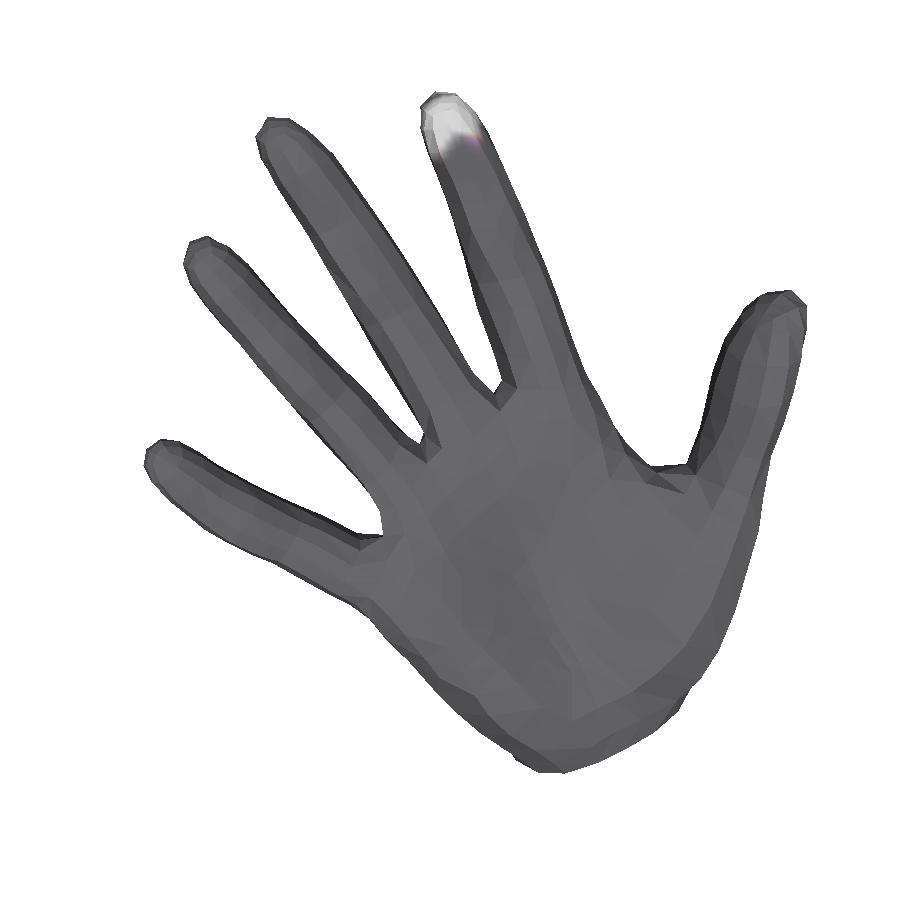} &

  \includegraphics[width=\sz\linewidth]{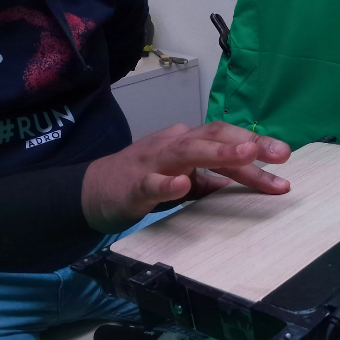} &
  \includegraphics[width=\sz\linewidth]{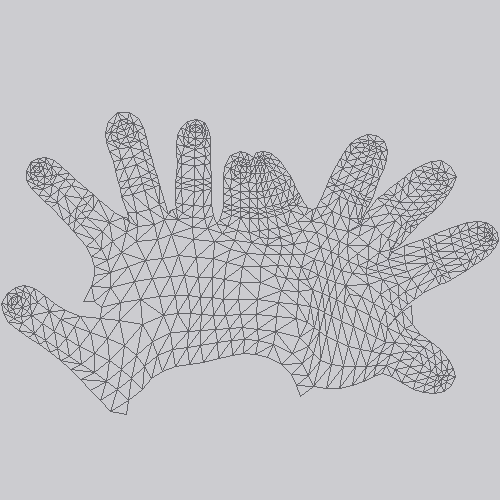} &
  \includegraphics[width=\sz\linewidth]{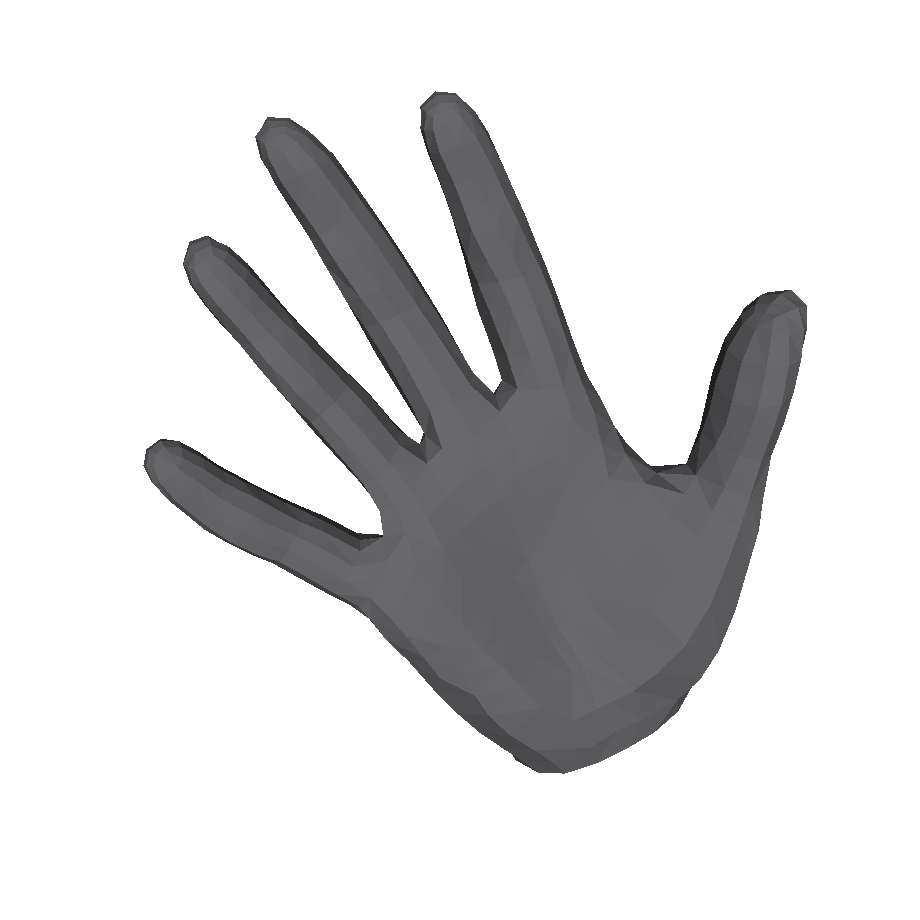} &
  
  \includegraphics[width=\sz\linewidth]{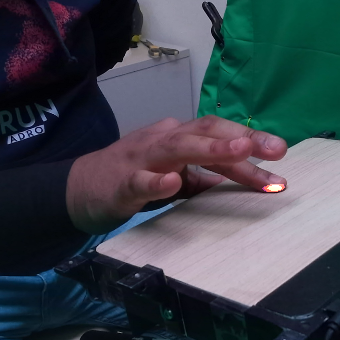} &
  \includegraphics[width=\sz\linewidth]{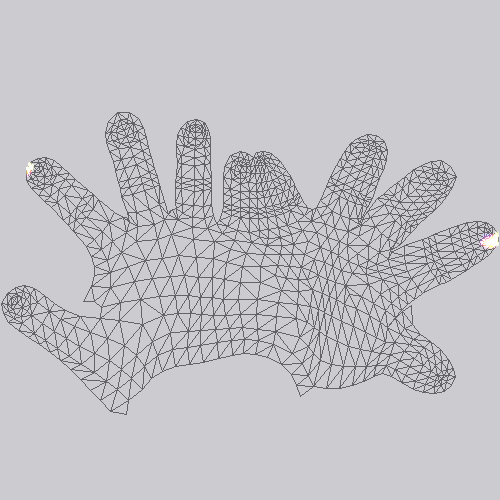} &
  \includegraphics[width=\sz\linewidth]{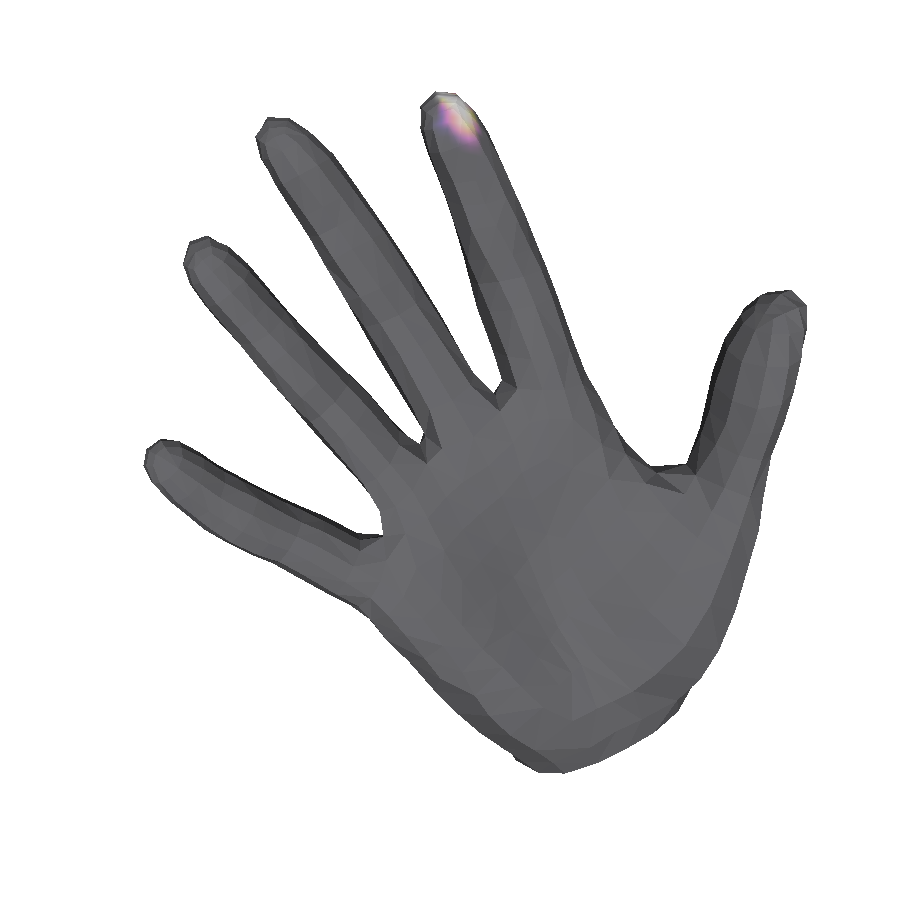} 
\\
   \includegraphics[width=\sz\linewidth]{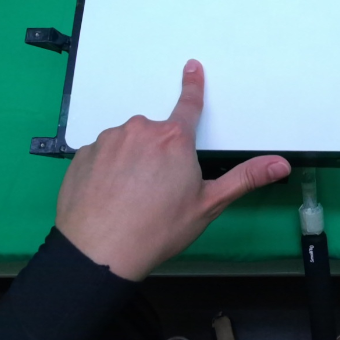} &
  \includegraphics[width=\sz\linewidth]{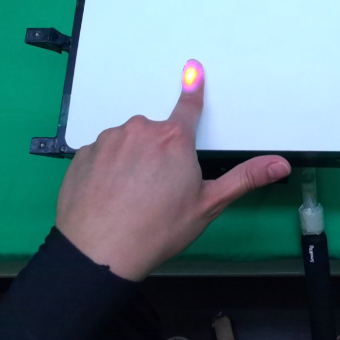} &
  \includegraphics[width=\sz\linewidth]{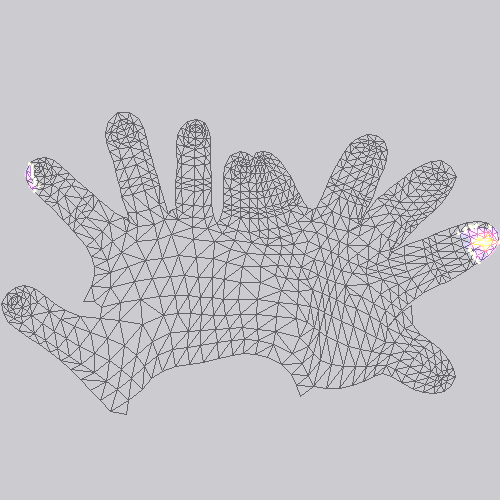} &
  \includegraphics[width=\sz\linewidth]{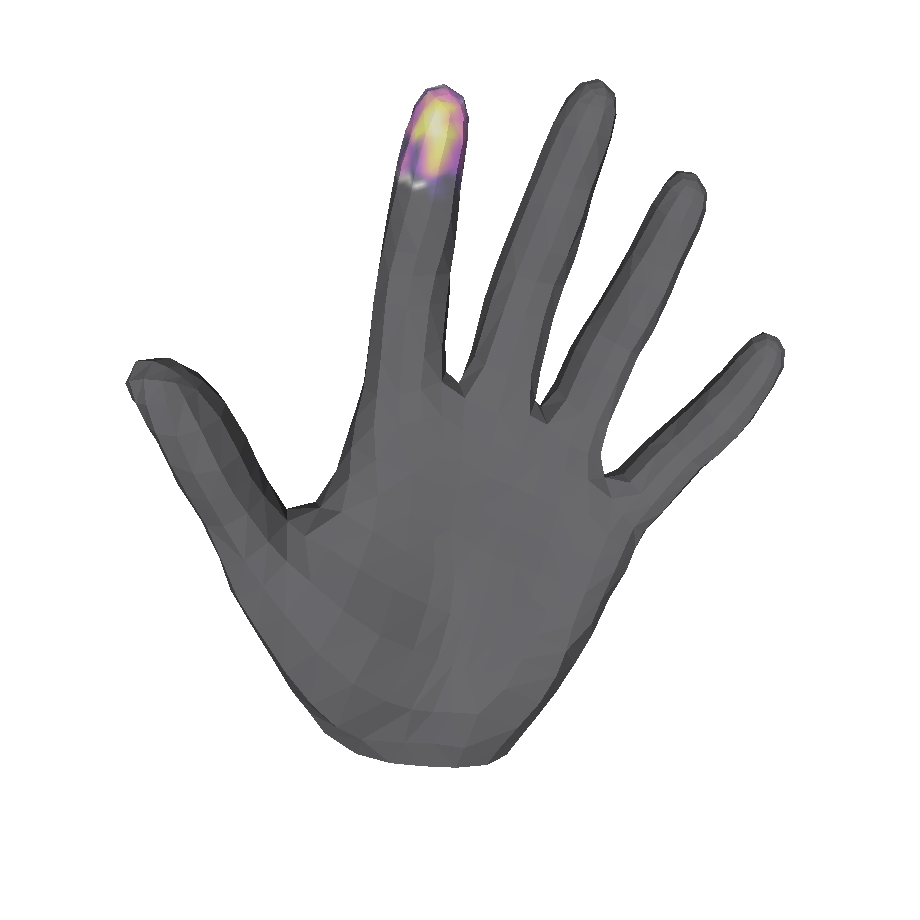} &
  \includegraphics[width=\sz\linewidth]{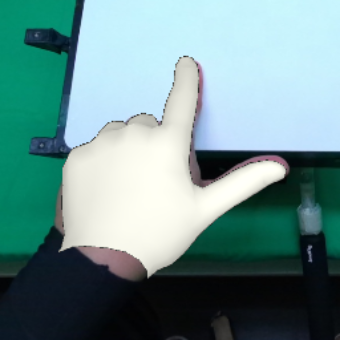} &
  
  \includegraphics[width=\sz\linewidth]{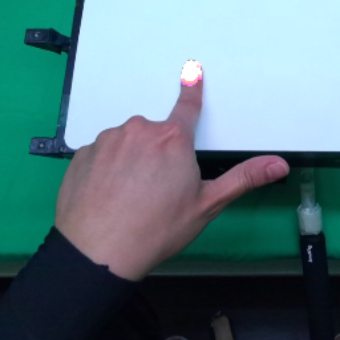} &
  \includegraphics[width=\sz\linewidth]{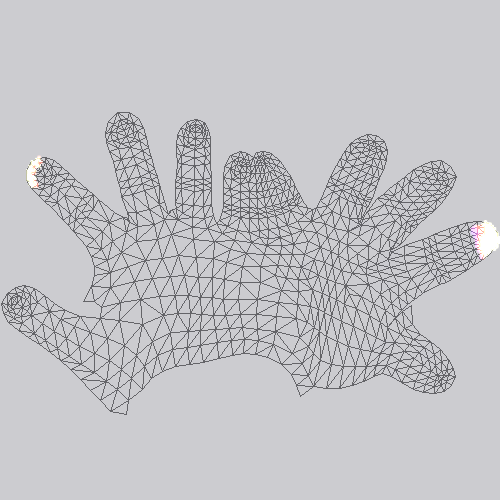} &
  \includegraphics[width=\sz\linewidth]{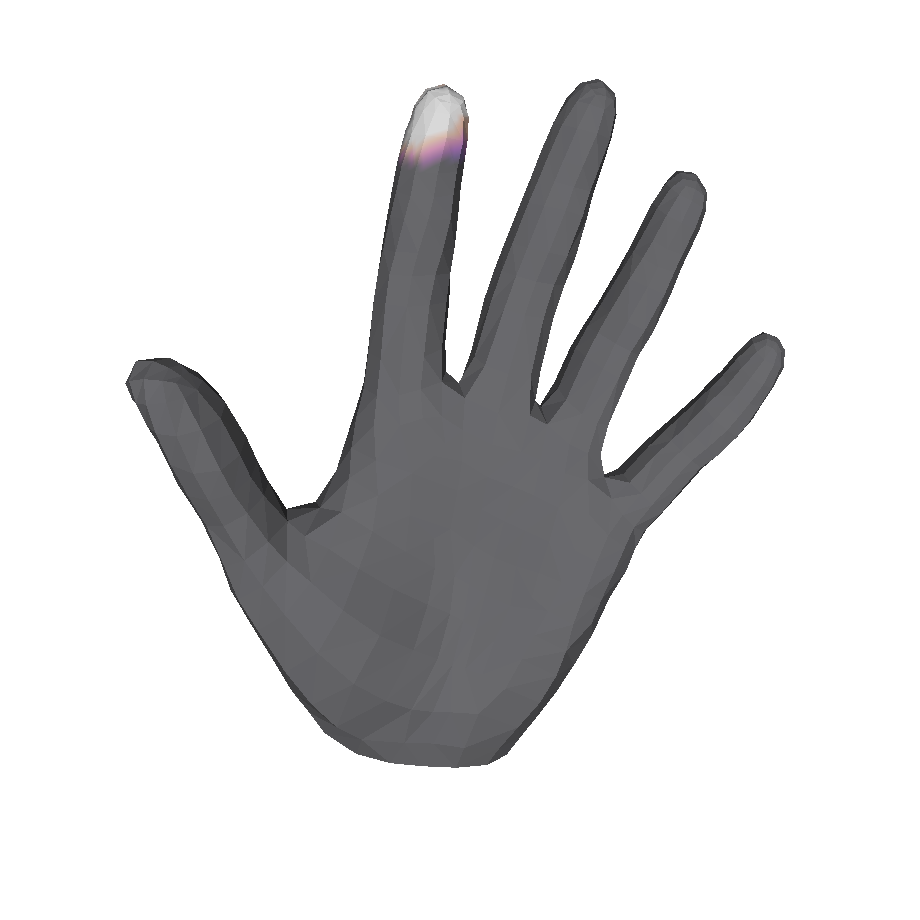} &

  \includegraphics[width=\sz\linewidth]{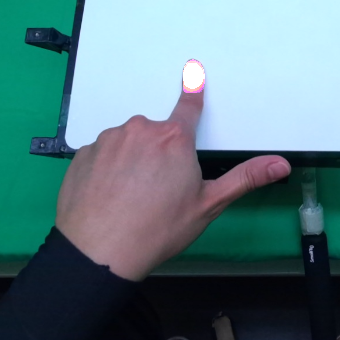} &
  \includegraphics[width=\sz\linewidth]{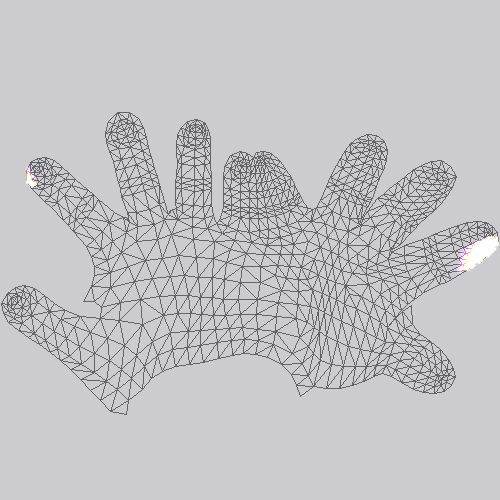} &
  \includegraphics[width=\sz\linewidth]{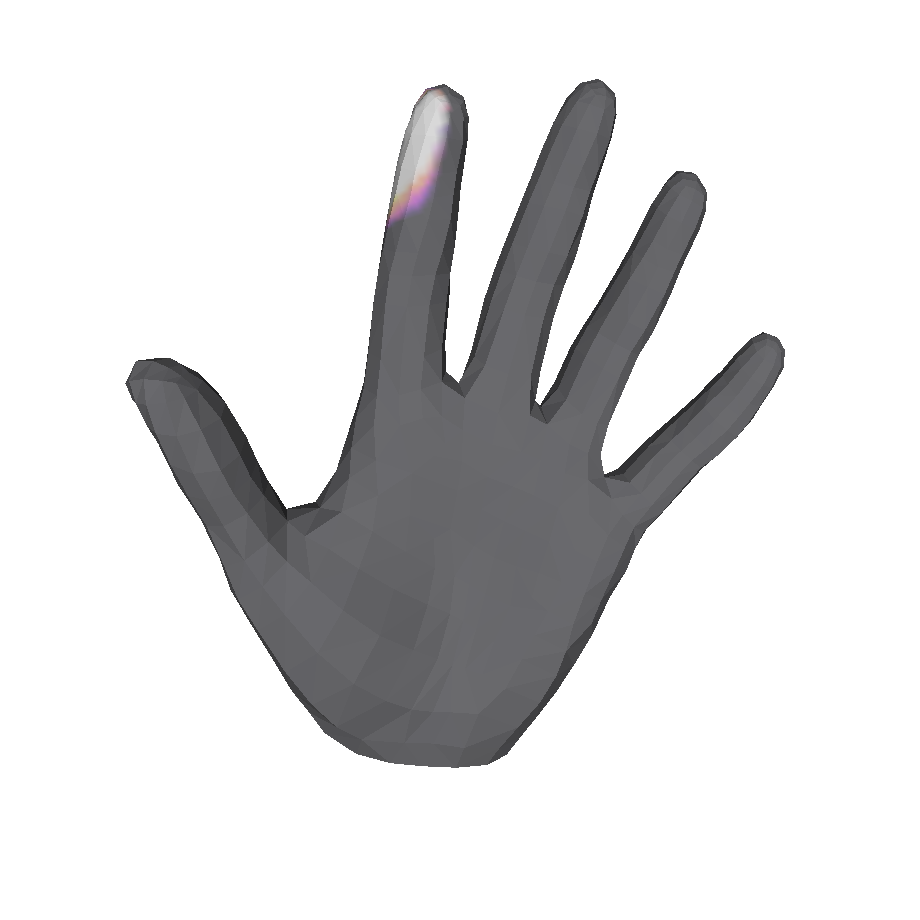} &
  
  \includegraphics[width=\sz\linewidth]{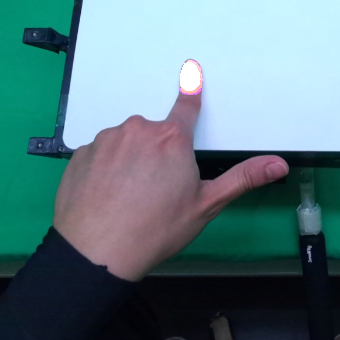} &
  \includegraphics[width=\sz\linewidth]{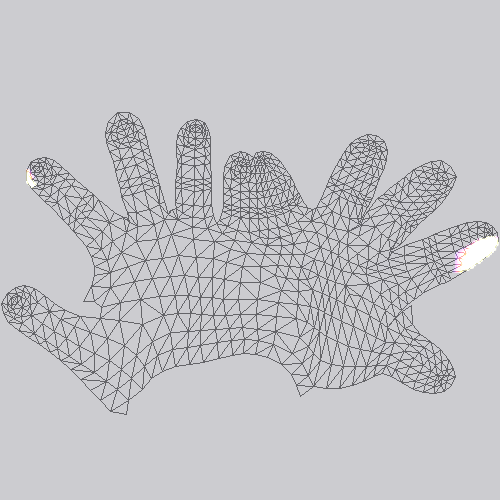} &
  \includegraphics[width=\sz\linewidth]{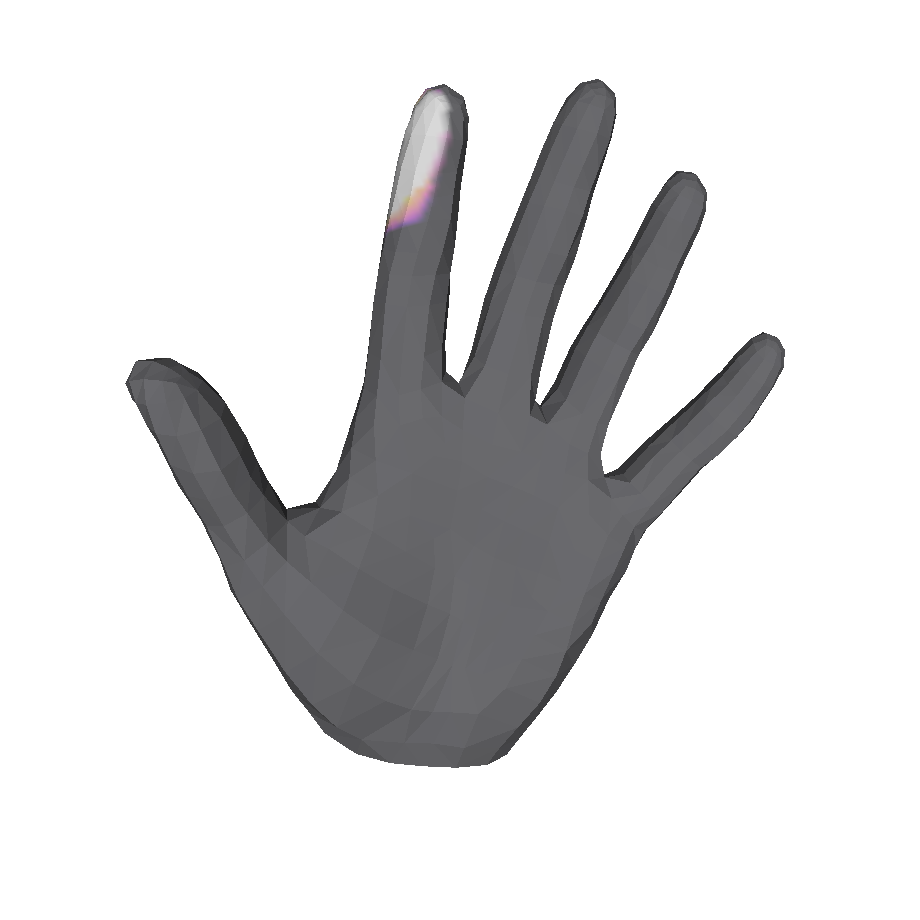} 
  
 \\
   \includegraphics[width=\sz\linewidth]{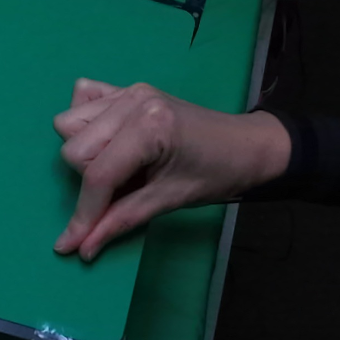} &
  \includegraphics[width=\sz\linewidth]{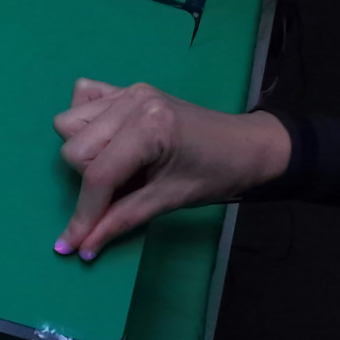} &
  \includegraphics[width=\sz\linewidth]{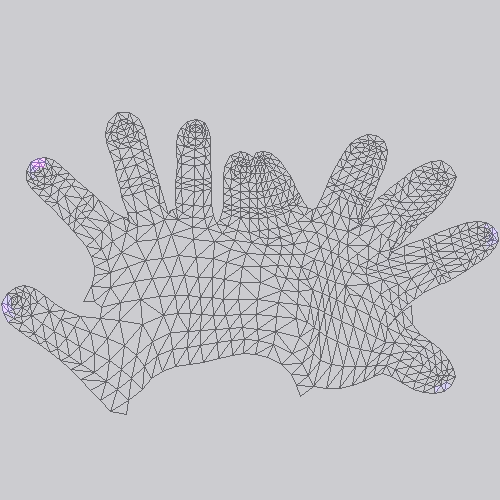} &
  \includegraphics[width=\sz\linewidth]{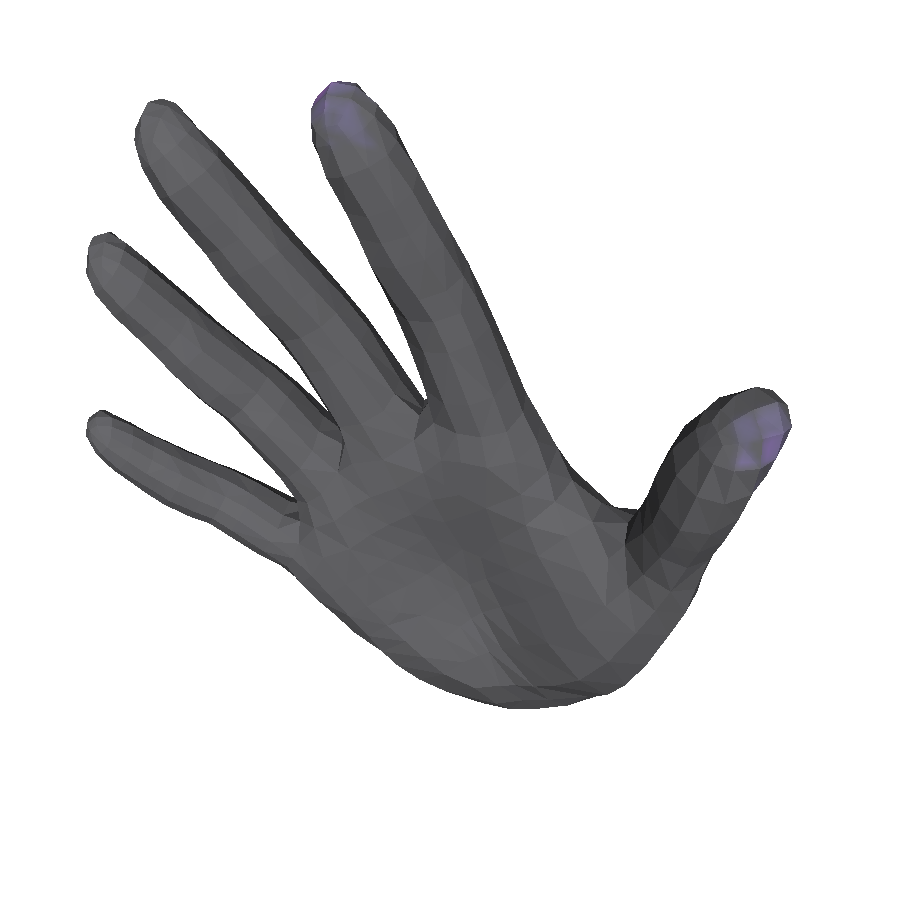} &
  \includegraphics[width=\sz\linewidth]{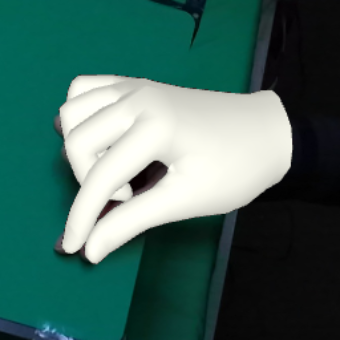} &
  
  \includegraphics[width=\sz\linewidth]{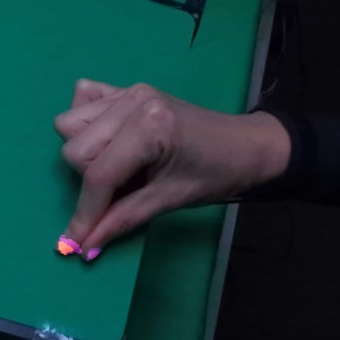} &
  \includegraphics[width=\sz\linewidth]{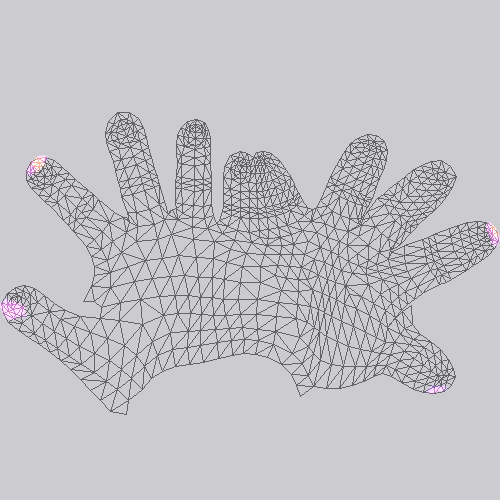} &
  \includegraphics[width=\sz\linewidth]{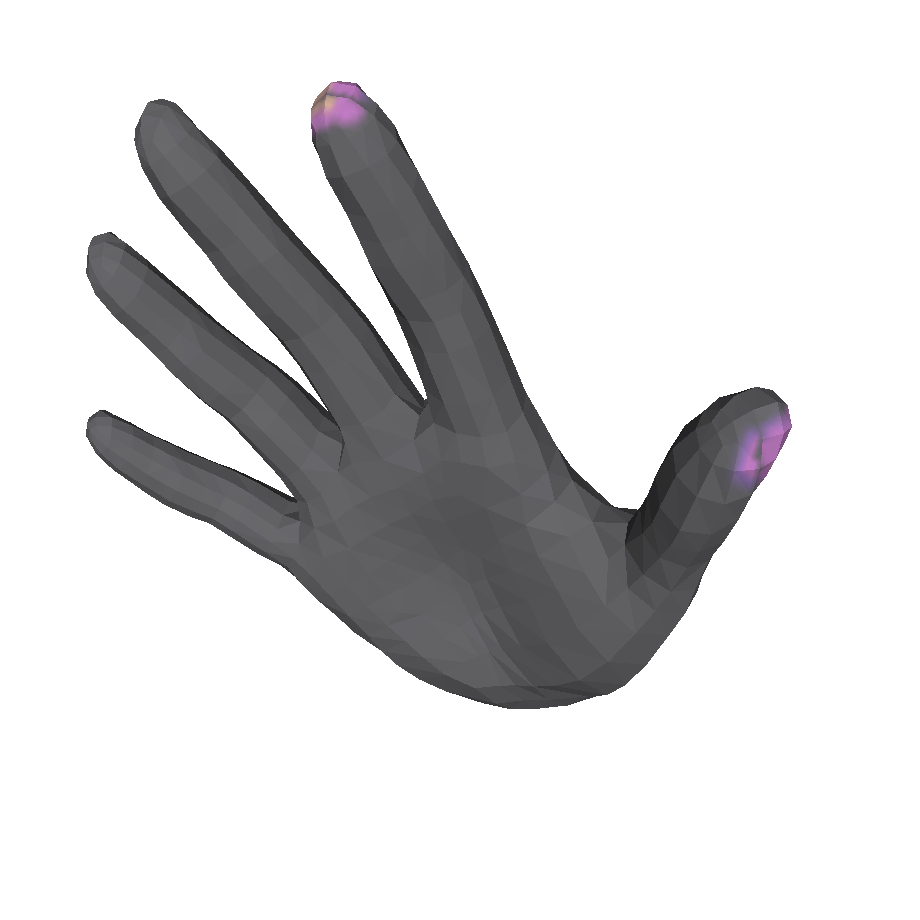} &

  \includegraphics[width=\sz\linewidth]{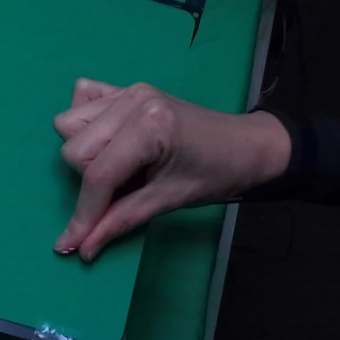} &
  \includegraphics[width=\sz\linewidth]{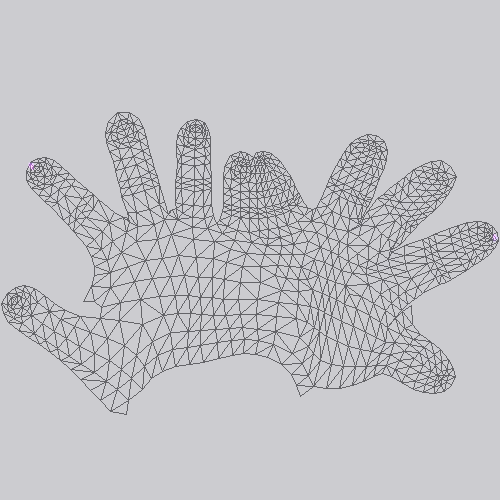} &
  \includegraphics[width=\sz\linewidth]{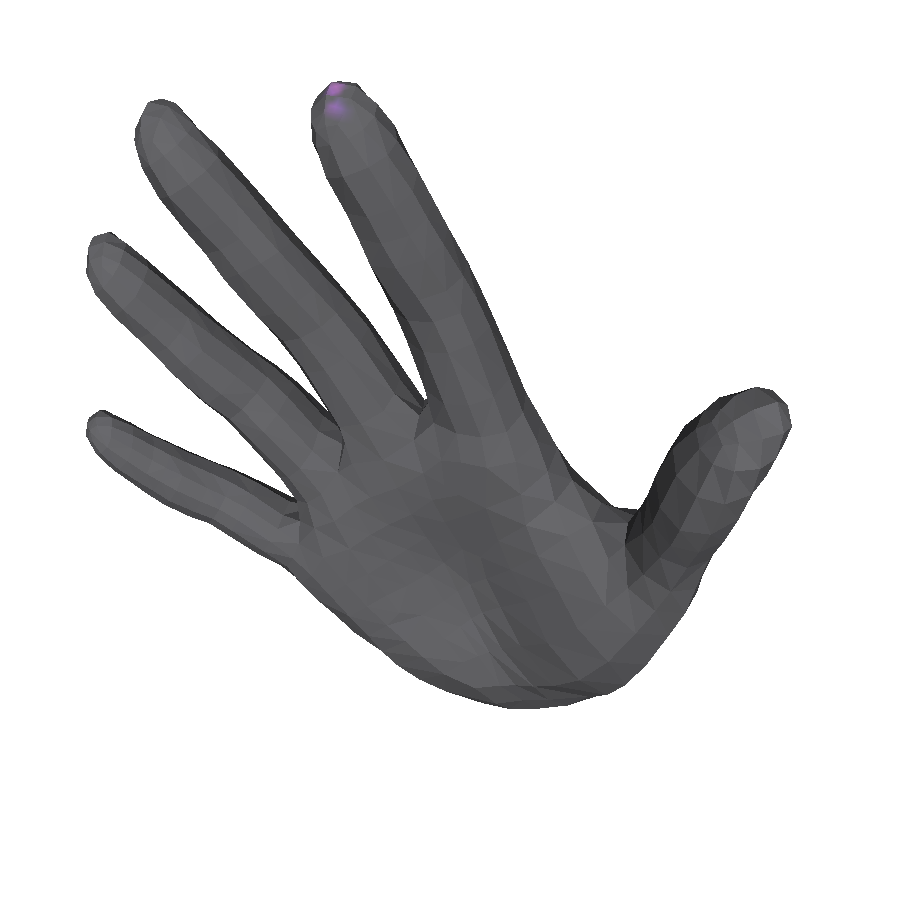} &
  
  \includegraphics[width=\sz\linewidth]{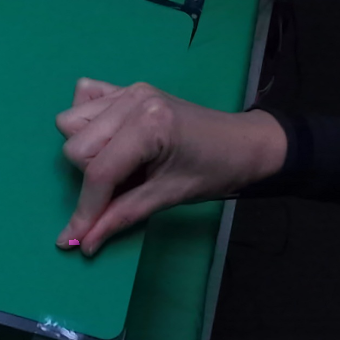} &
  \includegraphics[width=\sz\linewidth]{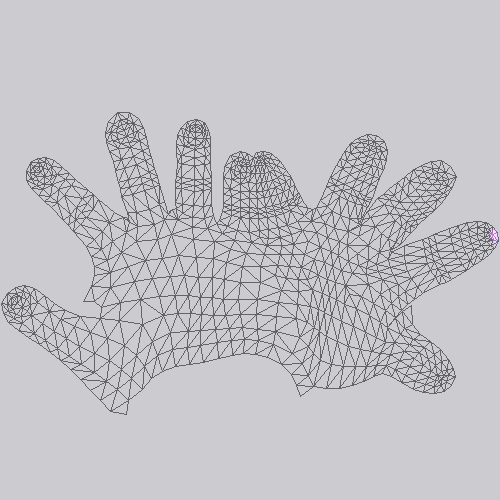} &
  \includegraphics[width=\sz\linewidth]{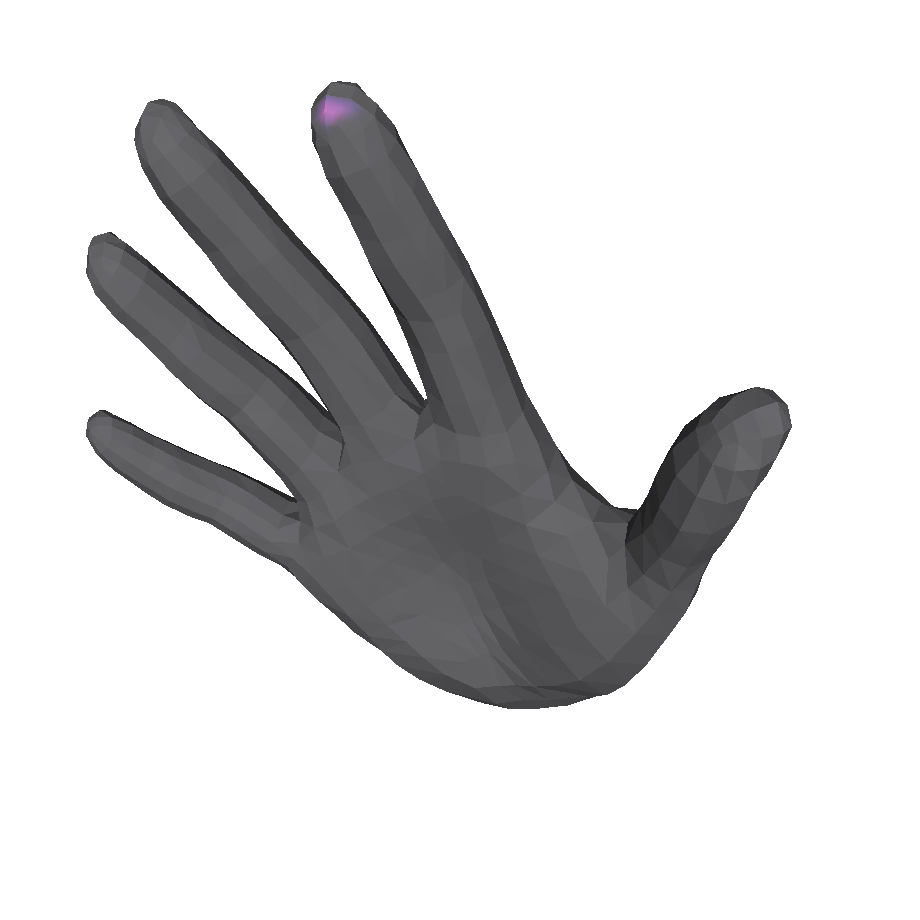} 
  
 \\
    \includegraphics[width=\sz\linewidth]{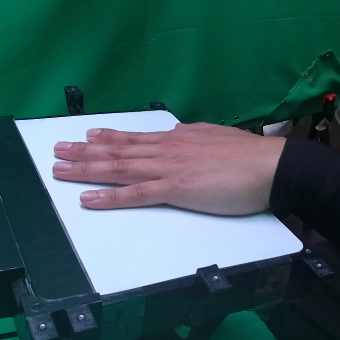} &
  \includegraphics[width=\sz\linewidth]{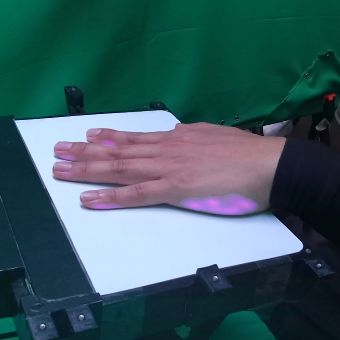} &
  \includegraphics[width=\sz\linewidth]{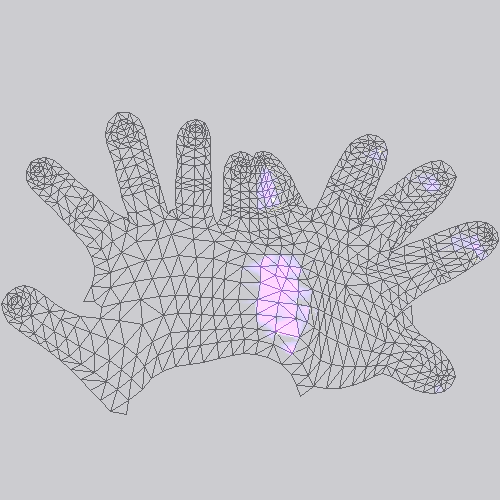} &
  \includegraphics[width=\sz\linewidth]{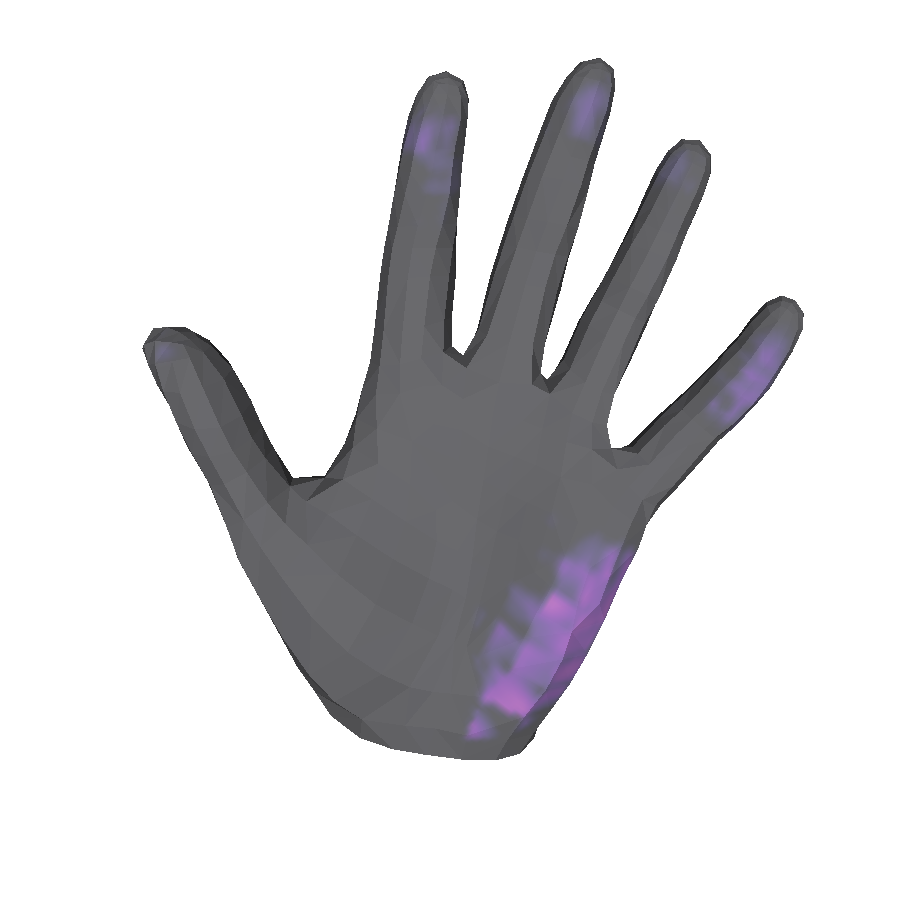} &
  \includegraphics[width=\sz\linewidth]{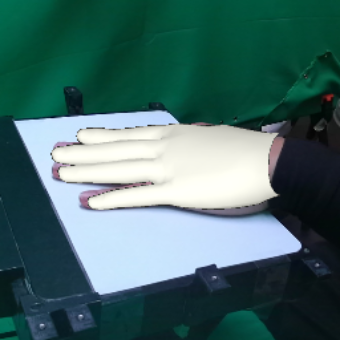} &
  
  \includegraphics[width=\sz\linewidth]{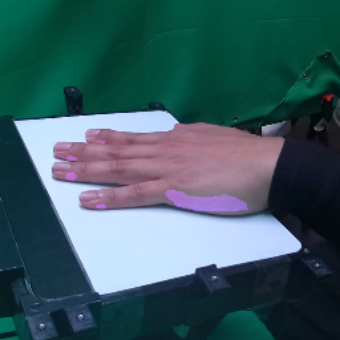} &
  \includegraphics[width=\sz\linewidth]{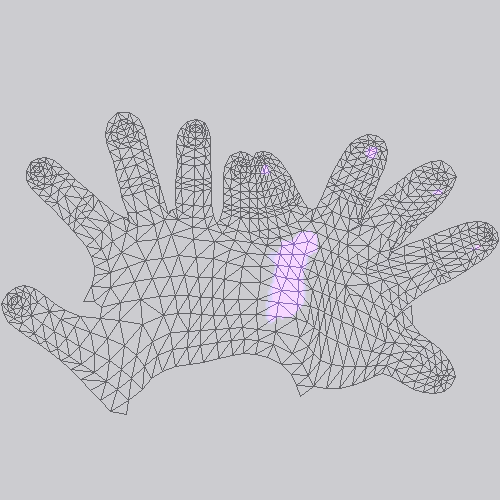} &
  \includegraphics[width=\sz\linewidth]{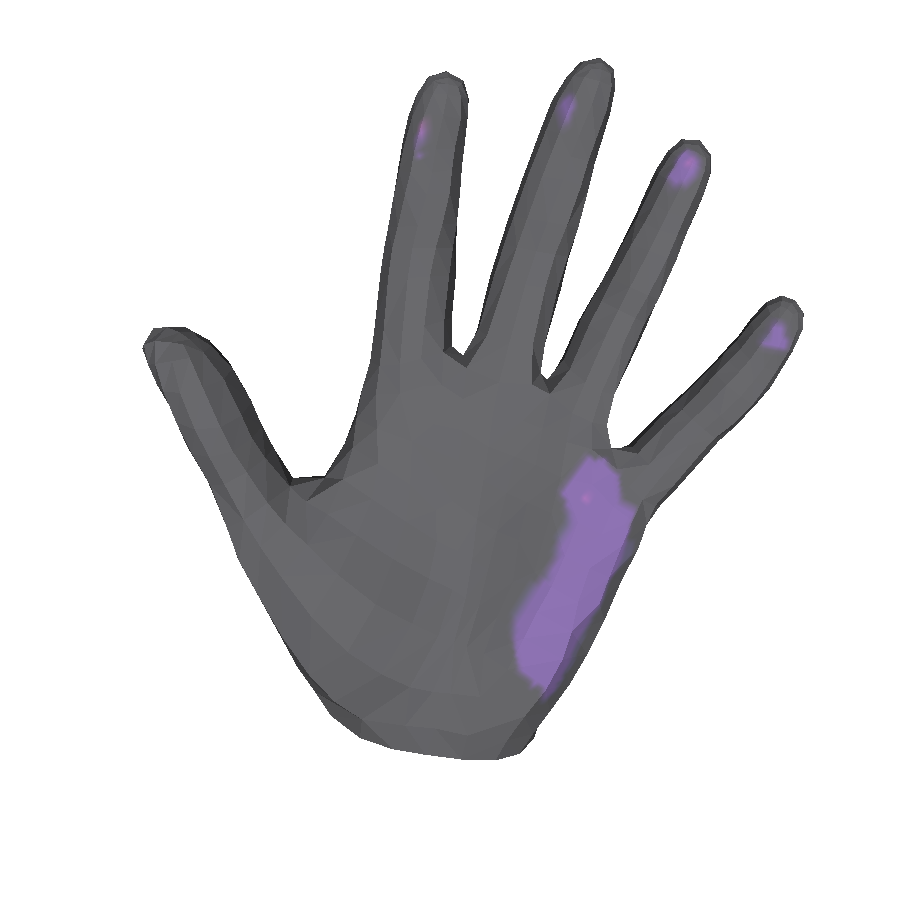} &

  \includegraphics[width=\sz\linewidth]{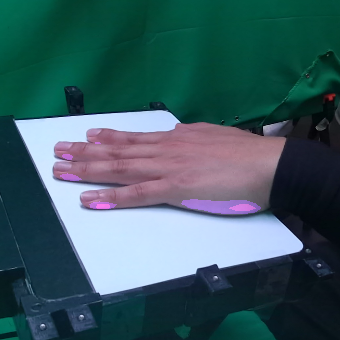} &
  \includegraphics[width=\sz\linewidth]{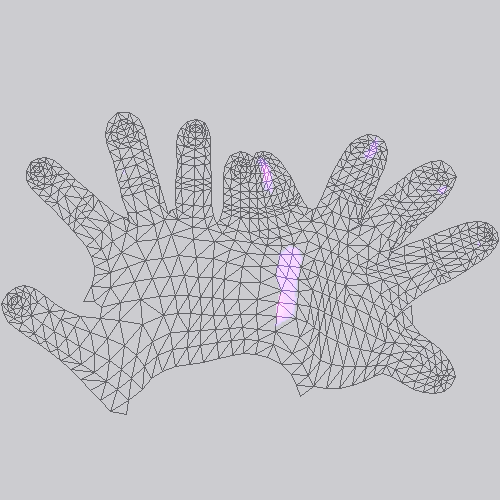} &
  \includegraphics[width=\sz\linewidth]{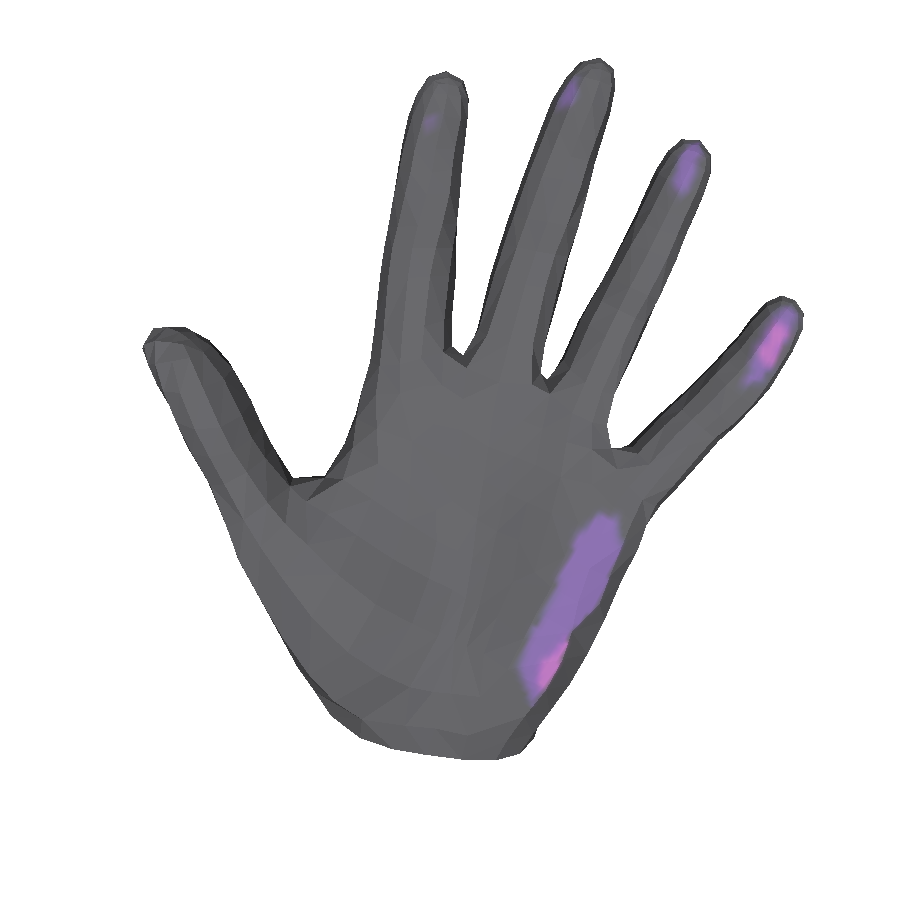} &
  
  \includegraphics[width=\sz\linewidth]{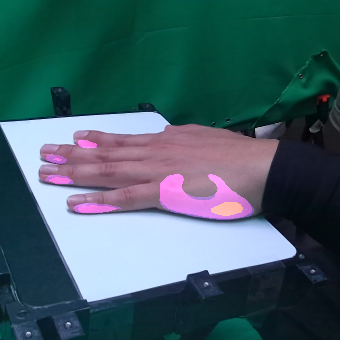} &
  \includegraphics[width=\sz\linewidth]{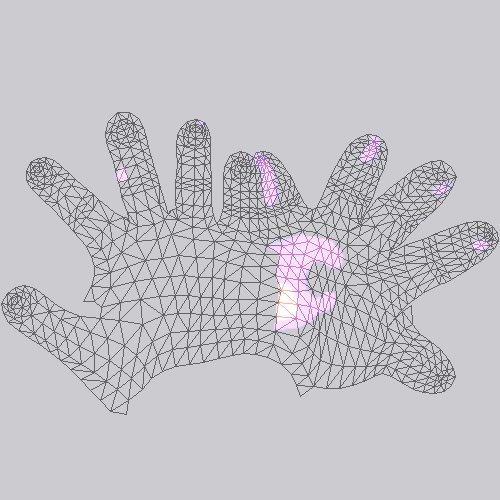} &
  \includegraphics[width=\sz\linewidth]{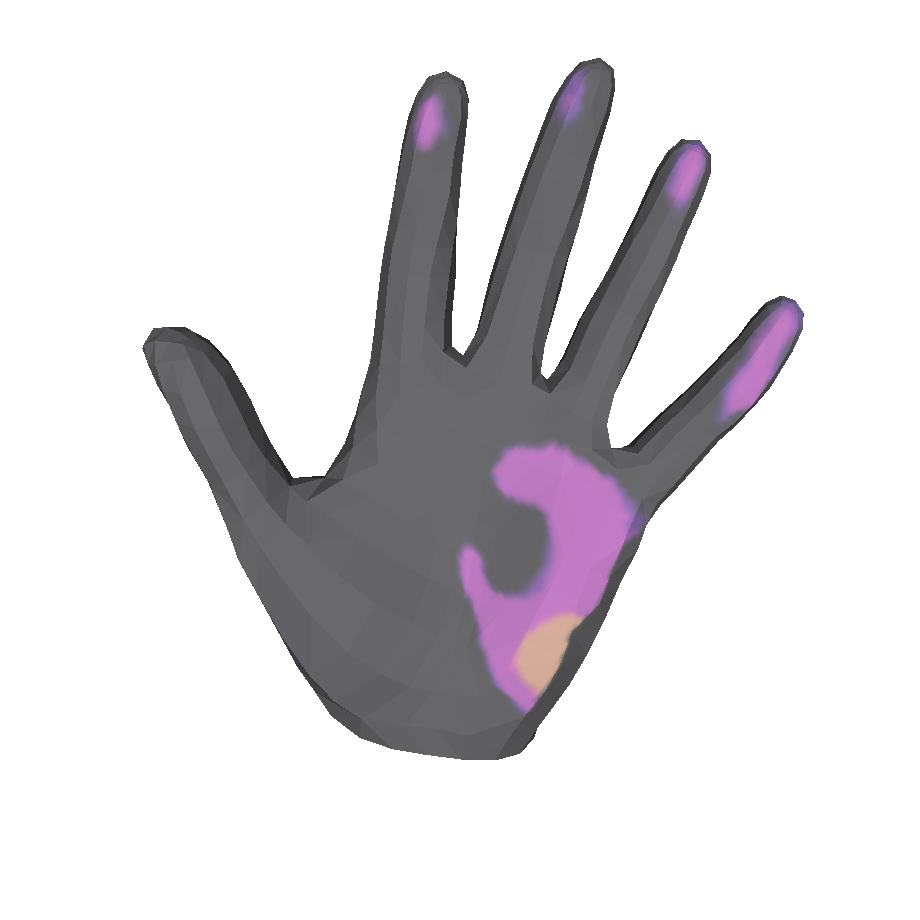} 
 \\
     \includegraphics[width=\sz\linewidth]{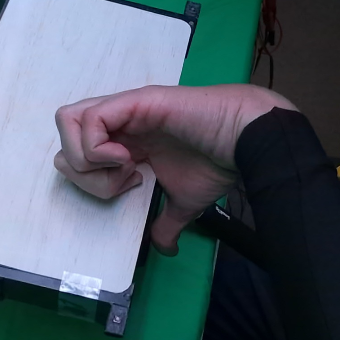} &
  \includegraphics[width=\sz\linewidth]{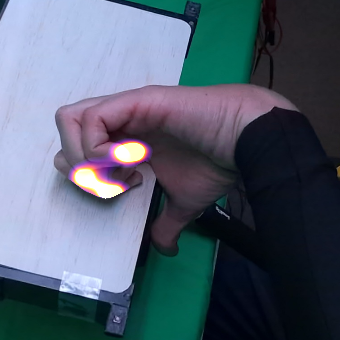} &
  \includegraphics[width=\sz\linewidth]{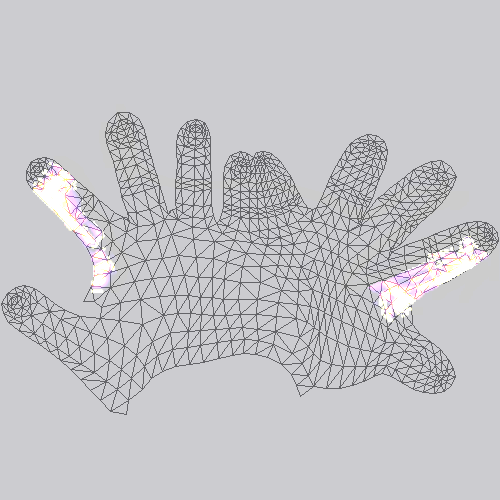} &
  \includegraphics[width=\sz\linewidth]{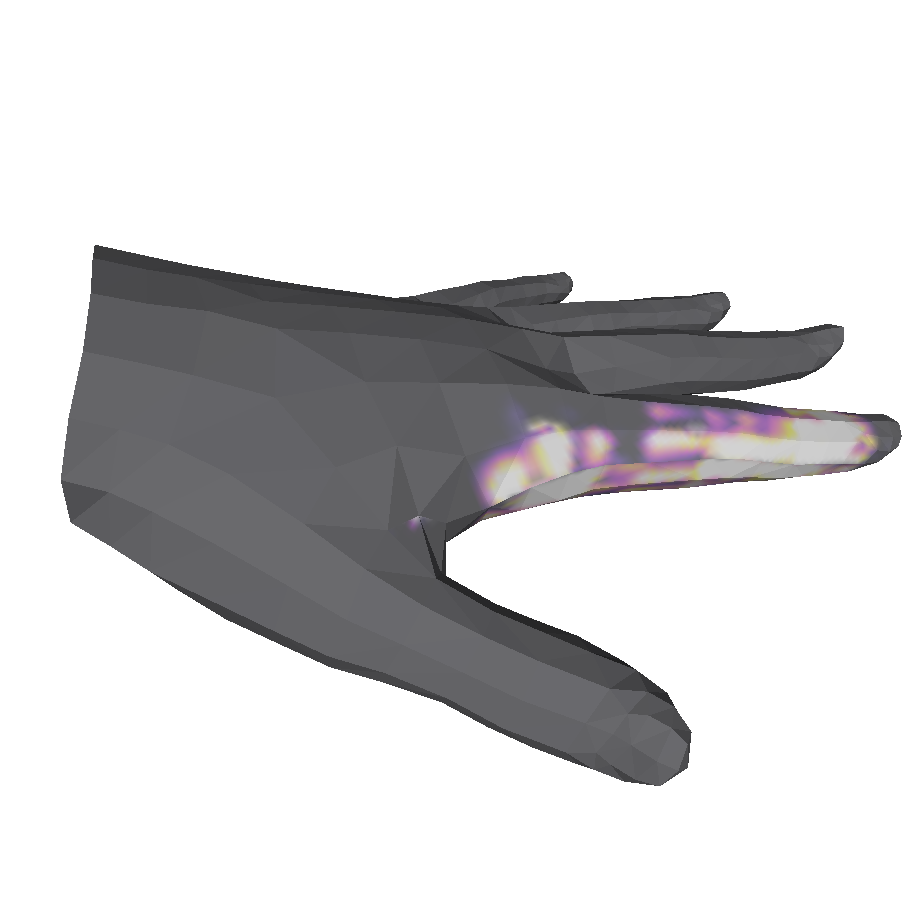} &
  \includegraphics[width=\sz\linewidth]{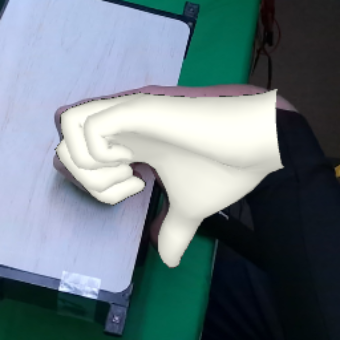} &
  
  \includegraphics[width=\sz\linewidth]{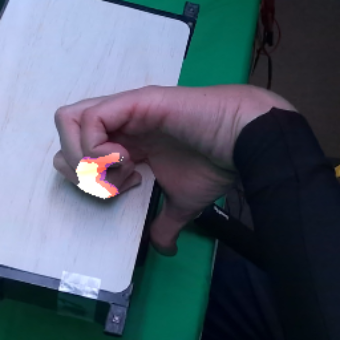} &
  \includegraphics[width=\sz\linewidth]{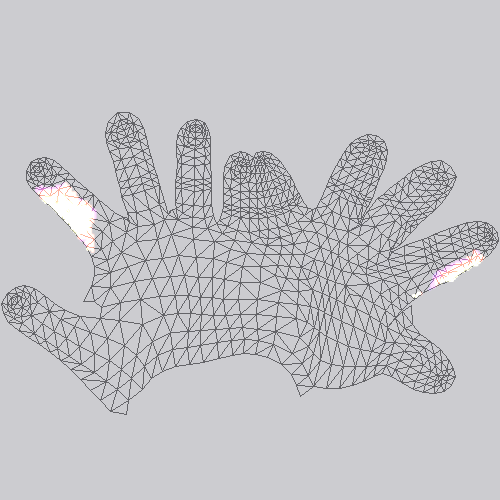} &
  \includegraphics[width=\sz\linewidth]{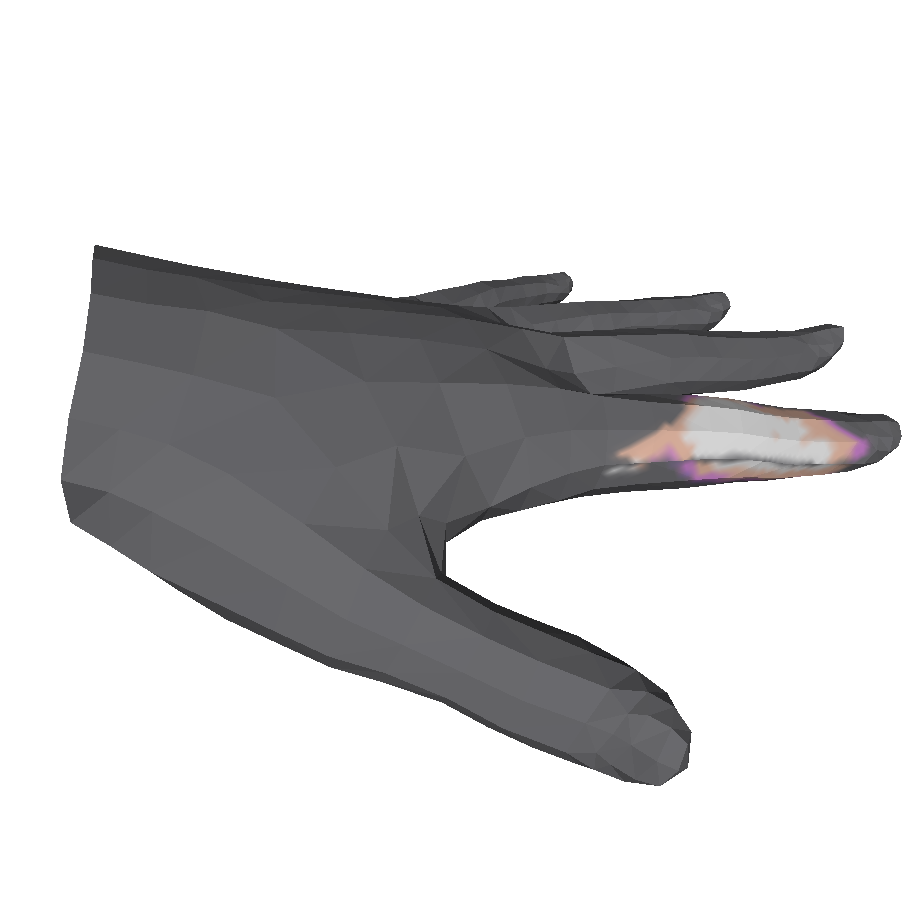} &

  \includegraphics[width=\sz\linewidth]{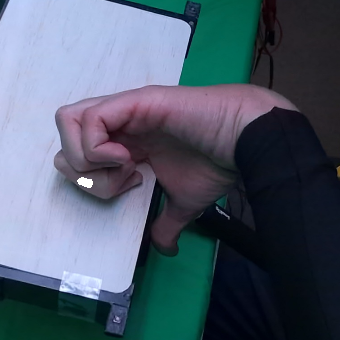} &
  \includegraphics[width=\sz\linewidth]{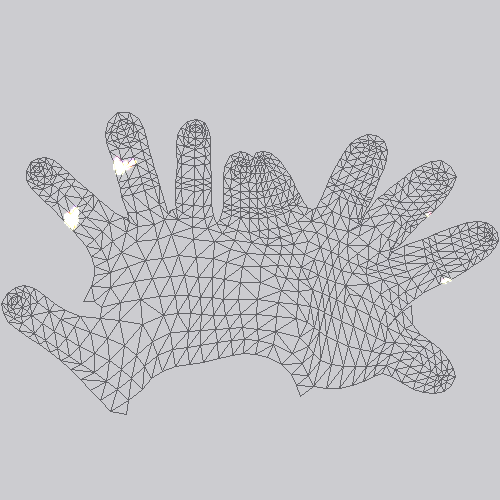} &
  \includegraphics[width=\sz\linewidth]{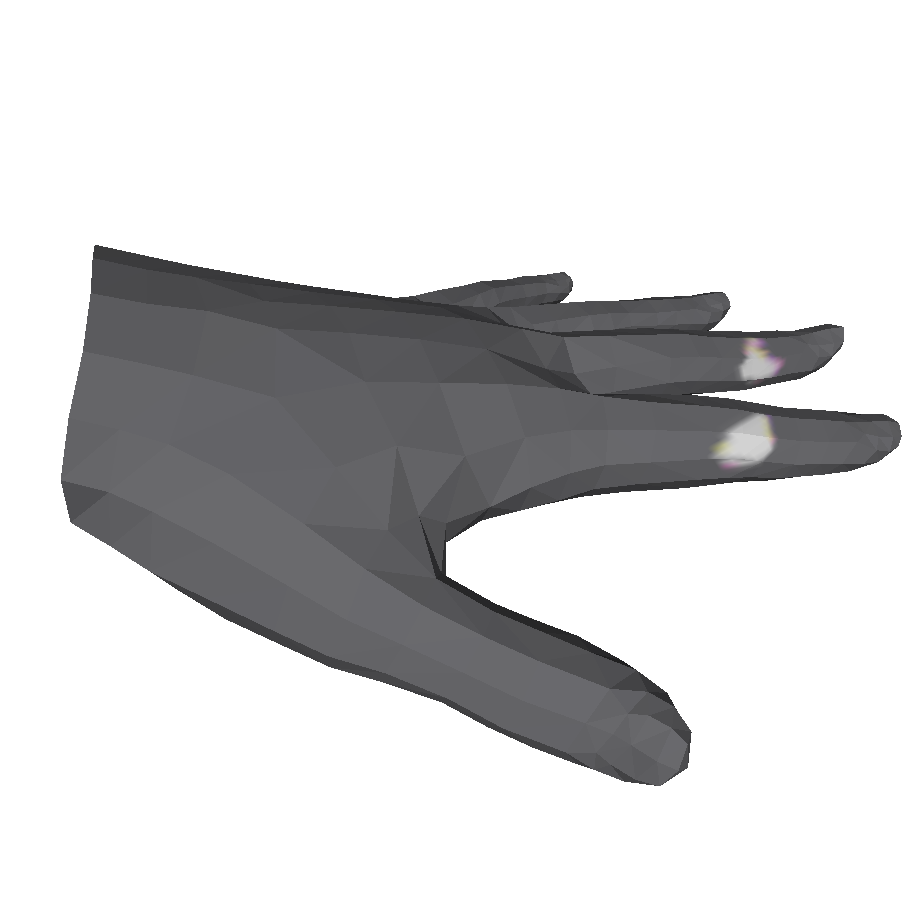} &
  
  \includegraphics[width=\sz\linewidth]{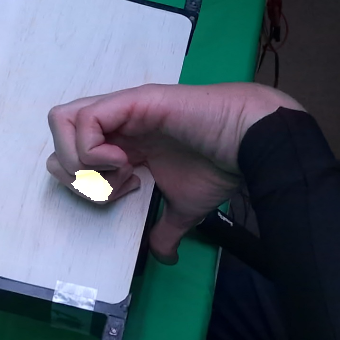} &
  \includegraphics[width=\sz\linewidth]{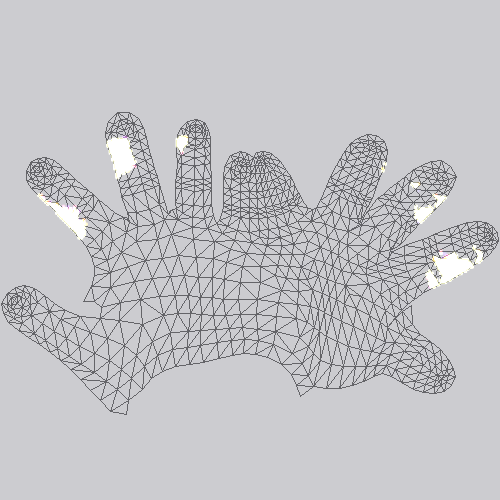} &
  \includegraphics[width=\sz\linewidth]{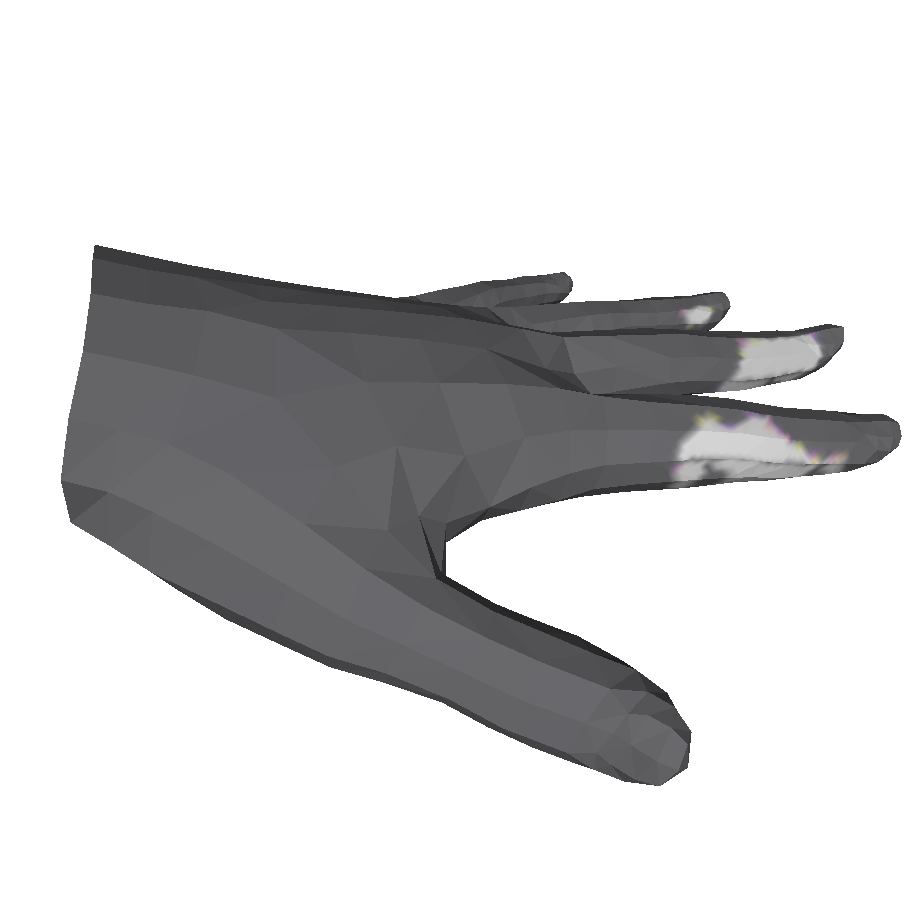} 
  \\
  Input &GT Pressure & GT UV & GT on Hand &  Mesh~\cite{hamer} & Pred. Pres. & Pred. UV &  On Hand
   &   Pred. Pres. & Pred. UV &  On Hand& Pred. Pres. & Pred. UV &  On Hand
  \\
     \multicolumn{5}{c}{}& \multicolumn{3}{c}{\scriptsize PressureFormer} &\multicolumn{3}{c}{\scriptsize PressureVision~\cite{grady2022pressurevision}} &\multicolumn{3}{c}{\scriptsize PressureVision~\cite{grady2022pressurevision}+ HaMeR~\cite{hamer}}

  \end{tabular}  
 \vspace{-2mm}
  \caption{\textbf{Qualitative comparison of UV Pressure.} We compare our PressureFormer model against the original PressureVision~\cite{grady2022pressurevision} and its extended version with additional hand keypoint inputs. For both PressureVision-based approaches, the UV pressure is obtained by baking the image-based pressure predictions onto the UV map of the hand mesh, using the hand mesh estimates provided by HaMeR~\cite{hamer}. }
  \label{fig:uv_rendering}
\vspace{-5pt}
\end{figure*} 

\subsubsection{Generalization of PressureFormer}

Employing a UV-pressure map can improve the generalization of hand contact and pressure prediction for more complex objects. Unlike estimating pressure on the image plane, which focuses on hand-surface interactions, UV-pressure mapping can highlight hand-centric pressure by directly predicting pressure on the hand vertices.

\begin{figure}[t]
  \centering
  \tiny
  \label{fig:uv_coarse_qualitative}
  \setlength{\tabcolsep}{0.1pt}
  \newcommand{\wsz}{0.18}
  \begin{tabular}{ccccc}

   \includegraphics[width=\wsz\linewidth]{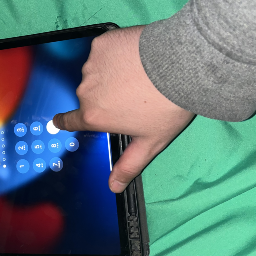} &
  \includegraphics[width=\wsz\linewidth]{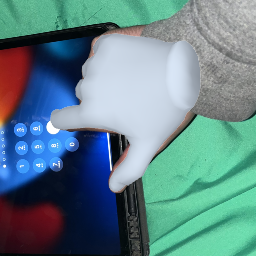} &
  \includegraphics[width=\wsz\linewidth]{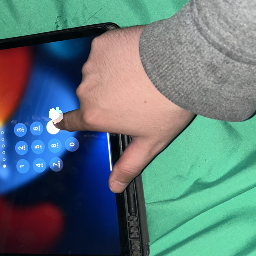} &
  \includegraphics[width=\wsz\linewidth]{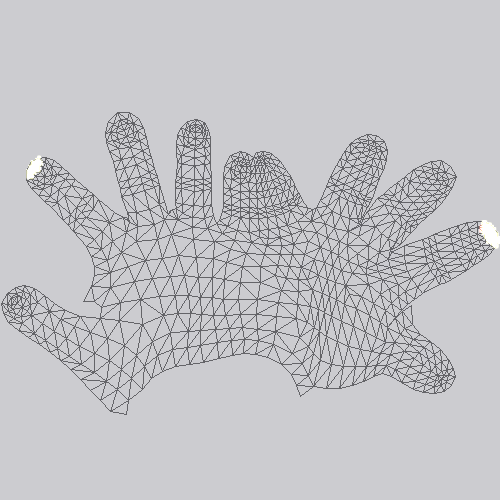} &
  \includegraphics[width=\wsz\linewidth]{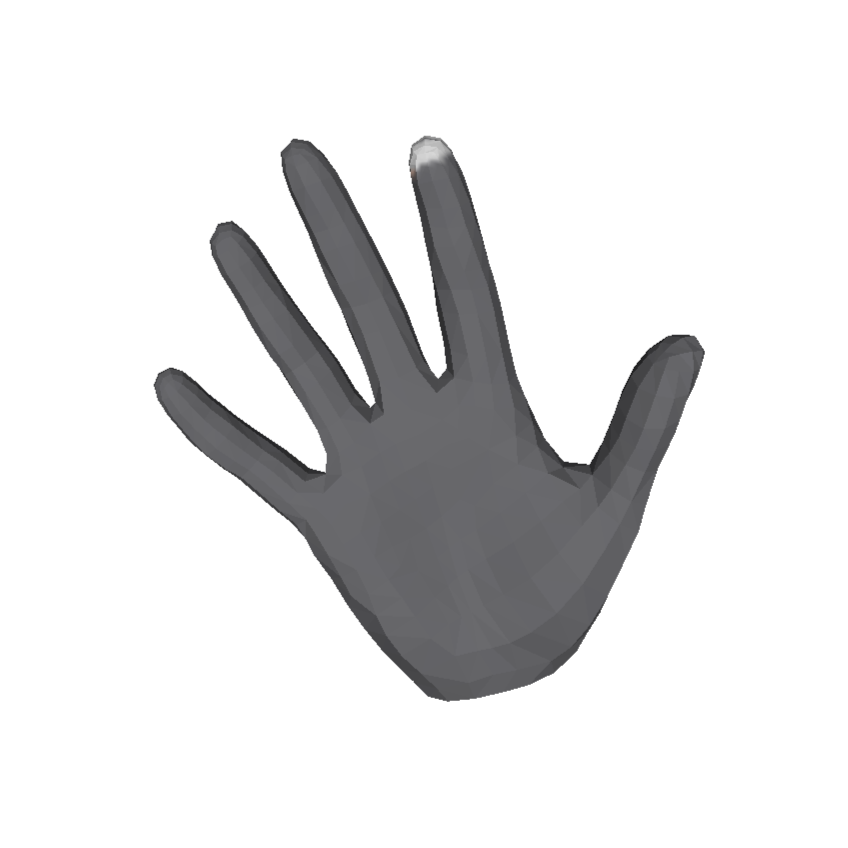} 
  \\
   \includegraphics[width=\wsz\linewidth]{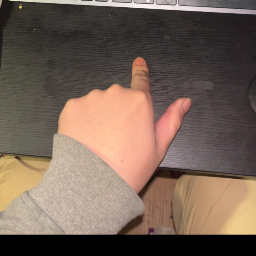} &
  \includegraphics[width=\wsz\linewidth]{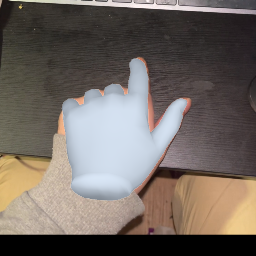} &
  \includegraphics[width=\wsz\linewidth]{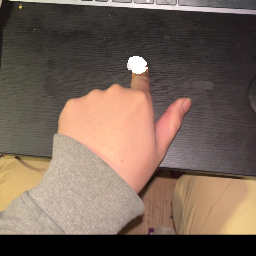} &
  \includegraphics[width=\wsz\linewidth]{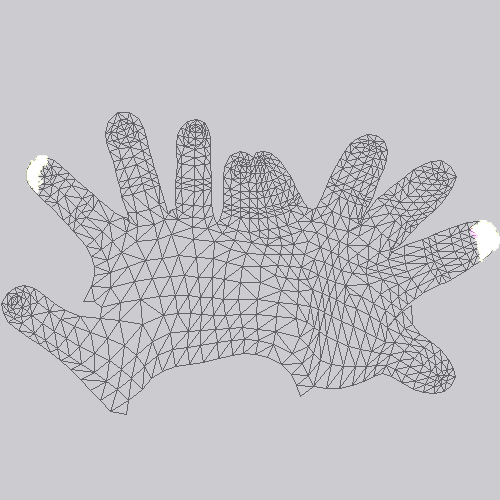} &
  \includegraphics[width=\wsz\linewidth]{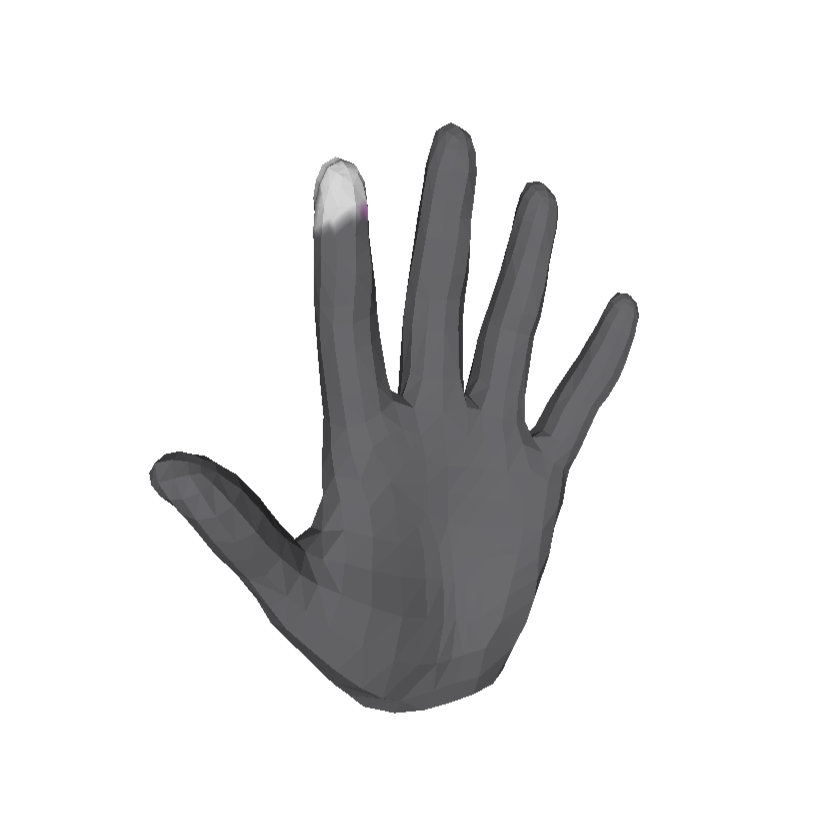} 
  \\
     \includegraphics[width=\wsz\linewidth]{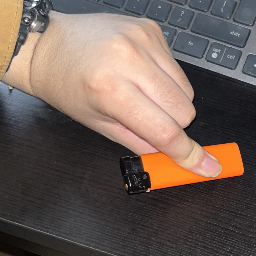} &
  \includegraphics[width=\wsz\linewidth]{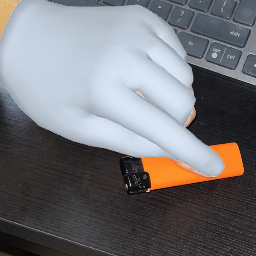} &
  \includegraphics[width=\wsz\linewidth]{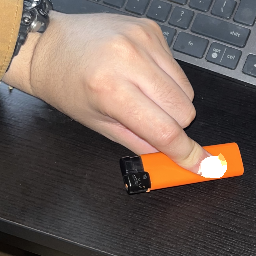} &
  \includegraphics[width=\wsz\linewidth]{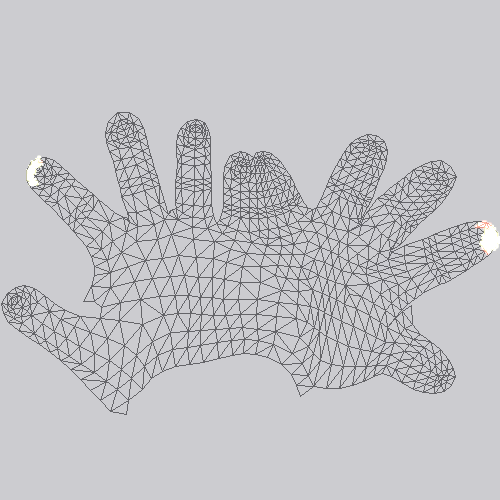} &
  \includegraphics[width=\wsz\linewidth]{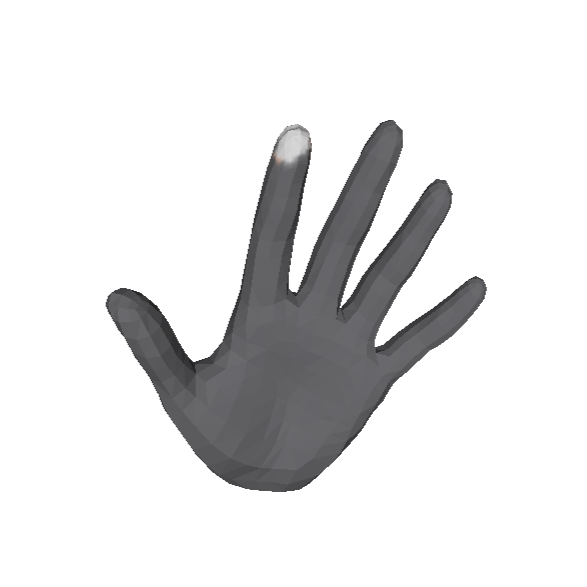} 
  \\
     \includegraphics[width=\wsz\linewidth]{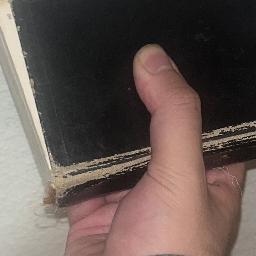} &
  \includegraphics[width=\wsz\linewidth]{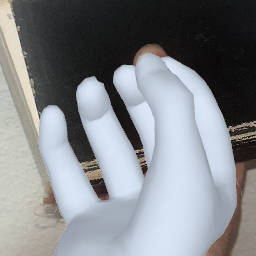} &
  \includegraphics[width=\wsz\linewidth]{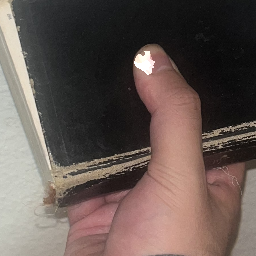} &
  \includegraphics[width=\wsz\linewidth]{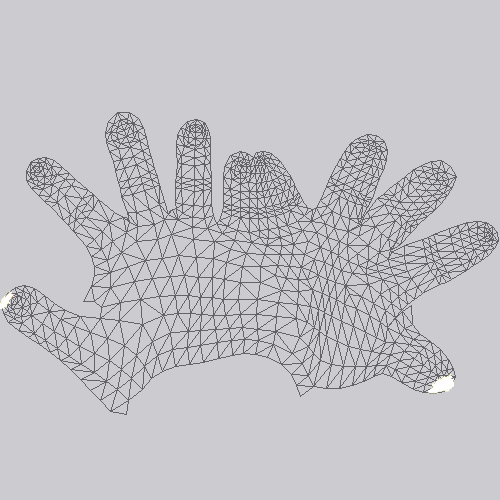} &
  \includegraphics[width=\wsz\linewidth]{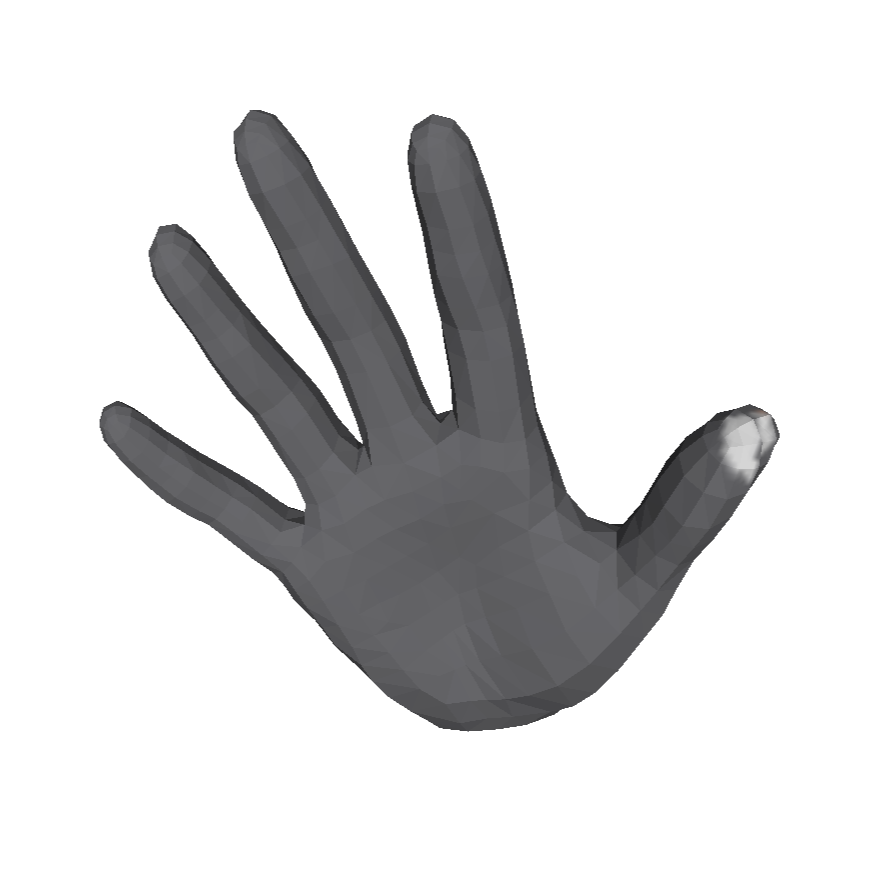} 
  \\
     \includegraphics[width=\wsz\linewidth]{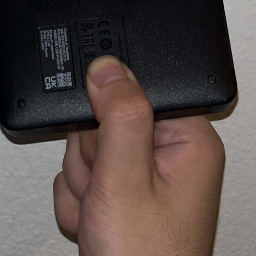} &
  \includegraphics[width=\wsz\linewidth]{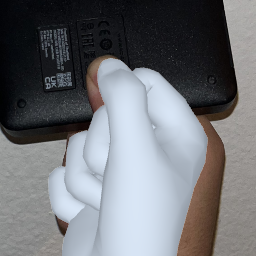} &
  \includegraphics[width=\wsz\linewidth]{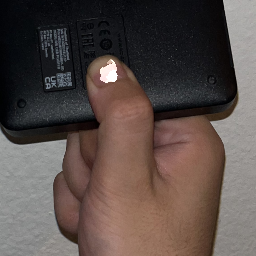} &
  \includegraphics[width=\wsz\linewidth]{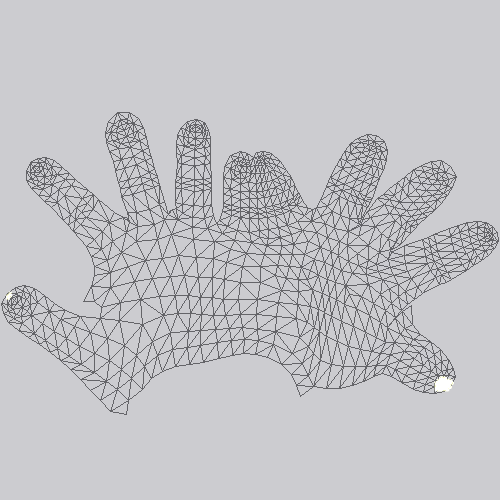} &
  \includegraphics[width=\wsz\linewidth]{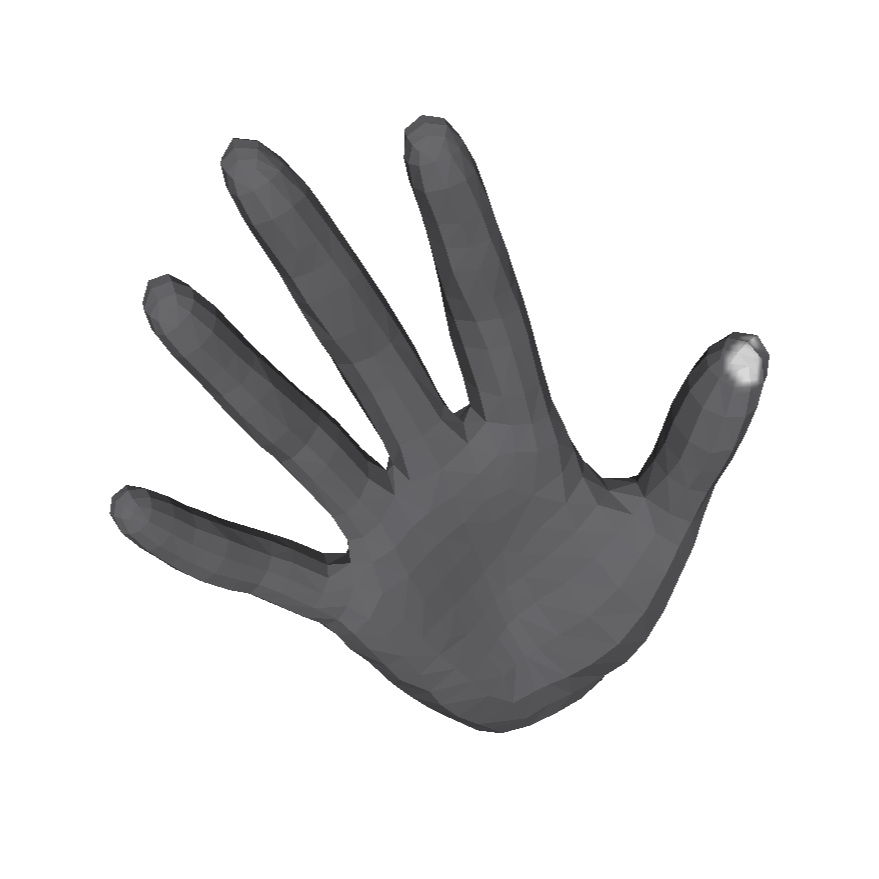} 
  \\
     \includegraphics[width=\wsz\linewidth]{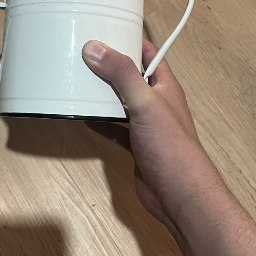} &
  \includegraphics[width=\wsz\linewidth]{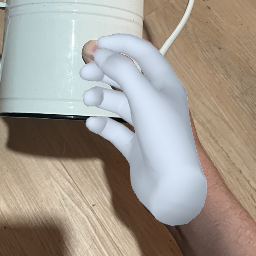} &
  \includegraphics[width=\wsz\linewidth]{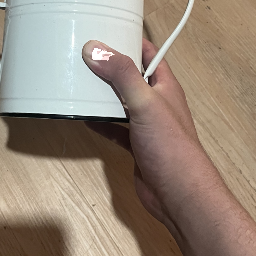} &
  \includegraphics[width=\wsz\linewidth]{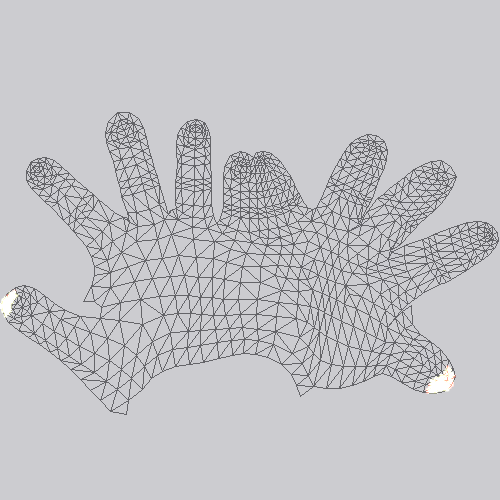} &
  \includegraphics[width=\wsz\linewidth]{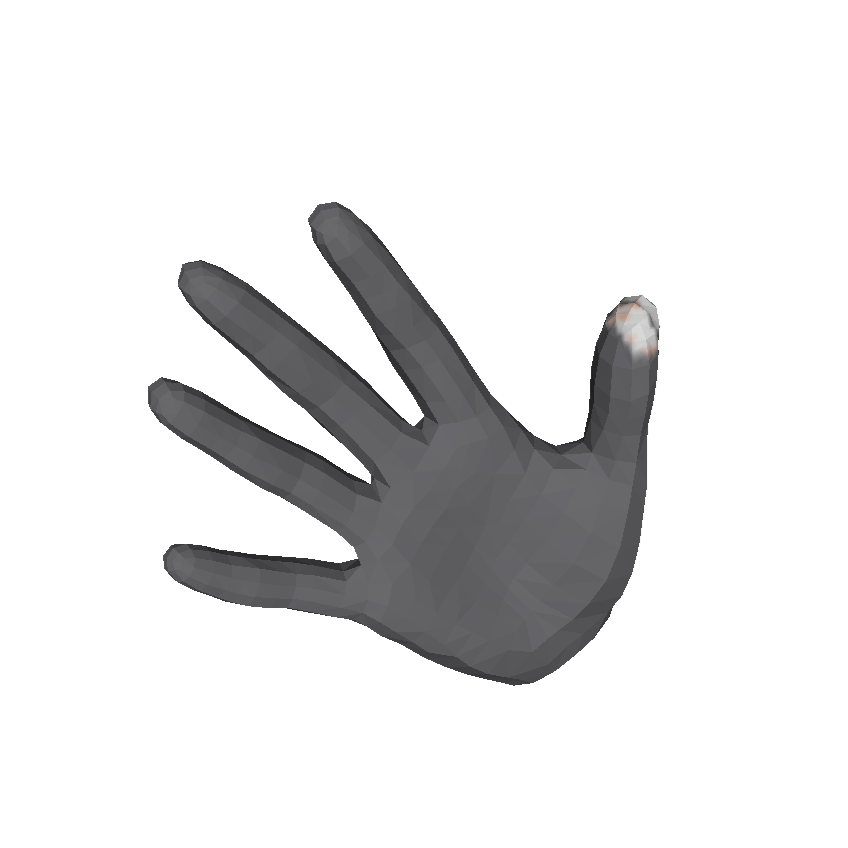} 
  \\
     \includegraphics[width=\wsz\linewidth]{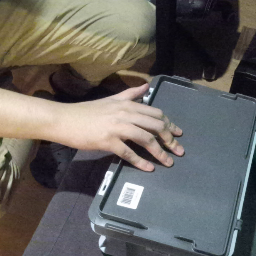} &
  \includegraphics[width=\wsz\linewidth]{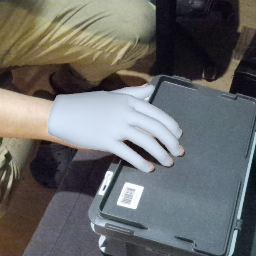} &
  \includegraphics[width=\wsz\linewidth]{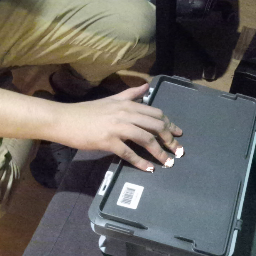} &
  \includegraphics[width=\wsz\linewidth]{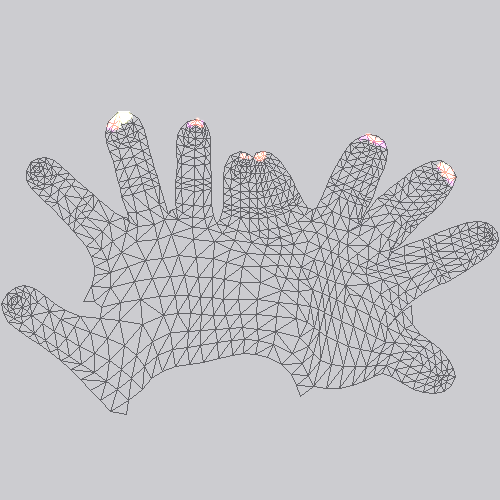} &
  \includegraphics[width=\wsz\linewidth]{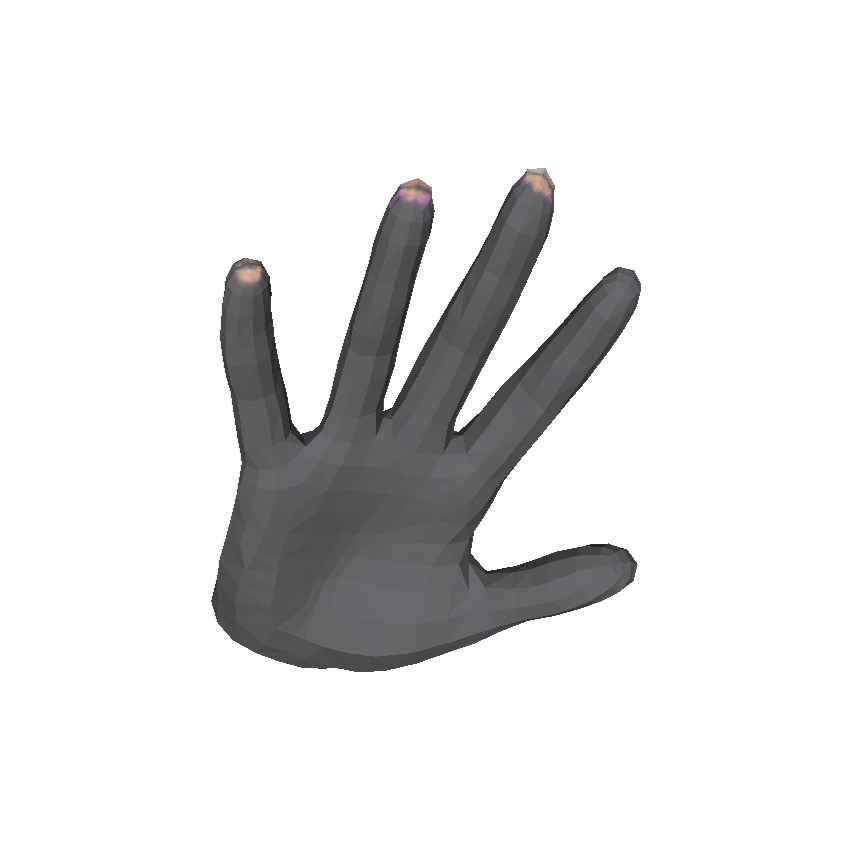} 
  \\
     \includegraphics[width=\wsz\linewidth]{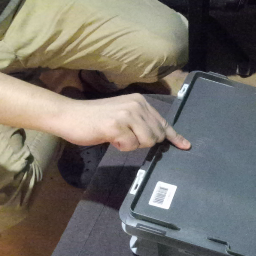} &
  \includegraphics[width=\wsz\linewidth]{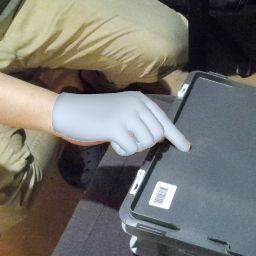} &
  \includegraphics[width=\wsz\linewidth]{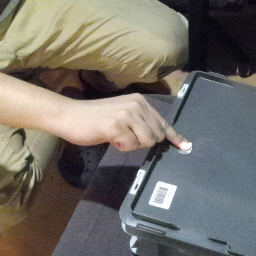} &
  \includegraphics[width=\wsz\linewidth]{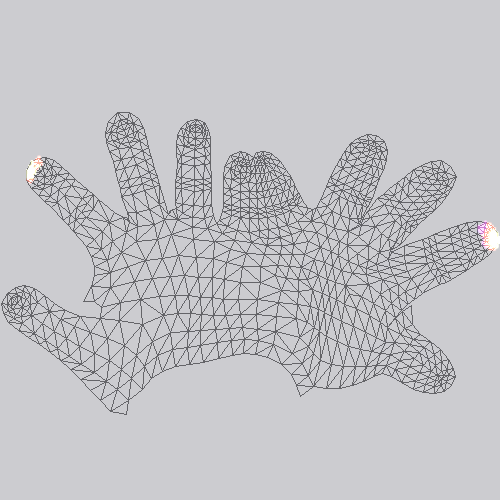} &
  \includegraphics[width=\wsz\linewidth]{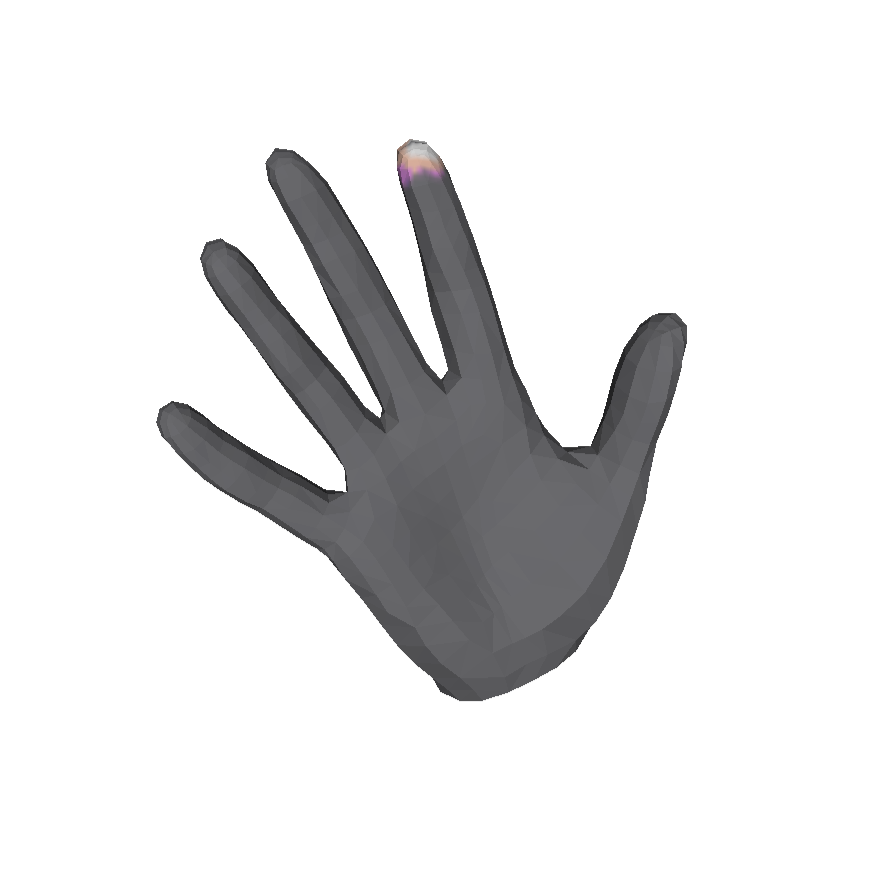} 
  \\
     \includegraphics[width=\wsz\linewidth]{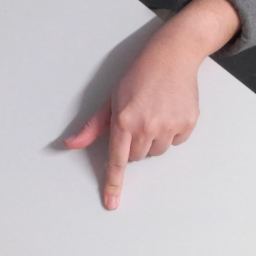} &
  \includegraphics[width=\wsz\linewidth]{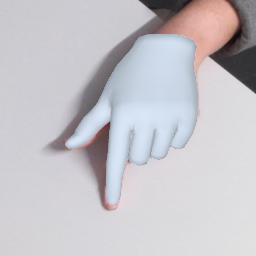} &
  \includegraphics[width=\wsz\linewidth]{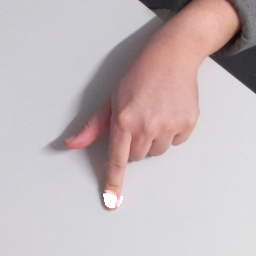} &
  \includegraphics[width=\wsz\linewidth]{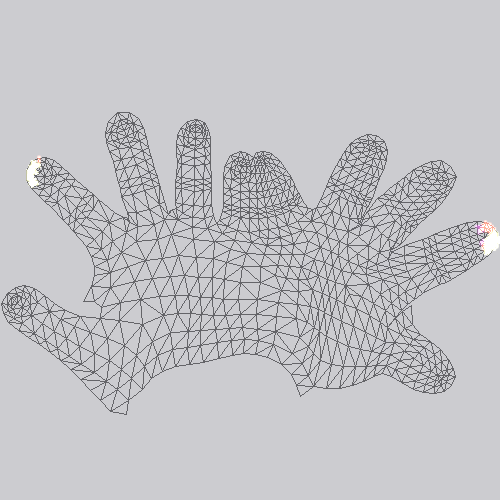} &
  \includegraphics[width=\wsz\linewidth]{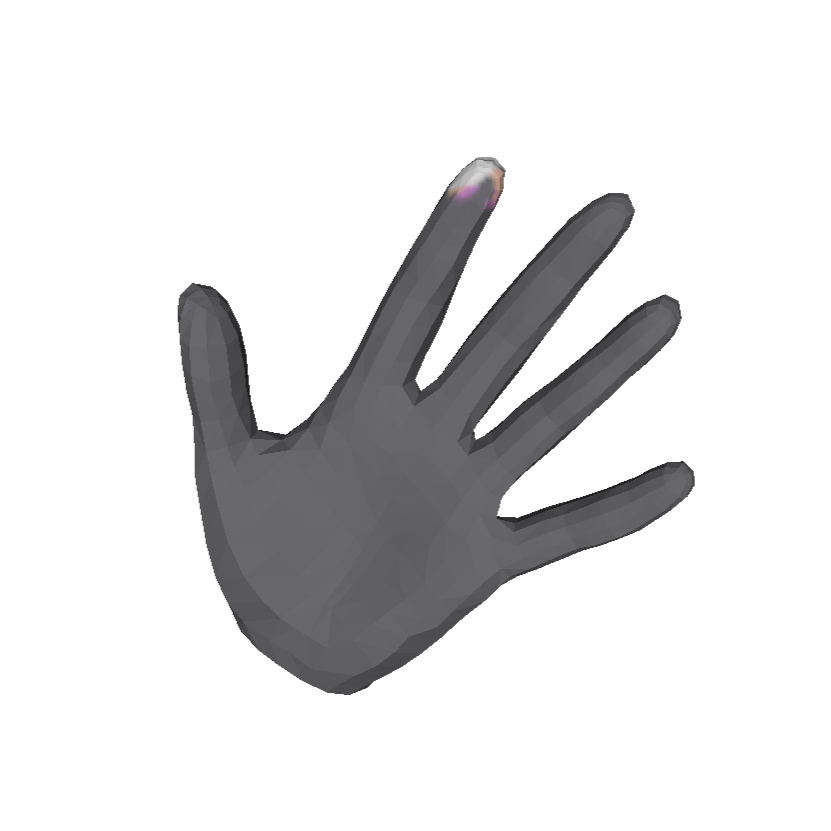} 
  \\

  Input & Mesh~\cite{hamer} & Pred. Pres. & Pred. UV &  On Hand
  \\ \cline{3-5}
     \multicolumn{2}{c}{}& \multicolumn{3}{c}{PressureFormer}

  \end{tabular}  
 \vspace{-2mm}
  \caption{\textbf{Qualitative evaluation of PressureFormer on diverse, real-world examples featuring various objects and scenes.} Despite being trained exclusively on EgoPressure, the model recognizes pressure regions during corresponding contact events,  demonstrating its potential for generalization.}
\label{fig:wild}
\vspace{-5pt}
\end{figure}

Our model, PressureFormer, utilizes the pretrained HaMeR~\cite{hamer} as its backbone to extract hand vertices and image features from the vision transformer tokens. 
This enables our approach to effectively handle diverse hand poses while integrating hand-centric image texture information encoded in the vision transformer tokens.
The qualitative results, shown in Figure~\ref{fig:wild}, demonstrate PressureFormer's ability to generalize to unseen objects and environments.

\begin{figure*}[htbp]
    \normalsize
\begin{equation}
 \lossren(\mesh) = \sum_{i = 0}^{C} [ \lambda_{M}(\underbrace{1-\mathrm{IoU}(\rens^i(\mesh),\maskgt^i)}_{\textbf{Mask IoU Loss}~ \mathcal{L}_M(\mesh)} )+ \lambda_{A} \underbrace{\mathrm{MSE}(\renf^i(\mesh, \texture), \imgt^i)}_{\textbf{Appearance Loss}~ \mathcal{L}_A(\mesh)} + \lambda_{D} \underbrace{(1 - \frac{| \min(\rend^i(\mesh), \depthgt^i) |_1}{| \max(\rend^i(\mesh), \depthgt^i) |_1})}_{\textbf{Depth Volumetric IoU Loss}~ \mathcal{L}_D(\mesh)}].
\label{eq:render_loss}
    \end{equation}
\end{figure*}

\section{Details and Evaluation of Annotation Method}

\subsection{Optimization Objectives}
In this section, we describe the optimization objectives necessary for complete implementation in conjunction with the objectives described in the main paper.

\subsubsection{Render Objective}
Since hand mesh \(\mesh\) is the only rendered object across all camera views, we use pseudo groundtruth mask \(\maskgt^i\) from Segment-Anything (SAM)~\cite{sam} to extract relevant regions, appearance \(\imgt^i = \bm{I}^i_{\text{in}} \otimes \maskgt^i\) and depth \(\depthgt^i = \bm{D}_{\text{in}}^i \otimes \maskgt^i\), from input RGB image \(\bm{I}_{\text{in}}^i\) and depth \(\bm{D}_{\text{in}}^i\). 
For the optimization of the rendered appearance \(\renf^i(\mesh, \texture)\), a single texture is shared across all camera views within an input batch of several consecutive frames, which ensures that the mesh \(\mesh\) remains consistent across different cameras and consecutive frames. 
The rendering loss \(\mathcal{L}_\mathcal{R}\) across all $C$ cameras is represented in Eq.~\ref{eq:render_loss}

\textbf{Depth Volumetric IoU}~$\mathcal{L}_D(\mesh)$~\cite{grady2022pressurevision} is defined in the third term of Equation~\ref{eq:render_loss}. We apply it to the ground truth and rendered depth. 
In Table~\ref{dataset_metrics}, we show these two losses: \textbf{Depth Volumetric IoU Loss}~$\mathcal{L}_D(\mesh)$ and \textbf{Mask IoU Loss}~$\mathcal{L}_M(\mesh)$ on the mesh from the initial input, i.e., $\pose_{ini}$ and $\tran_{ini}$, and two consecutive annotation stages, \poseoptimization{} and \shaperefinement{}.
\begin{table}[ht]
\begin{center}

\caption{ \textbf{Losses by Stages.} We validate the quality of hand poses using two metrics, Depth Volumetric IoU Loss~$\mathcal{L}_D$ (Eq.~\ref{eq:render_loss}) and Mask IoU Loss~$\mathcal{L}_M$ (Eq.~\ref{eq:render_loss}), computed on 386,231 $\times$7 (static cameras) = 2,703,617 annotated frames. Of these, 2,192,633 (81\%) show the hand in contact with the touchpad. We report the results before (initial) and after each consecutive optimization step: \poseoptimization{} and \shaperefinement{}.}
\begin{adjustbox}{width=0.98\columnwidth,center}

\begin{tabular}{c|ccc|ccc}
\toprule
\hline
\multirow{2}{*}{\textbf{Category}} & \multicolumn{3}{c|}{$\mathcal{L}_D~\downarrow$} & \multicolumn{3}{c}{$\mathcal{L}_M~\downarrow$} \\ 
 & \textbf{Initial} & \textbf{Pose.} & \textbf{Pose. + Shape.} & \textbf{Initial} & \textbf{Pose.} & \textbf{Pose. + Shape.} \\ \hline
Overall         & 0.4443 & 0.1759 & 0.1317 & 0.3887 & 0.1165 & 0.0558 \\ \hline
With Contact    & 0.4444 & 0.1752 & 0.1309 & 0.3891 & 0.1167 & 0.0562 \\ \hline
Without Contact & 0.4441 & 0.1790 & 0.1351 & 0.3871 & 0.1158 & 0.0545 \\ \hline
\bottomrule
\end{tabular}
\label{dataset_metrics}
\end{adjustbox}
\end{center}

\vspace{-1em}
\end{table}

\textbf{}\subsubsection{Geometry Objective}
\label{sec:geo_obj}
The geometry objective (\(\lossgeo\)) is composed of several terms:

\begin{equation}
\lossgeo = \mathcal{L}_{\text{insec}} + \mathcal{L}_{\text{arap}} + \mathcal{L}_{\vec{\textbf{n}}} + \mathcal{L}_{\text{lap}} + \mathcal{L}_{\text{offset}}
\label{eq:geo_loss}
\end{equation}

The term \(\mathcal{L}_{\text{insec}}\) represents the mesh intersection loss, which utilizes a BVH tree to identify self-intersections within the mesh. Penalties are subsequently applied based on these detections~\cite{Karras:2012:MPC:2383795.2383801, Tzionas:IJCV:2016}.

The term \(\mathcal{L}_{\text{arap}}\), as-rigid-as-possible loss, as introduced in \cite{ARAP_modeling:2007}, promotes increased rigidity in the 3D mesh while distributing length alterations across multiple edges. The variation in edge length is determined relative to the mesh from the last epoch of \poseoptimization{} as

\begin{equation}
\mathcal{L}_{\text{arap}} = \frac{1}{|\mathit{E}|} \sum_{v^* \in \mesh^*} \sum_{\bm{e}^* \in \mathit{E}(v^*, u^*)} | \|\bm{e}^*\| - \|\bm{e}^p\| |,
\label{eq:arap}
\end{equation}

where \(\mathit{E}(v^*, u^*)\) is the edge connecting vertex \(v^*\) and \(u^*\) in the set of all edges \(\mathit{E}\), and the edge \(\bm{e}^p\) is formed by the corresponding vertices \(v^p\) and \(u^p\) in the mesh without vertex displacement \(\vdisp\).

The mesh vertices \(\mathbf{V_{\mesh^*}}\) are smoothed by the Laplacian mesh regularization \(\mathcal{L}_{\text{lap}}\)~\cite{laplacian}, and the normal consistency regularization \(\mathcal{L}_{\vec{\textbf{n}}}\) smooths normals on the displaced mesh. Finally, the vertex offset term \(\mathcal{L}_{\text{offset}}\) is calculated by \(\|\vdisp\|^2\).

\subsubsection{Depth Culling}
\begin{figure}[tb]
\centering
   \includegraphics[width=0.5\textwidth]{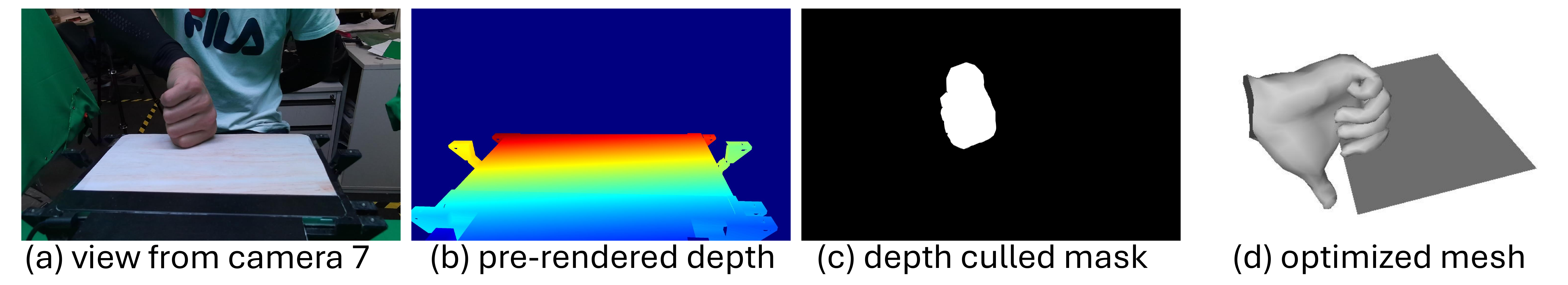}
   \caption{\textbf{Depth Culling.} (a) In the view of Camera 7, the thumb is behind the touchpad. (b) We compare the rendered depth of hand ~$\mathcal{R}_D$ and pre-rendered depth map of scene obstacles \(D_o\), and (c) cutout the part which has a larger depth value than \(D_o\). 
   The thumb rendered in blue color is cutout due to the depth culling. (d) The collision box is rendered in 3D.}
   \label{fig:depth_culling}
\end{figure}

In some sequences, hands may be partially occluded by the Sensel Morph touchpad from certain camera views, which can hinder the convergence of the optimization process for the total rendered mask. To address this issue, we have modeled the touchpad and its pedestal. We pre-generate the depth map \(D_o\) to represent these scene obstacles. Subsequently, we perform simple depth culling with the rendered depth \(\rend\) by generating a culling mask \(M_{dc} = \mathbb{I}(D_o > \rend)\). This allows us to create cutouts on the rendered depth \(\rend\), the appearance \(\renf\), and the mask \(\rens\), which together represent the hand parts in front of the scene obstacles. After initial tests, we noticed that this depth culling encourages the intersection of the hand mesh and the touchpad to reach lower mask IoU loss~$\mathcal{L}_M$. Therefore, we add a collision box of the touchpad into mesh intersection loss~$\mathcal{L}_{insec}$ to penalize this intersection. We show an example in Figure~\ref{fig:depth_culling}.

\subsubsection{Temporal Continuity}
Our optimization considers consecutive captures consisting of 7 RGB-D and one pressure frame in batches of size \(B\) to ensure temporal continuity of annotated hand poses across timestamps. We apply regularization on the approximated second-order derivative of the hand joint positions \(\mathbf{J}\), which are regressed from the MANO mesh. The temporal continuity regularization is:

\begin{equation}
\mathcal{L}_{\text{temp}} = \frac{1}{B-2} \sum_{i=1}^{B-2} \left\| \mathbf{J}_{i+2} - 2\mathbf{J}_{i+1} + \mathbf{J}_i \right\|_2.
\label{eq:temp_con}
\end{equation}

\subsection{Evaluation of Annotation Fidelity}

\subsubsection{Manual Annotation and Inspection}
\begin{figure}[tb]
\centering
   \includegraphics[width=0.50\textwidth]{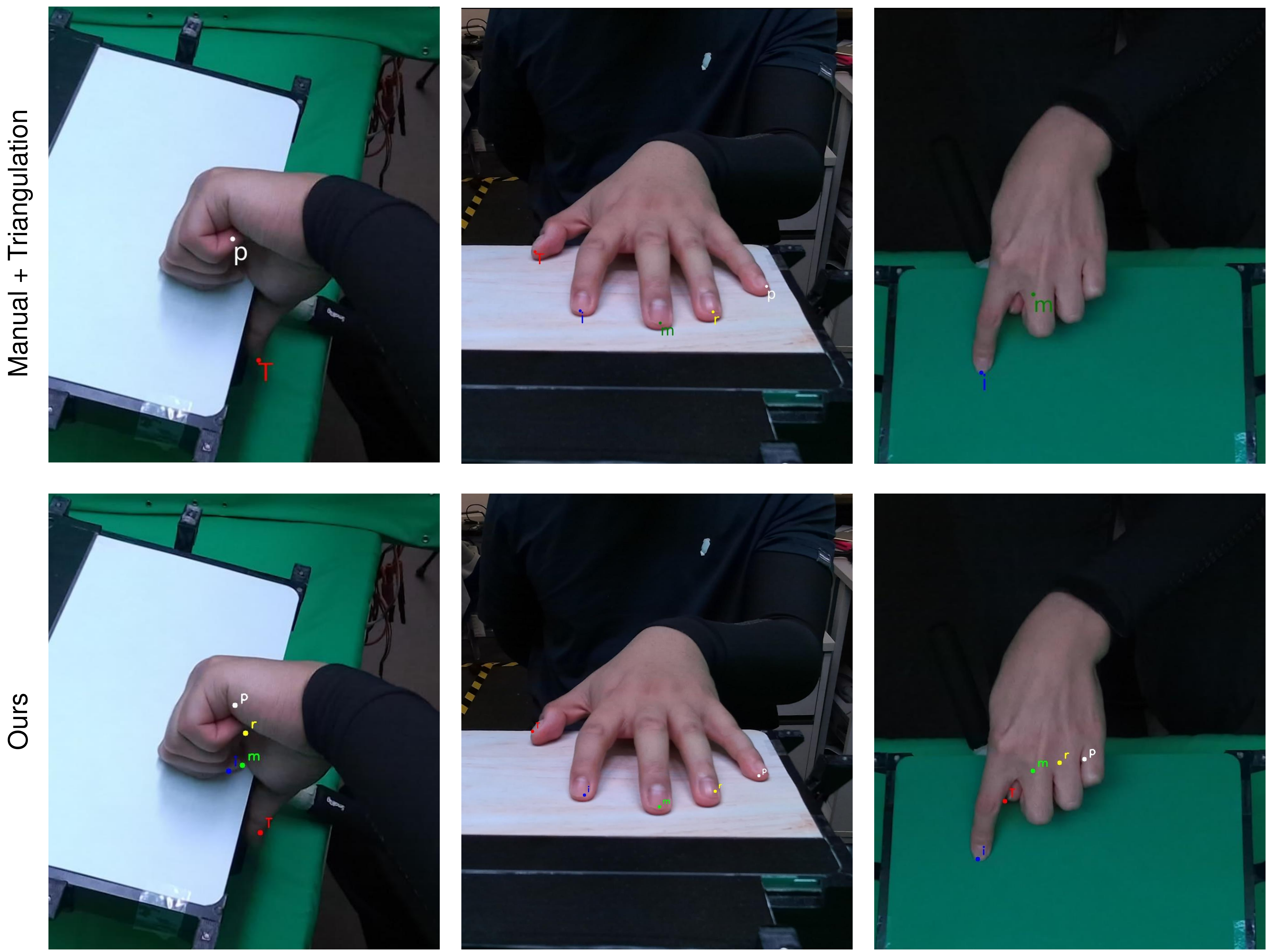}
   \caption{\textbf{Manual Verification Examples}. We demonstrate our annotation is accurate compared to the manual annotations.
   \textbf{(above)} We re-project the triangulated nail tips. We only triangulated them when they are visible in at least 2 views.
   \textbf{(bottom)} We re-project our 3D annotations which also show invisible nail tips as well. }
   \label{fig:manual}
\end{figure}

To verify the quality of the hand poses from our annotation method, we manually annotated 300 randomly selected sets of 7 static views and one egocentric view (300$\times$8 = 2400 frames). We annotated all the \textbf{visible} nail tips in the camera views, resulting in \textbf{7176} 2D points. These 2D nail tips were then triangulated to obtain 3D points. After applying a threshold of 2 pixels on the re-projection error to exclude inconsistent manual annotations, we obtained \textbf{1114} 3D points that were visible in at least two camera views. In Table~\ref{tab:3dtips}, we report the distance error of the hand tips obtained from our annotation method relative to the 3D tip positions based on the manual annotations. We also include an ablation study of our approach. Qualitative results of the manual annotations and our method are shown in Figure~\ref{fig:manual}.

\begin{table}[tb]
\begin{center}
\caption{\small  \textbf{Quantitative evaluation of our annotation method compared to 3D tip positions triangulated from manual annotations.} We conduct an ablation study for the different loss terms, including appearance loss~$\mathcal{L}_A$ (Eq.~\ref{eq:render_loss}), depth volumetric IoU loss~$\mathcal{L}_D$ (Eq.~\ref{eq:render_loss}), mesh intersection loss~$\mathcal{L}_{insec}$ (Sec.~\ref{sec:geo_obj}), as-rigid-as-possible loss~$\mathcal{L}_{arap}$ (Sec.~\ref{sec:geo_obj}), Laplacian smoothness~$\mathcal{L}_{lap}$ (Sec.~\ref{sec:geo_obj}), normal consistency regularization~\(\mathcal{L}_{\vec{\textbf{n}}}\) (Sec.~\ref{sec:geo_obj}), and vertex offset regularization~$\mathcal{L}_{offset}$ (Sec.~\ref{sec:geo_obj}). We demonstrate that each loss term contributes to our optimization performance.
}
\vspace{0.5em}
\begin{adjustbox}{width=0.5\textwidth,center}
\begin{tabular}{l |c c c c c c c c }
\toprule
\hline
Losses & Ours   &   w/o $\mathcal{L}_A$ &   w/o $\mathcal{L}_D$  &  w/o $\mathcal{L}_{insec}$  &  w/o $\mathcal{L}_{arap}$ &w/o $\mathcal{L}_{lap}$ &  w/o $\mathcal{L}_{\vec{\textbf{n}}}$  &   w/o $\mathcal{L}_{offset}$ \\

\hline
$\mathcal{L}_D ~\downarrow$ &\textbf{0.1251} & 0.1404 & 0.1797 & 0.1368 & 0.1347 & 0.1396 &  0.1349 & 0.1477\\
$\mathcal{L}_M~\downarrow$ &\textbf{0.0488} & 0.0662 & 0.0724 & 0.0619 & 0.0597 & 0.0654 &  0.0653 & 0.0728\\
3D tips error~[mm] $~\downarrow$    &\textbf{5.68}   & 7.66   & 8.45   &  7.99  & 8.61   & 8.49   &  8.39   & 8.28            \\
\hline
\bottomrule
\end{tabular}
\label{tab:3dtips}
\end{adjustbox}
\end{center}

\end{table}

\subsubsection{Comparison to learning-based model}

Compared to the state-of-the-art 3D hand pose estimator, HaMeR, our optimization-based method offers significant advantages, enabling the creation of high-quality annotations for our dataset. As shown in Figure~\ref{fig:ego_hamer_comparison}, although hand poses from HaMeR~\cite{hamer} appear plausible from a top view, side views expose inaccuracies and scale ambiguities. In contrast, our annotation method produces robust and consistent results across all camera views.
In Table~\ref{table:baselines_results}), we demonstrate that the baseline model with our high-quality 3D hand poses improves hand pressure estimation compared to using HaMeR's~\cite{hamer} predictions. 

To further evaluate annotation quality, we provide the validation results comparing the triangulation of predicted nail tips with manual annotations across static views in Table~\ref{tab:hamer_triangulation_results}.
Additionally, Figure~\ref{fig:qualitative_results_hamer_vs_our} presents a qualitative comparison of pressure estimation incorporating additional poses from HaMeR~\cite{hamer} and our ground truth annotations. The results emphasize the importance of the high-fidelity hand pose annotations from our optimization method, both quantitatively and qualitatively, and highlight the necessity of advancing hand pose and pressure map estimation in future research.

\begin{figure}[t]
  \centering
  \scriptsize
  \setlength{\tabcolsep}{0.1pt}
  \newcommand{\csz}{0.25}
  \begin{tabular}{cccc}

   \includegraphics[width=\csz\linewidth]{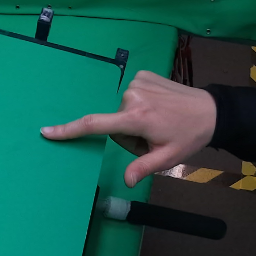} &
  \includegraphics[width=\csz\linewidth]{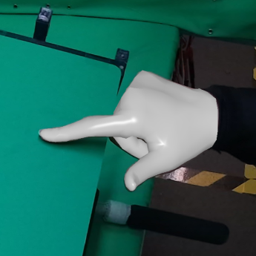} &
  \includegraphics[width=\csz\linewidth]{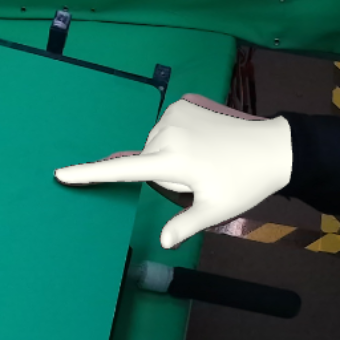} &
  \includegraphics[width=\csz\linewidth]{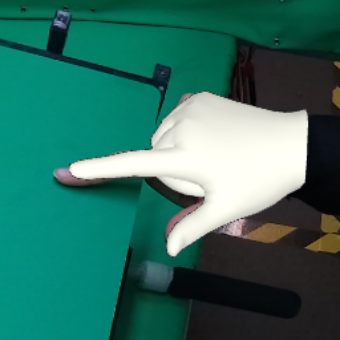} 
\\

 \includegraphics[width=\csz\linewidth]{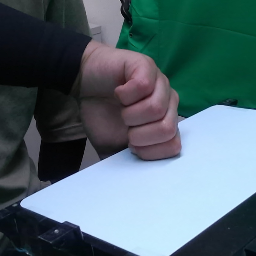} &
  \includegraphics[width=\csz\linewidth]{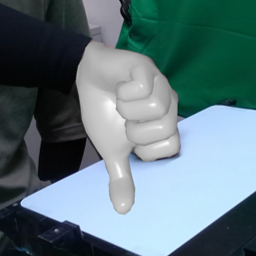} &
  \includegraphics[width=\csz\linewidth]{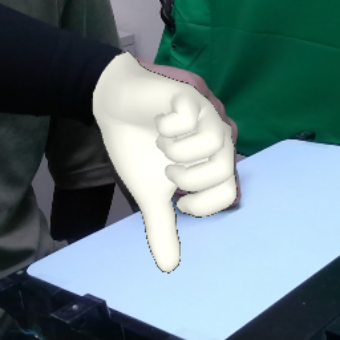} &
  \includegraphics[width=\csz\linewidth]{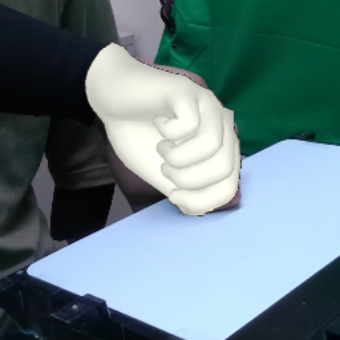} 

  \\
   \includegraphics[width=\csz\linewidth]{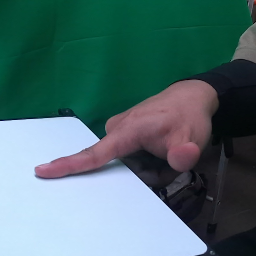} &
  \includegraphics[width=\csz\linewidth]{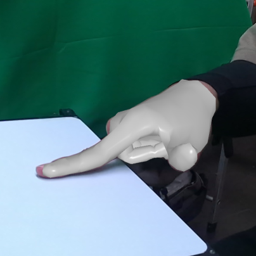} &
  \includegraphics[width=\csz\linewidth]{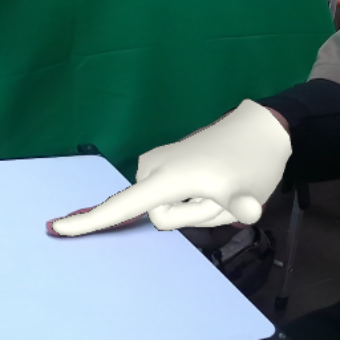} &
  \includegraphics[width=\csz\linewidth]{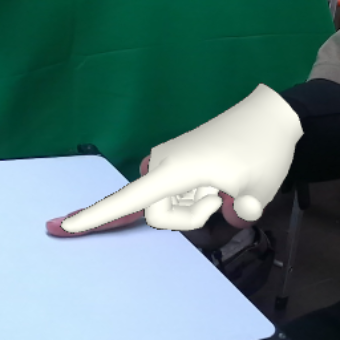} 
  \\
     \includegraphics[width=\csz\linewidth]{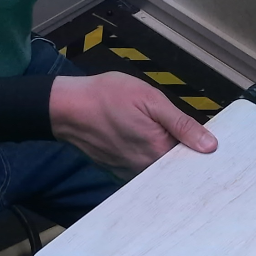} &
  \includegraphics[width=\csz\linewidth]{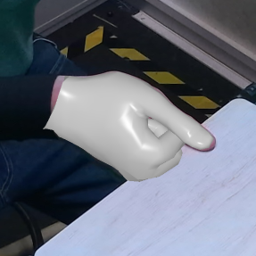} &
  \includegraphics[width=\csz\linewidth]{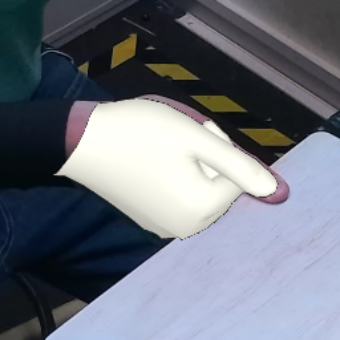} &
  \includegraphics[width=\csz\linewidth]{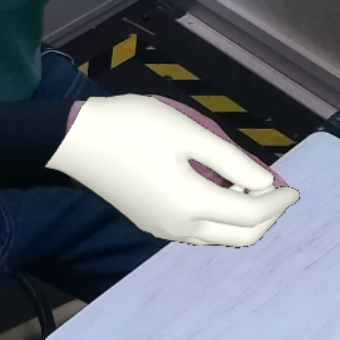} 
  \\
     \includegraphics[width=\csz\linewidth]{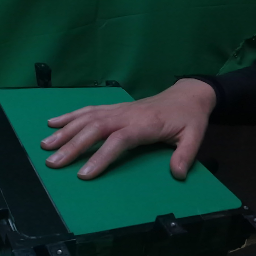} &
  \includegraphics[width=\csz\linewidth]{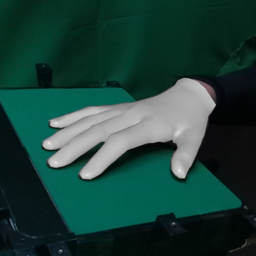} &
  \includegraphics[width=\csz\linewidth]{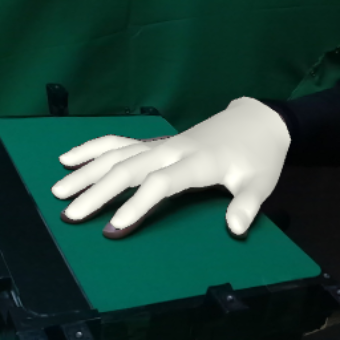} &
  \includegraphics[width=\csz\linewidth]{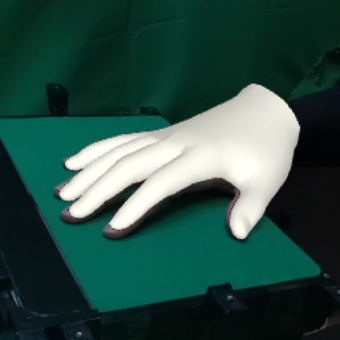} 
  \\

  Input & Our Annotation & Fine-tuned & Pretrained \\ \cline{3-4}
  & &  \multicolumn{2}{c}{\scriptsize HaMeR~\cite{hamer}}
  \
       
  \end{tabular}  
 \vspace{-2mm}

  \caption{ \textbf{Hand pose prediction and ground truth pose visualization for each camera.} We fine-tune HaMeR~\cite{hamer} on our dataset, demonstrating improved detail in hand pose estimation, particularly in scenarios where the hand interacts with a surface.
  }
\label{fig:finetune_hamer}
\vspace{-5pt}
\end{figure}

\begin{figure*}
\centering
\begin{tabular}{cccccc}  
\multicolumn{6}{l}
{\footnotesize \hspace{3.5em}  Egocentric  \hspace{7.5em}   Ours  \hspace{7.5em} HaMeR~\cite{hamer} \hspace{4.5em} Exocentric~(static)  \hspace{3.7em}  HaMeR~\cite{hamer}\hspace{4.0em}  Ours

}
\vspace{-0.1em}
\\
\multicolumn{6}{c}{

\includegraphics[width=1.0\textwidth]{"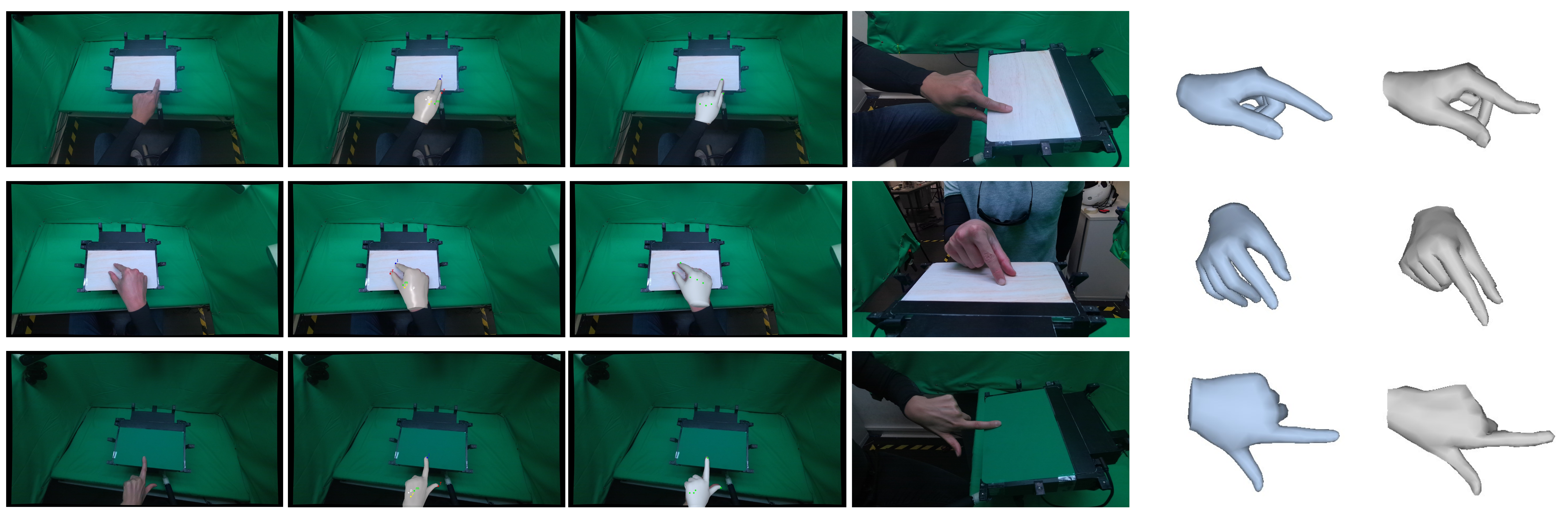"}
}
\end{tabular}

~   ~ \caption{\textbf{Comparison of the estimated hand mesh from HaMeR~\cite{hamer} and our annotation method in both egocentric and exocentric views.} While the projected hand mesh from HaMeR appears visually plausible from an egocentric perspective, observable differences in hand articulation and mesh deformations become apparent from the exocentric viewpoint of the static cameras.
}

\label{fig:ego_hamer_comparison}
\end{figure*}

\begin{figure*}
\centering
\small
\begin{tabular}{C{0.02\textwidth}C{0.12\textwidth}C{0.12\textwidth}C{0.12\textwidth}C{0.12\textwidth}C{0.12\textwidth}C{0.12\textwidth}C{0.02\textwidth}}
&            &            &\multicolumn{2}{c}{Pressurevision~\textbf{\cite{grady2022pressurevision} w. GT poses}}& \multicolumn{2}{c}{Pressurevision~\cite{grady2022pressurevision} w. HaMeR~\cite{hamer} poses}&
\\\addlinespace[-0.5ex]
\cmidrule[1pt](lr){4-5}\cmidrule[1pt](lr){6-7} \addlinespace[-0.5ex]
&Input Image & GT Pressure & Pose & Est. Pressure &Pose &Est. Pressure&
\\ \addlinespace[-0.5ex]
\multicolumn{8}{c}{   \includegraphics[width=0.9\textwidth]{"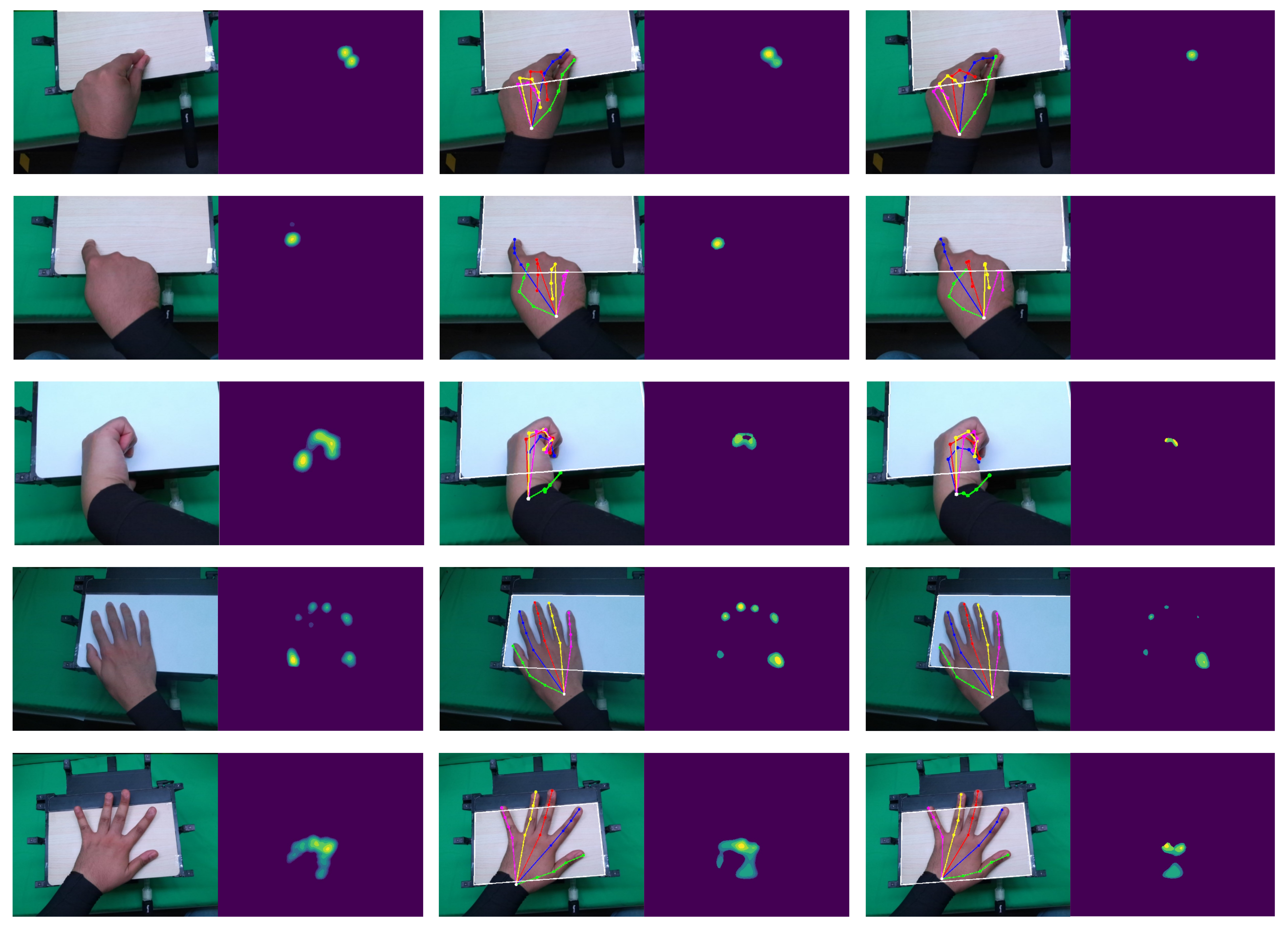"}
}
\end{tabular}

~   ~ \caption{\textbf{Qualitative results of the image-projected baselines on egocentric views, incorporating additional hand pose inputs using our annotations and predictions from HaMeR~[73].} We also reproject the area of the touchpad (indicated by white lines) to verify the egocentric camera pose.}

 \vspace{-5mm}
   \label{fig:qualitative_results_hamer_vs_our}
\end{figure*}

Finally, we report the results of the HaMeR method after fine-tuning on our dataset in Table~\ref{tab:hamer_finetuned} and in Figure~\ref{fig:finetune_hamer}. Although fine-tuning improves performance, there remains room for further enhancement. These results establish a solid baseline for tackling 3D hand pose estimation during hand-surface interactions in an egocentric view.

\begin{table}[h]
\centering
\caption{\textbf{Hand pose verification.} Triangulation is performed on the nail tips using HaMeR~\cite{hamer} predictions across all static cameras, compared against manual annotations.}

\resizebox{0.6\columnwidth}{!}{
\begin{tabular}{c|c c}
\toprule
\hline
&3D tips error [mm] & Std. \\
\hline
Ours& 5.68 & 4.9 \\
HaMeR~\cite{hamer} &12.37 &6.3\\ 
\hline
 \bottomrule
\end{tabular}
}
\label{tab:hamer_triangulation_results}
\end{table}

\begin{table}[h]
\centering
\caption{\textbf{Fine-tuning results} of HaMeR~\cite{hamer} on EgoPressure demonstrate improved hand pose accuracy, underscoring the value of our dataset for 3D hand pose estimation.}

\resizebox{1\columnwidth}{!}{
\begin{tabular}{c|c c}
\toprule
\hline
&MPJPE [mm] & Reconstruction Error [mm] \\
\hline
Finetuned HaMeR~\cite{hamer}& 10.75 &6.10\\
HaMeR~\cite{hamer}& 18.58 & 8.11\\ 
\hline
 \bottomrule
\end{tabular}
}
\label{tab:hamer_finetuned}
\end{table}

\begin{figure}[tb]
\hspace{-1.3em}
\begin{tabular}{lll}
\scriptsize 
\hspace{1.1em}Manual + Triangulation & 
\scriptsize \hspace{1.7em} Our annotation &
\scriptsize \hspace{1.5em} HaMeR~\cite{hamer}+ Triangulation \vspace{-0.2em}
\\

\multicolumn{3}{l}{
\centering
   \includegraphics[width=0.5\textwidth]{"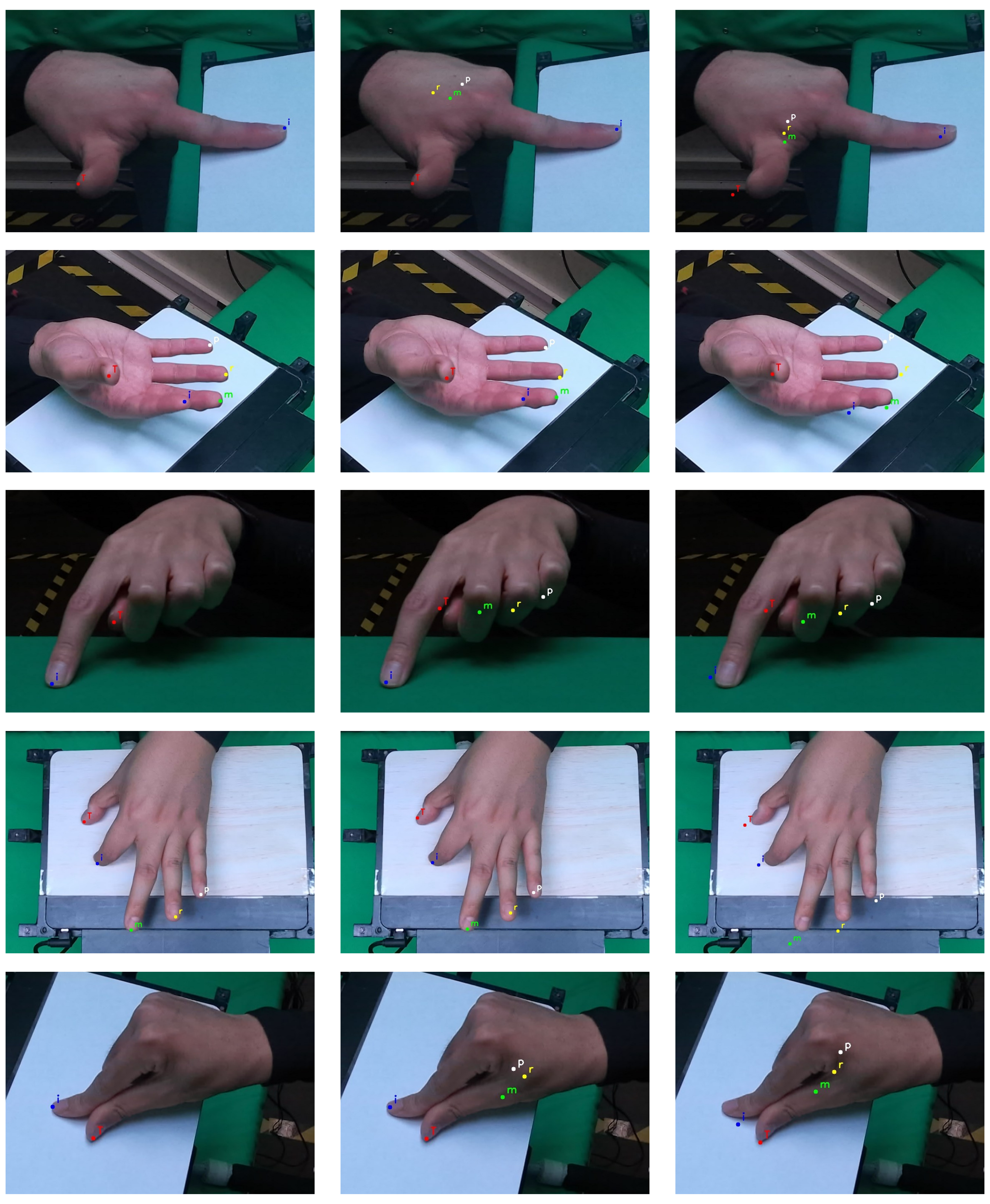"}}
  
   \end{tabular}
    \vspace{-1mm}
~   ~ \caption{\textbf{Qualitative comparison of reprojected nail tips} from our annotation method~(\textbf{center }) and triangulation of HaMeR~\cite{hamer} predictions~(\textbf{right}). The \textbf{left} column displays the reprojection of triangulated manually annotated visible tips.}

  \vspace{-3mm}
   \label{fig:triangulation_hamer_vs_our}
\end{figure}

\section{Extended Details about Dataset}

 \begin{figure}[ht]
    \centering
    \begin{subfigure}[b]{0.23\textwidth}
        \centering
        \includegraphics[width=\textwidth]{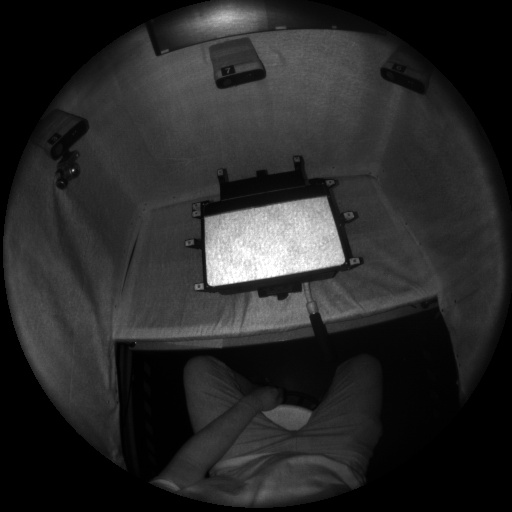} 
        \subcaption*{Marker deactivated}
        \label{fig:sub1}
    \end{subfigure}
   \hfill
    \begin{subfigure}[b]{0.23\textwidth} 
        \centering
        \includegraphics[width=\textwidth]{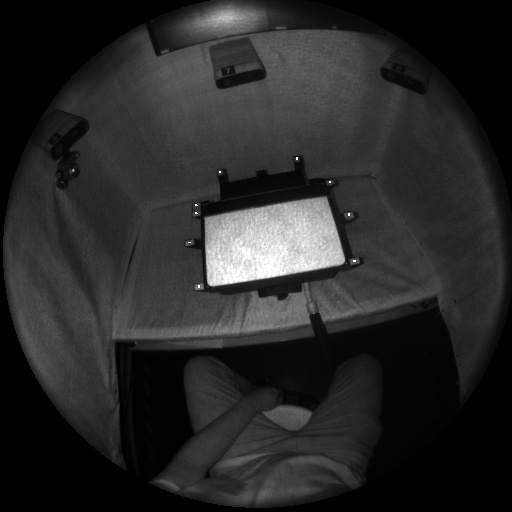} 
        \subcaption*{Marker activated}
        \label{fig:sub2}
    \end{subfigure}
    \caption{ Marker visibility in infrared frame of head-mounted egocentric camera}
    \label{fig:marker_visibilty}
\end{figure}

\subsection{Details about Gesture Description}

Table~\ref{tab:gestures} lists all gestures performed by a participant during the data collection, including which hands were used and how often each gesture was repeated.
We refer to the accompanying video for visual examples.

\begin{table}[h]
\caption{\textbf{List of gestures} performed by a participant during the data collection.}
\centering
\resizebox{1\columnwidth}{!}{
\begin{tabular}{ll|cccccccccccc}

& \textbf{Gesture} & \textbf{Left Hand} & \textbf{Right Hand} & \textbf{Number of Repetitions} \\ \hline
i. &calibration routine & \yesmark{} & \yesmark{} & - \\ 
ii. &draw word & \yesmark{} & \yesmark{} & 3 \\ 
iii. &grasp edge curled thumb-down & \yesmark{} & \yesmark{} & 5 \\ 
iv. &grasp edge curled thumb-up & \yesmark{} & \yesmark{} & 5 \\ 
v. &grasp edge uncurled thumb-down & \yesmark{} & \yesmark{} & 5 \\ 
vi. &index press high force& \yesmark{} & \yesmark{} & 5 \\ 
vii. &index press low force& \yesmark{} & \yesmark{} & 5 \\ 
viii. &index press no-contact & \yesmark{} & \yesmark{} & 5 \\ 
ix. &index press pull & \yesmark{} & \yesmark{} & 5 \\ 
x. &index press push & \yesmark{} & \yesmark{} & 5 \\ 
xi. &index press rotate left & \yesmark{} & \yesmark{} & 5 \\
xii. &index press rotate right & \yesmark{} & \yesmark{} & 5 \\ 
xiii. &pinch thumb-down high force& \yesmark{} & \yesmark{} & 5 \\ 
xiv. &pinch thumb-down low force& \yesmark{} & \yesmark{} & 5 \\ 
xv. &pinch thumb-down no-contact & \yesmark{} & \yesmark{} & 5 \\ 
xvi. &pinch zoom & \yesmark{} & \yesmark{} & 5 \\ 
xvii. &press cupped onebyone high force & \yesmark{} & \yesmark{} & 3 \\ 
xviii. &press cupped onebyone low force& \yesmark{} & \yesmark{} & 3 \\ 
xix. &press fingers high force& \yesmark{} & \yesmark{} & 5 \\ 
xx. &press fingers low force& \yesmark{} & \yesmark{} & 5 \\ 
xxi. &press fingers no-contact & \yesmark{} & \yesmark{} & 5 \\ 
xxii. &press flat onebyone high force& \yesmark{} & \yesmark{} & 3 \\ 
xxiii. &press flat onebyone low force& \yesmark{} & \yesmark{} & 3 \\ 
xxiv. &press palm high force& \yesmark{} & \yesmark{} & 5 \\ 
xxv. &press palm low force& \yesmark{} & \yesmark{} & 5 \\ 
xxvi. &press palm no-contact & \yesmark{} & \yesmark{} & 5 \\ 
xxvii. &press palm-and-fingers high force& \yesmark{} & \yesmark{} & 5 \\ 
xxviii. &press palm-and-fingers low force& \yesmark{} & \yesmark{} & 5 \\ 
xxix. &press palm-and-fingers no-contact & \yesmark{} & \yesmark{} & 5 \\ 
xxx. &pull towards & \yesmark{} & \yesmark{} & 5 \\ 
xxxi. &push away & \yesmark{} & \yesmark{} & 5 \\ 
xxxii. &touch iPad & \yesmark{} & \yesmark{} & 3 \\ 

\end{tabular}
}
\label{tab:gestures}
\vspace{-1em}
\end{table}

\subsection{Dataset Comparisons}

Table~\ref{tab:priorwork_all} provides a comprehensive comparison of our proposed dataset. Among existing public datasets focusing on contact or hand-object pose estimation,
\dataset{} is the first dataset to combine egocentric video data of hand-surface interactions with ground-truth contact and pressure information, as well as high-fidelity hand poses and meshes.

\begin{table*}[h]
\centering
\caption{Comparison between \dataset{} and extended list of hand-contact datasets.}
\resizebox{0.88\textwidth}{!}{
\begin{tabular}{l|ccccccccccccc}

\textbf{Dataset} & \textbf{frames} & \textbf{participants} & \textbf{hand pose} & \textbf{hand mesh} & \textbf{markerless}  & \textbf{real} & \textbf{egocentric} & \textbf{multiview} & \textbf{RGB} & \textbf{depth} & \textbf{contact} & \multicolumn{2}{c}{\textbf{pressure}} \\
 &  &  &  &  &   &  &  &  &  &  &  &  {surface} & {hand}   \\ \hline
\textbf{EgoPressure (ours)}                                   & 4.3M & 21 & \yesmark{} & \yesmark{} & \yesmark{} & \yesmark{} & \yesmark{} & \yesmark{} & \yesmark{} & \yesmark{} & Pressure sensor & \yesmark{}     &  \yesmark{}              \\ 
ContactLabelDB~\cite{grady2024pressurevision++} & 2.9M & 51 &  \nomark{} & \nomark{} & \yesmark{} 	& \yesmark{}  & \nomark{} & \yesmark{} & \yesmark{} & \nomark{}            & Pressure sensor & \yesmark{}	     &  \nomark{}      \\
PressureVisionDB~\cite{grady2022pressurevision} & 3.0M & 36 & \nomark{} & \nomark{} & \yesmark{} 	& \yesmark{}  & \nomark{} & \yesmark{} & \yesmark{} & \nomark{}            & Pressure sensor & \yesmark{} 	     &  \nomark{}           \\
ContactPose~\cite{brahmbhatt2020contactpose} & 3.0M & 50 & \yesmark{} & \yesmark{} & \yesmark{} 	& \yesmark{}  & \nomark{} & \yesmark{} & \yesmark{} & \yesmark{}           & Thermal imprint & \nomark{}  	     &  \nomark{}           \\
GRAB~\cite{taheri2020grab} & 1.6M & 10 & \yesmark{} & \yesmark{} & \nomark{} 						& \yesmark{}  & \nomark{} & \nomark{} & \nomark{} & \nomark{}              & Inferred from Pose & \nomark{}      &  \nomark{}         \\
ARCTIC~\cite{fan2023arctic} & 2.1M & 10 & \yesmark{} & \yesmark{} & \nomark{} 						& \yesmark{} & \yesmark{} & \yesmark{} & \yesmark{} & \yesmark{}           & Inferred from Pose & \nomark{}      &  \nomark{}             \\
H2O~\cite{kwon2021h2o}  & 571k & 4 & \yesmark{} & \yesmark{} & \yesmark{} 							& \yesmark{} & \yesmark{} & \yesmark{} & \yesmark{} & \yesmark{}           & Inferred from Pose & \nomark{}      &  \nomark{}            \\
OakInk~\cite{yang2022oakink} & 230k & 12 & \yesmark{} & \yesmark{} & \yesmark{} 						& \yesmark{} & \nomark{} & \yesmark{} & \yesmark{}  & \yesmark{}       & Inferred from Pose & \nomark{}      &  \nomark{}              \\
OakInk-2~\cite{zhan2024oakink2} & 4.01M & 9 & \yesmark{} & \yesmark{} & \yesmark{} 						& \yesmark{} & \yesmark{} & \yesmark{} & \yesmark{} & \nomark{}        & Inferred from Pose & \nomark{}      &  \nomark{}               \\
DexYCB~\cite{chao2021dexycb} & 582k & 10 & \yesmark{} & \yesmark{} & \yesmark{} 						&	 \yesmark{} & \nomark{} & \yesmark{} & \yesmark{} & \yesmark{}     & Inferred from Pose & \nomark{}      &  \nomark{}                  \\
HO-3D~\cite{hampali2020honnotate} & 103k & 10 & \yesmark{} & \yesmark{} & \yesmark{} 					& \yesmark{} & \nomark{} & \yesmark{} & \yesmark{} & \yesmark{}        & Inferred from Pose & \nomark{}      &  \nomark{}              \\                           
TACO~\cite{liu2024taco} & 5.2M & 14 & \yesmark{} & \yesmark{} & \yesmark{} 								& \yesmark{} & \yesmark{} & \yesmark{} & \yesmark{} & \yesmark{}       & Inferred from Pose & \nomark{}      &  \nomark{}                \\
Affordpose~\cite{jian2023affordpose} & 26.7k & - & \yesmark{} & \yesmark{} & \yesmark{} 				& \nomark{} & \nomark{} & \yesmark{} & \nomark{} & \nomark{}           & Inferred from Pose & \nomark{}      &  \nomark{}             \\
AssemblyHands~\cite{ohkawa2023assemblyhands} & 3.03M & 34 & \yesmark{} & \nomark{} & \yesmark{} 	& \yesmark{} & \yesmark{} & \yesmark{} & \yesmark{} & \nomark{}            & \nomark{} & \nomark{} 		         &  \nomark{}      \\
ContactArt~\cite{zhu2023contactart} & 332k & - & \yesmark{} & \yesmark{} & \yesmark{} 				& \nomark{} & \nomark{} & \yesmark{} & \yesmark{} & \yesmark{}             & Simulated Pose & \nomark{}  	     &  \nomark{}          \\
HOI4D~\cite{liu2022hoi4d} & 2.4M & 9 & \yesmark{} & \yesmark{} & \yesmark{} 						& \yesmark{} & \yesmark{} & \nomark{} & \yesmark{} & \yesmark{}            & Inferred from Pose & \nomark{}      &  \nomark{}            \\
YCBAfford~\cite{corona2020ganhand} & 133k & - & \yesmark{} & \yesmark{} & \yesmark{} 				& \nomark{} & \nomark{} & \nomark{} & \nomark{} & \nomark{}                & Simulated Pose & \nomark{} 	     &  \nomark{}       \\
ObMan~\cite{hasson2019learning} & 154k & - & \yesmark{} & \yesmark{} & \yesmark{} 					& \nomark{} & \nomark{} & \nomark{} & \yesmark{} & \yesmark{}              & Simulated Pose & \nomark{}  	     &  \nomark{}     \\
FPHAB~\cite{garcia2018first} & 100k & 6 & \yesmark{} & \nomark{} & \nomark{} 						& \yesmark{} & \yesmark{} & \nomark{} & \yesmark{} & \yesmark{}            & \nomark{} & \nomark{}  		     &  \nomark{}         \\
HA-ViD~\cite{zheng2024ha} & 1.5M & 30 & \nomark{} & \nomark{} & \yesmark{} 							& \yesmark{} & \nomark{} & \yesmark{} & \yesmark{} & \yesmark{}            &\nomark{} & \nomark{} 			     &  \nomark{}       \\
Ego4d~\cite{grauman2022ego4d} & 3670~hours & 923 & \nomark{} & \nomark{} & \yesmark{} 				& \yesmark{} & \yesmark{} & \nomark{} & \yesmark{} & \nomark{}             & \nomark{}  & \nomark{} 		     &  \nomark{}        \\
EPIC-KITCHEN-100~\cite{damen2022rescaling} & 20M & 37 & \nomark{} & \nomark{} & \yesmark{} 			& \yesmark{} & \yesmark{} & \nomark{} & \yesmark{}  & \nomark{}            & \nomark{}& \nomark{} 			     &  \nomark{}       \\
Ego-Exo4D~\cite{grauman2023ego} & 1422~hours & 740 & \yesmark{} & \nomark{} & \yesmark{} 			& \yesmark{} & \yesmark{} & \yesmark{} & \yesmark{} & \nomark{} 			& \nomark{} & \nomark{}  			&  \nomark{} 	\\
\end{tabular}
}
\label{tab:priorwork_all}
\end{table*}

\subsection{Details about Active IR Marker}

\begin{figure}[ht]
 \centering
     \includegraphics[width=0.45\textwidth]{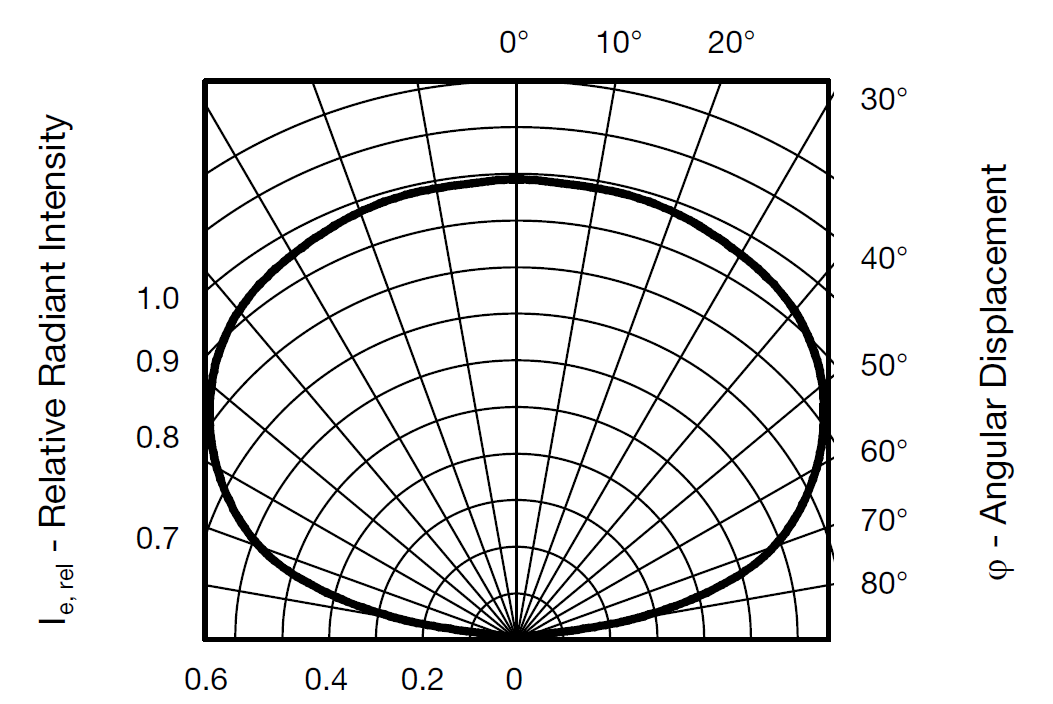}
     \caption{\textbf{Relative Radiant Intensity vs. Angular Displacement
} The marker enable a good visible radiant intensity of beam angle till 150 degree, which ensures good visibility in egocentric infrared camera}

    \label{fig:beam_angular}
\end{figure}

We use active IR Marker, operating similarly to passive markers, these markers emit their own infrared light, allowing for a much smaller and more precise form factor—often appearing as tiny light dots in the filtered infrared image. This reduces the impact of lens distortion on tracking accuracy. Moreover, these markers are programmable, providing crucial control over their activation and deactivation, which is vital for synchronization within our system. We utilize the infrared led with large beam angle~(see Fig.~\ref{fig:beam_angular}) as active infrared marker.

\begin{figure}[ht]
 \centering
     \includegraphics[width=0.45\textwidth]{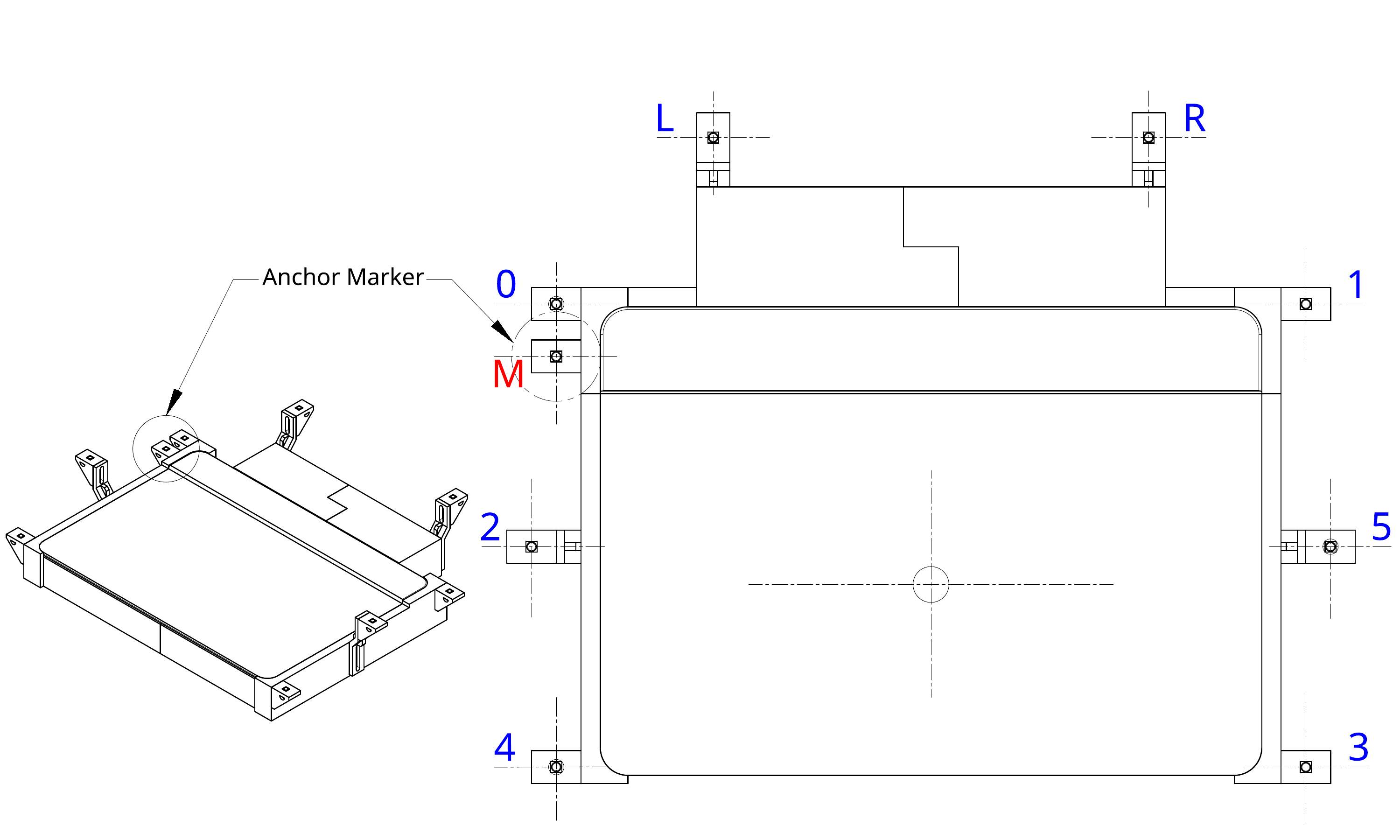}
\caption{\textbf{Layout of the Active Markers.} The indices of the markers are aligned with the pseudocode provided in Algorithm~\ref{alg:identify_marker}. Starting from the asymmetrical anchor marker \textbf{M}, all markers can be identified by computing their relative distances and considering their spatial relationships.}

    \label{fig:layout_marker}
\end{figure}

An asymmetrical layout with markers can be uniquely identified from any viewpoint within the upper hemisphere above the marker arrangement. This distinctive configuration enables robust and accurate real-time tracking using filtered infrared images, where the markers appear as light dots with a radius of several pixels. The process is detailed in the pseudocode presented in Algorithm~\ref{alg:identify_marker}.The effectiveness of this layout in facilitating accurate marker identification and pose estimation is further illustrated in Figure \ref{fig:layout_marker}, where the spatial arrangement of markers is depicted. Furthermore, this procedure can be generalized to other asymmetrical layouts.

The Perspective-n-Points (PnP) algorithm is used to compute the camera pose of the egocentric camera based on the identified markers in the infrared frame. 
In the experiment, the reprojection error for pose computed from well-identified markers remained below an average of 0.4 pixels. To ensure clarity and reliability in recognition, we applied a threshold value of 1 pixel to filter out frames potentially containing ambiguities in marker recognition during the recording. Additionally, for frames where tracking was lost, spherical linear interpolation (Slerp) is employed to estimate camera pose, thereby maintaining continuity and accuracy in the tracking data.

\begin{algorithm}
\footnotesize
\caption{Identify Marker}
\label{alg:identify_marker}

\begin{algorithmic}[1]
\Procedure{IdentifyMarkers}{filtered IR image}
    \State Extract marker coordinates $(u, v)$ from the filtered IR image 
    \State Compute all pairwise distances among markers
    \State Identify the pair with the smallest distance, initially labeled as $0$ and $M$
    \State Compute the vector from $M$ to $0$
    \State Count the number of markers on each side of the vector line $M-0$
    \If{more markers lie on the right of the vector}
        \State Confirm start point as $M$, endpoint as $0$
    \Else
        \State Swap, set start point as $0$ and endpoint as $M$
    \EndIf
    \State Identify $2$ and $4$ as markers aligned with $M-0$, on the same side relative to $M$
    \State Check distances from $M$ to $2$ and $4$ to determine which is closer
    \State Identify $L$ as the marker closest to the line extending through $(0, M, 2, 4)$ and on the same side as $0$
    \State Compute the centroid of all markers
    \State Draw a line from $0$ through the centroid
    \State Identify $3$ as the marker isolated on its side of the centroid line
    \State Identify $5$ as the closest marker to the line $(0-\text{centroid})$ not already labeled
    \State Determine $1$ and $R$ by their proximity to line $(2-5)$, with $1$ being closer

\EndProcedure
\end{algorithmic}
\end{algorithm}

\subsection{Details about Devices' Synchronization in Dataset Acquisition}

The Sensel Morph operates with zero buffer and maintains a stable 8 ms delay at 120 fps, whereas the Azure Kinect cameras function at 30 fps, capturing high-resolution RGB images and a depth map. Due to the high recording performance of the Azure Kinect, frames are initially stored in the device's cache, making it impractical to rely on the OS timestamp at the frame's arrival on the host computer for synchronization with Sensel Morph pressure data.

All cameras can be externally synchronized via a 30 Hz triggering signal from the Raspberry Pi CM4, ensuring simultaneous frame capture. However, an initial frame loss (1–3 frames) occurs at the start of recording due to device-specific issues. Since the absolute value of device ticks has no inherent meaning, it is unclear how many frames were lost before the first received frame. Relying on device tick differences for synchronization could therefore introduce a misalignment of 1–3 frames between cameras.

To address this, the programmable features of active infrared markers and the precise global OS timestamp synchronization (within 1 ms) between the two host computers and the Raspberry Pi CM4, facilitated by the Precision Time Protocol (PTP), are utilized. The Raspberry Pi CM4, equipped with basic electrical components~(see Fig.\ref{fig:marker_visibilty}) at the start of the next exposure cycle, providing a reliable synchronization point that compensates for the initial missing frames. The exact global OS timestamp of the marker activation is clearly recorded~(see Fig.~\ref{fig:diagram}).

By calculating the real OS timestamp for all frames based on the offset from device ticks, starting from the frame where the marker first appears, precise synchronization is achieved. This approach effectively aligns RGBD images and pressure data, optimizing data integration across the multi-modal sensor system. Moreover, this synchronization mechanism using an external active optical identifier is efficient and economical, making it generalizable to other multi-sensor systems, such as motion capture systems with external head-mounted cameras, that rely on different OS timestamp sources.
 \begin{figure}[ht]
 \centering
     \includegraphics[width=0.4\textwidth]{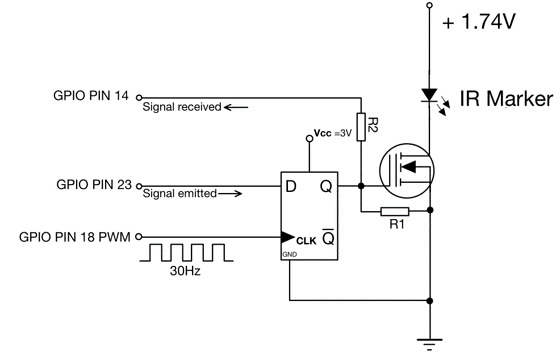}
     \caption{\textbf{Basic Electrical Elements Implementation}, We use a D-type flip-flop and N-channel MOSFET to ensure the IR marker will be activate by the next beginning of exposure after receiving signal from PIN 23. And PIN 14 will monitor the activation to obtain its timestamp}

    \label{fig:simple_circle}
\end{figure}
 \begin{figure}[ht]
 \centering
     \includegraphics[width=0.45\textwidth]{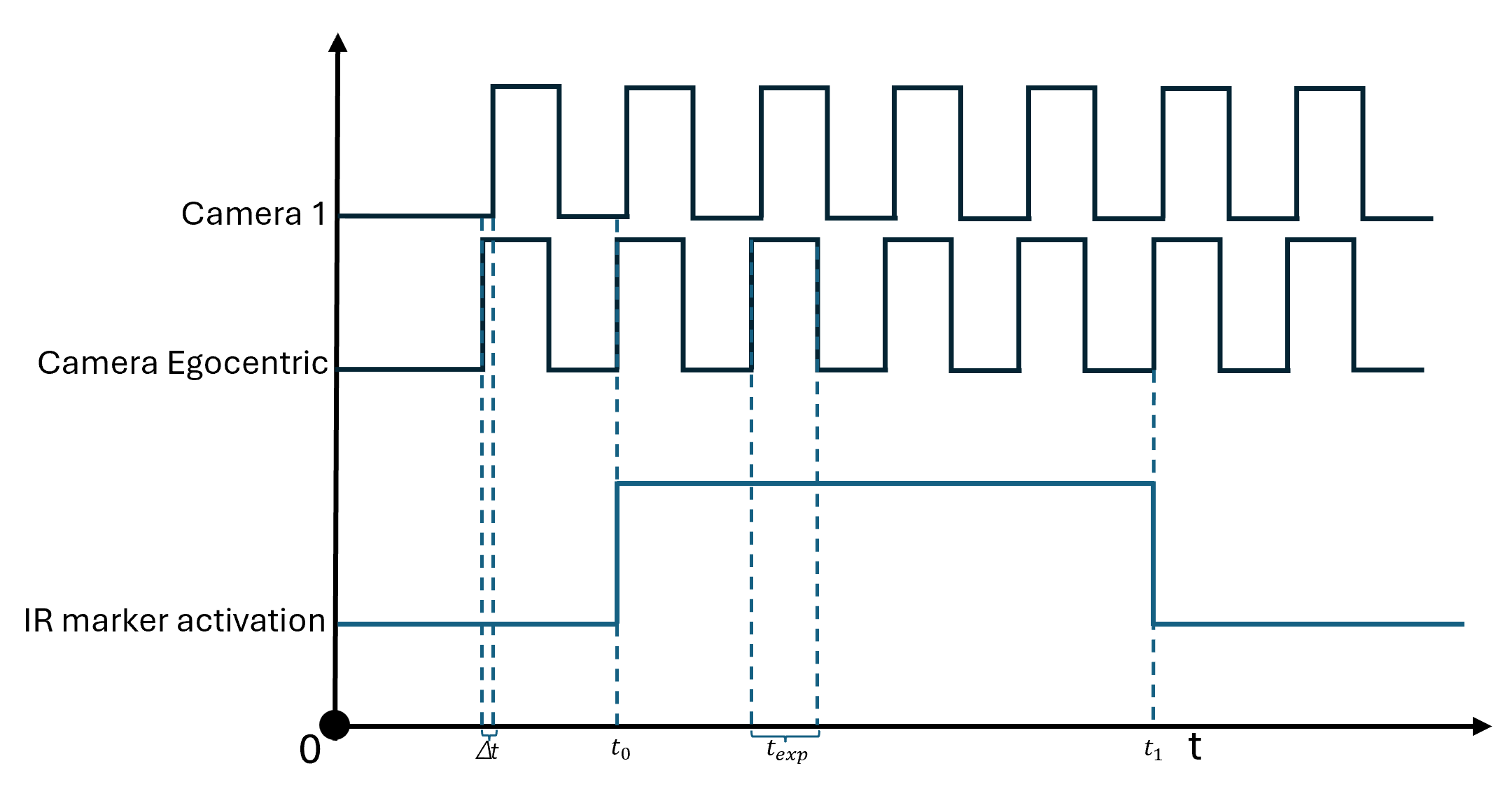}
     \caption{\textbf{Synchronization Diagram} We set head-mounted egocentric camera to align with 30 Hz triggering signal emitted by the Raspberry Pi CM4, this signal will also go to PIN 18 as clock frequency of D-type Flip-flop~ Fig.~\ref{fig:simple_circle}). Then exposure~$t_{exp}$ of all cameras is same.
     The other static cameras 1 to 7 will have a delay~$\Delta~t$ to triggering signal to avoid interference of infrared light. The marker will be activate at $t_0$~(around 300 milliseconds after start recording), which we know its global OS timestamp, then it will be visible to all camera at next exposure cycle. As verification, we deactivate marker by the very end of recording at the timestamp~$t_1$, then the marker will be invisible for all cameras in the next frame capture. The good synchronization will have equal frame number between $t_0$ and $t_1$ for all cameras.   }

    \label{fig:diagram}
\end{figure}
\section{Limitations}
Although EgoPressure serves as a foundational study for understanding pressure from an egocentric view, several challenges remain unresolved. These challenges are categorized into three main areas.

First, measuring pressure while interacting with general objects presents a challenge. Our current data capture is confined to sensing pressure on flat surfaces. While we are optimistic that future research will expand to include a wider variety of objects, sensing pressure on arbitrary surfaces poses significant challenges, as it would require extensive instrumentation of the user's hands, hindering natural interaction and introducing visible artifacts in the captured data.
Instrumenting objects for pressure sensing remains an ongoing research area, with recent advancements primarily in basic contact detection~\cite{brahmbhatt2020contactpose}.
However, we anticipate that our annotation method will extend naturally to more complex objects and interactions as these challenges are addressed.
PressureVision++~\cite{grady2024pressurevision++} explores weak labels to infer pressure on more complex objects. However, it only considers fingertip interactions and its evaluation of pressure regression remains limited to flat surfaces due to the challenges of acquiring precise pressure.
We present a qualitative evaluation of PressureFormer on a wider variety of objects in Figure~\ref{fig:wild}.

Second, the current dataset was only captured in an indoor setting.
Our data capture setup is optimized for acquiring high-fidelity annotations of hand-surface interactions. To increase the diversity of background environments to improve generalization to real-world settings, we have added green overlays to the background of our data capture rig and to the pressure pad.
This allows for background replacement and has been successfully demonstrated to enhance commercial in-the-wild hand tracking~\cite{han2022umetrack, zimmermann2019freihand}.

Finally, the current setup only considers single-hand interactions. Incorporating scenarios involving the use of both hands would be a natural extension of our work.

Further addressing these challenges in future research would improve pressure estimation in real-world scenarios and broaden its applicability.

\section{Ethical Considerations}
The recording and use of human activity data involve important ethical considerations. 
The EgoPressure project has received approval from  ETH Zürich Ethics Commission as proposal EK
2023-N-228.
This approval includes both the data collection and the public release of the dataset.
All participants provided explicit written consent for recording their sessions, creating the dataset, and releasing it (see accompanying consent form). 
All demographic information (such as sex, age, weight, and height) along with the sensor and video data are pseudonymized, assigning a numeric code to each participant.
Personal data (sex, age, weight, and height) is stored separately from the sensor and video data, and is accessible only to the primary researchers involved in the study.
We have not captured or stored any images of the participant's face.

\begin{figure*}
\centering
\begin{tabular}{llllllllll}  

\vspace{-0.1em}
&\hspace{1.55em} \tiny Input & \hspace{1.55em} \tiny Press.Vis~\cite{grady2022pressurevision}&\hspace{0.34em}\tiny {\textbf{\cite{grady2022pressurevision} w. GT Keypoints}} & \hspace{1.0em}\tiny GT& & \hspace{3.0em} \tiny Input &\hspace{1.5em} \tiny Press.Vis~\cite{grady2022pressurevision}& \hspace{0.08em} \tiny {\textbf{\cite{grady2022pressurevision} w. GT Keypoints}} & \hspace{1.0em} \tiny GT\\
\multicolumn{10}{c}{

\includegraphics[width=0.85\textwidth]{"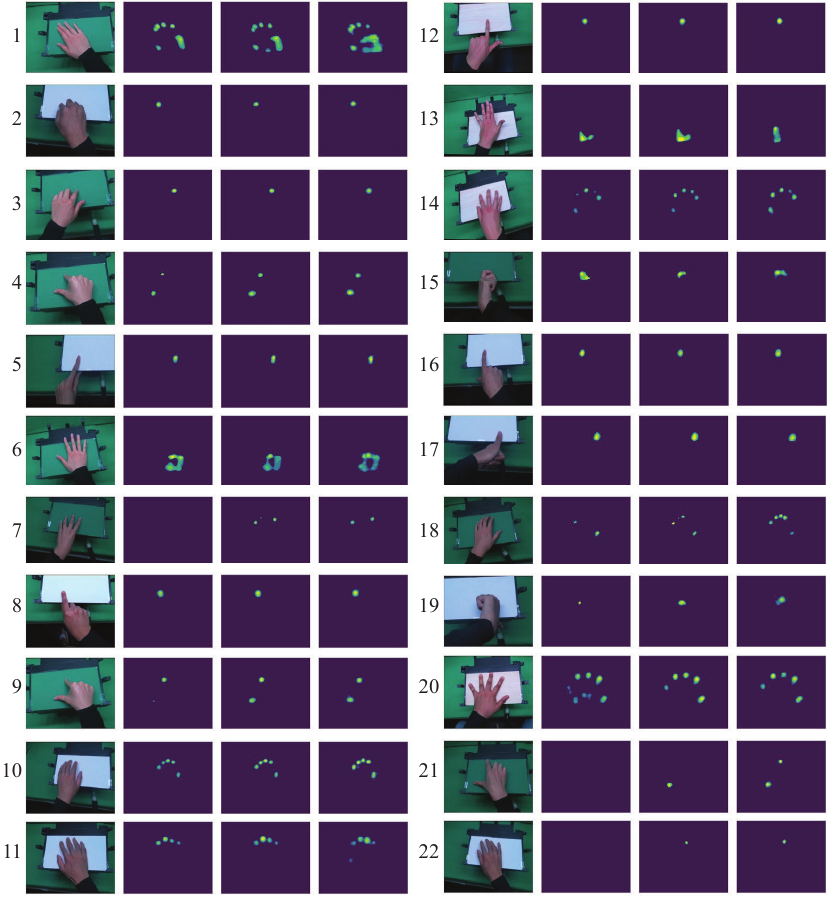"}
}
\end{tabular}

\resizebox{0.75\textwidth}{!}{
\begin{tabular}{cc|ccccccccccc}
\hline
 & & \textbf{1} & \textbf{2} & \textbf{3} & \textbf{4} & \textbf{5} & \textbf{6} & \textbf{7} & \textbf{8} & \textbf{9} & \textbf{10} & \textbf{11}  \\ \hline
\multirow{2}{*}{MAE $\downarrow$}  & \textbf{Press.Vis. ~\cite{grady2022pressurevision}} & 129.6 & 8.2 & 14.3 & 25.3 & 6.7 & 61.9 & 9.3 & 8.3 & 23.9 & 39.1 & 37.1 \\ 
& \textbf{\cite{grady2022pressurevision} w. GT poses} & 116.4 & 5.3 & 11.8 & 17.3 & 4.8 & 62.2 & 5.1 & 6.2 & 13.5 & 30.6 & 35.6 \\ \hline
\multirow{2}{*}{Contact IoU $\uparrow$} & \textbf{Press.Vis. ~\cite{grady2022pressurevision}} & 46.9 & 68.4 & 57.4 & 39.0 & 84.0 & 63.8 & 0.0 & 76.9 & 35.1 & 64.1 & 56.8 \\
& \textbf{\cite{grady2022pressurevision} w. GT poses} & 55.2 & 79.9 & 63.9 & 67.8 &86.8 & 65.6 & 58.7 & 82.6 & 73.2 & 72.1 & 65.4 \\   \hline
\end{tabular}
}

\vspace{0.05cm}

\resizebox{0.75\textwidth}{!}{
\begin{tabular}{cc|ccccccccccc}
\hline
& & \textbf{12} & \textbf{13} & \textbf{14} & \textbf{15} & \textbf{16} & \textbf{17} & \textbf{18} & \textbf{19} & \textbf{20} & \textbf{21} & \textbf{22} \\ \hline

\multirow{2}{*}{MAE $\downarrow$} & \textbf{Press.Vis. ~\cite{grady2022pressurevision}} & 6.2 & 50.1 & 39.4 & 54.7 & 11.7 & 10.7 & 32.0 & 37.0 & 67.4 & 21.5 & 8.29 \\
& \textbf{\cite{grady2022pressurevision} w. GT Keypoints} & 4.7 & 51.7 & 35.6 & 30.4 & 10.0 & 10.3 & 24.9 & 32.6 & 35.5 & 14.7 & 4.89 \\ \hline
\multirow{2}{*}{Contact IoU $\uparrow$} & \textbf{Press.Vis. ~\cite{grady2022pressurevision}} & 86.2 & 38.4 & 54.2 & 24.2 & 79.9 &76.6 & 17.6 & 1.2 & 54.1 & 0.0 & 0.0 \\ 
& \textbf{\cite{grady2022pressurevision} w. GT Keypoints} & 86.3 & 48.4 & 61.7 & 43.8 & 83.1 & 79.6 & 30.3 & 35.2 & 76.3 & 42.2 & 43.6 \\\hline
 \bottomrule
\end{tabular}
}

~   ~ \caption{Qualitative comparison of pressure maps inferred using PressureVisionNet~\cite{grady2022pressurevision} and our trained model with additional hand poses as input on representative cases across various gestures. The bottom table presents MAE [Pa] and Contact IoU [\%] for pressure maps inferred using PressureVisionNet~[1] and our trained model on selected samples shown in the Figure.
}

   \label{fig:more_qualitative}
\end{figure*}

\begin{figure*}
\centering

\small
\begin{tabular}{ccccccc}

& \hspace{0.1\linewidth}\scriptsize{Input} &\hspace{0.035\linewidth} \scriptsize{Anno. Pose} & \hspace{0.04\linewidth} \scriptsize{GT} & \hspace{0.03\linewidth} \scriptsize{\cite{grady2022pressurevision} w. GT Keypoints}
&\hspace{-0.01\linewidth} \scriptsize{Press.Vis.~\cite{grady2022pressurevision}} &
\\
\multicolumn{7}{c}{\includegraphics[width=0.65\textwidth]{"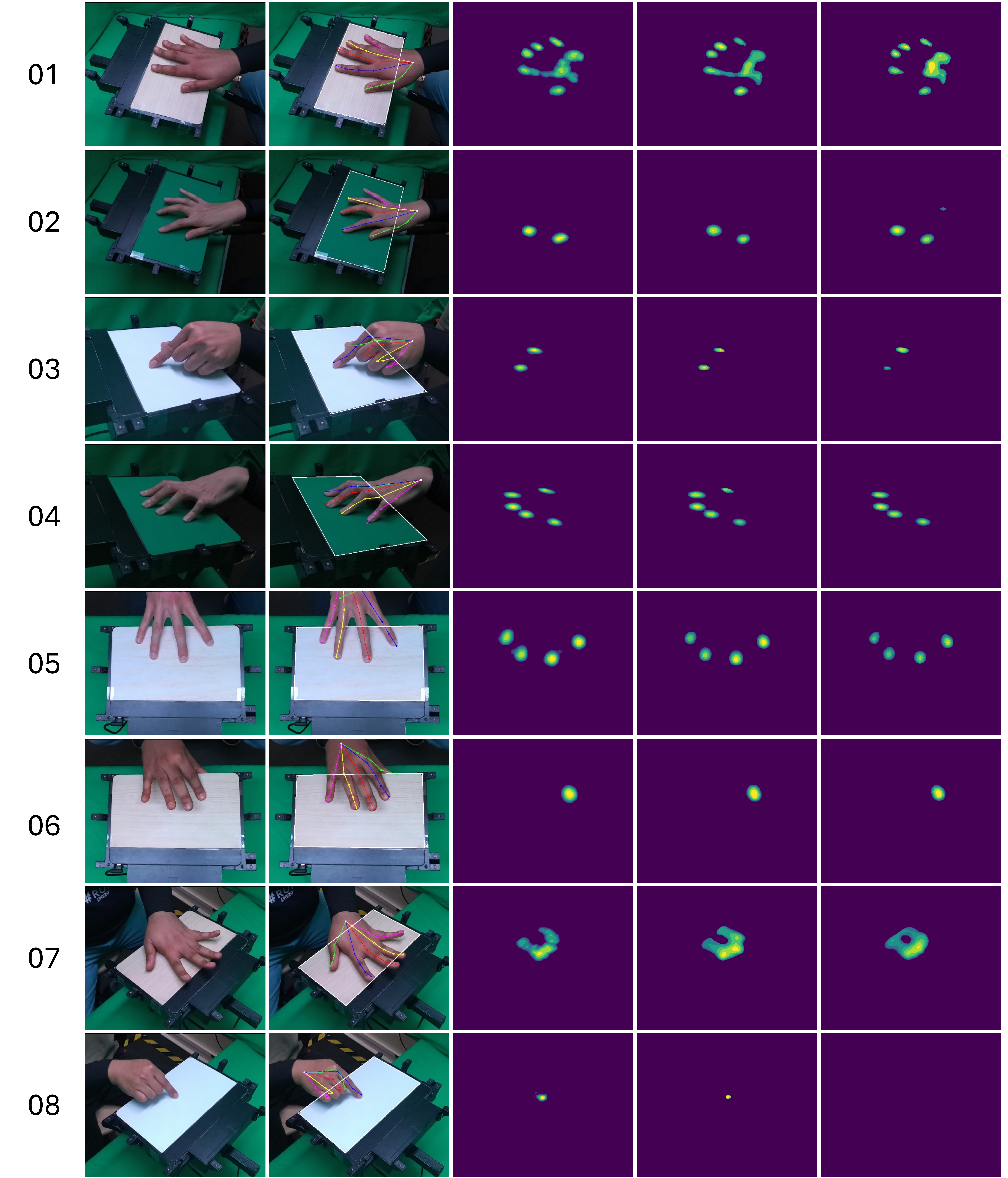"}}
\end{tabular}
   \vspace{-2mm}
\resizebox{0.65\textwidth}{!}{
\begin{tabular}{cccccccccc}
\hline
& & \textbf{01} & \textbf{02} & \textbf{03} & \textbf{04} & \textbf{05} & \textbf{06} & \textbf{07} & \textbf{08} \\ \hline

\multirow{2}{*}{MAE~[Pa] $\downarrow$} & \textbf{Press.Vis.~\cite{grady2022pressurevision}}  & 136.5  &114.8&22.3&57.1&56.4&51.3&71.6&22.1 \\
& \textbf{\cite{grady2022pressurevision} w. GT Keypoints} & 117.0  & 106.2 &20.5&50.5&49.9&46.4&50.6&9.1 \\ \hline
\multirow{2}{*}{Contact IoU~[\%] $\uparrow$} & \textbf{Press.Vis.~\cite{grady2022pressurevision}} &65.6  & 74.9&51.9&63.1&59.2&77.8&59.5&0.0 \\ 
& \textbf{\cite{grady2022pressurevision} w. GT Keypoints} & 78.1  & 77.4&58.1&69.2&65.1&83.2&67.7&31.1\\\hline
&  Camera \# &2 &2&3&3&4&4&5&5  \\\hline
 \bottomrule
\end{tabular}

}
   
~   ~ \caption{Comparison of pressure maps estimated by PressureVisionNet ~\cite{grady2022pressurevision} and our adapted model, using separate training and validation sets, both consisting of images from camera views 2, 3, 4, and 5. }

   \label{fig:cam2345val2345}
\end{figure*}
\begin{figure*}
\centering
  \small
\begin{tabular}{ccccccc}
\multicolumn{7}{l}{
\hspace{0.094\linewidth} \scriptsize{Input} \hspace{0.06\linewidth} \scriptsize{Anno. Pose}  \hspace{0.07\linewidth} \scriptsize{GT}   \hspace{0.03\linewidth} \scriptsize{\cite{grady2022pressurevision} w. GT Keypoints}
 \hspace{0.02\linewidth} \scriptsize{Press.Vis.~\cite{grady2022pressurevision}} }\\
\multicolumn{7}{c}{\includegraphics[width=0.65\textwidth]{"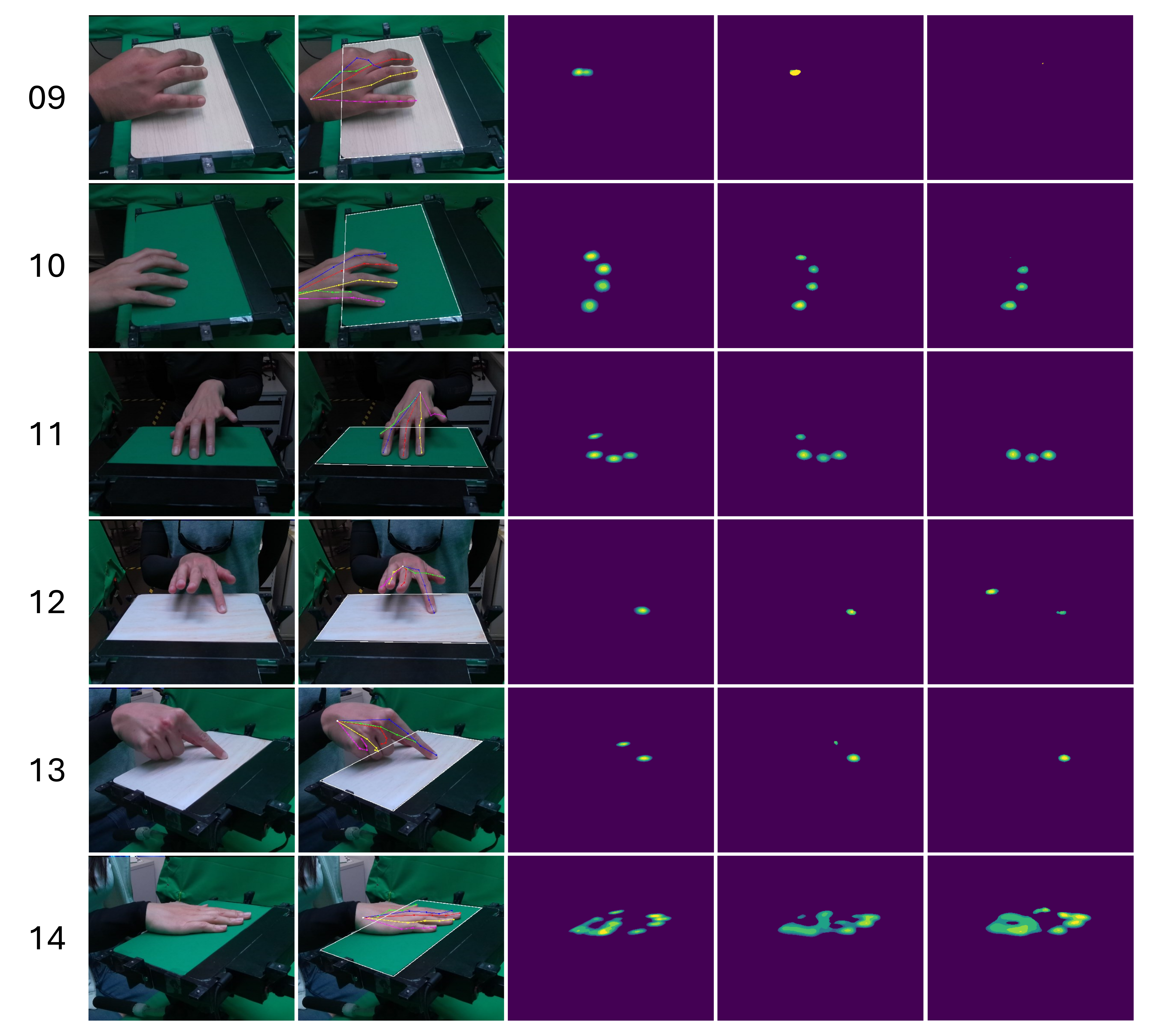"}}
  \end{tabular}
   \vspace{-1mm}

\resizebox{0.5\textwidth}{!}{
\begin{tabular}{cccccccc}
\hline
& & \textbf{09} & \textbf{10} & \textbf{11} & \textbf{12} & \textbf{13} & \textbf{14}  \\ \hline

\multirow{2}{*}{MAE $\downarrow$} & \textbf{Press.Vis. ~\cite{grady2022pressurevision}}  &52.8&77.4&119.5&76.5&62.6&199.6 \\
& \textbf{\cite{grady2022pressurevision} w. GT keypoints} &23.4&70.8&102.7&43.7&52.8&182.0\\ \hline
\multirow{2}{*}{Contact IoU $\uparrow$} & \textbf{Press.Vis.~\cite{grady2022pressurevision}} &0.0&43.6&60.1&17.4&48.2& 49.3 \\ 
& \textbf{\cite{grady2022pressurevision} w. GT keypoints} &38.5&56.0&68.9&43.5&55.7&57.1\\\hline
&  Camera \# &1&1&7&7&6&6  \\\hline
 \bottomrule
\end{tabular}

}
~   ~ \caption{Comparison of pressure maps estimated by PressureVisionNet~\cite{grady2022pressurevision} and our adapted model, evaluated using input images from cameras 1, 6, and 7. The models are the same as in Figure~\ref{fig:cam2345val2345}, which are trained  on images from camera views 2, 3, 4, and 5.}

   \label{fig:cam2345val167}
\end{figure*}

\begin{figure*}[h]
\centering
   \includegraphics[width=0.88\textwidth]{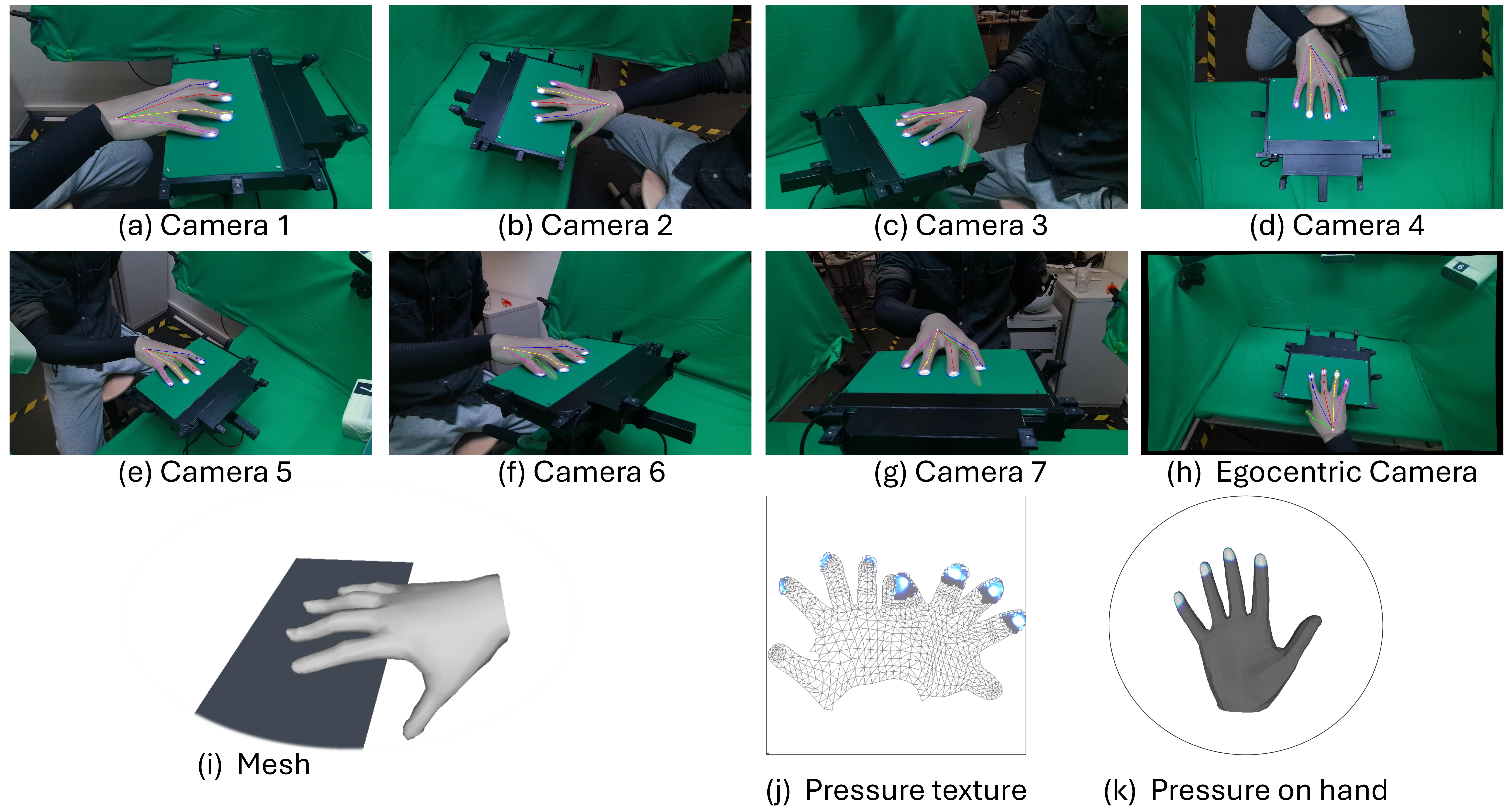}
   \caption{\textbf{Example of Annotation 1} Right hand with gesture: grasp edge with uncurled thumb down}
   \label{fig:ff1}
\end{figure*}
\begin{figure*}[h]
\centering
   \includegraphics[width=0.88\textwidth]{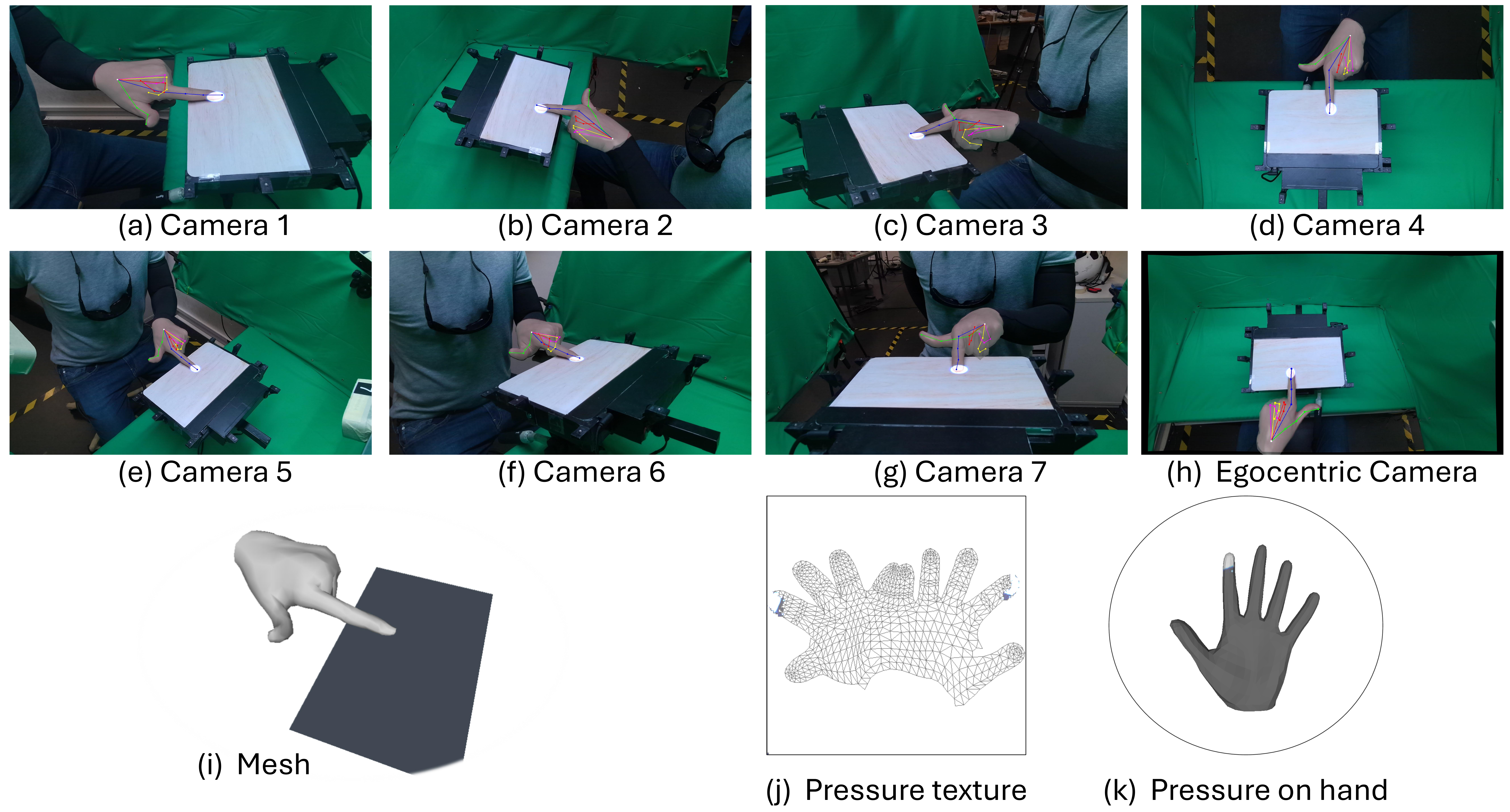}
   \caption{\textbf{Example of Annotation 2} Left hand with gesture: index press with high force }
   \label{fig:ff2}
\end{figure*}

\begin{figure*}[h]
\centering
   \includegraphics[width=0.88\textwidth]{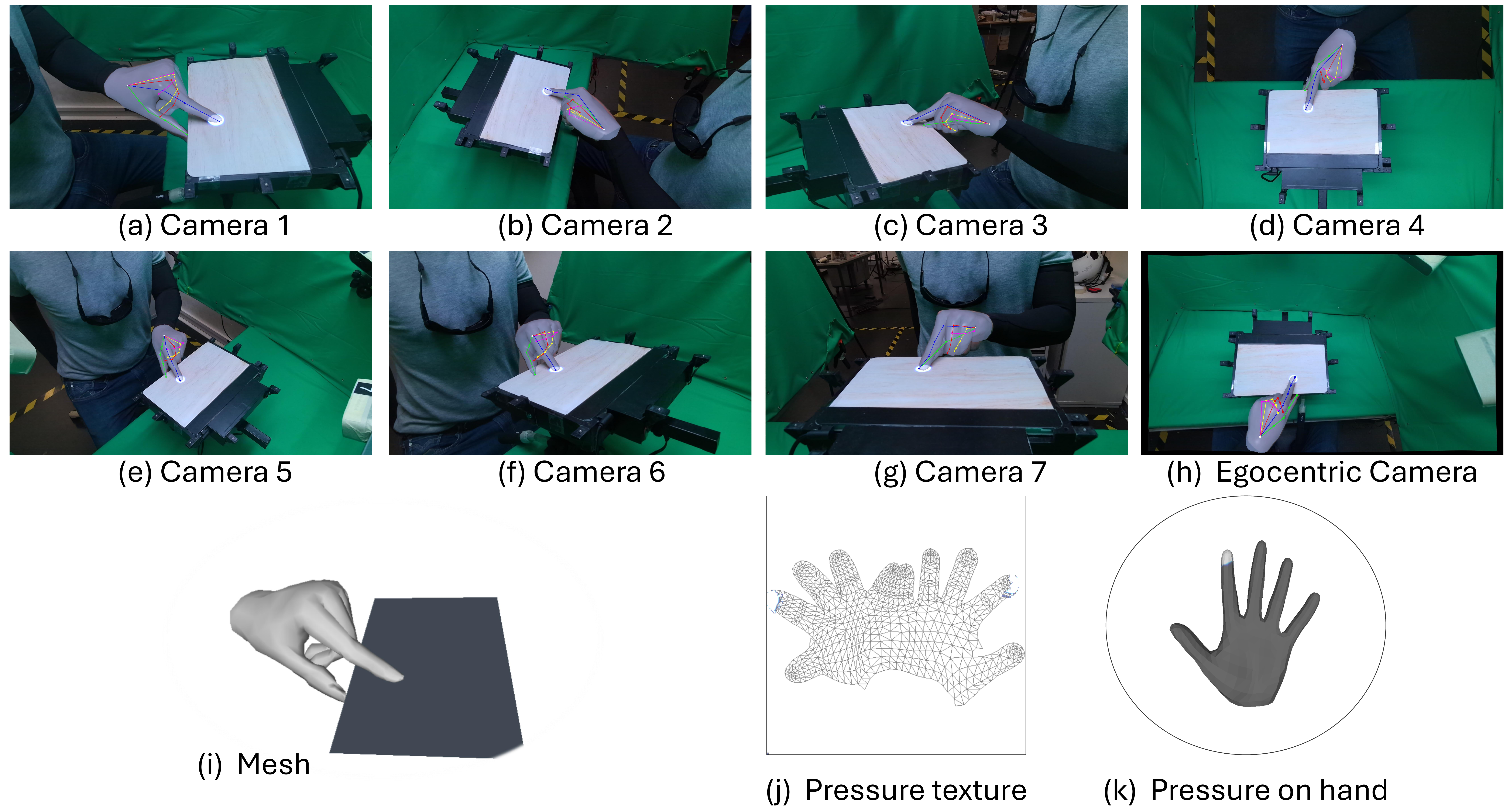}
   \caption{\textbf{Example of Annotation 3} Left hand with gesture: pinch thumb down on the edge with high force}
   \label{fig:ff3}
\end{figure*}

\begin{figure*}[h]
\centering
   \includegraphics[width=0.88\textwidth]{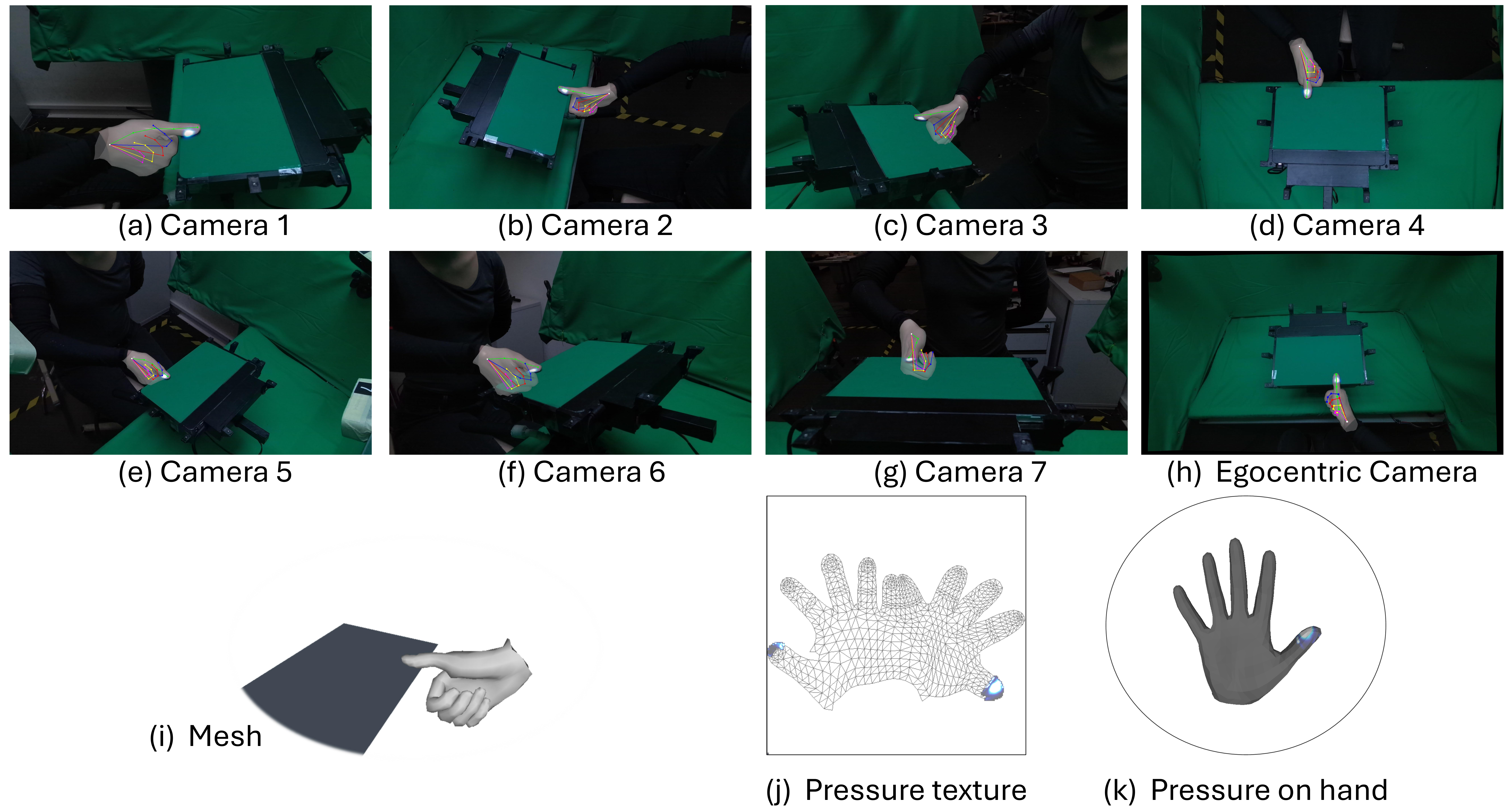}
   \caption{\textbf{Example of Annotation 4} Right hand with gesture: grasp edge with curled thumb up}
   \label{fig:ff4}
\end{figure*}

\begin{figure*}[h]
\centering
   \includegraphics[width=0.88\textwidth]{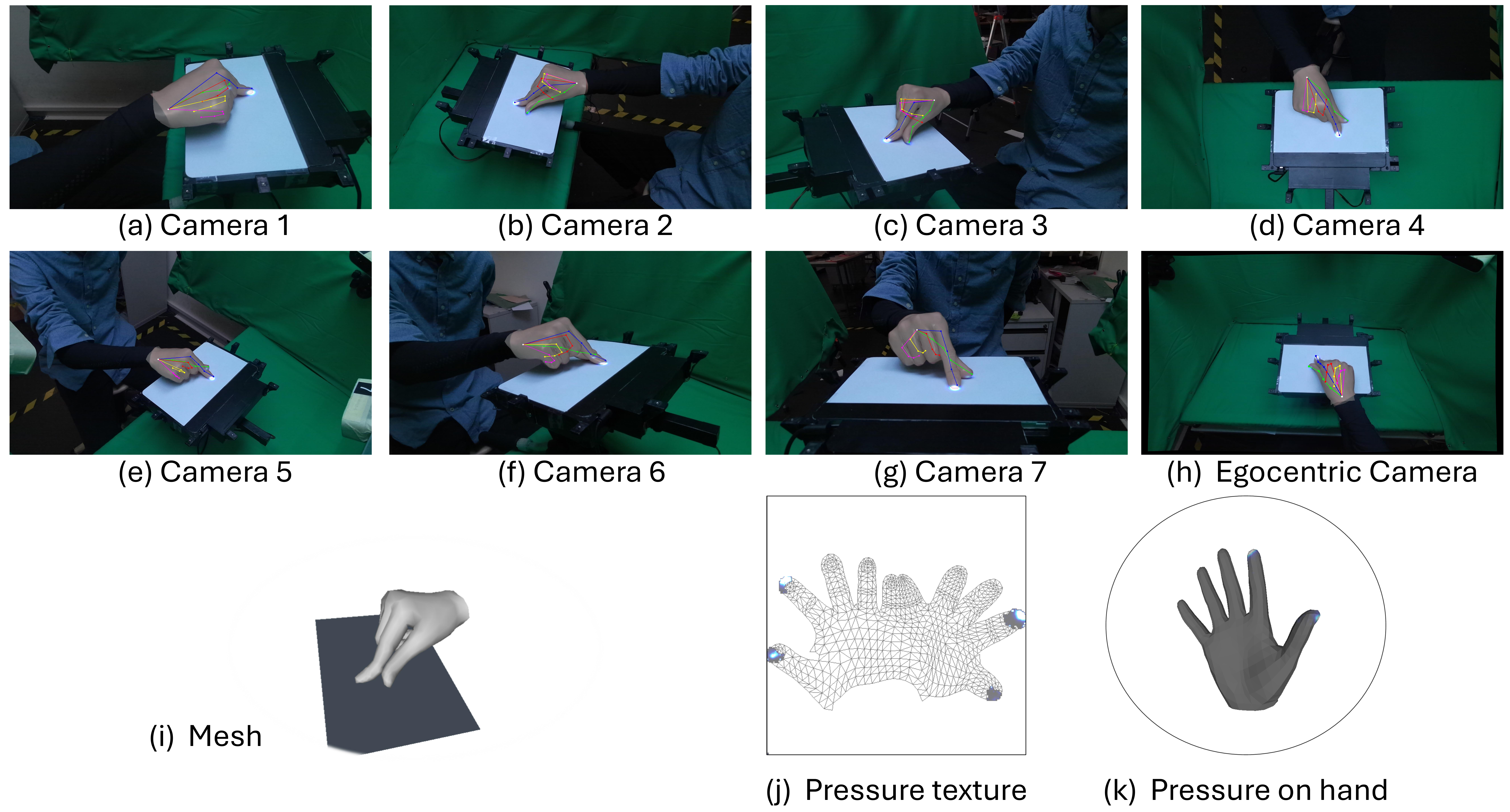}
   \caption{\textbf{Example of Annotation 5} Right hand with gesture: pinch finger zoom in and out}
   \label{fig:ff5}
\end{figure*}

\begin{figure*}[h]
\centering
   \includegraphics[width=0.88\textwidth]{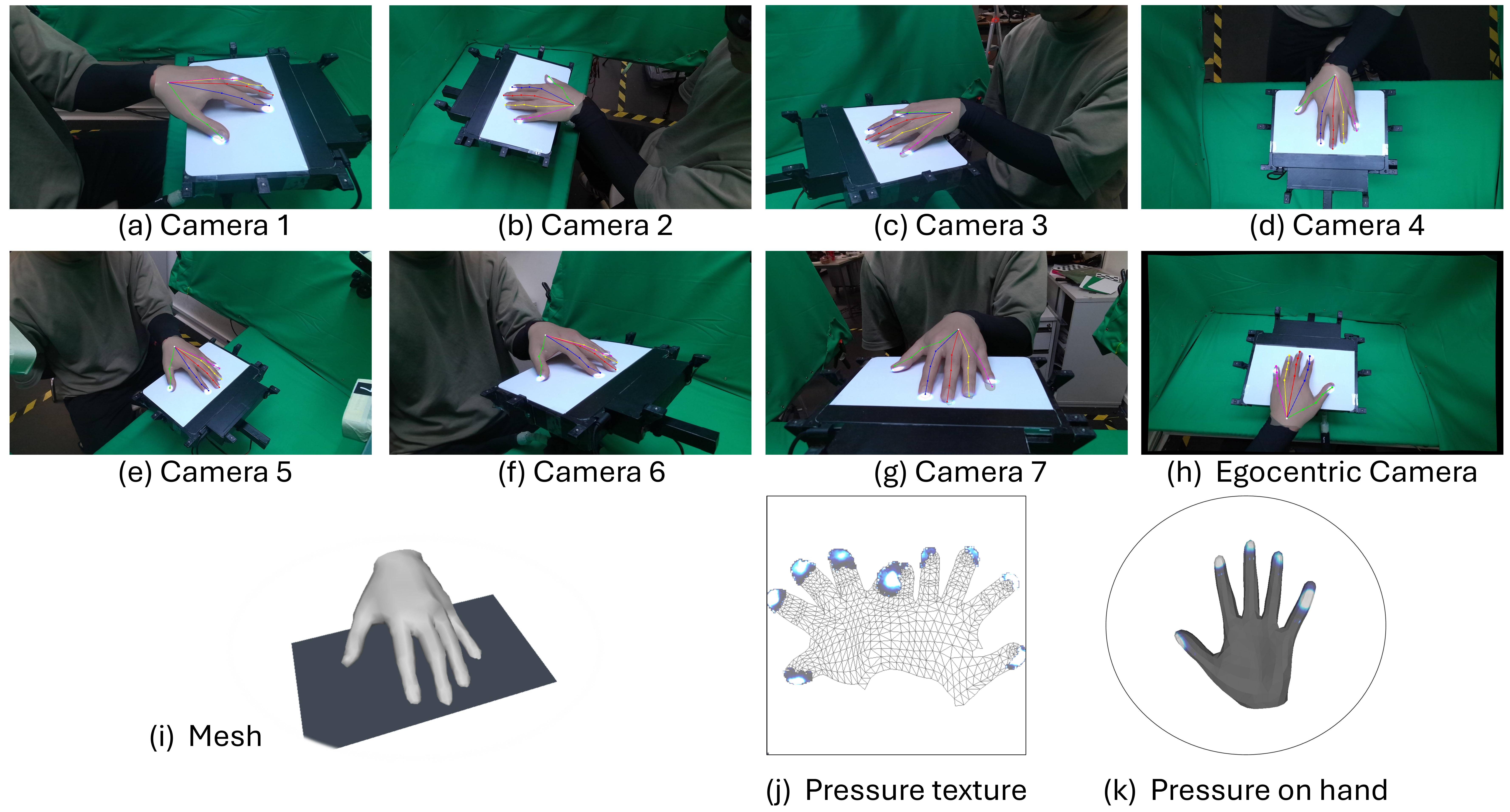}
   \caption{\textbf{Example of Annotation 6} Left hand with gesture: pull all finger towards participant}
   \label{fig:ff6}
\end{figure*}

 \clearpage
 \clearpage
{
    \small
    \bibliographystyle{ieeenat_fullname}
    \bibliography{main}
}


\end{document}